# Extraction of Clinical Information from the Non-Invasive Fetal Electrocardiogram

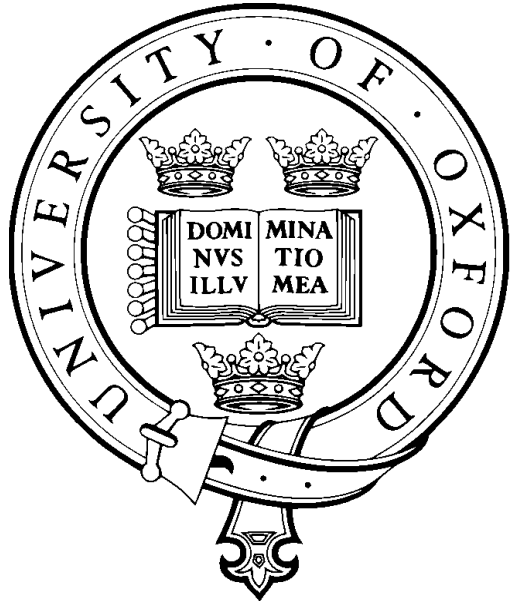

Joachim Behar

Balliol College

Supervised by
Prof. Gari D. Clifford

University of Oxford

A thesis submitted for the degree of

*Doctor of Philosophy*

Michaelmas 2014

This thesis is dedicated to the memory of:

My dearest grand mother, Lea Afriat and my dearest grand father, Fradji Behar.

<div style="text-align: right;">
Leelou Nichmat Lea bat Esther Afriat.  
Leelou Nichmat Fradji ben Eliazar Behar.
</div>

# Acknowledgements

This thesis would not have been possible without the help of numerous people and funding bodies. I would like to thank:

My supervisors Prof Gari D. Clifford and Dr Julien Oster. In particular I want to thank Gari for giving me the opportunity to do a DPhil in his research group. I want to thank both Gari and Julien for their guidance, advice and continuous support over the years. Coming from a very general Engineering background, you have very much been my scientific parents in introducing me to the beautiful world of biosignal processing.

My colleagues and friends at the institute of biomedical engineering. In particular I want to thank Alistair Johnson, Tingtng Zhu, Aoife Roebuck, David Springer, Lisa Stroux, Arvind Raghu, Thanasis Tsanas, Qiao Li and all the other department friends with whom I have been learning and living over the past three years. My thank to Fernando Andreotti (TU-Dresden) with whom I have been collaborating (and sharing many beers with!) over the last months of my DPhil. My thank to Dr Ikaro Silva and Dr George Moody (MIT) with whom I have organised the Physionet Challenge 2013 and for the many constructive discussions we have had on the topic. I also want to express my gratitude to the IT and admin teams at the department and in particular to Dr Vernon Bailey for his dedication to helping students with their computer problems.

Gari, Dr David Clifton and Prof Michael Chappell for giving me the numerous opportunity to demonstrate and teach courses on the topics of numerical methods, biosignal processnig and machine learning. This had a tremendous impact on my DPhil experience in both deepening my understanding in these topics and also developing my teaching skills. I very much enjoyed teaching these courses and hope I will be given the opportunity to pursue it in the future.

My thank to my examiners, Prof Maarten De Vos and Prof Christian Jutten, who provided me with many useful recommendations to improve the quality of the manuscript as well as numerous ideas for future research.

My family: my father Elie, my mother Daniele and my brother David for their continuous support over the years and in particular to my parents for giving me the opportunity to go to university for my bachelor, MSc and DPhil. I am very aware of the chance I had to attend the institutions I went to in France (ENSMSE), Switzerland (EPFL) and England (Oxford).

Finally I want to thank my generous financial benefactors: The Balliol French Anderson scholarship, the EPSRC and MindChild Medical Inc. North Andover. This thesis would not have been possible without the funding provided by these different entities.

# Abstract

**Extraction of Clinical Information from the Non-Invasive Fetal Electrocardiogram**
Joachim Behar
Thesis submitted for the degree of Doctor of Philosophy
Balliol college
Michaelmas 2014

Estimation of the fetal heart rate (FHR) has gained interest in the last century; low heart rate variability has been studied to identify intrauterine growth restricted fetuses (prepartum), and abnormal FHR patterns have been associated with fetal distress during delivery (intrapartum). Several monitoring techniques have been proposed for FHR estimation, including auscultation and Doppler ultrasound.

This thesis focuses on the extraction of the non-invasive fetal electrocardiogram (NI-FECG) recorded from a limited set of abdominal sensors. The main challenge with NI-FECG extraction techniques is the low signal-to-noise ratio of the FECG signal on the abdominal mixture signal which consists of a dominant maternal ECG component, FECG and noise. However the NI-FECG offers many advantages over the alternative fetal monitoring techniques, the most important one being the opportunity to enable morphological analysis of the FECG which is vital for determining whether an observed FHR event is normal or pathological.

In order to advance the field of NI-FECG signal processing, the development of standardised public databases and benchmarking of a number of published and novel algorithms was necessary. Databases were created depending on the application: FHR estimation with or without maternal chest lead reference or directed toward FECG morphology analysis. Moreover, a FECG simulator was developed in order to account for pathological cases or rare events which are often underrepresented (or completely missing) in the existing databases. This simulator also serves as a tool for studying NI-FECG signal processing algorithms aimed at morphological analysis (which require underlying ground truth annotations).

An accurate technique for the automatic estimation of the signal quality level was also developed, optimised and thoroughly tested on pathological cases. Such a technique is mandatory for any clinical applications of FECG analysis as an external confidence index of both the input signals and the analysis outputs.

Finally, a Bayesian filtering approach was implemented in order to address the NI-FECG morphology analysis problem. It was shown, for the first time, that the NI-FECG can allow accurate estimation of the fetal QT interval, which opens the way for new clinical studies on the development of the fetus during the pregnancy.

# Contents

# List of Figures

# List of Tables

**Glossary**

ADFECGDB: artificial FECG database
ADECGDB: Physionet abdominal and direct FECG database
AE: absolute error
AECG: abdominal electrocardiogram. This term refers to the electrocardiogram recorded on the maternal abdomen
AQTDB: artificial QT database
AR: auto-regressive model
BSS: blind source separation
Challenge: refers to the MIT-Physionet /Computing in Cardiology Challenge
CI: confidence interval
dB: decibels
ECG: electrocardiogram
EFM: electronic fetal monitoring
EM: expectation maximisation
EMG: electromyogram
EKF: extended Kalman filter
ESN: Echo-State Neural network
E1-E5: Challenge event 1-5
FECG: fetal electrocardiogram. This term refers to the fetal component of the abdominal electrocardiogram mixture
FFT: fast Fourier transform
FHR/MHR: fetal/maternal heart rate
FN: false negative
FP: false positive
FQRS/MQRS: fetal/maternal QRS
FQT/MQT: fetal/maternal QT
FSE: fetal scalp electrode
$F_1$: accuracy measure for FQRS detection
HR: heart rate
HRV: heart rate variability
ICA: independent component analysis
ICU: intensive care unit
IIR: infinite impulse response
IUGR: intrauterine growth restricted
KF: Kalman filter
LMS: least mean square
MECG: maternal electrocardiogram. This term refers to the maternal component of the abdominal electrocardiogram mixture or to the scalp ECG
MSE: mean square error
$MA$: muscle artifact
NI-FECG: non-invasive fetal electrocardiogram
NIFECGDB: Physionet non-invasive FECG database
PCA: principal component analysis
PCG: phonocardiogram
PLA: probabilistic label aggregator
PNIFECGDB: private non-invasive ECG database
PPV: positive predictive value/positive predictivity
PVC: premature ventricular contraction
PSFECGDB: private scalp FECG database
RLS: recursive least square
RMS: root mean square
RNN: recurrent neural network
RQTDB: real QT database
RR: intervals between R-peaks in the ECG signal
RSA: respiratory sinus arrhythmia
SNR: signal to noise ratio
SECG: scalp electrocardiogram. This term refers to the scalp electrocardiogram recorded by placing an electrode on the fetal scalp
SQI: signal quality index
SVM: support vector machine
Se: sensitivity
Sp: specificity
TN: true negative
TP: true positive
TS: template subtraction

# Part I

# Context and state of the art

# Chapter 1

# Introduction

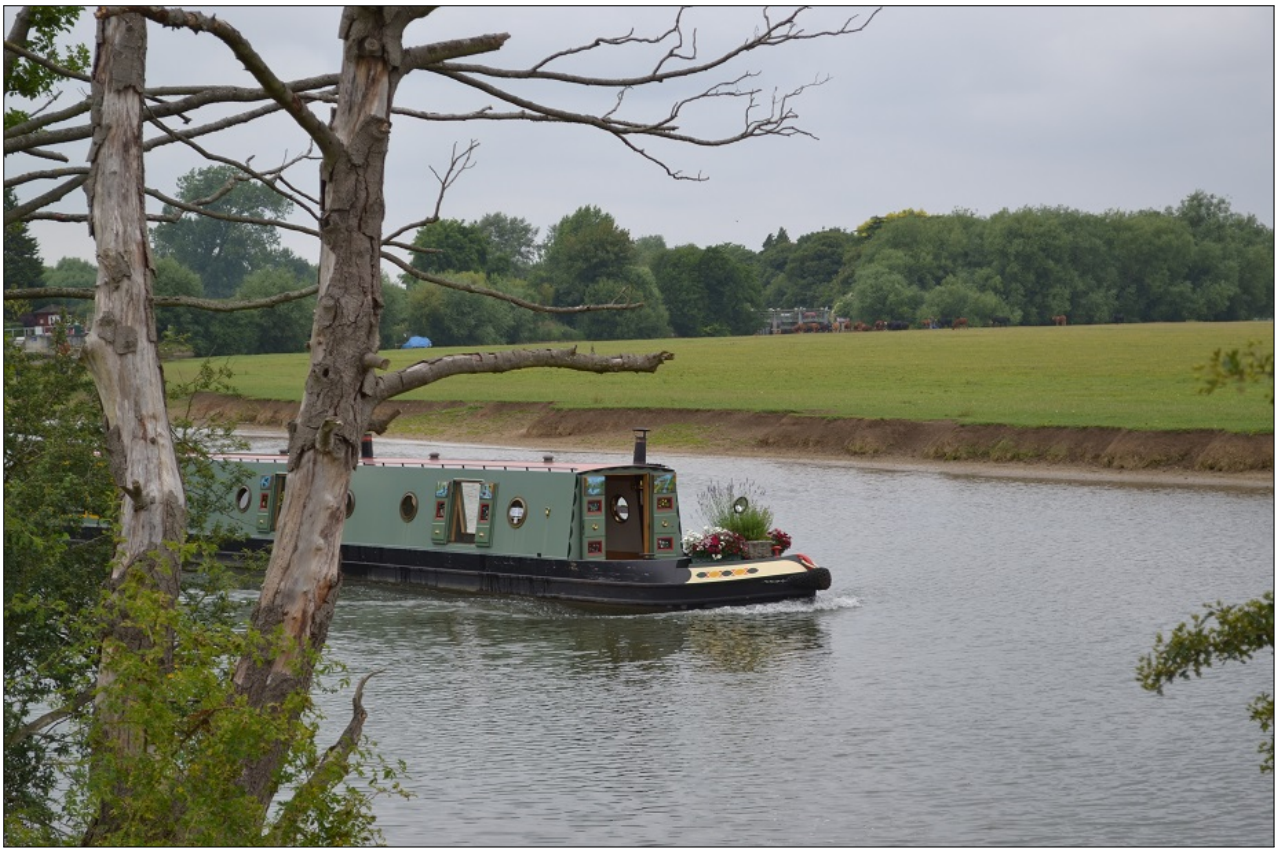

Figure 1.1: Oxford was historically a port on the river Thames and the section that passed through Oxford is called the Isis. Today, a number of barges are settled along the river or access Oxford via the Isis.

## 1.1 Motivation

In England and Wales, neonatal mortality (deaths under 28 days) was 2.9 deaths per 1,000 live births in 2011 [1]. Worldwide, an estimated 2.65 million stillbirths occur yearly, of which 98% occur in countries of low and middle income and with more than 45% in the intrapartum period [2] (i.e. occurring during childbirth). Thus, there is a need for effective monitoring techniques that can provide information on the fetal health during the pregnancy and at delivery. In addition, heart defects are the leading cause of birth defect-related death [3]. It is estimated that 9 in 1000 babies born in the UK have a congenital birth disease [4]. Nowadays, more and more cases of congenital heart disease are diagnosed prepartum during a routine ultrasound scan. However, there are some cardiac defects that cannot be identified with this modality [4] and there is interest in prepartum technology that could provide additional information about the cardiac health of the fetus.

Since the late 19th century, late decelerations of the fetal heart rate (FHR) following the peak contraction or thereafter, are known to be associated with fetal distress [5]. Intermittent observations of fetal heart sounds (auscultation) became standard clinical practice by the mid-20th century. The first FHR monitors were developed more than 50 years ago and became widely available by the mid-1970s [6]. Continuous FHR monitoring was expected to result in a dramatic reduction of undiagnosed fetal hypoxia, but disillusionment rapidly set in as studies showed that the outputs of FHR monitors were often unreliable and difficult to interpret. This lead to a large increase in the rates of a painful and expensive cesarean section, higher prevalence of postnatal depression [7], and post-operative pain negatively affecting breastfeeding and infant care [8]. There is still little evidence that reductions in adverse outcomes were attributable to the use of FHR monitors.

The non-invasive fetal ECG (NI-FECG) recorded on the maternal abdomen represents an alternative to Doppler ultrasound recording, which could provide a more accurate estimation of FHR and additional information related to the electrical activity of the fetal heart that could be obtained through the study of the FECG morphology. However, the FECG is difficult to extract from the abdominal signal mixture which has restricted its use to date.

## 1.2 Contributions and impact

This thesis, focuses on the signal processing techniques that can be used to extract clinically relevant information from the NI-FECG. Two features are of interest in this work: fetal heart rate and fetal QT (FQT). For successful extraction of both features it is necessary to accurately detect the fetal QRS (FQRS) location. This is because the FHR is directly derived from the FQRS location and QT measurement techniques use the QRS location as an anchor point. Thus, a large part of this thesis is concerned with implementing and benchmarking techniques for FQRS detection. A particular emphasis is given to the methodology for tuning and assessing these algorithms. Following the work on FQRS detection, manual measurement of the FQT from the abdominal ECG is, for the first time, compared against the FQT measured on the scalp ECG. The contributions of this thesis are summarised below:

- Review of existing NI-FECG extraction approaches. A thorough and comprehensive review of NI-FECG extraction approaches was conducted. See Chapter 3 and the corresponding journal papers [1, 2].

- Introduction of the Echo State Neural network as an adaptive filtering technique for its application to NI-FECG extraction and benchmark with existing techniques. See Chapters 6 and the corresponding journal paper [1].

- Co-development of the Physionet Challenge 2013 which addressed the topic of NI-FECG extraction. By defining a rigorous protocol for assessing participants algorithms and providing a dataset for benchmarking the many techniques on the same data, the Challenge clarified which approaches to NI-FECG were promising. See Chapters 4 and 7 and journal papers [2, 3].

- Clarification of the field; a number of publications on the topic of NI-FECG extraction have been written over the past decades but with limited insights into how these methods compare. This is due both to the lack of quantitative analysis and to the inability to benchmark the algorithms on the same data. See Chapters 3, 6 and 7 and the corresponding journal papers [1, 2].

- Demonstrated that improvement in FQRS detection was possible when using a higher baseline wander cut-cuff frequency, $f_b$, and also that combining different categories of NI-FECG extraction techniques was improving the overall performance. These elements were important in winning the Physionet challenge 2013 events E1/E2 and finishing 2nd or 3rd for all the other events.

- Development of the *fecgsyn* simulator build on the *ecgsyn* ECG model [4, 5]. The simulator can be used to generate non-stationary abdominal ECG signals and include the modelling of a variety of physiological events such as contractions and accelerations/decelerations. The known ground truth can be used to evaluate NI-FECG algorithms for FHR and morphological analysis. See Chapter 4 and the corresponding journal papers [6, 7].

- Accurate QT measurement by comparing the manually annotated NI-FECG from a commercial monitor to the scalp ECG (used as the reference). For the first time QT measurement obtained from the extracted NI-FECG was compared and proved to match the scalp QT. See Chapter 8 and the conference paper [8].

- Novel joint Bayesian extraction approach to enable the study of beat-to-beat morphological analysis of the NI-FECG. The framework builds on the work of Sameni *et al.* [9]. It was shown that accurate QT intervals can be derived from the abdominal fetal ECG. An important discussion in defining standards for evaluating NI-FECG extraction algorithms aimed at morphological analysis was also conducted. See Chapter 8, Appendix A and the corresponding conference paper [10].

- Development of signal quality indices and evaluation of them on large databases (45086, 10 sec segments) of normal and pathological adult ECG. This work has already been used by other researchers in the field, as well as enabled myself and colleagues to win Phase III of the Physionet Challenge 2014 (on the topic of *Robust Detection of Heart Beats in Multimodal Data*). See Chapter 5, Appendix E and the corresponding journal papers [11, 12, 13] as well as the Physionet Challenge 2014 conference paper [14].

- Open source code and teaching material; most of the code used in this thesis is available on `Physionet.org` or on my personal webpage (`http://joachim.behar.perso.neuf.fr/Joachim/`). The code is posted with the hope that it will both save time for any interested researcher in the field as well as pushing forward the research from this thesis.

## 1.3 Thesis outline

The objective of this thesis was to investigate and create novel fetal ECG analysis tools for clinical practice that can be used easily by a medical professional and provide actionable medical information. The thesis was divided into three logical parts: I) Context and state of the art; II) Development of databases and tools and III) novel methods for NI-FECG extraction.

### 1.3.1 Part I: Context and state of the art

The purpose of this part is to introduce the context of this thesis and motivate the development of NI-FECG extraction methods. State of the art techniques for NI-FECG extraction are reviewed and later used in Part III of the thesis for benchmarking purposes.

**Chapter 2: Fetal ECG background** This chapter introduces the problem of fetal monitoring together with the methods that are currently used in clinical practice and research for that purpose. Advantages and constraints of the different methods are discussed. A special focus is given to the development of FECG monitors and to the information that can be harnessed from this physiological signal. The chapter concludes by motivating the use of the NI-FECG.

**Chapter 3: Review of NI-FECG extraction methods** There have been a number of algorithms developed over the past decade for NI-FECG extraction from the abdominal recordings. The theory (i.e. mathematical background) behind a number of published NI-FECG extraction methods is presented in this chapter. These methods are benchmarked in Chapter 6 and 7. This chapter reviews the 'state of the art' in the field of signal processing for NI-FECG extraction.

### 1.3.2 Part II: Databases and tools

It was necessary to develop new databases given the limited number of freely available data and the lack of reference annotations. In addition, a competition was organised to enable people to evaluate their method on the same dataset and following a strict evaluation protocol (train, validation, test sets). An abdominal ECG simulator, *fecgsyn*, was created to account for the lack of pathological examples and absence of some physiological phenomena (such as contraction). These events are the most important on which to evaluate the algorithms because they are the ones where failure of the monitoring system could have dramatic clinical consequences. Yet, there is a complete absence of these kind of annotated events in real data. *fecgsyn* is the first model to provide realistic data for the evaluation of NI-FECG extraction algorithms in relatively rare but important clinical scenarios. It is also essential to exclude the noisy chest and abdominal ECG channels in the context of NI-FECG extraction. This is because the separation methods can be negatively affected by the presence of poor quality signals. A variety of signal quality indices were studied in Chapter 5 for that purpose. Experiments combining signal quality indices with NI-FECG algorithms were performed in Chapter 6.

**Chapter 4: Databases, labelling and statistics** A very limited number of open source NI-FECG databases exists. This chapter starts with a review of what fetal ECG databases were freely accessible in the field before 2013, and the results of which motivated the creation of a larger open source database which was used in the Physionet/Computing in Cardiology Challenge 2013 addressing the problem of *Noninvasive fetal ECG*. This chapter presents the baseline databases and annotations used in the thesis for evaluating the NI-FECG extraction methods. The additions made to the *ecgsyn* ECG simulator [4] are also presented. The novel generator, called *fecgsyn*, served the purpose of generating important cases that are under represented or absent from all public data, but were nevertheless essential as they are encountered in clinical practice. The extraction algorithms should be able to cope with these rhythms. The statistics for assessing the algorithm's performance are also presented.

**Chapter 5: ECG signal quality** When dealing with biosignals it is important to differentiate good quality signals (that will carry some relevant information) from bad quality signals (that will result in inaccurate inference). This chapter focuses on evaluating a variety of signal quality indices on a large database of adult ECG. The best signal quality index is further used in Chapter 6 in the context of NI-FECG extraction and also was instrumental in winning the Physionet Challenge 2014 addressing the problem of *Robust Detection of Heart Beats in Multimodal Data*.

### 1.3.3 Part III: New methods for NI-FECG extraction

Different NI-FECG electrodes configuration are appropriate depending on the intended use. A series of methods requiring a limited number of channels were evaluated in Chapter 6. Limited channel availability corresponds to ambulatory or point of care prenatal or intrapartum diagnostic monitoring, particularly in resource-constrained locations. In Chapter 7, a series of algorithms were evaluated when using a total of four abdominal channels and no maternal reference channel. The motivation for using this channel setting was three-fold. First, any given abdominal channel can be too noisy or placed at a location which is non-optimal (due to the variable fetal heart location) and thus results in a poor NIFECG extraction. Redundant channel recording is an obvious way to mitigate such an issue. Second, the usage of a chest channel can be culturally inappropriate in some instances and thus evaluating methods that make no use of a chest ECG channel is meaningful. Thirdly, similar to the adult ECG, it is possible to extract much more information from the ECG than just the heart rate. There has been a very limited number of attempts at doing so with the NI-FECG because of the challenge in reconstructing the morphology of the FECG extracted from a set of abdominal sensors but also due to the lack of methodology for evaluating such algorithms. Chapter 8 focuses on the extraction of the fetal QT interval.

**Chapter 6: Fetal QRS detection with maternal chest reference** In this chapter the benchmarking of novel and existing adaptive filtering methods is presented. These methods use an available maternal chest electrode, as the input of an adaptive filter, and regressed on the abdominal electrodes in order to cancel the maternal ECG (MECG). The performance of the algorithms is assessed when considering only one available abdominal channel (by opposition to multi channel extraction). A particular focus is given to the optimisation procedure for these algorithms.

**Chapter 7: Fetal QRS detection with no maternal chest reference** The lack of available open source databases is an issue for evaluating NI-FECG algorithms. Indeed, it is important, for objective assessment of the algorithms, to make sure that the researchers have access to the same data for designing and tuning their algorithms while evaluating them on a hidden dataset. This chapter presents the results of the Physionet Challenge 2013, which addressed the topic of *non-invasive fetal ECG*. A variety of extraction algorithms, from a total of 53 international teams were evaluated on the same database of signals following a strict evaluation protocol (training, validation and hidden test set), allowing objective comparative assessment of the performances. This chapter summarises the approach undertaken when participating in the Challenge and presents the results that were achieved.

**Chapter 8: FECG Morphology analysis** This chapter introduces the joint Bayesian extraction approach denoted 'Dual Extended Kalman filter' (EKFD). The framework and equations are presented, along with the motivation for using the EKFD in the context of NI-FECG morphological analysis. Finally, the EKFD is evaluated on synthetic and real data to extract the fetal QT interval.

**Chapter 9: Conclusion and future work** This chapter summarises the finding of this research as well as highlights promising research avenues.

**Appendices** This series of appendices present the mathematical background behind the main algorithms used in this thesis as well as the open source code and software that were released as the output of this thesis.

## 1.4  List of publications

The work from this thesis was published in a series of seven journal articles and 14 conference articles, with another five journal articles in preparation or submission. The journal papers, that are in preparation, are referenced starting with a * symbol. These include journal papers related to ECG signal quality [1, 2, 3], crowd-sourcing of medical annotations [4, 5], the FECG model *fecgsyn* [6, 7] and NI-FECG extraction [8, 9, 10, 11, 12]:

The corresponding conference papers were published: signal quality [1, 2, 3], crowd-sourcing of medical annotations [4, 5], the FECG model *fecgsyn* [6, 7] and NI-FECG extraction [8, 9, 10, 11, 12, 13, 14]:

## 1.5   Clarification on contribution

Some of the work conducted during this thesis was collaborative or built on existing work. For clarity, the core thesis is mainly limited to my work with adequate references to the tools/existing work when indicated. Precision on contribution in joint work was specified in the appendix sections. More specifically: Appendix B presents the PLA model developed with Tingting Zhu; Appendix E.1 presents the code of the *fecgsyn* simulator developed with Fernando Andreotti and the associated graphic user interface *fecgsyngui* developed with Mohsan Alvi; Appendix E.3 on the random search toolbox developed with Alistair Johnson; Appendix E.4 on the Challenge 2014 work, product of the joint work with Alistair Johnson and Julien Oster.

# Chapter 2

# Fetal ECG background

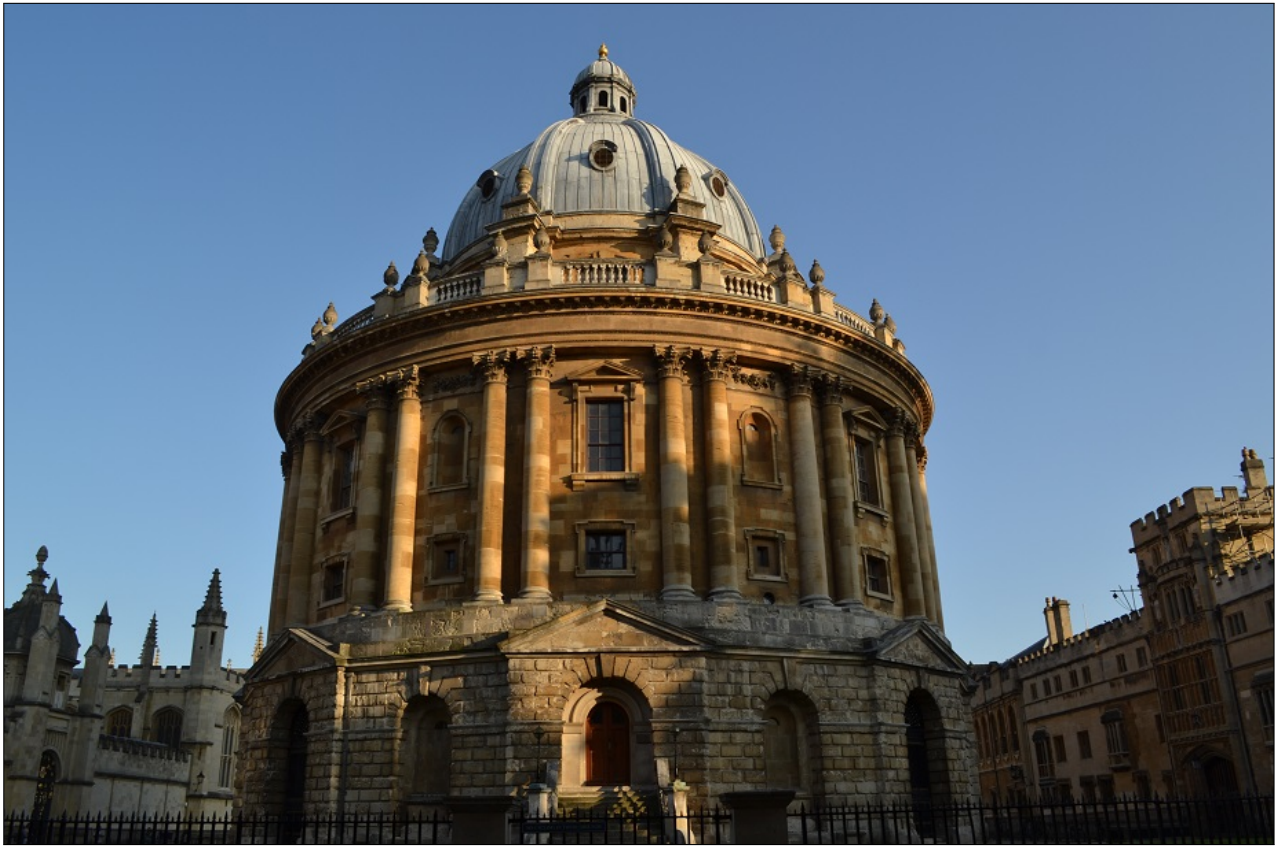

Figure 2.1: The Bodleian library is the principal research library of the University of Oxford. It is the largest library in Britain after the British library and the host of over 11 million items. A library is a place of quiet study and the host of all historical research that is needed for conducting a review.

## 2.1 Introduction

This chapter introduces the physiological concepts that are necessary to understand the remainder of this thesis. A review of the monitoring methods that are currently used in clinical practice and research is also provided. Advantages and constraints of the continuous fetal monitoring methods are discussed, and a special focus is given to the development of the fetal electrocardiogram modality and to the information that can be extracted from this physiological signal.

## 2.2 Background physiology

### 2.2.1 Fetal development

The heart starts its development with the formation of a primary tube that will separate into the four cardiac chambers and paired arterial trunks that form the adult heart [15]. Figure 2.2 shows the main stages of this development during gestation. The heart is the first functional organ to form in vertebrate embryos, and starts beating at the end of the first month. During the first two trimesters of pregnancy the fetus moves rather often, with an estimated frequency of once every 4 minutes (min) at 8-20 weeks gestational age and every 5 min at 20-30 weeks gestational age [16]. The movements are less frequent closer to term when fetal movement in utero is restricted by its own size. Figure 2.3 shows the differences in fetal presentation at delivery together with the prevalence of each of them. Most fetuses will be in the vertex position (96.8%) i.e. head down directed toward the birth channel.

The development stage of the heart together with fetal position and movement have an impact on the strength, orientation and non-stationary features of the fetal electrocardiogram (FECG) recorded from the maternal abdomen.

### 2.2.2 The fetal electrocardiogram

Einthoven used a string glavanometer in 1901 to record the electrical activity in the adult heart for the first time. Five years later, Cremer [20] performed the first non-invasive fetal ECG (NI-

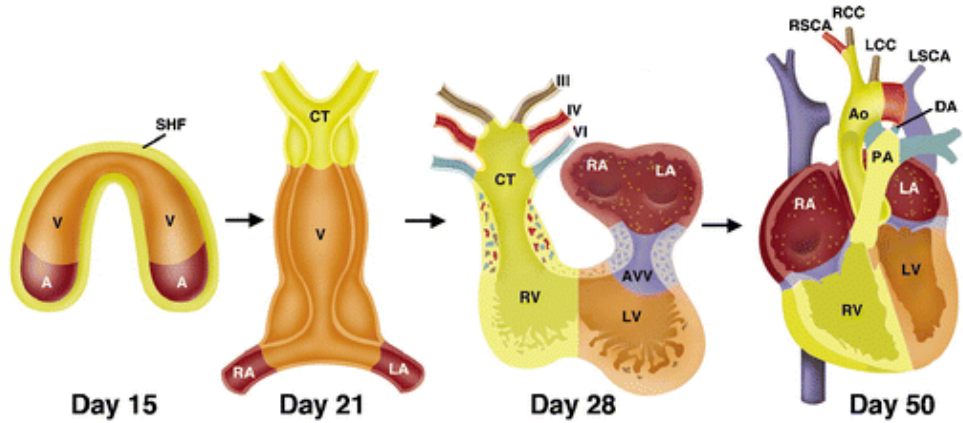

Figure 2.2: Development stages of the fetal heart during gestation. Adapted from Srivastava *et al.* [17].

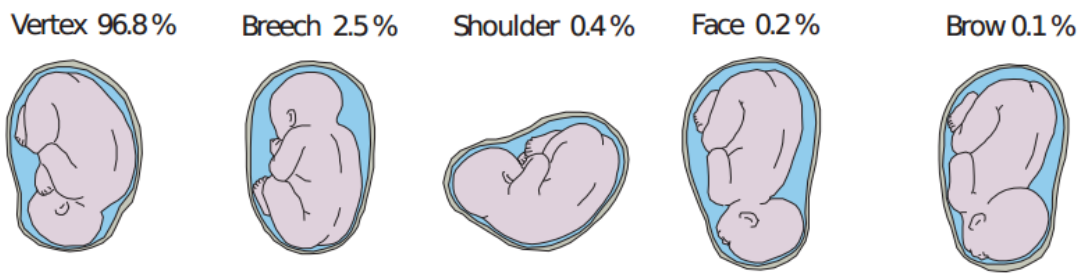

Figure 2.3: Fetal presentation and incidence in utero. Statistics from Symonds *et al.* [18] and graphics adapted from Stinstra [19].

FECG) measurement. In 1953, Smyth [21] performed the first scalp fetal electrocardiogram (SECG) by attaching an electrode to the presenting part of the fetus during labour.

The SECG can only be placed at the very last stage of the pregnancy (intrapartum) and is associated with a small risk of infection; therefore it is not used routinely. Moreover, only one differential electrode is possible, and will define the projection of the resultant cardiac activity onto a specific lead axis. Thus the three dimensional electrical field emanating from the fetal heart is unavailable, and only singletons can be monitored. Conversely, the NI-FECG is non-invasive, and can theoretically be performed at earlier stages of the pregnancy (although with a weaker field strength). However, the NI-FECG always manifests as a mixture of (significant) noise, fetal activity (from each fetus) and a much larger amplitude of maternal activity (Figure 2.4a). The signals overlap in both the time and frequency domain (Figure 2.4) and therefore accurate extraction and analysis of the FECG waveform is challenging.

The fetal electrocardiogram is similar to the adult ECG in that it contains the different characteristic waves (P, QRS and T-waves) as originally defined by Einthoven. The distinct waves represent the contraction and relaxation of the atria and ventricles caused by the propagation of the electrical activity through the heart [23]. The contraction of the myocardium is known as the *depolarisation* and the relaxation phase is known as *repolarisation*. Some important waveform delineations are depicted in Figure 2.5.

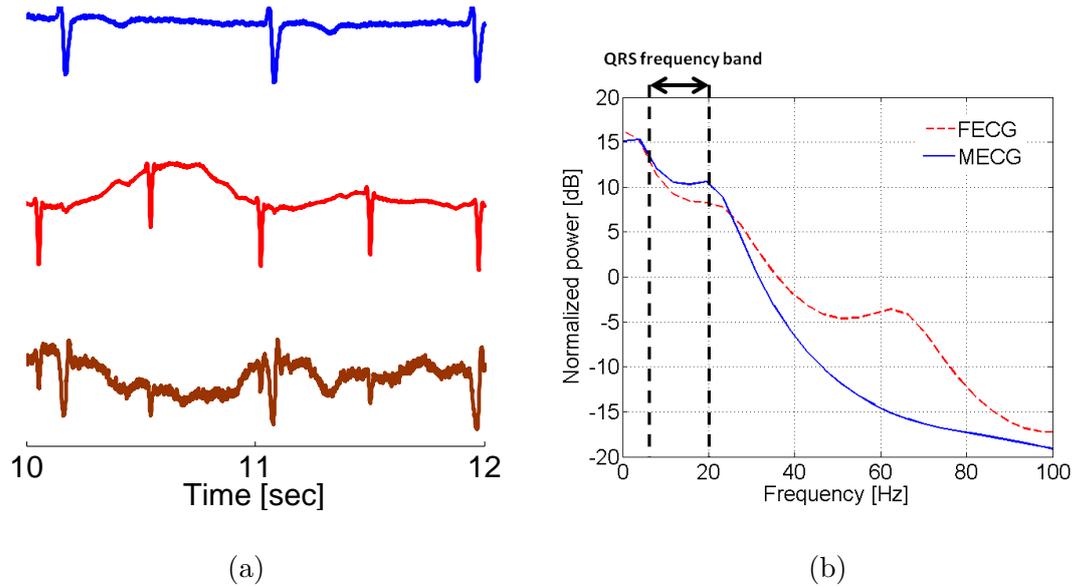

Figure 2.4: Frequency and temporal overlap of the MECG and FECG signals. (a) From top to bottom: example of maternal chest ECG, fetal scalp ECG and abdominal ECG. Note that the AECG contains a mixture of both MECG and FECG and that some FQRS are overlapping with the MQRS in the time domain. To produce (a) a notch filter at 60 Hz was used to make the FECG visible on the abdominal channel. (b) Power spectral density distribution (Burg method [22], order 20) for 5 min of scalp electrode ECG and 5 min of adult ECG. Notice the frequency overlap between the adult and fetal scalp ECG signals particularly in the frequency band of the QRS.

The electrical currents propagate through the heart as follows: the sino-atrial node acts as a pacemaker; when fired, the electrical impulse spreads through the atrial myocardium; the currents propagate through the right atrium and reach the atrioventricular node, through which the currents pass and propagate into the ventricular myocardium via the *bundle of His* and the *Purkinje fibres*. Figure 2.6 illustrates the electrical pathway through the heart and how this propagation translates into the characteristic waves that we observe on the ECG trace. The electrical current and associated potential generated by the heart is the result of the opening and closing of ionic channels at a cellular level. The synchronous activation of the myocardial cells creates an electrical field that propagates through the abdomen. The measurement of the potential on the body surface corresponds to the propagated electrical

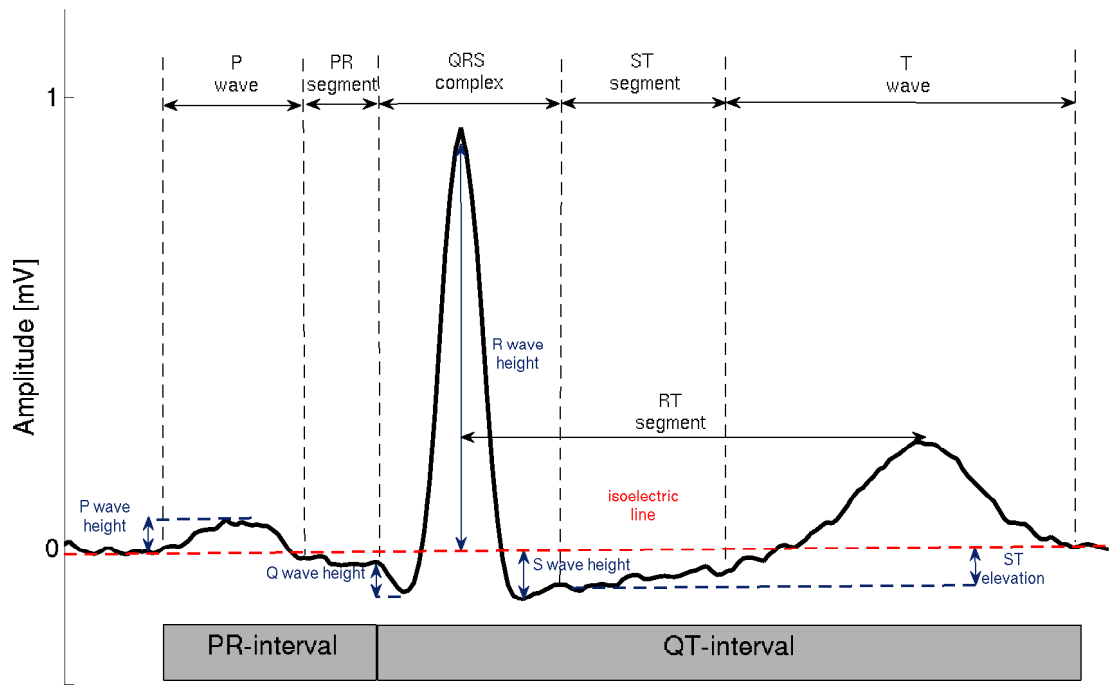

Figure 2.5: ECG waveform with characteristic waves. Note that the PR segment is sometimes called the PQ segment.

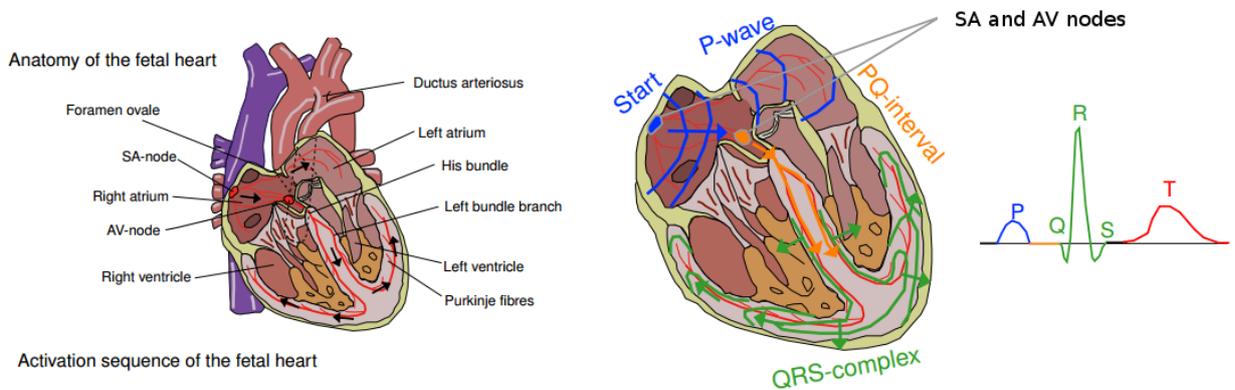

Figure 2.6: Fetal heart anatomy and electrical activation sequence with the corresponding waves recorded on the ECG. Image adapted from Stinstra [19].

activity from the myocardium to the body surface.

Despite the large similarities between the adult and fetal heart, there are some differences in HR and morphology of their ECG [9]; the FHR is usually higher than the adult one, with the normal range varying over the duration of the pregnancy. The heart rate variability is lower for the fetus than for an adult [24] and also evolves during gestation, becoming more and more complex. Finally, it is usually more difficult to clearly identify some of the FECG waves. For example, the T wave is often characterised as being weak (i.e. low amplitude) for the fetus [25].

Some aspects of the FECG waveform which differ from the adult ECG include signal strength and cycle length. The fast evolving nature of the fetal heart means that the signal processing techniques have to handle a large degree of inter-and intra-subject variability. In addition, the weaker field strength of the fetal heart and the recorded abdominal mixture make the accurate morphological reconstruction of the FECG waveform a challenging task.

## 2.3 Electronic fetal monitoring

Electronic fetal monitoring (EFM) techniques can be invasive or non-invasive with intermittent or continuous assessment. Current techniques include fetal phonocardiography, Doppler ultrasound, cardiotocography (CTG), fetal magnetocardiography (FMCG) and fetal electrocardiography. From 20 weeks onwards the heart can be heard without amplification [26], and can be monitored using Doppler ultrasound, the FECG, and the FMCG [27]. Around the 28th-32nd week of gestation, a thin layer called the vernix caseosa is formed and dissolves in the 37th-38th week in normal pregnancies [19]. This layer is highly non-conductive and thus limits the extraction of the NI-FECG. A number of studies [28, 29] have observed the adverse effect of this layer on the performance of the NI-FECG extraction. Doppler ultrasound is routinely used for FHR monitoring during pregnancy and delivery. However, despite being a non-invasive modality, the ultrasound is not passive and it has not been demonstrated that ultrasound irradiation exposure is completely safe for the fetus [30]. In addition, while using Doppler ultrasound, it can happen that the maternal HR (MHR) is picked up instead of the fetal one because of the misorientation of the ultrasound transducer. Nelson *et al.* [31] reported 5 cases out of 10000 monitored pregnancies that were fatal for the fetus where the MHR was recorded by the ultrasound instead of the FHR. Table 2.1 lists the three main non-invasive continuous EFM techniques and presents their advantages and drawbacks.

Since the introduction of the EFM in the mid 1970s and the uncertainty around the benefit of this method, the medical community has been very cautious about the adoption of any alternative method or adjunct to the CTG. In practice, this has resulted in large clinical studies targeted at assessing the benefits of novel monitoring methods. For example, fetal pulse oximetry (FPO) was recently introduced (first FDA approved device in 2000 [32]) and clinical studies were run to assess its clinical benefit. Studies on FPO showed that using this monitoring method in addition to the CTG was not associated with significant difference in neonatal outcomes [33, 32] and conclusions on whether it reduced cesarean frequency could not be drawn.

| Method | System | Gestational age | Comments |
| --- | --- | --- | --- |
| CTG | Cardiotography; ultrasound transducer and uterine contraction pressure-sensitive transducer | $\geq 20$ weeks | - contraction monitoring with a pressure transducer<br>- smoothed HR time series<br>- no beat-to-beat data and cardiac function descriptor limited to HR<br>- not passive; ultrasound irradiation |
| FMCG | Fetal magnetocardiogram. Detection of the fetal heart's magnetic field through SQUID sensors positioned near the maternal abdomen | $\geq 20$ weeks | - expensive<br>- requires skilled personnel<br>- morphological analysis of the FMCG easier than FECG because of higher SNR<br>- no long term monitoring possible to date because of apparatus size and cost |
| FECG | Standard ECG electrodes with varying skin preparation methods | $\geq 20$ weeks with dip from 28th to 37th weeks | - low cost<br>- does not required skilled user<br>- continuous monitoring possible<br>- FHR and possibly morphological analysis<br>- low SNR |

Table 2.1: non-invasive continuous fetal monitoring modalities. SNR: signal to noise ratio.

In clinical practice there are two main methods currently in use for continuous fetal monitoring: the CTG, which is the most widespread, and the SECG. However, neither method is ideal; the CTG only provides an estimate of the FHR while the SECG is invasive. The NI-FECG has been suggested as an alternative monitoring method that would combine the advantages of both the CTG and SECG i.e. be non-invasive while providing an accurate FHR, as well as the additional information on the electrical activity of the heart which are contained in the ECG. For this reason, the NI-FECG has been the subject of a lot of interest in the scientific community over the past decades [26].

## 2.4  FECG and clinical applications

Despite significant advances in the recording and analysis of adult clinical electrocardiography, such as improved signal processing techniques and the potency of digital processors, few advances have been made in the extraction and analysis of NI-FECG. This is partly due to the relatively low signal-to-noise ratio (SNR) of the FECG compared to the MECG. This low SNR is partly caused by the various media between the fetal heart and the measuring electrodes, and the fact that the fetal heart is simply smaller. Moreover, there is a less complete clinical knowledge concerning fetal cardiac function and development than adult cardiology. Another significant barrier to the analysis of NI-FECG is the paucity of (public) gold standard databases with expert annotations and reference signals, such as independent measures of the FECG (through direct scalp electrodes), heart rate, ischemia and rhythm. The key features in fetal monitoring are rhythm-related and morphology related (e.g. ST and QT changes in the FECG).

### 2.4.1  Fetal heart rate

The range of normal fetal heart rate varies significantly with gestational age. The heart starts beating at the end of the first month of the pregnancy (see Figure 2.7), when a single hollow ventricle is created (see Figure 2.2) and a ventricular rhythm is established at 60-80 bpm [23]. By the ninth week, the sino-auricular node has developed and the baseline FHR becomes 175 bpm. In the following weeks FHR declines and reaches between 110-160 bpm at delivery. There are a number of FHR patterns that have been studied and it is beyond the scope of this chapter to review this extensive literature on FHR interpretation. However, it is important to mention that the FHR variability reflects the autoregulation by the autonomic nervous system, which can be affected by factors such as the stage of pregnancy, fetal sleep state and drugs [23]. This makes its interpretation challenging.

The FHR can be used antepartum for the screening of intrauterine growth restricted fetuses [34], and it can be used at delivery as an indicator of fetal distress [35] for monitoring fetuses at risk. This analysis is, however, curently limited by the use of Doppler ultrasound which provides an estimate of the FHR and does not allow the study of beat-to-beat variability.

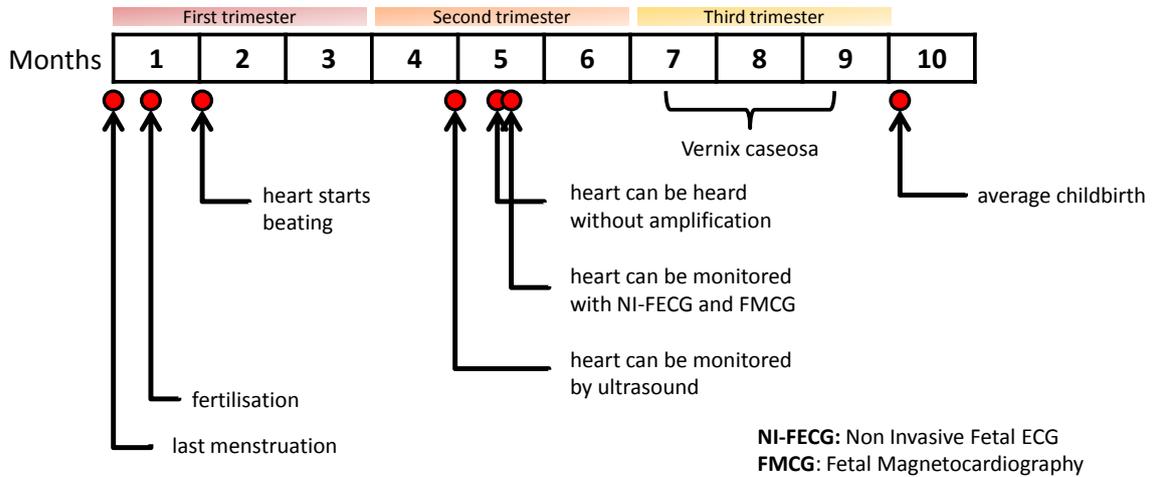

Figure 2.7: Prenatal development timeline with key landmark relevant to fetal monitoring. From 20 weeks onwards the heart can be heard without amplification [26], and can be monitored using Doppler ultrasound, the FECG, and the FMCG [27]. However the vernix caseosa forms around the 28th-32nd week and dissolves in the 37th-38th week in normal pregnancies [19] limiting NI-FECG effectiveness recording during this period.

The most accurate method for measuring FHR is the direct fetal ECG using a fetal scalp electrode. Unfortunately, it is common for the scalp electrode to lose contact because of maternal and fetal movement [36]. The NI-FECG can be used throughout the second half of pregnancy with no risk, and could provide a more accurate FHR signal than its ultrasound counterpart.

### 2.4.2 Morphological analysis

The ECG allows for interpretation of the electrical activity of the heart far beyond the analysis of only HR and heart rate variability (HRV). However, morphological analysis of the FECG waveform is usually not performed in clinical practice, with the exception of the STAN monitor (Neoventa Medical, Goteborg, Sweden) which uses an invasive scalp electrode.

There exists a number of measures, for ECG morphological analysis, that have been studied for their applications in fetal monitoring (see Chapter 6 of Symonds' book for a good summary [23]). These include: QRS complex (looking at the width and shape), R/S ratio (for fetal vector cardiography), PR/FHR ratio (inverse correlation between the PR interval and heart rate which is changing in hypoxic events), P wave morphology (inversion, notching and disappearance), PR interval, QT interval and ST-segment. A summary of the temporal and morphological intervals in the FECG is given in Table 2.2. In the next paragraphs a particular focus is given to the T/QRS ratio (which has been used in the STAN monitor) and the QT measurement (the extraction of which was studied in this thesis):

**T/QRS-ratio**: The interest in using the T/QRS ratio as a proxy for the ST segment originates from an animal experiment by Greene *et al.* [37] in 1982, where the authors examined 10 chronically instrumented fetal lambs, 115 days to term. The study showed that the normal T/QRS ratio was less than 0.30, whereas it was in the range of 0.17 to 0.59 for eight of the lambs after inducing hypoxia and reverted to normal with normoxia. The SECG-based STAN monitor uses the T/QRS ratio as a proxy for the ST segment deviation. Recently, the use of the STAN analyser, together with competency based training on fetal monitoring, showed a significant decrease in the number of cesarean sections at St George's Maternity Unit (St George's hospital, London, UK). At the same time, hypoxic ischaemic encephalopathy and early neonatal death decreased slightly [38]. However, a recent Cochrane study [39] reviewed six trials that compared the effect of analysing SECG waveforms during labour with alternative methods used for fetal monitoring, and showed that no significant difference in primary outcomes were achieved using the STAN ST proxy (based on five trials using different version of the STAN monitor with a total of 15,338 women). This suggests that the STAN proxy measure for computing ST is not accurate enough, or that the ST measure does not provide significant information to improve fetal monitoring.

**QT-segment**: In adults, changes in the T-wave occur due to myocardial ischemia. For example, it has been shown that acute myocardial ischemia will modify the duration of the QT interval and increase repolarisation heterogeneity [40]. Thus the FQT interval has been of much interest in the monitoring of fetal hypoxia. A recent study by Oudijk *et al.* [41] has shown that a significant shortening of the QT interval was associated with intrapartum hypoxia resulting in metabolic acidosis, whereas in normal labour these changes do not occur.

The assumption of these studies is that heart abnormalities or fetal hypoxic suffering will manifest in the FECG waveform. This is supported by observations in adults and animal models. However, because of the difficulty in performing accurate morphological analysis on the FECG (whether obtained through the scalp electrode which has one single electrode or through ECG sensors place on the maternal abdomen), a limited number of studies have managed to show a significant improvement in fetal outcomes or lower the proportion of cesarean deliveries. To some extent this is what Symonds *et al.* wrote in 201 "The issue of the value of current use of the FECG morphological characteristics and time intervals for the prediction of fetal compromise remains promising but unresolved" [23]. It is accurate to say that in 2014, 13 years later, the problem still remains unresolved.

| Temporal parameters | |
|---|---|
| Parameters name | Definition |
| P wave duration | duration between the onset and end of the P-wave |
| PR segment | duration between P and R peaks |
| RT segment | duration between R and T peaks |
| QRS complex | duration between the onset of the Q wave and end of the S wave |
| T wave | duration between onset and end of the T wave |

| Morphological parameters | |
|---|---|
| Parameters name | Definition |
| P wave height | defined as amplitude from isoelectric line to peak of R-wave |
| P wave area | area between the P wave and the isoelectric line |
| PR segment elevation | amplitude from the end of the P wave to start of the Q wave |
| Q wave height | defined as amplitude from isoelectric line to peak of Q-wave |
| R wave height | defined as amplitude from isoelectric line to peak of R-wave |
| S wave height | defined as amplitude from isoelectric line to peak of S-wave |
| ST segment elevation | amplitude between the end of the S wave (J point) to start of the T wave |
| T wave height | defined as amplitude from isoelectric line to peak of T-wave |
| T wave area | area between T wave and isoelectric line |
| T/QRS-ratio | ratio of height of T wave over height of R wave |

Table 2.2: Definition of temporal and morphological intervals in the FECG. (Table adapted from Symonds *et al.* [23] p 66). See Figure 2.5 for a pictorial representation of a subset of these wave delineations.

### 2.4.3 Contraction and breathing effort

Information about the timing of contractions is used to interpret the FHR pattern [18]. These contractions are usually detected using an external tocodynamometer or internal uterine pressure transducer. It is also possible to record the uterine contractile activity by using the ECG recorded on the abdomen in order to sense the electrical impulses generated by the myoelec-trical activity [42]. Figure 2.8 shows an example of abdominal ECG where the contraction event causes the quality of the abdominal signal to decrease. This effect is positive in that it suggests that the effect of the contraction can be captured by looking at the associated noise on the abdominal signal. However, it does also make the job of any NI-FECG extraction algorithm more difficult.

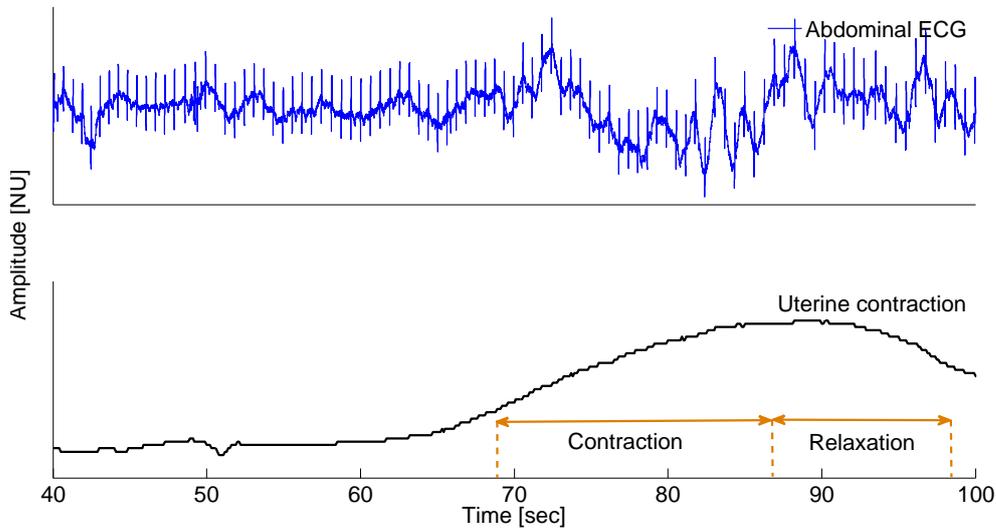

Figure 2.8: Example of a uterine contraction and its effect on the quality of the abdominal ECG signal. Note the degradation in the abdominal ECG quality after 70 seconds, which corresponds to the onset of the contraction recorded by an intrauterine pressure transducer.

Respiratory movements of the fetus were first observed by Ahlfeld in 1888 [43]. It became clearer with later studies that the breathing-like fetal movement were necessary for stimulating the development of the lung. As the oxygen is supplied by the mother during pregnancy, the lungs appear to serve no oxygenation-related purpose until birth. These movements are observed after 10-weeks of gestation [44] and at a rate of 57.2 breaths per minute (brpm, 30-33 weeks of gestation) and 47.9 brpm (37-40 weeks) for healthy fetuses [45]. In adults, breathing rate extraction from the ECG is well studied and the method is used in commercial applications. It would be theoretically possible to extract the breathing rate from the FECG signal in the same way as adult ECG. However, this has not been thoroughly studied to date.

### 2.4.4 The NI-FECG industry today

To date there are only two NI-FECG devices (to the author's knowledge) that have recently obtained FDA clearance and regularly published scientific papers on NI-FECG analysis: the Monica AN24 monitor (Monica Healthcare, Nottingham, UK) and the Meridian monitor from MindChild Medical (North Andover, MA). Both monitors have proved to be accurate in detecting the FHR, and early works on extracting morphological information have been published for the Meridian monitor [46, 47]. These recent advances in the field are very exciting. However, these studies are limited in number and population size, and the positive outcomes of these devices on fetal monitoring are yet to be established.

### 2.4.5 Conclusion

In medical practice 1-D Doppler ultrasound is usually employed to measure the FHR, but this requires frequent repositioning of the ultrasound transducer and is less accurate than the scalp electrode method. Moreover, little progress has been made in the use of FHR to provide clinically actionable information. In contrast, some studies have shown that the FECG morphology was promising in identifying abnormalities (e.g. [41]). Enabling the use of beat-to-beat and morphological information derived from the NI-FECG would open a whole area of research on ante and intrapartum fetal monitoring.

## 2.5 Summary and conclusion

In summary NI-FECG has the potential to provide:

- Rhythm information using the extracted FHR. The NI-FECG can provide a more accurate FHR estimation than Doppler ultrasound and opens up research opportunities into beat-to-beat variability analysis.

- ECG morphological information, such as PR, QT intervals and ST level. Studies of these parameters have proven to be clinically relevant in adults and animal models, and thus have the potential to add information to the monitoring of the fetal heart.

- Contraction monitoring by sensing the electrical impulses generated by the myometrial activity [42]. Knowledge of contraction occurrence helps to interpret the FHR in clinical practice.

- Fetal respiration, fetal movement (as suggested in [9]) and fetal orientation.

The challenges for extracting this information from the NI-FECG signal were identified:

- Relatively low signal-to-noise ratio of the FECG compared to the MECG. This is due to the size of the fetal source which is much smaller than its maternal counterpart.

- Fetal non-stationarities, such as fetal movement, which are frequent at the earlier stages of pregnancy or fetal respiratory efforts.

- Need for algorithms that either do not require the intervention, or require minimal intervention from a medical professional.

- The limited feasibility in recording the NI-FECG from the formation of the vernix caseosa layer and until it dissolves.

- Metrics that provide a confidence measure related to the extraction performances and avoid medical errors such as confounding the FHR and MHR.

- Algorithms that enable the reconstruction of the FECG waveform (as opposed to being restricted to fetal R-peak detection) from the NI-FECG mixture.

Commercial applications for NI-FECG monitoring are in their early days, and there is a growing interest in improving their performance and prospect in order to reach the point where they will benefit fetal monitoring by providing actionable information to the clinician. As such, both because it is a non-invasive technique and because of the information the NI-FECG has the potential to provide, there is a strong motivation for pushing forward the research in this area.

# Chapter 3

# Review of NI-FECG extraction methods

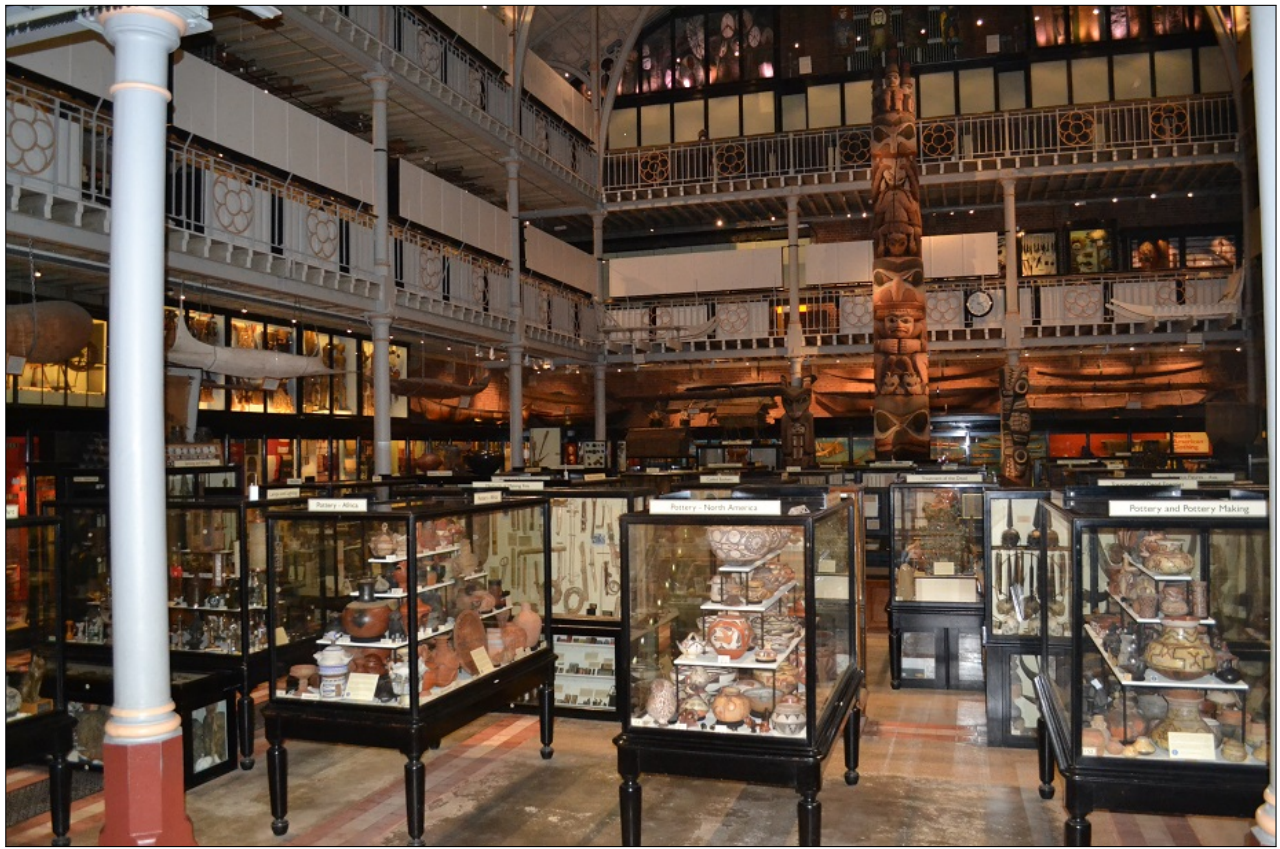

Figure 3.1: The Oxford Pitt Rivers museum displays the archaeological and anthropological collections of the university. It was created in 1884 by Lt-General Augustus Pitt Rivers. The high density of objects that can be found in this museum makes a careful and timely review necessary in order to take the most of this historical shambles.

## 3.1 Introduction

The primary feature that any algorithm must extract from the abdominal electrocardiogram (AECG) signal mixture is the fetal QRS complex (FQRS) location. This peak detection is used for computing the FHR, detecting rhythm abnormalities, or further used as an anchor point for extracting features from the fetal electrocardiogram (FECG) waveform. Ascertaining the location of the FQRS is simplified by first separating the FECG from the AECG, and several approaches have been suggested in the literature. These include: principal component analysis [48], independent component analysis [49, 50], or periodic component analysis ($\pi$CA) [51] which makes use of the ECG's periodicity. In essence, these approaches are a form of blind source separation (or in the case of $\pi$CA semi-blind source), which aim to separate the underlying statistically independent sources into three categories: maternal ECG (MECG), FECG and noise. Other techniques, which operate in lower dimensions (i.e. using a lower number of, or single, abdominal channel) include adaptive filtering [52], template subtraction [48, 53, 54, 55] and Kalman filtering (KF) [9, 56]; see Sameni and Clifford [26] for a good overview of the methods. Despite many interesting theoretical frameworks, the robustness of most of these methods has not been sufficiently quantitatively evaluated and little progress has been made in their use. This is mainly due to three factors: (1) the lack of gold standard databases with expert annotations; (2) the underdeveloped methodology for assessing the algorithms; (3) the absence of open source code makes the re-implementation of the original algorithms prone to errors, and makes objective benchmarking difficult if not impossible.

The purpose of this chapter is to review and introduce the non-invasive FECG (NI-FECG) extraction methods that were used in this thesis for FQRS detection. The presented methods were divided between the temporal methods (referring to methods that operate in the observation domain) and the spatial methods (referring to methods that are used in the source domain after they have gone through some blind source separation transformation). A particular emphasis is given to the assumptions upon which the methods rely.

## 3.2 Temporal methods

### 3.2.1 Template subtraction

Five template subtraction (TS) techniques were implemented following the works of [48, 55, 53, 54]. These algorithms have various degrees of adaptability, but all are based on the construction of an MECG template cycle which is subtracted from the subsequent MECG cycle of the AECG mixture. This process is performed for each individual abdominal channel. The different methods are denoted TS, $TS_c$ [53], $TS_m$ [54], $TS_{lp}$ [55] and $TS_{pca}$ [48] and are described in the following sections.

#### 3.2.1.1 Simple template subtraction

TS corresponds to the simplest TS technique: an MECG template cycle is built centred on the maternal R-peaks, with a fixed duration of 0.20 sec, 0.10 sec and 0.40 sec for the P, QRS and T wave respectively. $TS_c$ and $TS_m$ were implemented as described in Martens *et al.* [54]. In the case of $TS_c$, the template MECG cycle was scaled with a constant and the resulting template was subtracted from each individual MECG cycle. In the case of $TS_m$, the scaling procedure was performed for the P, QRS, and T waves independently. In the case of $TS_c$, the MECG template $\underline{t}$ was scaled with a constant $a$ in order to reduce the mismatch between the template and each individual MECG cycle $\underline{m}$[1]. The scaling constant $a$ was selected to minimise the mean square error (MSE, $e^2$) between $\underline{m}$ and $\underline{t}$ (i.e. solving $e^2 = min(||\underline{t}a - \underline{m}||^2)$) [54]. In the case of $TS_m$, the scaling procedure was performed for the P, QRS, and T waves independently. The problem becomes:

$$e^2 = min(||\mathbf{T}\,\underline{a} - \underline{m}||^2), \tag{3.1}$$

---

[1] In other words, $\underline{m}$ corresponds to one incoming MECG cycle whereas $\underline{t}$ is a template MECG cycle built by selecting and averaging a number of different MECG cycles. $\underline{t}$ can be updated online over time with incoming MECG cycles.

$$\mathbf{T} = \begin{bmatrix} \underline{\mathbf{t}}_P & 0 & 0 \\ 0 & \underline{\mathbf{t}}_{QRS} & 0 \\ 0 & 0 & \underline{\mathbf{t}}_T \end{bmatrix}, \quad \underline{\mathbf{a}} = (\mathbf{T}^T\mathbf{T})^{-1}\mathbf{T}^T\underline{\mathbf{m}} \quad (3.2)$$

where $\underline{\mathbf{a}} \in \Re^3$ is the scaling vector and $\underline{\mathbf{t}}_P, \underline{\mathbf{t}}_{QRS}, \underline{\mathbf{t}}_T$ are the parts of the template corresponding to the P, QRS and T wave respectively. $TS_{lp}$ was implemented as described in [57]: the template ECG was built by weighting the former cycles in order to minimise the MSE (as opposed to the other TS methods where the weights of the different cycles are equal). For all of these methods, the most important parameter to optimise was the number of cycles to use in order to build the MECG template (denoted nbC). In order to account for the non-stationary nature of the ECG signal, the template was updated by integrating cycles as they were processed, while removing the contribution of the oldest cycles. Misdetections were handled by rejecting incoming cycles that did not match with the template. For that purpose, the Pearson's correlation coefficient between the template cycle and an incoming cycle was computed:

$$r_{tc} = \frac{\sum_{i=1}^{n}(\underline{\mathbf{t}}_i - \bar{\mathbf{t}})(\underline{\mathbf{c}}_i - \bar{\mathbf{c}})}{\sqrt{\sum_{i=1}^{n}(\underline{\mathbf{t}}_i - \bar{\mathbf{t}})^2}\sqrt{\sum_{i=1}^{n}(\underline{\mathbf{c}}_i - \bar{\mathbf{c}})^2}} \quad (3.3)$$

where $\underline{\mathbf{t}}$ and $\underline{\mathbf{c}}$ are the template and incoming ECG cycle respectively, and $\bar{\mathbf{t}}$ and $\bar{\mathbf{c}}$ correspond to their mean. An incoming cycle with $r_{tc} \leq 0.8$ was rejected (i.e. not used to update the template cycle $\underline{\mathbf{t}}$).

### 3.2.1.2 PCA based template subtraction

PCA aims to identify a meaningful basis to re-express a dataset. It can be used for dimensionality reduction, source separation and visualisation. In this work, $TS_{pca}$ is an implementation of the method introduced by Kanjilal *et al.* [48]. The design matrix $\mathbf{M}$ can be written:

$$\mathbf{M} = \begin{bmatrix} m_{11} & \cdots & m_{1q} \\ \vdots & \ddots & \vdots \\ m_{p1} & \cdots & m_{pq} \end{bmatrix} \quad (3.4)$$

$\mathbf{M} \in \Re^{p \times q}$ with $p$ variables and $q$ observations, and where each row corresponds to one MECG cycle with all the maternal R-peaks lying in the same column. Singular value decomposition (SVD) can be used to obtain the principal components (PCs) of $\mathbf{M}$ and the components representing the maternal contribution to the cycles can be subtracted to the overall cycles:

$$\mathbf{M}_{res} = \mathbf{M} - \sum_{j=1}^{r} \underline{\mathbf{w}}_j \underline{\mathbf{w}}_j^T \mathbf{M} \qquad (3.5)$$

where $r$ is the number of PCs representing the maternal contribution (the higher $r$, the more adaptive the method) and $\underline{\mathbf{w}}_j \in \Re^p$ corresponds to the direction of the $jth$ PC. FQRS detection can then be performed on the reconstructed residual from $\mathbf{M}_{res}$. In a similar manner as for the other TS techniques, the design matrix can be updated by integrating new cycles and removing the oldest from the design matrix using the correlation coefficient from equation 3.3. This way the PCA basis can be updated to take into account the non-stationary nature of the ECG signal. Figure 3.2 illustrates the $\text{TS}_{pca}$ method on a real example. Further explanation on the mathematical basis of PCA can be found in Appendix C.

### 3.2.2 Adaptive filtering

The classical method for removing noise from a corrupted signal is to filter it. The filter can be fixed (i.e. its transfer function is constant) or adaptive. In the context of FECG extraction, adaptive noise cancellation [52] is commonly used to suppress noise from the mixture of signals in the AECG. The AECG $y(n)$ is treated as the sum of the signal of interest, the FECG $s(n)$, and noise $\eta(n)$, i.e. $y(n) = s(n) + \eta(n)$. $\eta(n)$ corresponds to the combination of the MECG, other physiological signals such as muscle noise and artifacts such as movement. As the signal recorded on the chest does not have an FECG component due to its location, it serves as an observation of the noise and as a reference for the noise cancelling field. The abdominal noise $\eta(n)$ is adaptively removed by a filter whose coefficients $\underline{\mathbf{w}} = [w_1, ..., w_N]$ form a finite impulse response filter, with $N$ being the number of coefficients or weights that are recursively updated in order to minimise an error signal $e(n)$. Thus, the goal of the adaptive filter methods is to learn a model with input $u(n)$ and output $\hat{\eta}(n)$, where $\hat{\eta}(n)$ matches the target signal $y(n)$ as closely as possible in the least mean square error sense. By filtering $u(n)$ to match $y(n)$, the resultant signal $\hat{\eta}(n)$ will primarily resemble the major source of noise in the abdominal mixture, the MECG. Subtracting $\hat{\eta}(n)$ from the abdominal mixture results in the suppression of the MECG. This process is represented in Figure 3.3.

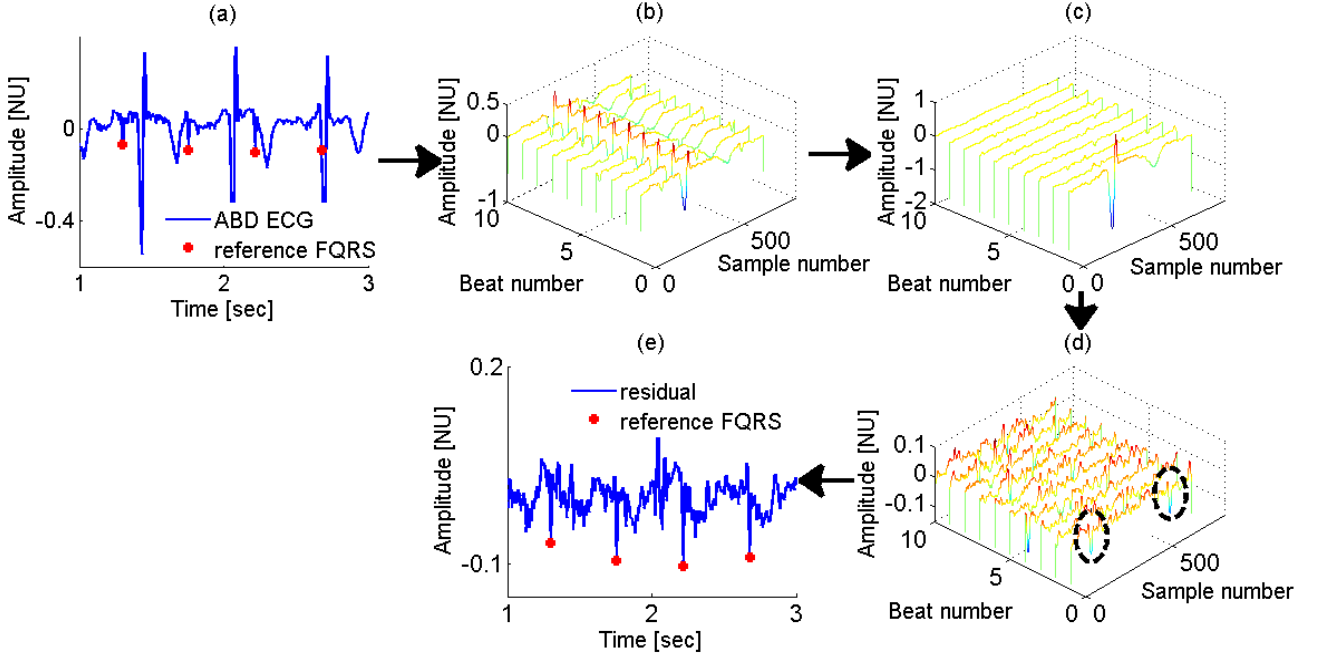

Figure 3.2: TS$_{pca}$ method. (a) 3 sec of AECG mixture (record a39, first AECG) with FQRS position (red dots); (b) stack of MECG cycles (format of the design matrix $\mathbf{M}$ with $p = 10$ cycles and $q = 700$ data points). (c) PCA transformation of the MECG stack in (b). Note that most of the information is contained in the first 2-3 principal components (PCs); (d) residual after subtracting the first and second PCs contribution from the ECG stack in (b). Note that the MECG has been cancelled and only the residuals containing the FQRS are left. In circled black, two FQRS are highlighted on the residual of the first cycle; (e) time reconstructed residual signal ('de-stacking').

### 3.2.2.1 The LMS adaptive filter

The least mean square (LMS) adaptive filter applied to NI-FECG extraction was first published by Widrow *et al.* in 1975 [52], but only qualitative results were demonstrated. LMS is used to find filter coefficients that minimise the MSE $e^2(n)$ between the filter output $\hat{\eta}(n)$ and the desired response or target $y(n)$. Let $\underline{\mathbf{u}}(n) = [u_1(n - N + 1), .., u_1(n)]^T$, $\forall\, n > N$ be a segment of the input signal (with $N$ being the last $N$ input samples), $\underline{\mathbf{w}}(n) = [w_1(n), ..., w_N(n)]$ be the filter weights, and $e(n) = y(n) - \underline{\mathbf{w}}^T \underline{\mathbf{u}}(n)$ the error rate at each step. The optimal weight vector $\underline{\mathbf{w}}_o$, also called the Wiener weight vector, is given by $\underline{\mathbf{w}}_o = \boldsymbol{R}^{-1}\boldsymbol{P}$ where $\boldsymbol{R}$ is the input correlation matrix and $\boldsymbol{P}$ is the cross correlation between the desired response $y(n)$ and $\underline{\mathbf{u}}(n)$.

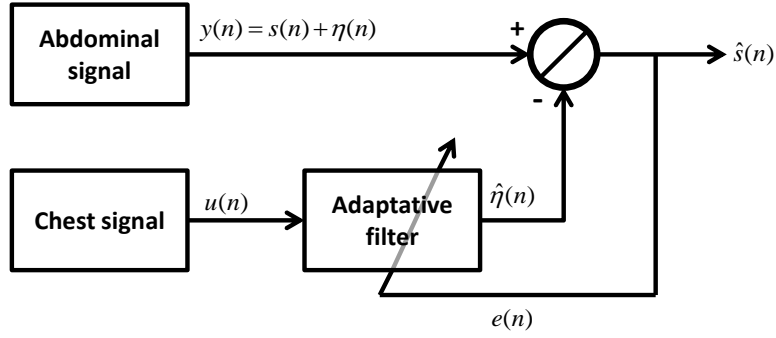

Figure 3.3: Adaptive noise cancelling block diagram in the case of one reference input $u(n)$. On the diagram : the FECG $s(n)$, the noise $\eta(n)$, the abdominal ECG $y(n) = s(n)+\eta(n)$, the chest signal $u(n)$, the estimated noise $\hat{\eta}(n)$, the estimation error $e(n)$ and the output signal $\hat{s}(n)$. $n$ corresponds to a time index.

LMS algorithms aim to estimate the filter weights that minimise the MSE. For that purpose the LMS adaptive filter from Widrow *et al.* [52] uses the steepest descent gradient method in order to adaptively adjust the value of the weight parameters at each time step. In this method the new estimate of the weight vector $\underline{\mathbf{w}}(n+1)$ is equal to the present estimate of the weight vector $\underline{\mathbf{w}}(n)$ plus a change proportional to the negative gradient at the nth iteration [58]. Thus the weight update is given by: $\underline{\mathbf{w}}(n+1) = \underline{\mathbf{w}}(n) - \mu \nabla E[e^2(n)]$. By assuming that the true gradient of the error can be approximated by the instantaneous gradient of the error, it can be written: $\nabla E[e^2(n)] \approx \nabla e^2(n) = -2e(n)\underline{\mathbf{u}}(n)$. The weight update equation then becomes $\underline{\mathbf{w}}(n+1) = \underline{\mathbf{w}}(n) + \mu 2 e(n)\underline{\mathbf{u}}(n)$. The key LMS adaptive algorithm steps can be summarised:

$$\hat{\eta}(n) = \underline{\mathbf{w}}^T(n-1)\underline{\mathbf{u}}(n), \tag{3.6}$$

$$e(n) = y(n) - \hat{\eta}(n), \tag{3.7}$$

$$\underline{\mathbf{w}}(n) = \underline{\mathbf{w}}(n-1) + \mu e(n)\underline{\mathbf{u}}(n), \tag{3.8}$$

where 3.6 gives the filter prediction, 3.7 is used to evaluate the error and 3.8 is used to update the filter weights at each sample $n$. Note that there are two parameters to set in order to design the adaptive filter: the filter length $N$ and the step size $\mu$. Both these parameters must be optimised on a training set. In the context of this work, the use of the LMS technique implicitly assumes that there is a linear relationship between the maternal waveform recorded on the chest and on the abdomen.

#### 3.2.2.2 The RLS adaptive filter

The recursive least square (RLS) algorithm minimises the total squared error between the desired signal and the target signal. In contrast to the LMS, which only considers the current error value to adapt its coefficients, the RLS considers the total error from the beginning of the signal to the incoming data point. The forgetting factor, $\lambda \in [0\ 1]$, defines what proportion of past data contribute to the filter coefficient update. In the extreme case of $\lambda = 1$ all past data contribute equally, and as $\lambda$ approaches zero only the most recent datapoints have a higher weight. This translates into finding the parameters so that the following 'loss-function', $\epsilon(n)$, is minimised:

$$\epsilon(n) = \sum_{i=1}^{n} \beta(n,i)e(i)^2 = \sum_{i=1}^{n} \beta(n,i)[y(i) - \hat{\eta}(n,i)]^2 \tag{3.9}$$

with $\hat{\eta}(n,i) = \underline{\mathbf{w}}^T(n)\underline{\mathbf{u}}(i)$ and where $\beta(n,i) = \lambda^{n-i}$ in the case of the exponentially weighted least squares solution. The RLS algorithm updates the filter coefficients at each iteration [59, 60] as follows:

$$e(n) = y(n) - \underline{\mathbf{w}}^T(n-1)\underline{\mathbf{u}}(n) \tag{3.10}$$

$$\underline{\mathbf{k}}(n) = \frac{\mathbf{P}(n-1)\underline{\mathbf{u}}(n)}{\lambda + \underline{\mathbf{u}}(n)^T \mathbf{P}(n-1)\underline{\mathbf{u}}(n)} \tag{3.11}$$

$$\mathbf{P}(n) = (\mathbf{I} - \underline{\mathbf{k}}\,\underline{\mathbf{u}}(n)^T)\mathbf{P}(n-1)\frac{1}{\lambda} \tag{3.12}$$

$$\underline{\mathbf{w}}(n) = \underline{\mathbf{w}}(n-1) + \underline{\mathbf{k}}(n)e(n), \tag{3.13}$$

where $\mathbf{P} = \varphi(n)^{-1}$ with $\varphi(n) = \sum_{i=1}^{n} \lambda^{n-1}\underline{\mathbf{u}}(i)\underline{\mathbf{u}}(i)^T$ and $\mathbf{I}$ is the identity matrix. There are two important parameters in the RLS: the forgetting factor $\lambda$ and the number of filter coefficients $N$. RLS tends to converge faster than LMS and is usually more accurate. However, this is at the price of a higher computational complexity. In the context of this work and similar to the LMS approach, the RLS filter assumes that a linear relationship exists between the maternal waveform recorded on the chest and the abdomen.

### 3.2.3 Kalman filter

Bayesian filtering is a probabilistic approach that recursively estimates the posterior distribution $P(\underline{\mathbf{x}}|\underline{\mathbf{y}}_{1:k})$ of a hidden state random variable $\underline{\mathbf{x}}$ at each time $k$ by using incoming measurements $\{\underline{\mathbf{y}}_k\}$ and a mathematical process model. The parameters are supposed to be random variables that evolve according to an *evolution equation* (the mathematical process model) and observed through measurements that are related to the state by a *measurement equation*. The dynamical system is written:

$$\begin{cases} \underline{\mathbf{x}}_k = \mathbf{G}_{k-1}\underline{\mathbf{x}}_{k-1} + \underline{\mathbf{w}}_{k-1} & (evolution\ equation) \\ \underline{\mathbf{y}}_k = \mathbf{H}_k\underline{\mathbf{x}}_k + \underline{\mathbf{v}}_k & (measurement\ equation), \end{cases} \quad (3.14)$$

where $\mathbf{G}_{k-1}$ is the state transition matrix applied to the previous state $\underline{\mathbf{x}}_{k-1}$, $\{\underline{\mathbf{w}}_k\}$ and $\{\underline{\mathbf{v}}_k\}$ correspond to the process and observation noise which are assumed to be white, zero-mean, uncorrelated ($E[\underline{\mathbf{v}}_k\underline{\mathbf{w}}_j^T] = 0$) and have covariance matrices $\mathbf{Q}_k = E[\underline{\mathbf{w}}_k\underline{\mathbf{w}}_k^T]$ and $\mathbf{R}_k = E[\underline{\mathbf{v}}_k\underline{\mathbf{v}}_k^T]$ respectively. It is further assumed that the components of the noise processes are uncorrelated i.e. $\mathbf{R}_k$ and $\mathbf{Q}_k$ are diagonal. $\mathbf{H}_k$ is the observation matrix that maps the state space into the observed space. The KF estimates the state $\underline{\mathbf{x}}_k$ based on the knowledge of the system dynamics and the noisy measurements $\{\underline{\mathbf{y}}_k\}$.

The dynamical model used for the KF is the one introduced by McSharry et al. [4]. The model equations can be discretised by considering a small sampling period $\delta$, such that $\delta \to 0$:

$$\begin{cases} \theta_{k+1} \equiv (\theta_k + \omega\delta) \bmod 2\pi \\ z_{k+1} = z_k - \sum_{i=1}^{N} \delta\frac{\alpha_i\omega}{b_i^2}\Delta\theta_{i,k}exp(-\frac{\Delta\theta_{i,k}^2}{2b_i^2}) + \eta_k, \end{cases} \quad (3.15)$$

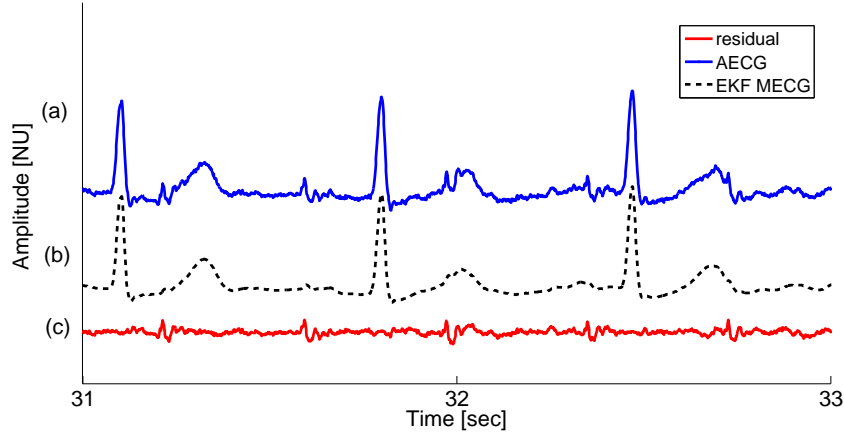

Figure 3.4: Illustration of the EKF used for MECG cancellation. (a) initial ECG mixture, (b) EKF mother filtered ECG, (b) residual ((c) = (b)-(a)). AECG: abdominal ECG.

where $\theta_k$ and $z_k$ are the discrete phase and amplitude, $\Delta\theta_{i,k} = \theta_k - \xi_i$ and $\eta$ is a perturbation term that corresponds to random additive noise, modelling the error made by substituting the model for a real ECG. The extended Kalman filter (EKF), which is an adaptation of the KF for non-linear systems, was used to filter the MECG while considering FECG as noise. The filtered MECG is then subtracted from the AECG mixture, leaving the residual signal containing the FECG and noise. Figure 3.4 shows an example of AECG where the MECG is filtered out using the EKF. FQRS detection can then be performed on the residual. The EKF was used as in Zaunseder *et al.* [61], and the covariance matrices were built as in Sameni *et al.* [9] while multiplying the **R** and **Q** covariance matrices by gain terms, written $G_R$ and $G_Q$ respectively. Estimation of the Gaussian parameters was performed automatically by identifying the P, QRS and T waves of the template cycle, and randomly initialising the Gaussian centres around these fiducials. Non-linear curve fitting was performed to find the Gaussian parameters that minimised the RMS error between the template cycle and the cycle as mapped by the Gaussian functions. This was repeated up to one hundred times (stopping criteria RMS<5%) and the best set of Gaussian (i.e. the set associated with the lowest RMS error) parameters was selected.

The EKF approach is similar to a TS approach in that an estimated MECG cycle is subtracted to each individual beat. However, the EKF has no constraints on P, QRS and T wave lengths, and thus performs a continuous estimation of the MECG (and consequently

FECG contained in the residual), compared to the TS techniques which removed the MECG cycle on a window with a given length around the MQRS. In addition, the EKF framework is more adaptive, and would allow for a better estimation of the MECG in highly non-stationary scenarios. This method is referred as TS$_{EKF}$. An extended discussion on the ECG Bayesian filtering framework and its historical use in the context of ECG processing can be found in Appendix A.

## 3.3 Spatial methods

In the context of NI-FECG extraction, blind source separation (BSS) algorithms aim to separate the underlying sources into three groups: MECG, FECG and noise. The standard BSS model assumes the existence of $p$ independent signals $\underline{\mathbf{s}} = [s_1^k, s_2^k, ..., s_p^k]$ and the observation of as many mixtures $\underline{\mathbf{m}} = [m_1^k, m_2^k, ..., m_p^k]$, where $k$ is the time instant. The mixtures are assumed to be linear and instantaneous combinations of the sources i.e. $m_i^k = \sum_{j=1}^{p} a_{i,j} s_j^k$ [62]. The mixing equation can be defined as $\underline{\mathbf{m}} = \mathbf{A}\underline{\mathbf{s}}$, where $\underline{\mathbf{m}} \in \Re^p$ is the observation vector at a time instant (i.e. containing the AECG signals) and $\underline{\mathbf{s}} \in \Re^p$ is the source vector and $\mathbf{A} \in \Re^{p \times p}$ is called the mixing matrix. $\underline{\mathbf{s}}$ and $\underline{\mathbf{m}}$ are vectors of random variables. BSS aims at recovering $\underline{\mathbf{s}}$ by using the observed data $\underline{\mathbf{m}}$.

A key assumption of traditional PCA and ICA techniques is that of a linear stationary mixing matrix between these sources. Although the stationary aspect of the mixing matrix could be unrealistic over long recordings, in practice the matrix can be regenerated for successive short time periods. The original signals are projected into the 'source' domain, where the channels representing the MECG and noise can be cancelled. The resulting back-projected signals should primarily consist of FECG components. Alternatively, the analysis (QRS detection, morphological parameters extraction) can be performed in the source domain. In this thesis the following terms will be used: 'observation domain signals' will refer to the signals recorded on the abdomen and 'source domain signals' will refer to the abdominal transformed signals after using a BSS algorithm.

### 3.3.1 Principal component analysis

The design matrix $\mathbf{M} \in \Re^{p \times k}$ was built with the rows corresponding to each individual channel and the columns to the observations through the abdominal channels at each time instant (with a total of $k$ times). PCA looks to decorrelate the dataset i.e. remove second order dependencies. It assumes that the signals coming from the various sources are linearly mixed, that large variance represents interesting structures and that the PCs are orthogonal. Figure 3.5 shows an example of applying PCA on a set of AECG signals, where no FECG was visually identifiable in the time domain. An extended discussion and derivation of PCA can be found in Appendix C[2].

### 3.3.2 Independent component analysis

Two of the main limitations of PCA are that the axis of the new basis are orthogonal and that the method looks to remove second order dependencies (decorelation). However, there is no reason for the MECG and FECG to be orthogonal in the observation space, and looking for independence rather than deccorelation might be a stronger criterion for separating the FECG from the abdominal mixture. Traditional ICA algorithms rely on the assumption that the source signals are statistically independent and non-Gaussian, while using higher order statistics than PCA to separate the multivariate signal into its additive sub-components. Classical ICA algorithms include JADE [63] and FastICA [64]. An extended presentation of ICA can be found in Appendix D.

## 3.4 Summary and conclusion

A number of NI-FECG extraction techniques were carefully reviewed in this chapter. These will be used either independently or in combination with each other in the thesis. The in-tention behind the above listing and selection was to identify different types of approaches that were introduced for NI-FECG extraction, in order to implement the corresponding algorithms and draw conclusions with respect to their performances. In addition, the added value in combining a subset of these techniques will be studied in Chapter 7.

---

[2] Also note that PCA is used here as a BSS technique to isolate the different sources contained in the abdominal recordings. This is different from its use in 3.2.1.2 were PCA was used to 'describe' a template maternal cycle in term of the PCs obtained from a stack of maternal ECG cycles. In $TS_{pca}$ and for a given MECG cycle in the observation domain, the data were projected on the PCs and the first few PCs that describe the MECG cycle can then be cancelled in order to remove the maternal contribution from each individual MECG cycle.

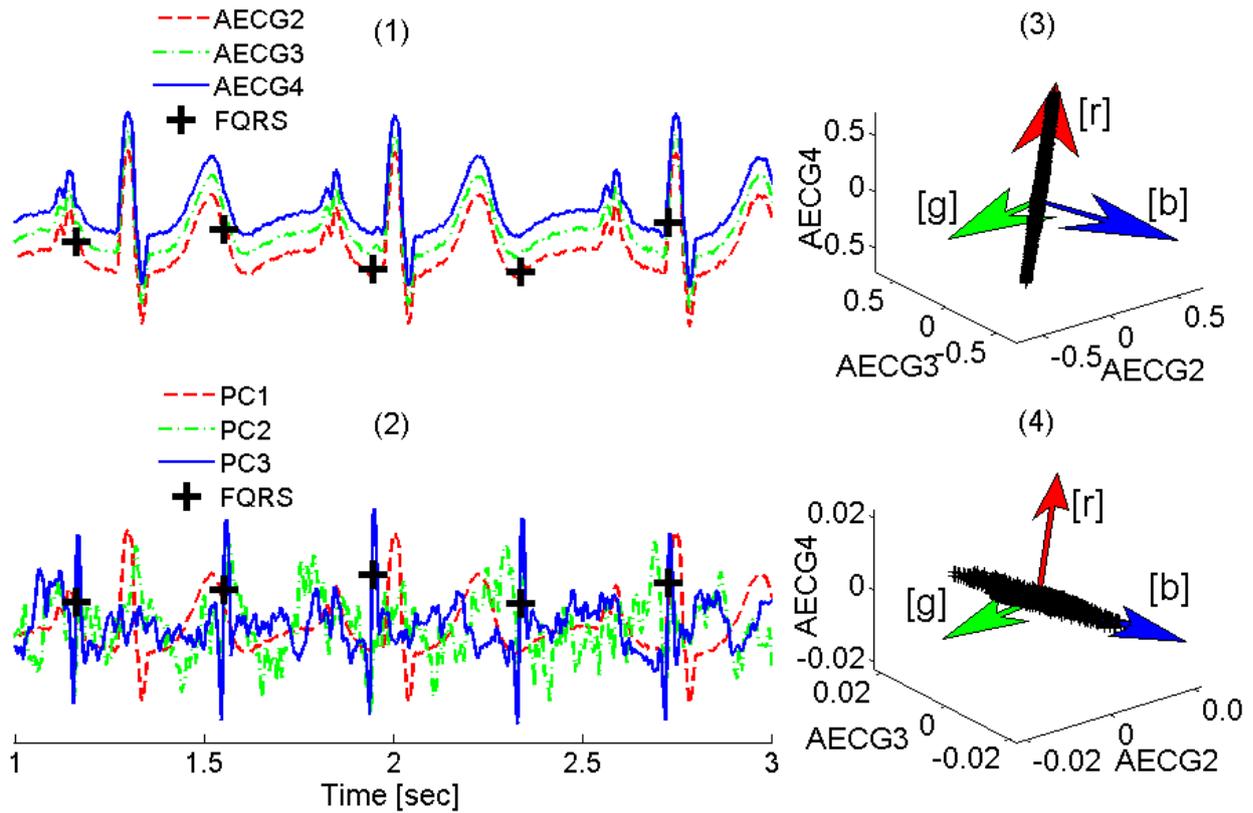

Figure 3.5: Applying PCA on three abdominal channels of record a57. (1) 3 sec of the AECG signal, with crosses representing the reference FQRS location. Each channel has been slightly shifted vertically for visualisation purposes. Note that the FECG cannot be visually spotted in the time domain; (2) principal components (PCs) normalised by the square root of their corresponding eigenvalue. Note the blue continuous line which corresponds to the extracted FECG; (3) 3D representation of the 3 AECG channels. The red axis [r] carries the most information (highest variance, corresponding to the MECG) and the 3 PCs are orthogonal; (4) 3D representation of the 3 AECG channels when the contribution of the first PC (along the red arrow [r]) has been removed. Note the difference in axis range between (3) and (4). Projection of the datapoints along the blue arrow [b] will reveal the FECG component in blue continuous line in (2); [r] red, [g] green, [b] blue.

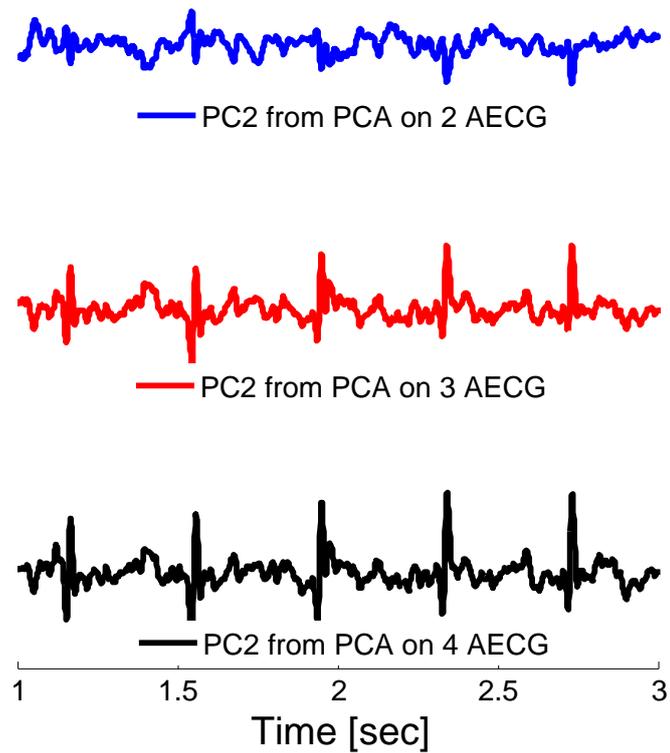

Figure 3.6: Second PC extracted using PCA when using 2, 3 and 4 AECG channels respectively (i.e. with p=2 or 3 or 4 in the design matrix). This figures illustrates that by adding AECG channels the extracted FECG SNR increases. In other words, that better extraction of the FECG is made possible by increasing the number of AECG channels. [r] red, [g] green, [b] blue.

# Part II

# Databases and tools

# Chapter 4

# Databases, labelling and statistics

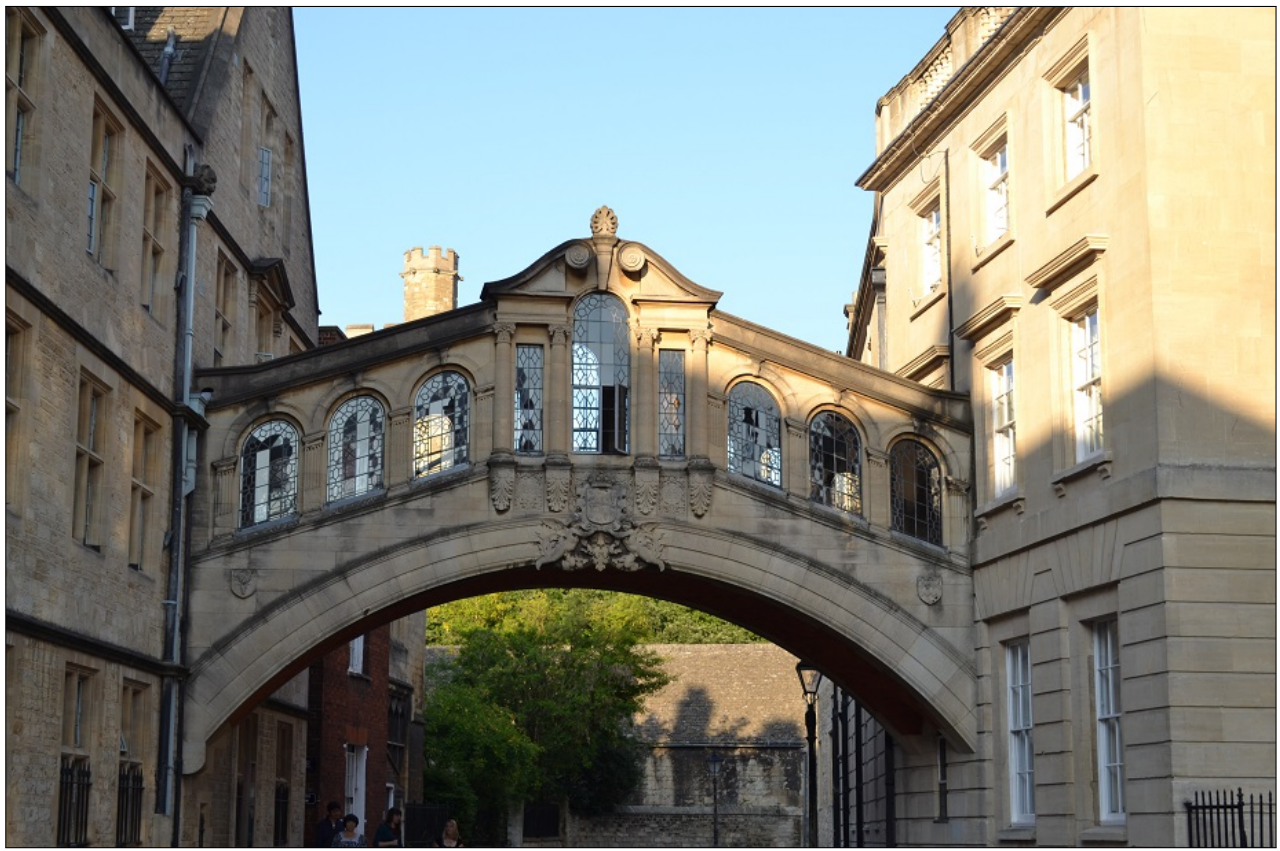

Figure 4.1: Hertford Bridge, also known as the Bridge of Sighs, connects two buildings of Hertford College. A bridge is a structure that is built to support a road, a path. It enables a connection to be made between two points. The mechanics behind such buildings are complex and if the backbone of the construction does not hold then the whole structure collapses. Experimental research is similar in that if the methods are not carefully defined then the conclusions drawn out of the experiments might well collapse.

## 4.1 Introduction

This chapter introduces the real and artificial databases used in the thesis, together with the techniques employed for obtaining the reference annotations ('ground truth'). The statistical measures used for evaluating the non-invasive fetal electrocardiogram (NI-FECG) extraction methods are also presented. The fetal ECG synthetic model, *fecgsyn*, used to generate the artificial databases developed in the context of this thesis is also described.

## 4.2 Databases

Until the PhysioNet/Computing in Cardiology Challenge 2013 (*the Challenge*) there were three public NI-FECG databases:

- The Daisy database consisted of 8 ECG channels (5 abdominal and 3 thoracic), lasting for 10 sec, using a sampling frequency, $fs$, of 250 Hz, without reference annotations.

- The Non-Invasive Fetal Electrocardiogram Database (NIFECGDB), available on PhysioNet [65], $fs = 1$ kHz includes 55 multichannel AECG recordings taken from a single subject (21 to 40 weeks of gestation) and recorded using a g.BSamp Biosignal Amplifier (GTech GMBH, Austria). No reference annotations were available.

- Abdominal and Direct Fetal Electrocardiogram Database (ADFECGDB), available on PhysioNet [65] $fs = 1$ kHz with 5 min of recordings (4 abdominal channels and the scalp ECG) from 5 women in labour (38 to 41 weeks of gestation) and with reference FQRS annotation derived from the SECG. Data were recorded using the KOMPOREL fetal monitoring system (ITAM, Zabrze, Poland).

It is important to note that these three databases were of low dimension (in term of the number of recordings, individuals, and abdominal channels available), and few of the recordings had reference annotations. When annotations were available, they only consisted of the FQRS complex location from a single annotator. In this section, the following new databases created for this thesis are introduced:

- CRDB: chest reference database. This database will be used in Chapter 6 and consists of 262 min of AECG from 9 pregnant women, with 37452 FQRS reference peak location annotations. Data were recorded at $fs = 1$ kHz. Each record had one reference chest channel.

- CDB: Challenge database. This database will be used in Chapter 7 and consists of 447 min of data from five different databases, with FQRS reference. Data were recorded at $fs = 1$ kHz.

- AQTDB: artificial QT database. This database will be used in Chapter 8. It consists of 220 min of simulated data (9 different vectocardiogram morphologies) with MQRS, FQRS and FQT references. Data were generated at $fs = 1$ kHz.

- RQTDB: real QT database. This database will be used in Chapter 8. Data were recorded at $fs = 1$ kHz from 22 pregnant women; with 28 abdominal channels, scalp reference electrode and the output of the MindChild medical monitor were available. One segment of 4-5 min of data was taken for each subject, totalling 105 min of data. Fetal QT were annotated on the SECG by three medical experts and used as the reference annotations.

In the following subsections the above mentioned databases are described further.

## 4.2.1 Chest reference database

The Chest Reference Database (CRDB) is comprised of two datasets. The first was taken from the NIFECGDB [65] consisting of 55 records. Each record contained two chest channels, and three to four abdominal channels with different electrode configurations (electrode position was varied in order to improve the SNR). All signals were sampled at 1 kHz with 16-bit resolution. A bandpass filter (0.01 Hz-100 Hz) and a notch filter (50 Hz) were applied to the data during acquisition.

A total of 14 records were manually selected where the FQRS complexes were visible on at least one channel (so that manual annotation of the FQRS was feasible). The gestational age for these records ranged from 22 to 40 weeks. One minute of signal, 30 seconds after the start of the record, was extracted for all available channels for each record[1]. A total of 2148 FQRS were manually annotated, using the channel with the most visible FECG, for each record. These markers were considered to be the reference. For homogeneity purposes, only the first three abdominal channels were kept so that all records had two chest and three abdominal channels. The three abdominal channels were considered independently for FECG extraction, thus providing $3 \times 14 = 42$ min of annotated data (i.e. $3 \times 2148 = 6444$ reference FQRS). Data were of good quality with variable FECG/MECG SNR and some minor artifacts which were not discarded. This first dataset is denoted $CRDB_1$.

The second dataset is composed of a subset of records from a private, commercial database[2]. Each record consisted of 28 abdominal channels, one maternal chest channel, and the invasive scalp electrocardiogram (SECG) signal. All signals were sampled at 1 kHz with 16-bit resolution. The chest channel, as well as a subset of four abdominal channels where a FECG trace was visible, were manually selected for processing. A total of eleven 5-minute records from 8 pregnant women were used. An energy QRS detector based upon that of Pan and Tompkins (P&T) [66] was applied to the scalp electrode, and the corresponding markers were used as the reference (more details on the QRS detector can be found in Appendix E). For a given record, each abdominal channel was considered independently for FECG extraction thus providing $11 \times 5 \times 4 = 220$ minutes of annotated data (overall $4 \times 7752 = 31008$ reference FQRS annotations). Some artifacts were present in a few records and the FECG/MECG SNR was variable. This second dataset was denoted $CRDB_2$.

In the work conducted in Chapter 6, $CRDB_1$ constitutes the training database and $CRDB_2$ the test database. Summary statistics for the fetal and maternal HR in $CRDB_2$ are reported in Table 4.1. Note that the data in the training set ($CRDB_1$) and the test set ($CRDB_2$) were recorded with different hardware, following a separate protocol, at different stages of pregnancy for different subjects. This allows for reassessment of whether the algorithms are sufficiently flexible to work with data that have similar, but not equivalent, recording

---

[1] The set of the 14 records selected from the NIFECGDB was: 154, 192, 244, 274, 290, 323, 368, 444, 597, 733, 746, 811, 826, 906.

[2] The study was approved by The Institutional Review Boards at Summa Health System (RP#12018) and Brigham and Women's hospital (RP#2010-P-002778/1).

Table 4.1: HR reference statistics for CRDB$_2$ records

|  | FHR stats (bpm) | | | | MHR stats (bpm) | | | |
| --- | --- | --- | --- | --- | --- | --- | --- | --- |
|  | Mean | std | Max | Min | Mean | std | Max | Min |
| 123a | 130 | 5 | 139 | 119 | 66.7 | 2.74 | 76.3 | 61.2 |
| 123b | 131 | 6.19 | 144 | 117 | 63.8 | 2.12 | 68.5 | 59.2 |
| 125 | 145 | 5.44 | 147 | 126 | 80.9 | 2.77 | 87.3 | 74.8 |
| 126 | 159 | 2.43 | 163 | 155 | 78.9 | 5.36 | 96 | 69.5 |
| 172a | 154 | 6.35 | 159 | 135 | 91.5 | 4.32 | 102 | 83.6 |
| 172b | 142 | 10.8 | 153 | 112 | 93.6 | 6.6 | 109 | 81 |
| 210 | 157 | 5.51 | 170 | 145 | 90.8 | 6.77 | 108 | 80.2 |
| 261 | 132 | 3.54 | 142 | 122 | 79.3 | 2.69 | 89.7 | 74.2 |
| 265 | 129 | 2.57 | 135 | 125 | 64.9 | 3.42 | 74 | 59.3 |
| 299a | 148 | 6.92 | 159 | 123 | 83.4 | 5.65 | 100 | 74.8 |
| 299b | 147 | 13.1 | 157 | 95.5 | 80.2 | 5.57 | 97.2 | 72.1 |

All the values use trimmed estimates of the FHR distribution, i.e. they exclude the most extreme values (2.5% on either tail). This is due to inherent artifacts in the data causing misleading statistics if not excluded.

configurations. It is particularly pertinent to assess the adaptability of the extraction methods to an unseen set of signals from distinct subjects, as there are many free parameters which are tuned on the training set database. Furthermore, the NIFECGDB was used as the training set (rather than the test set) because the reference FQRS fiducial markers were not as accurate as for the private database, which had a simultaneous fetal scalp signal.

As only one chest channel was available in CRDB$_2$, only one of the two available chest channels was used for CRDB$_1$. Each abdominal channel was considered individually for processing.

### 4.2.2 Physionet Challenge 2013 Database

Despite many interesting theoretical frameworks, the robustness of most of the methods for NI-FECG extraction in the literature has not been sufficiently evaluated quantitatively. This is due to two main factors: (1) the lack of gold standard databases with expert annotations and (2) the methodology for assessing the algorithms. The Physionet Challenge 2013 attempted to address these issues by creating a database, the CDB, that consisted of 447, one-minute abdominal signals with reference FQRS annotations. This is the largest publicly available dataset for NI-FECG signal processing to date.

There were two key questions in the Challenge: "can accurate FHR measurements be performed using a set of non-invasive abdominal ECG electrodes?" and "can an accurate fetal QT measure be performed in an automated way using the extracted signal?"

Table 4.2: Physionet Challenge 2013 FECG database description.

| Database Name | N records |
|---|---|
| ADFECGDB [67] | 25 |
| NIFECGDB [65] | 14 |
| AFECGDB [6] | 20 |
| PNIFECGDB | 340 |
| PSFECGDB | 48 |
| Total | 447 |

The Challenge was organised under the supervision of Prof Gari D. Clifford from the Oxford *Intelligent Patient Monitoring* group (IPM) and George Moody from the MIT *Laboratory for Computational Physiology* (LCP). The technical work (database formatting, scoring system, sample code etc.) was performed by Ikaro Silva (LCP) and Joachim Behar (IPM).

The datasets used for the Challenge were obtained from five different sources, (see Table 4.2), yielding a total of 447, 1-min records. Two out of the five datasets (ADFECGDB and NIFECGDB) had been made public [65, 67], and the AFECGDB was artificially generated using the *fecgsyn* dipole model described in Section 4.5. The other two datasets (PNIFECGDB, the private non-invasive FECG database, and PSFECGDB) were donated to PhysioNet for this Challenge (the PSFECGDB was not made public, and used only for scoring open source algorithms in Events 1 and 2 described in Section 4.4.2). The gold standard used for the initial stage of the Challenge consisted of reference annotations from the datasets. For the non-invasive datasets, the annotations were obtained from FECG QRS estimates derived manually, or through additional abdominal maternal ECG leads that were not available to competitors; for the PSFECGDB and ADECGDB, the reference FQRS were obtained from the scalp ECG; for the AFECGDB the FQRS reference were known from the simulation parameters. A subset of both the initial and final reference annotations were manually verified by two of the Challenge organisers, Ikaro Silva and Joachim Behar, although some minor errors in annotations persisted.

The data were recorded with different hardware, following separate protocols and on a number of pregnant women. This allows for the reassessment of whether the algorithms were sufficiently flexible to work with data that have similar, but not equivalent, recording configurations and on unseen patients.

All records were formatted to have a 1 kHz sampling frequency, one minute duration, and four channels of non-invasive abdominal maternal MECG leads. The databases in Table 4.2 were re-arranged into three datasets for the Challenge:

- Set A: 75 records, both records and expert annotations were made public

- Set B: 100 records, only the records were made public

- Set C: 272 records, both records and expert annotations were withheld from the public so that they could not train their algorithms on this dataset

Joachim formatted and provided the PSFECGDB, NIFECGDB (manual annotation of each individual FQRS was necessary for this database), AFECGDB and the ADFECGDB. In order to objectively compare the approach presented in this thesis (see Chapter 7) for NI-FECG extraction to the other Challenge participants, only the official data provided to the Challenge participants was used for training.

### 4.2.3 Artificial QT Database

Signals were generated using the *fecgsyn*, introduced further in Section 4.5. The open source simulator was used to generate single channel maternal-fetal ECG mixtures with realistic amplitudes, morphology, beat-to-beat variability, heart rate changes, correlated noise and positional (rotation and translation-related) movements in the fetal and maternal heart due to respiration. A set of 20, one-minute signals were generated for the training set and 200, one-minute signals were generated for the test set. Parameters for the model were randomly sampled from a Gaussian distribution modelling various physiological parameters (see Table 4.3). The single electrode position was randomly selected from a set of 32 positions to simulate different views of the heart electrical activity. A set of nine vectorcardiograms (VCG) with associated Gaussian parameters was available in *fecgsyn*. The first four (1-4) were used to generate the training set and the last five (5-9) were used to generate the test set dataset. The split was performed in order to ensure that the beat morphology was different in training and test. Reference MQRS and FQRS were obtained from the ECG simulator.

Table 4.3: Parameters for the abdominal ECG simulator.

| Params | Definition | 95% CI |
|---|---|---|
| MRES | maternal breathing rate | [0.2  0.3] Hz |
| FRES | fetal breathing rate | [0.80  0.95] Hz |
| FHR | fetal mean heart rate | [120  160] bpm |
| MHR | maternal mean heart rate | [70  110] bpm |
| SNRmn | signal to noise ratio (maternal to noise) | [6  18] dB |
| SNRfm | signal to noise ratio (fetal to maternal) | [-15  -5] dB |

### 4.2.4 Real QT Database

FECG data were recorded from 22 full-term women in labour with singleton fetuses. Data were recorded simultaneously using a NI-FECG monitor (Mindchild Medical, North Andover, MA) and an invasive fetal scalp electrode (GE Corometrics). Data were recorded at $fs = 1$ kHz with 16-bit accuracy. All women delivered newborns with five-minute Apgar scores above six and none of the fetuses were exposed to selective serotonin reuptake inhibitors (SSRI) medication in-utero.

Each recording period was divided into one minute segments for the analysis. Segments were selected so that they had a relatively stable FHR. 'Stable' was taken, subjectively, to be that the heart rate did not change by more than 20 bpm during the one-minute segment. This procedure was implemented to ensure that the FQT interval was approximately stable over each one minute segment, which is necessary when doing any averaging of ECG cycles. Indeed there exist a relationship between the QT length and the heart rate [68]. Since a few preselected one minute segments did not match the FHR criteria because of the presence of an acceleration, the corresponding sub-segments (generally lasting between 10-15 seconds) were replaced by random noise. This ensured that the annotators were not annotating in areas with large changes in heart rate. See Figure 4.2 for an example of the annotation interface.

The QT interval is defined as the time interval between the Q wave onset and the end of the T wave in the heart's electrical cycle. Three paediatric cardiologists were presented with three sets of data to annotate by visual analysis. Visual analysis means that the cardiologists annotated the Q-onset and T-end according to their experience and understanding of adult ECGs. The first set (SET0) was considered a training set, and was used to train the cardiologists to use the interface. The annotation interface was built on top of the Physionet Lightwave interface (see Section 4.3 for more details). This online system either presented the cardiologists with a rhythm strip one minute long, or presented them with a single ECG waveform created by automatically averaging a series of ECG waveforms. Each cardiologist was trained on the interface individually during an online training session on a set of examples (SET0).

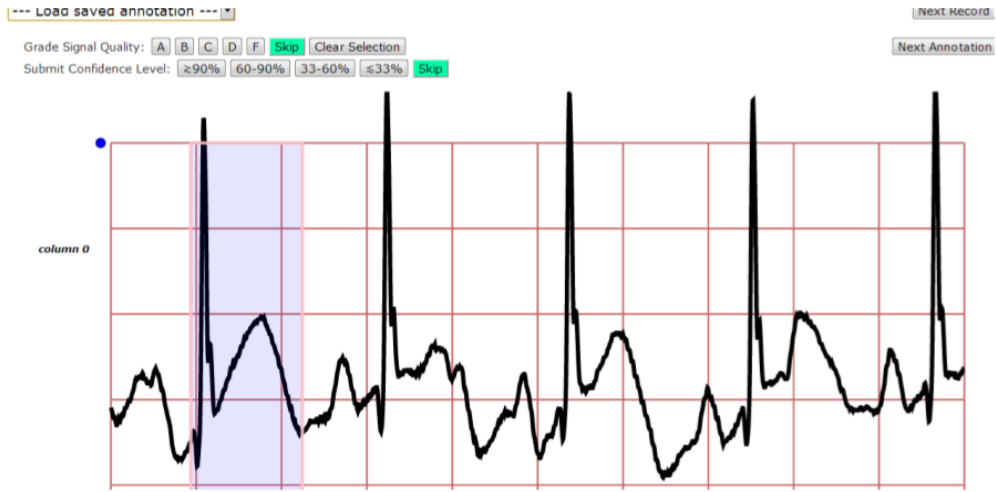

Figure 4.2: Annotation interface. A FQT interval is being annotated (shaded area) on a scalp ECG segment (black bolded line).

The next set (SET1) contained 210 one minute segments (105 NI-FECG and the corresponding 105 SECG segments). Each cardiologist was instructed to annotate five QT intervals per one-minute segment. The third set (SET2) included 210 averaged FECG waveforms (105 NI-FECG and 105 corresponding SECG). One annotation per waveform was requested. In each set, the data were randomised so that no two segments or averaged waveform from the same patient followed each other. An example of an annotation made on a rhythm strip segment is demonstrated in Figure 4.2. Figure 4.3 shows an example of SECG and NI-FECG (output by the MindChild monitor), as well as the corresponding templates built by averaging the FECG cycles.

Because the abdominal and scalp recordings were not aligned in time and no time stamp was available, it was necessary to align the data. This was performed in two steps: (1) the HR times series derived from the NI-FECG output by the MindChild monitor and the HR derived from the SECG were matched using cross correlation. This provided a gross alignment constant; (2) refinement was obtained by manual adjustment i.e. by visually checking that the FQRS from the abdominal signals were aligned with the scalp ones.

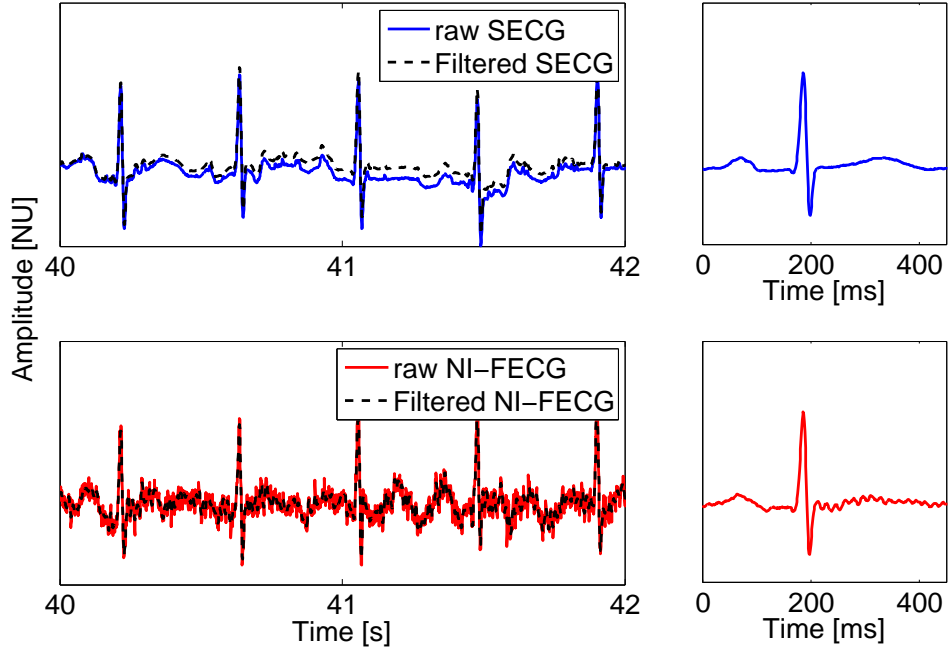

Figure 4.3: Example of scalp ECG and NI-FECG extracted by the MindChild monitor, as well as the corresponding templates built by averaging the ECG cycles.

## 4.3 Labelling

For many biosignal processing applications, the performance of an algorithm or system must be evaluated against some reference or 'gold standard' annotations. It is common to have one or multiple 'experts annotators' evaluate the data when this ground truth is not readily available. However, significant intra- and inter-observer variability, as well as human biases limit the accuracy of this 'gold standard' [69]. In the context of ECG annotations, this variability can be the result of the intrinsic difficulties in interpreting the signals (for example, due to the signal quality) and the level of training/experience of the annotator. The work presented in this section is the result of collaborative work with Zhu *et al.* and was published in [70, 71]. This section introduces the Crowd-label interface and the Probabilistic Label Aggregator (PLA) algorithm that were used to crowd source the fetal QT annotations. A more in depth presentation of the CrowdLabel interface and the PLA algorithm as well as specifics on the authors contribution to this work can be found in Appendix B.

### 4.3.1 CrowdLabel interface

PhysioNet provides a signal viewer called LightWAVE [72][3]. The CrowdLabel annotation system was adapted from LightWAVE version 0.38, and consists of a user authentication interface, a LightWAVE client, a back-end server database which mirrors Physiobank, and a local server which stores annotations provided by users. The configuration of the back-end server follows the LightWAVE's infrastructure, which utilises a common gateway interface application to retrieve data from local data repository (including PhysioBank and private databases) and a user interface for display. Figure 4.2 shows a screenshot of the annotation interface where a FQT is being annotated. Interested readers are referred to Zhu *et al.* [71] for more details on the features of the interface and backbone code.

### 4.3.2 Probabilistic Label Aggregator

The probabilistic model by Raykard *et al.* [73] was adapted to combine crowd-sourced medical QT annotations. The resulting model is referred as the Probabilistic Label Aggregator (PLA). In the context of QT annotations, it is assumed that $R$ annotators have annotated a series of $N$, QT observations. We write $z_i$ the true QT annotation for each individual record i ($i \in [1; R]$) and $y_i^j$ the annotation of annotator $j$ performed on record $i$. Furthermore, it is assumed that $z_i$ can be predicted using a linear regression model as $z_i = \underline{\mathbf{w}}^T \underline{\mathbf{x}}_i + \epsilon$, where $\underline{\mathbf{w}}$ is the regression vector and $\epsilon$ is a zero-mean Gaussian noise with precision (defined as the inverse of the variance) $\gamma$ and $\underline{\mathbf{x}}$ is a feature vector. The likelihood function of the problem can be derived (see Appendix B) and the expectation maximisation (EM) algorithm can be used to solve the corresponding maximum likelihood problem:

(1) The E-step estimates the expected true annotations for all records, $\hat{\underline{\mathbf{z}}}$, as a weighted sum of the provided annotations with their precisions $\lambda^j$:

---
[3]Available on: http://physionet.org/lightwave/

$$\hat{\underline{\mathbf{z}}} = \frac{\sum_{j=1}^{R} \lambda^j \underline{\mathbf{y}}^j}{\sum_{j=1}^{R} \lambda^j} \tag{4.1}$$

(2) The M-step is based on the current estimation of $\hat{\underline{\mathbf{z}}}$ and given dataset $\mathbf{D}$. The model parameters such as the regression coefficient $\underline{\mathbf{w}}$ and precision $\underline{\lambda}$ can be updated using equations (4.2) and (4.3) accordingly in a sequential order:

$$\frac{1}{\hat{\lambda}^j} = \frac{1}{N} \sum_{i=1}^{N} (y_i^j - \hat{\underline{\mathbf{w}}}^T \underline{\mathbf{x}}_i)^2 \tag{4.2}$$

$$\hat{\underline{\mathbf{w}}} = \left(\sum_{i=1}^{N} \underline{\mathbf{x}}_i \underline{\mathbf{x}}_i^T\right)^{-1} \sum_{i=1}^{N} \underline{\mathbf{x}}_i \hat{z}_i \tag{4.3}$$

with

$$\hat{z}_i = \frac{\sum_{j=1}^{R} y_i^j \lambda^j}{\sum_{j=1}^{R} \lambda^j}$$

Precision is initialised as being equal for all annotators (i.e. no prior on the ranking of the annotators) and as being equal to: $\hat{\underline{\mathbf{z}}} = \frac{1}{R} \sum_{j=1}^{R} \underline{\mathbf{y}}^j$. Then the E-step and M-step can be iterated until convergence of the parameters.

## 4.4 Statistics

### 4.4.1 Standard statistics

In accordance with the ANSI/AAMI guidelines [74], the sensitivity ($Se$) and positive predictive value ($PPV$) should be reported with:

$$Se = \frac{TP}{TP + FN}, \quad PPV = \frac{TP}{TP + FP}, \tag{4.4}$$

where $TP$, $FP$ and $FN$ are true positive (correctly identified FQRS), false positive (wrongly detected FQRS) and false negative (missed FQRS) detections respectively. For algorithm parameter optimisation, the $F_1$ statistic was used as the accuracy measure. In the context of binary classification, $F_1$ is defined as:

$$F_1 = 2 \cdot \frac{PPV \cdot Se}{PPV + Se} = \frac{2 \cdot TP}{2 \cdot TP + FN + FP}. \qquad (4.5)$$

From equation (4.5) one can observe that $FN$ and $FP$ play a symmetric role in penalising the accuracy measure $F_1$. The $F_1$ statistic is an average of the $Se$ and $PPV$ thus providing a good summary metric. The $F_1$ measure is a harmonic mean and is suited for situations when the average of rates is desired [75] (as opposed to an arithmetic mean)[4].

In the context of the work conducted in Chapter 7 for the Challenge 2013, the $F_1$ measure was computed on Set A while discarding the first and last 2 seconds of each 1 min segments. The following records were also discarded due to some inaccurate reference annotations (identified by visual inspection): $a33$, $a38$, $a47$, $a52$, $a54$, $a71$, $a74$. The overall statistics were calculated by averaging the $F_1$, $Se$, and $PPV$ (respectively) obtained from each individual record (i.e. computing the gross average statistics).

For the work performed in Chapter 6, the performance of the algorithms was also evaluated in terms of the fetal heart rate (FHR). FHR can be derived from the RR interval time series and compared with the reference trace HR derived from the reference annotations. Both the extracted FHR and the reference FHR were extracted as the reciprocal of the RR interval scaled by a factor of 60. At any given time, the extracted FHR was said to match the reference FHR if it was within ±5 beats per minute (bpm) with respect to the reference trace (Figure 4.4). The corresponding FHR measure is denoted $HR_m$. The tolerance of ±5 bpm was motivated by the industrial standards [79].

### 4.4.2 Physionet Challenge 2013 Statistics

The Challenge was organised into five events denoted $E$: a QT estimation event, and four RR/HR time series events. Table 4.4 defines the four RR/HR time series events, with $E1$ and $E2$ only used for the open source entries (evaluated on set C) and $E4$-$E5$ considered for the open source and close source entries (evaluated on set B).

---

[4]In order to get an idea of why the $F_1$ measure is preferable over the arithmetic mean, consider the following example: $Se = 1$ and $PPV \approx 0$ (which could correspond to a QRS detector annotating every single datapoint of the ECG signal as a R-peak) then the arrhythmic mean would be $\approx 0.5$ (i.e. 50% accuracy!) whereas we would have $F_1 \approx 0$ (which is certainly more representative of the outcome of the QRS detection algorithm). After I suggested the $F_1$ measure to be used at the Computing in Cardiology conference 2013, it was well adopted in the Challenge follow up Physiological Measurement special issue as for example in [76, 77, 78].

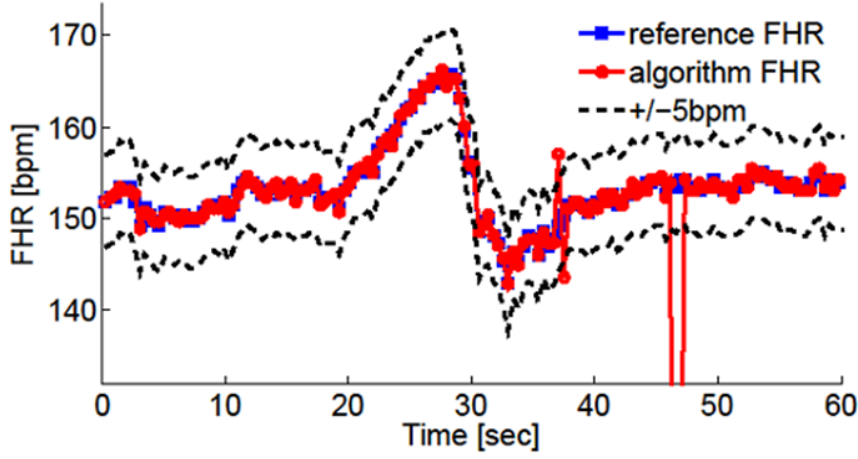

Figure 4.4: FHR trace. The plot displays the reference and extracted FHR for the first 60 sec of channel 4, record 172a. In black are the delineation at ±5 bmp with respect to the reference FHR trace.

The Challenge scores for the different events were defined as follows; scores for the FHR based events ($E1$ and $E4$) were computed from the differences between matched reference and test FHR measurements at 12 instances (i.e. one every 5 sec):

$$E1^k/E4^k = \frac{1}{12}\sum_{i=1}^{12}\Delta h_i^2 \qquad (4.6)$$

where $\Delta h_i = h_i^t - h_i^r$, $h_i^t$ the test FHR on the $ith$ segment, and $h_i^r$ is the reference FHR for the $ith$ segment. $E1^k/E4^k$ are the scores for event $E1/E4$ for record $k$. Scores for the RR events ($E2$ and $E5$) were computed from the differences between matched reference and test RR intervals as follow:

$$E2^k/E5^k = \sqrt{\frac{1}{M}\sum_{i=1}^{N}\Delta RR_i^2} \qquad (4.7)$$

where $\Delta RR_i = RR_i^t - RR_i^r$, $RR_i^t$ is the test $RR$ interval and $RR_i^r$ is the reference $RR$, $N$ is the number of test measurements and $M$ is the number of reference measurements. $E2^k/E5^k$ are the scores for event $E2/E5$ for record $k$. The score for the QT measure event ($E3$) was calculated as the square difference between matching reference and test QT intervals:

$$E3^k = \Delta QT^2 \tag{4.8}$$

where $\Delta QT = QT^t - QT^r$, $QT^t$ is the test $QT$ interval, and $QT^r$ is the reference $QT$. $E3^k$ is the score for event E3 for record $k$.

The purpose of the RR events was to assess whether an algorithm was able to extract the absolute FQRS position, i.e. the position of the fetal R-peak in the signal with respect to the reference fiducial markers. The purpose of the FHR events was to assess whether an algorithm provided clinically relevant information, regardless of where the fetal R-peaks were located (so the FQRS time series could be highly smoothed before computing the FHR). As such, the RR and FHR scores represented two distinct events (even if they were highly correlated as the results of the Challenge showed- see Section 7.4.3). The Challenge was divided into three phases, corresponding to three time periods, where participants were allowed to submit a certain number of entries (phase 1: 25-04-2013 to 01-06-2013, 3 entries; phase 2: 01-06-2013 to 25-08-2013, five entries; phase 3: 25-08-2013 to 05-09-2013, 1 entry).

The participants of the Challenge were expected to use set A to train their algorithms, while Sets B and C were used by the Challenge organisers to score them at the MIT LCP laboratory. It was not possible to score one record in Set A and two in Set B due to errors in the corresponding reference annotations. The training dataset (Set A), and the records for Set B, are publicly available on `Physionet.org`

$E1$ and $E2$ were scored on a private PhysioNet server running the participant's algorithm on set C. $E4$ and $E5$ were automatically scored on PhysioNet's web server by comparing the user's submitted annotation file with the expert annotations. The web based scoring interface in PhysioNet remains open for those wishing to compare their results with those from the official Challenge on $E4$ and $E5$. The scoring methods for the three different FECG estimation tasks are described in Table 4.4. Records that were not annotated by the competitors were given a very high penalty value. The WaveForm DataBase (WFDB) software package (version 10.5.19) was used to score the events related to FHR and RR Series. The final score for a given event was determined by the average score of all the records within the event's dataset. The source code used for all scoring remains available at `http://physionet/chalenge/2013`

and the source code from the open-source competitors can be found at `http://physionet.org/challenge/2013/sources/`.

| Estimation Task | Scoring Method | Units | Event |
|---|---|---|---|
| FHR Series | beat-by-beat classification error | (beats/minute)$^2$ | E1, E4 |
| RR Series | average root square error | milliseconds | E2, E5 |

Table 4.4: Scoring methods for the records of the Challenge 2013. E stands for event.

### 4.4.3 Discussion

It is possible to assess the error for FQRS detection using a window based metric (e.g. FQRS within ±50ms window) or using a distance based measures (e.g. absolute error or root mean square error between the detected R-peak and the true R-peak locations). Both approaches have their advantages and drawbacks.

Motivation for performing FQRS detection can be two-fold: i) obtaining an accurate estimate of the heart rate; ii) localising precisely the absolute position of the R-peaks in order to use them as anchor points for further processing such as reconstructing the morphology of the NI-FECG and automated QT and ST measurement. Some error measures better serve the purpose of i) whereas other will better serve ii). Ultimately the choice of the accuracy measure should be conditioned by the clinical problem i.e. it should reflect how well the algorithm is performing in identifying a pathology or an abnormal event, for example. This was not possible in this thesis given the absence of pathological cases. In order to clarify the purpose and advantages or limitations of the main statistics used in the thesis, a discussion together with a summary table (Table 4.5) are included below.

The Physionet *mxm* function was used for E2/E5 scoring; The purpose of the function is to compare the RR intervals of the extracted time series with the reference time series. However *mxm* does not take into account the distance between the reference position of the FQRS and the one outputted by the algorithm. As an example, when a QRS detector does not look for the R-peak location but say for the R-peak offset whereas the reference QRS correspond to the R-peak maxima then *mxm* will give the same output.

Following the ANSI guidelines the classical statistics for evaluating QRS detectors are:

| Statistics | Comments |
|---|---|
| $Se = \frac{TP}{TP+FN}$ | The proportion of true FQRS that have been detected out of the total number of true FQRS. This statistic tells how good an algorithms is at finding the true FQRS. The ANSI/AAMI guidelines [74] recommend to report $Se$ when evaluating a QRS detector. |
| $PPV = \frac{TP}{TP+FP}$ | The proportion of the true FQRS that have been detected out of all the detected FQRS. This statistic tells how good the algorithm is at identifying true FQRS out of all the detections it makes. The ANSI/AAMI guidelines [74] recommend to report the $PPV$ when evaluating a QRS detector. |
| $F_1 = 2 \cdot \frac{PPV \cdot Se}{PPV+Se}$ | The $F_1$ statistic is an average of the $Se$ and $PPV$ thus providing a good summary metric in the absence of TN. The $F_1$ measure is a harmonic mean and is suited for situations when the average of rates is desired [75]. It can be used as an accuracy measure when training an algorithm but also to summarise the final overall performance of it in accurately detecting the R-peak locations. |
| $E1^k/E4^k = \frac{1}{12}\sum_{i=1}^{12} \Delta h_i^2$ | Differences between matched reference and test FHR measurements at 12 instances for each one minute segment (i.e. one approximately every 5 sec). $\Delta h_i = h_i^t - h_i^r$, with $h_i^t$ the test FHR on the ith segment and $h_i^r$ the reference FHR for the ith segment. The purpose of this statistic is to assess the accuracy of an algorithm in evaluating the HR from the detected R-peaks. |
| $E2^k/E5^k = \sqrt{\frac{1}{M}\sum_{i=1}^{N} \Delta RR_i^2}$ | Computed from the differences between matched reference and test RR intervals. $\Delta RR_i = RR_i^t - RR_i^r$ with $RR_i^t$ the reference $RR$, $N$ is the number of test measurements and $M$ is the number of reference measurements. The purpose of this statistic is to assess an algorithm in its performance to accurately evaluate the RR interval. |
| $HR_m = N/M$ | Percentage of the HR within ±5 bpm of the reference HR. $M$ number of reference points, $N$ number of test points that are within ±5 bpm of their corresponding reference measurement. The purpose of this statistic is to evaluate the accuracy of an algorithm in extracting the heart rate with a tolerance of ±5 bpm as used in industrial practice [79]. Thus in this case the statistic is window based and not RMS based. |

Table 4.5: Summary of the statistics used for FQRS detection.

sensitivity ($Se$) and positive predictive value ($PPV$). A detected QRS is considered a true positive if it is within 150 ms from the reference annotation for adults and in this thesis a window of 50 ms was taken to account for the higher fetal HR typically encountered in adults. Following the ANSI guidelines the *bxb* Physionet function can be used to compute the $Se$ and $PPV$ statistics in order to score participants entries. Indeed, one cannot expect that every detector will choose the same sample as the QRS annotation, hence it is essential to define a time window for agreement between detectors (observers) in *bxb*. However this approach also has limitations: let's assume that we have two sets of annotations, a reference set with stable fiducial markers (i.e. every beat is marked at the same location on the QRS complex, say the R-wave peak), and another with highly variable markers (e.g. because its creator attempted unsuccessfully to mark QRS onsets, with an unreliable method). Assume, however, that the second set does contain an annotation for each beat, within *bxb*'s window of tolerance. Using *bxb*, it would be learnt that there was perfect agreement between the two sets- but nothing would be learnt about the accuracy of the RR intervals measured in the second set. It is therefore important to compare the 'interval measurements' and not the accuracy of the QRS detection and this is the intention behind using *mxm* in the E2/E5 Challenge scores.

Using the RMS error as a distance measure between two RR time series (event E1/E4 of the Challenge) might be highly sub-optimal in many cases. Figure 4.4 illustrates this statement by showing a real example where the reference and algorithms estimated FHR are mostly equivalent apart from a small interval at 47 sec. Because of this local error the usage of the RMS measure will drive the error for this record to a high value. The problem will only be present when the FHR time series outputted by the algorithm is not smoothed. However a smoothing function will also 'remove' the high/low increases in HR due to the presence of ectopic beats for example or might lead to an inaccurate smoothing in the case of high variability/inadequate calibration of the smoothing function parameters.

In conclusion, the statistic(s) to use very much depend on the intended purpose behind what is being evaluated:

- The $Se$, $PPV$, $F_1$ statistic allows for the assessment of whether or not the QRS are detected at the location they are intended to be with a tolerance of ±window. The length of this window should be shorter for fetal ECG than for adult ECG due to the higher FHR

- The statistic for $E2/E5$ using the Physionet *mxm* function allows to assess whether the RR intervals are correctly evaluated. Thus it assesses whether a QRS detector is consistent in the fiducial point that is picked up on the QRS waveform

- The statistic for $E1/E4$ allows to assess whether the detection can result in accurate FHR evaluation. It is however prone to outlier (RMS based statistic) and a measure such as the HRm might be preferable.

## 4.5 Fetal ECG synthetic model

### 4.5.1 Introduction

One of the main limitations of the Challenge 2013 (aside from the limited number of channels of maternal and fetal data available) was the absence of pathological cases (pregnancy with adverse outcome), or rare cases such as similar FHR and maternal heart rates (MHR) at times of abrupt heart rate changes. This motivated the development of an artificial NI-FECG simulator and inclusion of a limited number of artificially generated signals into the challenge database (20 in total) in order to model some of these events.

An artificial FECG-MECG mixture simulator can be used to create realistic abdominal recordings that model specific physiological phenomena. The simulator presented in this work, *fecgsyn*, is an extension of the ECG simulator introduced by McSharry *et al.* [4], later adapted by Sameni *et al.* [5] to reflect a single dipole per cardiac source. By decreasing the dipole magnitude and placing it at a certain location with respect to the maternal heart, a fetal-maternal mixture is generated. The dipole model is linearly related to the body surface potentials by a projection matrix that takes into account the temporal movements and rotations of the cardiac dipole. By superimposing two such projections with different amplitudes it is possible to simulate a realistic FECG-MECG mixture. However, several features of the mixing were not implemented in the previously published simulator. In particular, the non-stationary mixing due to respiration and fetal movement, as well as realistic heart rate changes (both normal and pathological) were not included in previous simulators.

This section introduces the physiological phenomena encompassed by the simulator, and the framework for generating the fetal-maternal mixtures on the abdominal ECG. Some specific cases are then described and simulated. Finally, a summary of limitations and possible future additions to the simulator are presented. The open source toolbox and associated graphical user interface released as the outcome of this work are presented in Appendix E.

### 4.5.2 Simulator description

Surface electrodes measure the electrical potential created by different sources that propagate throughout the volume conductor. These sources can be categorised as cardiac sources (i.e. maternal and fetal myocardia) or noise sources (e.g. muscle activity from movement or contractions). The developed simulator considers all sources as point source dipoles, that may be rotated and translated. The dipoles have two basic attributes: a vector, represented by three coordinates in the Cartesian coordinates system, and a location which, together with the electrodes' location, define the propagation matrix of the signals to the observation points. Table 4.6 lists the key parameters together with the physiological effects that were modelled. Figure 4.5 illustrates the modelling of the volume conductor (defined as a cylinder), fetal and maternal heart locations, as well as the electrode positions. The dipole frame is represented by three arrows (that depict the orientation of the fetal and maternal body axis), centred on each of the maternal and fetal hearts. Table 4.7 specifies the heart and electrode locations in polar coordinates (where $\theta$ is the angle in radians, $\rho$ is the radius, and $z$ the height), used for the illustration in Figure 4.5. The following phenomena, that affect the morphology and dynamics of the abdominal ECG, were modelled: noise (muscle noise, electrode motion, baseline wander), maternal respiration and fetal breathing movement, fetal movement, contractions, presence of ectopic beats and multiple pregnancies.

Table 4.6: Important simulator parameters. NS: noise, HR: heart rate, TR: trajectory, RS: respiration, GE: geometry, MA: muscle artifact, EM: electrode motion, BW: baseline wander. EB: ectopic beat.

|  | Parameters | Definition | Range/type |
|---|---|---|---|
|  | $fs$ | sampling frequency [Hz] | — |
| NS | $ntype$ | type of noise | $MA, EM, BW$ [80] |
|  | $SNR_{fm}$ | signal to noise ratio of the FECG relative to maternal interference | — |
|  | $SNR_{mn}$ | signal to noise ratio of the MECG relative to noise | — |
| HR | $mhr$ | maternal heart rate [bpm] | 70-140 |
|  | FHR | fetal heart rate [bpm] | 120-160 [24] |
|  | $macc$ | maternal acceleration/deceleration [bpm] | — |
|  | $facc$ | fetal acceleration/deceleration [bpm] | — |
|  | $macctype/facctype$ | maternal/fetal acceleration/deceleration type | 'none', 'nsr', 'tanh', 'mexhat', 'gauss' |
| TR | $ftraj$ | fetal heart trajectory to model fetal movement | 'none', 'linear', 'spline', 'helix' |
| RS | $mres$ | maternal respiration frequency [Hz] | $0.2 - 0.3$ |
|  | $fres$ | fetal respiration frequency [Hz] | $0.8 - 0.95$ [45] |
| GE | $mheart$ | maternal heart position in polar coordinates | — |
|  | $fheart$ | fetal heart position in polar coordinates | — |
|  | $elpos$ | electrode position in polar coordinates | — |
|  | $mvcg$ | maternal vectorcardiogram number | 1-9 |
|  | $fvcg$ | fetal vectorcardiogram number | 1-9 |
| EB | $mectb$ | ectopic beat for the maternal ECG | boolean |
|  | $fectb$ | ectopic beat for the fetal ECG | boolean |

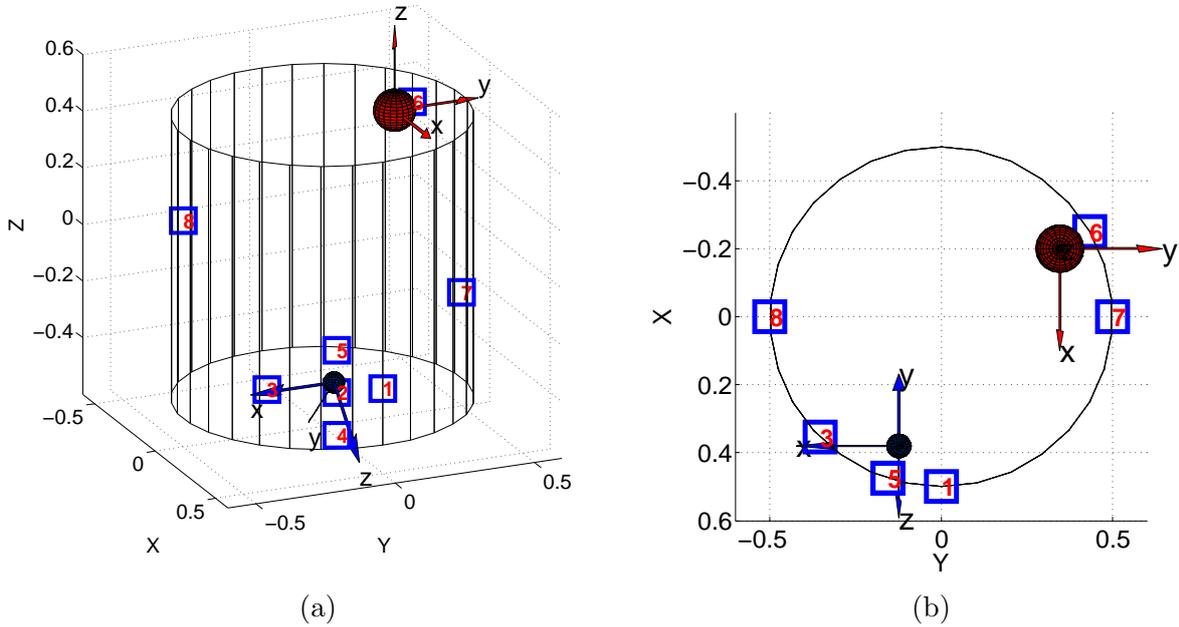

Figure 4.5: (a) Volume conductor representation. The two spheres represent maternal (large sphere) and the fetal (smaller sphere) hearts. The blue numbered squares show the location of the electrodes on the volume conductor, while the arrows represent the orientation of the fetal and maternal body axis which are defined by the heart orientation; (b) Top view of the volume conductor. Note that in (b) electrodes 2 and 4 are not represented because of the top view.

Table 4.7: Electrodes and hearts position for the example in Figure 4.5; $\theta$ is the angle (in rad), $\rho$ is the radius and $z$ the height in cylindrical coordinates.

| Electrode Variable | 1 | 2 | 3 | 4 | 5 | 6 | 7 | 8 | fheart | mheart |
|---|---|---|---|---|---|---|---|---|---|---|
| $\theta$ | 0 | $-\pi/10$ | $-\pi/4$ | $-\pi/10$ | $-\pi/10$ | $2\pi/3$ | $\pi/2$ | $3\pi/2$ | $2\pi/3$ | $-\pi/10$ |
| $\rho$ | 0.50 | 0.50 | 0.50 | 0.50 | 0.50 | 0.50 | 0.50 | 0.50 | 0.40 | 0.40 |
| $z$ | -0.30 | -0.30 | -0.3 | -0.45 | -0.15 | 0.40 | -0.20 | 0.20 | 0.40 | -0.30 |

### 4.5.2.1 The dipole model

Following the work of Sameni *et al.* [5] the myocardial electrical activity is represented by a time-varying vector which has its origin at the centre of the heart. The dipole vector can be written as $\underline{\mathbf{d}}(t) = x(t)\underline{\mathbf{e}}_x + y(t)\underline{\mathbf{e}}_y + z(t)\underline{\mathbf{e}}_z$, where $\underline{\mathbf{e}}_j, j \in \{x, y, z\}$, are basis vectors of the body axis. The torso potential, $\phi(t)$, generated by the dipole and sensed by an electrode located at a point, such that the vector from the dipole centre to this point is defined by $\underline{\mathbf{r}}$, can be written as:

$$\phi(t) = \frac{\underline{\mathbf{d}}(t) \cdot \underline{\mathbf{r}}(t)}{4\pi\sigma|\underline{\mathbf{r}}(t)|^3} = \frac{1}{4\pi\varepsilon_0}[x(t)\frac{r_x(t)}{|\underline{\mathbf{r}}(t)|^3} + y(t)\frac{r_y(t)}{|\underline{\mathbf{r}}(t)|^3} + z(t)\frac{r_z(t)}{|\underline{\mathbf{r}}(t)|^3}]. \quad (4.9)$$

where $\varepsilon_0$ is the permittivity of the volume conductor (here, considered to be isotropic and unitary). The tension measured at this electrode is then given by $\phi(t) - \phi_0(t)$, where $\phi_0(t)$ is the reference potential, which was taken to be located on the back of the cylinder at the polar coordinates [pi 0.5 -0.3]. Note that the potential is time dependent with respect to $\underline{\mathbf{d}}(t)$ and $\underline{\mathbf{r}}(t)$ which can vary with fetal movement and respiration, for example.

### 4.5.2.2 The ECG model

In order to generate realistic MECG and FECG waveforms, the dynamical ECG model introduced by McSharry *et al.* [4] was used. The model is built on the premise that a set of Gaussian kernel functions can be used to approximate ECG cycles. Figure 4.6 illustrates how an ECG cycle is approximated by a set of $N = 7$ Gaussian functions. This number of Gaussian was motivated in [81, 82] so that each asymmetric turning point (P and T-waves) be represented with two Gaussians and the QRS complex with a total of three Gaussians (one for each wave).

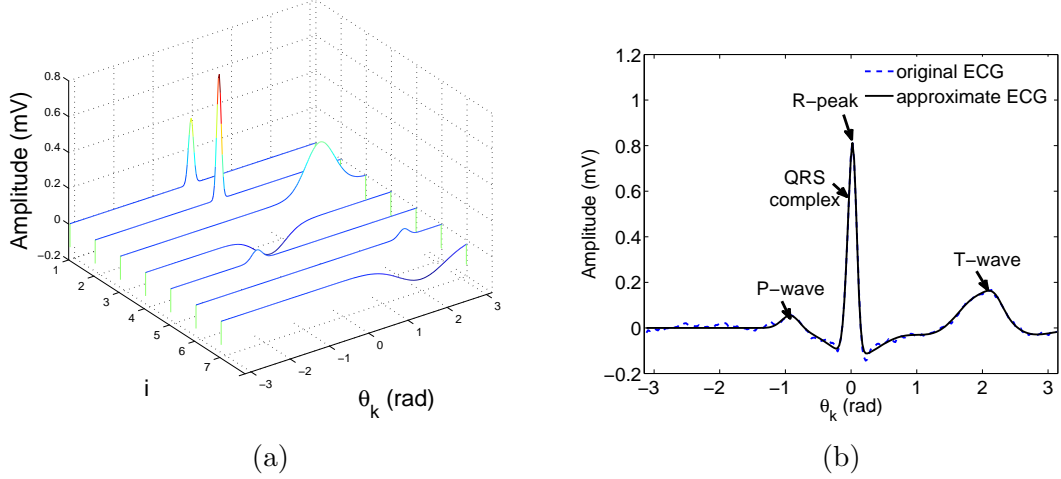

(a)                                (b)

Figure 4.6: ECG Gaussian fitting example: (a) The seven Gaussians (index i) used to represent an ECG cycle. (b) Original ECG cycle (dotted line) and the reconstructed ECG cycle obtained by summing the seven Gaussian functions from (a).

The set of equations that describes the model can be written:

$$\begin{cases} \dot{\theta} &= \omega \\ \dot{x} &= -\sum_{i=1}^{N} \frac{\alpha_i^x \omega}{(b_i^x)^2} \Delta\theta_i^x exp(-\frac{(\Delta\theta_i^x)^2}{2(b_i^x)^2}) \\ \dot{y} &= -\sum_{i=1}^{N} \frac{\alpha_i^y \omega}{(b_i^y)^2} \Delta\theta_i^y exp(-\frac{(\Delta x\theta_i^y)^2}{2(b_i^y)^2}) \\ \dot{z} &= -\sum_{i=1}^{N} \frac{\alpha_i^z \omega}{(b_i^z)^2} \Delta\theta_i^z exp(-\frac{(\Delta\theta_i^z)^2}{2(b_i^z)^2}) \,, \end{cases}$$ (4.10)

where $\Delta\theta_i^j \equiv (\theta - \xi_i^j) \mod 2\pi$, $j \in \{x, y, z\}$, $\alpha_i^j$, $b_i^j$, $\xi_i^j$ correspond to the peak amplitude, width and centre parameters of the *ith* Gaussian kernel functions respectively. The velocity of the circular trajectory of the heart dipole $\underline{\mathbf{d}}(t)$ is $\omega = 2\pi f$, where $f$ is the beat-to-beat heart rate (in $Hz$). The model parameters for the FECG-MECG mixture were derived from a set of VCG[5] from the PTB Diagnostic ECG Database [83] records (i.e. $\alpha_i^j$, $b_i^j$, $\xi_i^j$ parameters were derived for all three channels of the VCGs). For each cardiac signal, heart rate variability (HRV) is introduced by making $\omega$ time variant. Therefore, maternal and fetal signals can be produced in both physiological and pathological scenarios, which is important for testing the robustness of NI-FECG extraction techniques. Moreover, since each dipole is indepen-dent, more than one fetal heart dipole can be generated to simulate multiple pregnancies. Additionally, by changing the SNR between the maternal and fetal signal (SNR$_{fm}$), different mixtures with varying FECG amplitudes can be generated.

---

[5]The parameters were extracted from the following records: s0273lre, s0291lre, s0302lre, s0303lre, s0306lre, s0491_re, s0533_re from the PTBDB and two others sets of parameters were taken from [5].

#### 4.5.2.3 Forward propagation of the dipoles fields

A projection matrix built upon the cardiac dipole model (equation 4.9) was used to project the VCG waveforms onto the electrode positions on the surface of the volume conductor, therefore obtaining ECG signals. The projection matrix contains information about the permittivity of the conductor (assumed constant), dipole origin and relative location between observing electrodes and source. The projection of a single dipole can be described mathematically as:

$$\underline{\mathbf{s}}(t) = \mathbf{H}(t) \cdot \mathbf{R}(t) \cdot \underline{\mathbf{d}}(t) + \underline{\mathbf{w}}(t), \qquad (4.11)$$

where $\underline{\mathbf{s}}(t) \in \Re^M$ corresponds to the signal recorded on the $M$ ECG channels at time $t$, $\underline{\mathbf{d}}(t) = [x(t), y(t), z(t)]^T$ contains the three components of the dipole vector generated by the ECG model [4], $\mathbf{H}(t) \in \Re^{M \times 3}$ corresponds to the body volume conductor model (projection matrix) which is time-varying in cases where the dipole is translated along a given trajectory, $\mathbf{R}(t) \in \Re^{3 \times 3}$ is the rotation matrix for the dipole vector, and $\underline{\mathbf{w}}(t) \in \Re^M$ corresponds to the noise on each ECG channel at time $t$. Allowing the projection matrix, $\mathbf{H}(t)$, to be time-variant opens new possibilities for shifting dipoles, which are useful when modelling breathing-related heart motion and fetal activity-related movement. The resulting signals are given by:

$$\underline{\mathbf{s}}(t) = \mathbf{H}_m(t) \cdot \mathbf{R}_m(t) \cdot \underline{\mathbf{d}}_m(t) + \mathbf{H}_f(t) \cdot \mathbf{R}_f(t) \cdot \underline{\mathbf{d}}_f(t) + \underline{\mathbf{w}}(t), \qquad (4.12)$$

where the subscripts $m$ and $f$ indicate mother and fetus respectively. The forward projection matrices can be written as $\mathbf{H}_m = (h_{ij}^m)$, $\mathbf{H}_f = (h_{ij}^f)$ with $j \in \{x, y, z\}$ and $i$ represents the electrode number so that:

$$h_{ij}(t) = \frac{1}{4\pi\varepsilon_0} \frac{r_j^i(t)}{|\mathbf{r}^i(t)|^3}, \qquad (4.13)$$

where $\varepsilon_0$ is the permittivity of the volume conductor, $\underline{\mathbf{r}}^i$ is the vector going from the dipole to $ith$ electrode and $r_j^i$ is the $jth$ component of $\underline{\mathbf{r}}^i$. Note that the elements of the **H** matrices are time dependent.

#### 4.5.2.4 Modelling noise

Real noise signals from the PhysioNet Noise Stress Test Database [65, 80] (NSTDB) were used. The NSTDB contains 30 min, two-channel recordings of three noise sources: muscle artifact ($MA$), electrode motion ($EM$) and baseline wander ($BW$). However, there are three main limitations when using these records: (1) their length is limited to 30 min (thus sub-segments of these records would be re-used if the modelled signal has a greater length); (2) there are only two channels available (three orthogonal components are required to model a noise dipole); (3) this is noise from one individual and does not account for inter-individuals variability. To overcome these limitations, an auto-regressive (AR) model can be used to simulate the non-stationarity of the noise process.

The coefficients of the AR model (taken to be 12) were estimated over a randomly selected 20 sec segment of one of the NSTDB signals. This defined the AR filter's coefficients, associated poles and frequency response. In order to account for the uniqueness of the records available in the NSTDB and incorporate some variability, the poles from the identified AR filter were allowed to evolve slightly, following a random walk, while ensuring they stayed within the unit circle (ensuring the stability of the filter), and preserving symmetry between conjugated poles. In order to generate a third noisy channel, principal component analysis (PCA) was applied to both of the generated noise channels, with the first principal component considered to be the third channel. This ensures that the generated channel was not independent from the original two channels. Indeed the VCG is based on the concept of the dipole approximating the electromagnetic activity of the heart. The three VCG leads are correlated and are mainly contained in a two dimensional plane. In the same way the channels of the, VCG-like of the noise dipole, will exhibit some statistical dependence. This is an approximation that is realistic given that muscle recorded on different parts of the body will likely not be completely independent.

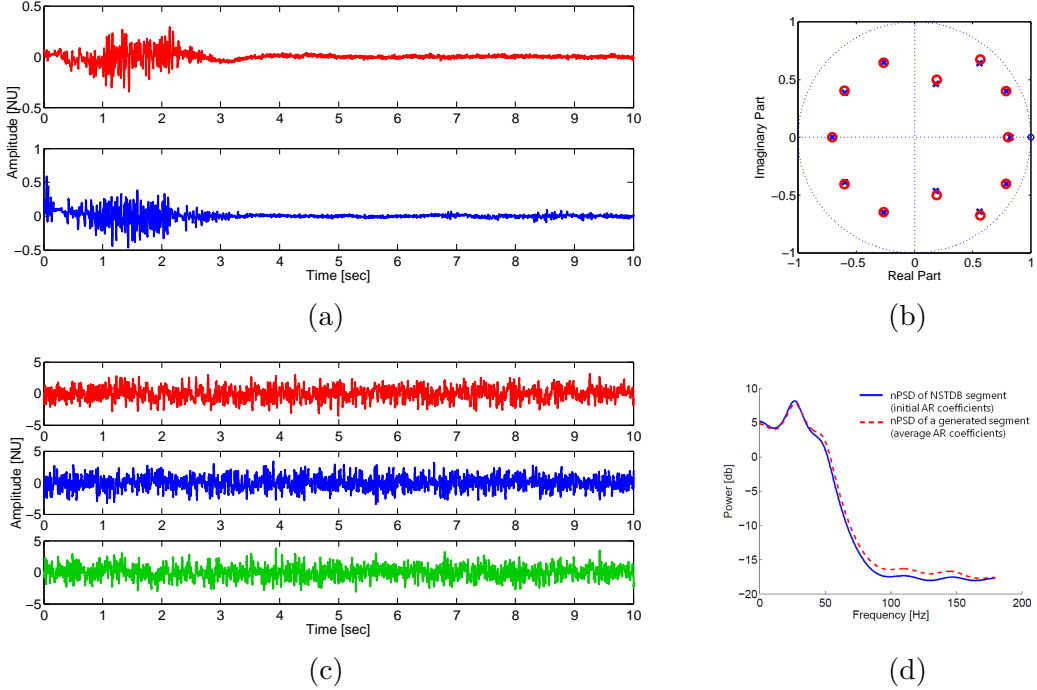

Figure 4.7: Noise generator. (a) Noisy segment selected in the $MA$ record of the NSTDB (2 channels available); (b) the AR coefficients (initial value: blue crosses and final values after leaving the coefficient evolve following a random walk: red circle); (c) artificially generated noise (the third channel corresponds to the first principal component of PCA applied on the first two channels), (d) the normalised power spectral density (nPSD) of the second channel of the NSTDB selected noise and the second generated noise channel.

Figure 4.7 illustrates a noisy segment from the muscle artifact record, $MA$, of the NSTDB, and the artificially generated noise together with the normalised power spectral density (nPSD). The nPSD was obtained by computing the frequency response from the transfer function using the initial AR coefficients (for obtaining the nPSD of the NSTDB segment) and the averaged, evolving AR coefficients (for obtaining the nPSD of the generated segment). Note how the nPSD shape was preserved while producing the artificial noisy segments. This behaviour is desirable as the intention is to simulate a specific physiological noise (e.g. $MA$) as characterised by its frequency content.

#### 4.5.2.5 Signal calibration

Cardiac and noise signals were calibrated in a similar fashion with respect to the maternal signal. The amplitude scaling gain, $p$, applied to fetal or noise signals before adding them to the signal mixture, is defined as:

$$y(t) = x(t) + p \cdot v(t) \tag{4.14}$$

$$p = \sqrt{\frac{P_x}{P_v}} \cdot 10^{\frac{-S}{20}} \tag{4.15}$$

where $S$ represents the SNR in decibels of the signal to be calibrated (i.e. -SNR$_{fm}$ or SNR$_{mn}$), $y(t)$ is the output signal (i.e. the abdominal mixture $\underline{s}$), $x(t)$ is the maternal signal and $v(t)$ is either the fetal signal (for S = -SNR$_{fm}$) or noise (for S = SNR$_{mn}$). $P_x$ represents the power of the MECG and $P_v$ the power of noise/fetal signal. In order to eliminate the influence of baseline wander from the $MA$ and $EM$ records on the $P_v$ calculation, they were high-pass filtered by a 10th order zero-phase Butterworth filter at 1 Hz cut-off frequency prior this calculation. Filtering was performed because the $MA$ and $EM$ records contained $BW$ noise, which usually has a high power (which would be taken into account when computing $P_v$) but is easier to remove than $MA$ or $EM$ noise.

Fetal calibration was performed as follows: the average power of the MECG and FECG projected onto the abdominal channels were computed. Given the averaged power and the SNR$_{fm}$, the coefficient $p$ from equation 4.15 was computed and the resulting mixture was obtained using equation 4.14 for each individual channel. With respect to noise, any subset of the $MA$, $EM$, $BW$ noises can be added to the abdominal mixture in a similar way. In the presence of multiple noise types, the parameter SNR$_{mn}$ specified the overall amount of noise to add to the MECG. This approach for noise calibration was chosen (instead of the other alternative to calibrate the amount of noise on each individual channel (as in Sameni *et al.* [5])) because of the need to model the varying SNR that is observed in practice, and that is dependent on the electrodes position, hearts and noise sources location.

#### 4.5.2.6 Modelling respiration

Respiration influences the morphology of the ECG by changing the position of the heart, but also the orientation of the cardiac dipole. The respiratory pattern was modelled as a variable sawtooth-like signal, $\beta(t)$. The respiration model was introduced in Stridh *et al.* [84] then applied in Petrenas *et al.* [85] for simulating f-waves in the ECG. In this work, the first three coefficients of the Fourier transform of the normalised sawtooth signal were used to model the waveform, while time-varying amplitude and frequency modulation was also introduced in order to account for the non-stationary behaviour of the physiological process. The model can be written as:

$$\beta(t) = \sum_{j=1}^{3} a_j(t)\, sin(j\omega_0 t + \frac{\Delta f_d}{f_m} sin(\omega_m t))\,, \quad (4.16)$$

$$a_j(t) = \frac{2}{j\pi}(a + \Delta a\, sin(\omega_a t))\,, \quad (4.17)$$

where $\omega_0 = 2\pi f_0$ is the fundamental angular frequency corresponding to the breathing rate, $\Delta f_d$ is the frequency deviation tolerated around $f_0$, $f_m$ is the modulation frequency (i.e. frequency at which the breathing wave is modulated), $a$ is the sawtooth amplitude, $\Delta a$ is the modulation amplitude and $\omega_a = 2\pi f_a$ where $f_a$ is the modulation frequency. For the simulations in this work the values were set to: $\Delta a = 0.3$, $f_a = 0.1$ Hz, $\Delta f_d = 0.05$ Hz and $f_m = 0.1$ Hz. Typical breathing rates are between 12 and 18 breaths per minute (brpm) for adults and ~50 brpm for a fetus[6]. Figure 4.8a shows an example of the respiratory waveform generated over a 15 sec period. The sawtooth-like function models respiration as a slowly increasing inspiration and rapidly declining expiration. The time-varying rotation matrix coupled with the breathing waveform was then computed using the following equations:

$$\psi_j = \psi_j^{max} \cdot \beta(t) + \psi_j^0, \quad (4.18)$$

$$\mathbf{R}(t) = \mathbf{R}_x(t) \cdot \mathbf{R}_y(t) \cdot \mathbf{R}_z(t)$$

$$= \begin{bmatrix} 1 & 0 & 0 \\ 0 & cos(\psi_x) & sin(\psi_x) \\ 0 & -sin(\psi_x) & cos(\psi_x) \end{bmatrix} \begin{bmatrix} cos(\psi_y) & 0 & sin(\psi_y) \\ 0 & 1 & 0 \\ -sin(\psi_y) & 0 & cos(\psi_y) \end{bmatrix} \begin{bmatrix} cos(\psi_z) & sin(\psi_z) & 0 \\ -sin(\psi_z) & cos(\psi_z) & 0 \\ 0 & 0 & 1 \end{bmatrix}$$

with $\beta(t)$ representing the sawtooth shape of the generated signal, $\psi_j$ the rotation angle around axis $j$ with $j \in \{x, y, z\}$, and $\psi_j^0$ the static (mean) orientation of the fetal heart with respect to the maternal heart (which is assumed to be aligned with the default Cartesian basis). Note that $\psi_j$ is time dependent but for notation simplicity the time index was omitted. $\psi_j^{max}$ is the maximum deviation of the rotation angle, $\psi_j$, in radians and taken to be $\psi_x^{max} = 0.2$, $\psi_y^{max} = 0.16$, $\psi_z^{max} = 0.14$ (order of magnitude inspired by Leanderson *et al.* [86]).

---

[6]Numerical values on healthy fetuses as found by [45] were 57.2 (30-33 weeks of gestation) and 47.9 brpm (37-40 weeks). Note that the fetus does not actually breath as the oxygen supply is coming from the mother. However, breathing movement is observed in the fetus after 10-weeks of gestation [44], and thus changes in the heart location (and consequently ECG amplitude modulation) is present.

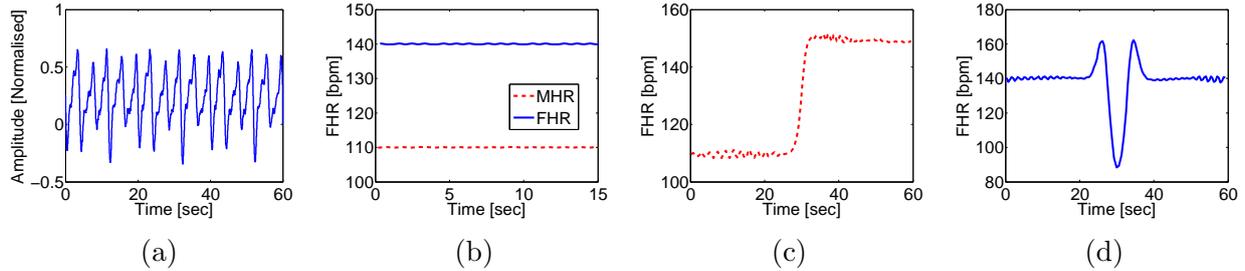

Figure 4.8: Example of simulated (a) respiration waveform; (b) heart rate; (c) fast heart rate increase; (d) umbilical cord effect on FHR.

The same sawtooth function was used to modulate the translation of the heart during breathing. According to Mcleish *et al.* [87] the heart position oscillates around 1-2 cm along the z-axis with the breathing activity of the diaphragm. This range is approximated by allowing the heart to translate its position by 5% of the height of the cylinder volume conductor for the maternal heart and 1.5% for the fetal heart.

#### 4.5.2.7 Modelling heart rate variability

The model for HRV introduced by [4] was used. The sympatho-vagal balance is defined as the balance between the effect of the sympathetic and parasympathetic nervous systems which act in opposite directions. The sympatho-vagal balance is believed to be reflected in the beat-to-beat changes of the cardiac cycle [88]. The RR time series power spectrum was modelled as a mixture of two Gaussians. The first one models the HR modulation with respiration (parasympathetic activity) known as respiratory sinus arrhythmia (RSA, $\sim$ 0.25 Hz for adults), and the second one models the Mayer peak for baroreflex regulation (sympathetic activity, $\sim$ 0.1 Hz)[7]. Figure 4.8b shows two HR time series (one for the FHR and one for MHR) generated using this model. Use of this model is referred to as normal

---

[7]Baroreflex regulation aims to maintain the blood pressure at nearly constant level.

sinus rythm (NSR) modulation.

Furthermore, two additional heart rate related effects were modelled: (1) a rapid decrease or increase of HR that models fast transition to prolonged deceleration caused by fetal head compression or uteroplacental insufficiency [89], and (2) the ischemic response due to umbilical cord compression. The first effect was modelled using a hyperbolic tangent increase or decrease in HR, while the latter was modelled using a 'Mexican-hat' function to modulate the FHR. The 'Mexican hat' function in the one-dimensional case is defined as:

$$f(t) = \frac{2}{\sqrt{3\sigma}\pi^{\frac{1}{4}}}(1 - \frac{t^2}{\sigma^2})exp(-\frac{t^2}{2\sigma^2}), \qquad (4.19)$$

where $t$ is the time index and $\sigma$, the standard deviation. Figure 4.8d shows an example of modelled FHR response to umbilical cord compression; the umbilical cord is initially only slightly compressed - there is partial obstruction of the umbilical vein. This results in a heart rate increase. As the obstruction increases, the umbilical artery becomes occluded resulting in a decrease of FHR, thus creating a 'shoulder' followed by a drop. As the umbilical cord is released, there is a FHR recovery with a possible slight increase before going back to the baseline value (second shoulder) [90].

### 4.5.2.8 Modelling fetal movement

Fetal movements were simulated by changing the position of the cardiac dipole within the volume conductor. This makes the projection matrix time variant since the distance between dipole and electrodes changes at every time instant. Different trajectory patterns were modelled, including linear, spline and helicoidal (see Table 4.6).

### 4.5.2.9 Modelling contractions

Uterine contractions are muscular activities that vary in intensity and periodicity during pregnancy. By using the $MA$ noise model (Section 4.5.2.4) and modulating it by an envelope function (e.g. Gaussian) for simulating contraction's intensity increase and decay, it is possible to produce a realistic estimate of the contraction signals. Umbilical cord compression might happen during a uterine contraction, with this compression generating a typical FHR pattern, modelled using the Mexican-hat function described in Section 4.5.2.7. With the developed model, early and late decelerations can be modelled by time shifting the HRV and noise modulating functions. Early deceleration corresponds to the FHR going below the baseline heart rate. In general the onset, minimum and recovery of the deceleration are in phase with the onset, peak and recovery of the contraction. In the case of a late deceleration, the onset, minimum and recovery of the deceleration happen after the contraction [89]. Examples are depicted in Figure 4.9. During those episodes, it is possible that a cross-over between MHR and FHR occurs which might make the extraction of the FECG more difficult. In addition, the muscular noise associated with a contraction will affect the quality of the abdominal mixture. This effect is depicted in Figure 4.9 for the simulated data and an example on real data was presented in Figure 2.8 in Chapter 2.

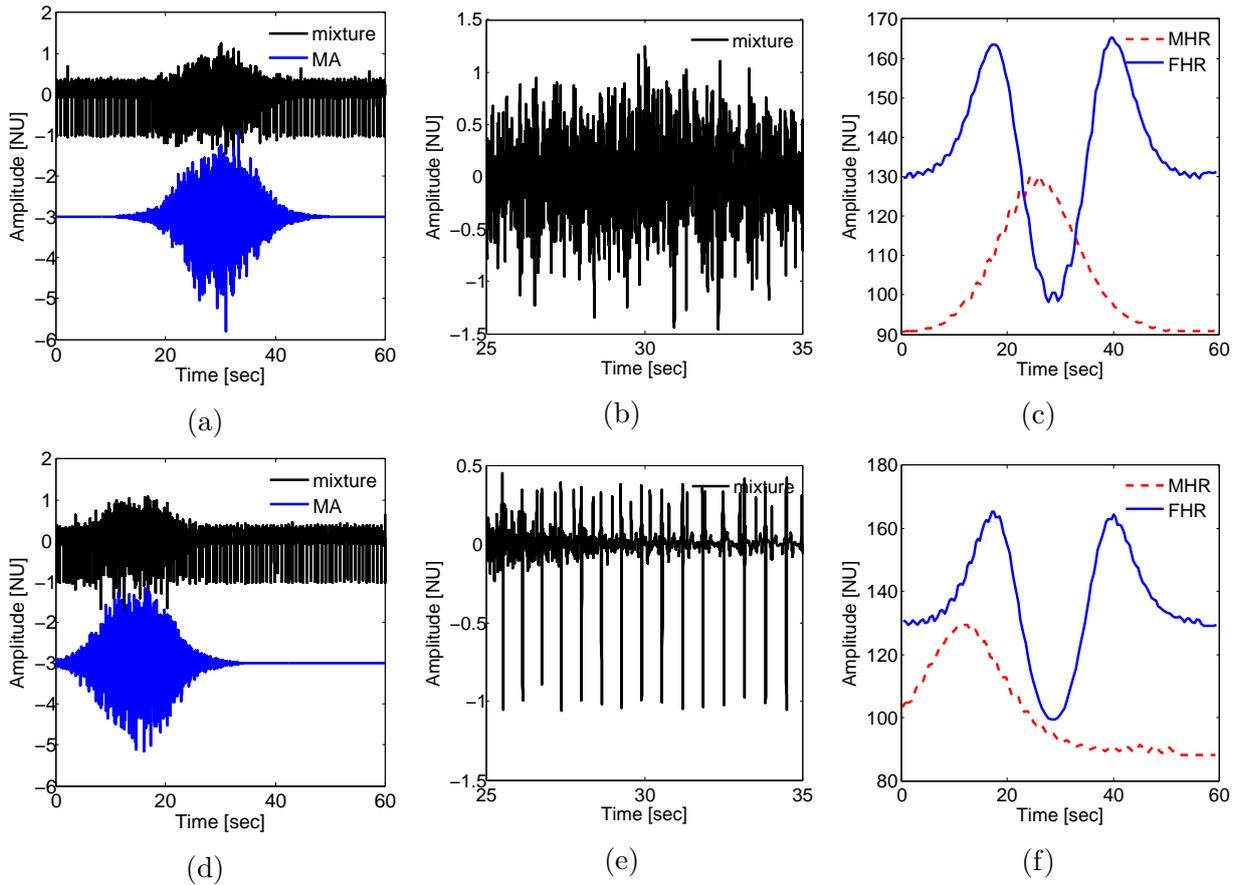

Figure 4.9: (a) FECG-MECG mixture (60 sec) with a contraction in the centre; (b) 10 sec excerpt from (a); (c) the corresponding FHR and MHR for plots (a), representing an early deceleration; (d) signal mixture (60 sec) with a contraction at the beginning of simulation; (e) 10 sec segment from (d); (f) is the corresponding FHR and MHR for plots (d), representing a late deceleration.

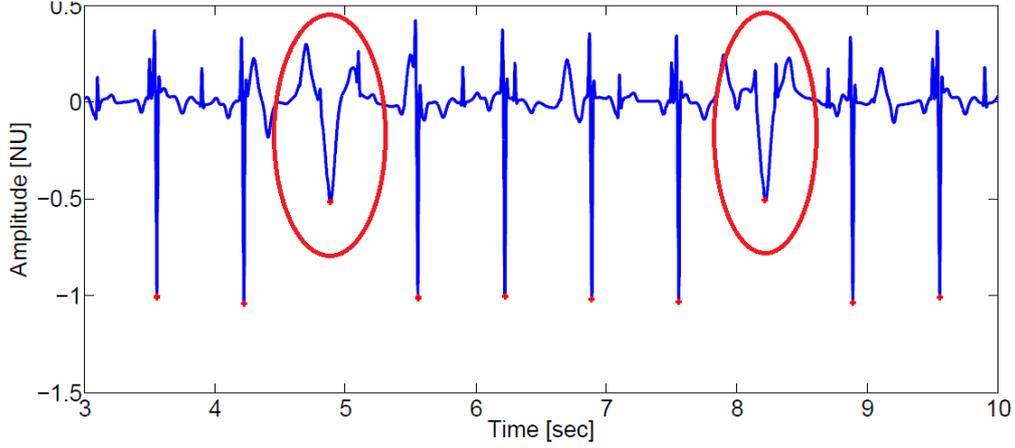

Figure 4.10: Example of ectopic beats generated on the abdominal ECG mixture.

#### 4.5.2.10 Modelling ectopic beats

Ventricular ectopic beats or premature ventricular contractions (PVC) are common, even in healthy individuals [91]. According to Kennedy *et al.* [92] ectopic beats occur in 1% of normal patients undergoing standard ECG measurements and in 40 to 75% of patients assessed by Holter ECG (during 24 or 48h). The consequences of these beats and arrhythmia for both mother and fetus are rarely analysed in NI-FECG literature. The model used was introduced by Clifford *et al.* [93] and subsequently used in Oster *et al.* [94]. It makes use of a Hidden Markov Model to switch between normal and ectopic beats. The state transition matrix, $\mathbf{T}$, is described as:

$$\mathbf{T} = \begin{bmatrix} rn & 1 - rn \\ 1 - re & re \end{bmatrix} \qquad (4.20)$$

where $rn = 0.7 + 0.1 \cdot \mathcal{U}(0,1)$ is the probability of the incoming beat being chosen as normal given that the previous beat was normal and $re = 0.2 + 0.1 \cdot \mathcal{U}(0,1)$ is the probability of the incoming beat being chosen as PVC given that the previous beat was PVC. $\mathcal{U}(0,1)$ is the uniform distribution over the interval [0 1]. Gaussian parameters for the PVC beats were taken from [94]. Figure 4.10 shows an example of ectopic beats generated on the abdominal ECG mixture.

#### 4.5.2.11 Modelling multiple pregnancies

The model can be generalised to multiple pregnancies by adding additional terms to equation 4.12. As an example in the case of twins:

$$\underline{\mathbf{s}}(t) = \mathbf{H}_m(t) \cdot \mathbf{R}_m(t) \cdot \underline{\mathbf{d}}_m(t) + \mathbf{H}_{f_1}(t) \cdot \mathbf{R}_{f_1}(t) \cdot \underline{\mathbf{d}}_{f_1}(t) + \mathbf{H}_{f_2}(t) \cdot \mathbf{R}_{f_2}(t) \cdot \underline{\mathbf{d}}_{f_2}(t) + \underline{\mathbf{w}}(t), \quad (4.21)$$

where $f_1$ is used to designate the first fetus and $f_2$ the second. It is then possible to assess how source separation algorithms would perform on multiple pregnancies cases.

### 4.5.3 Discussion

The simulator introduced in this section is not intended to be a substitute for clinical experimentation. In fact, it is meant to account for the inherent incompleteness of the current NI-FECG databases and simulators. Its purpose is to facilitate the rapid stress-testing of NI-FECG algorithms under highly variable conditions, with a complete gold standard. Essentially, the main purpose is to provide a common open-source tool for benchmarking extraction/detection algorithms. This tool allows the comparison between methods in the literature to extract the NI-FECG and FQRS complexes (which is a pre-requisite for clinical studies). The tool can be used to provide a preliminary hurdle for stress testing algorithms and for proof-of-concept purposes. However, it is not an alternative to evaluating any technique on real data.

To date there is no study known to the authors that assessed the performance of the NI-FECG extraction algorithms with respect to electrode placement. In many instances, researchers made use of large arrays of electrodes or positioned the electrodes manually in order to achieve an acceptable extraction. The simulator presented here can be used to better understand how electrode position, fetal position and orientation can influence the extraction algorithms.

The simulator can be used to generate a series of challenging scenarios such as contractions, fetal heart rate dips below the mother's heart rate, the effect of ectopy, cases of non-stationary mixes and the influence that the number of channels has on the performance of an algorithm (e.g. blind source separation). The analysis of these cases can be used to illustrate the limitations of existing NI-FECG extraction algorithms, such as various template subtraction techniques and blind source separation (see for example the methods presented in [1, 2]).

The simulator also permits the testing of different conditions concomitantly which may have an impact on each other (e.g. a uterine contraction is expected to affect the fetal heart rate). Since the fetal heart rate response depends on the state of the fetus, there is no rule how this interaction should manifest. However, the ECG simulator allows the simulation of important physiological events, which is the relevant aspect for FECG extraction methods. The user may adjust the simulator to embed rules to reflect this.

The volume conductor was chosen as a cylinder and the conduction properties were considered homogeneous. The vernix caseosa is a thin layer that forms around 28th-32nd weeks and dissolves in 37th-38th weeks in normal pregnancies [19]. It is highly non-conductive thus limiting successful NI-FECG extraction. There are few studies which describe the properties of the tissues involved in FECG conduction [95, 96]. The overall permittivity of the material between the fetal heart and the sensors also changes with gestational week, amount of placenta and presence of vernix caseosa. A more advanced modelling of how this thin layer affects permittivity is required; however, the effects of such layer were not modelled in this thesis. Nevertheless, a simple way of modelling this effect with the simulator in its current state, is to change the value of the $\text{SNR}_{fm}$ in order to decrease the FECG power into the mixture. However, the intention in this work was to keep the model simple but realistic, to evaluate the effects that the physiological changes may have on the NI-FECG extraction algorithms. Thus the choice was driven by the need for simplicity. Logical extensions of the simulator could include a more realistic volume conductor with inhomogeneous conduction properties. These extensions can further be integrated upon the toolbox that was released on `Physionet.org`

One of the limitations of the ECG simulator presented in this paper is that the morphology of the ECG cycle will vary linearly with HR (see equation 4.10 $\dot{\theta} = \omega$). Although the PR, ST segments, the QRS complex and the T-wave do vary with HR [97], this variation is likely to be non-linear. One way of accounting for this problem would be to change the phase wrapping of the ECG model to be, at least, piece-wise linear for each individual segment. Moreover, the noise sources provided by the NSTDB were only measured from one individual. Although our AR model generates different noise time series, their power spectrum is roughly the same. In order incorporate more variability and improve the noise generator, it would be useful to have recordings from more individuals. The uterus activity could be modelled better with some real electrohysterography measurements. The simulator could be further improved by simulating atrial fibrillation in the MECG [85, 94]. All these suggested additions to the simulator can be incorporated into the released version of the code.

## 4.6 Summary

This section presented the databases used in this thesis as well as the background work that was necessary to create or annotate these data. A total of four databases with reference annotations were formatted for the experiments presented in Chapters 6, 7 and 8. The statistics used to report the results in these chapters were also presented. Finally, the *fecgsyn* simulator, used to generate the AQTDB as well as a subset of data of the CDB, was introduced.

# Chapter 5

# ECG signal quality

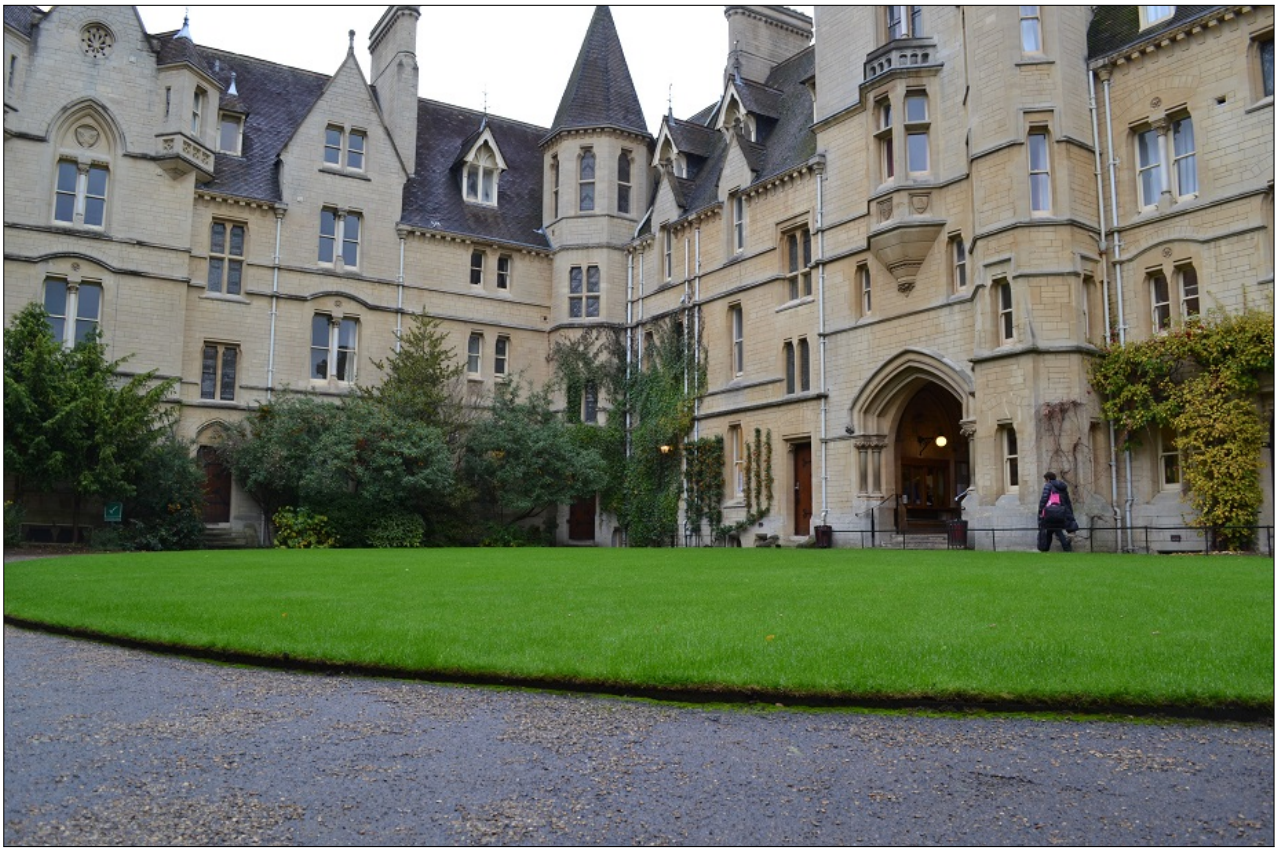

Figure 5.1: It is traditional for colleges in Oxford to have a *quad* (or quadrangle) which is usually a rectangular place surrounded by the college buildings. Most *quads* have a small grass garden which is kept perfectly cut and brushed to give some visual effect. It is forbidden for students or visitor to walk on it as it could affect its quality.

## 5.1 Introduction

It is very common for the electrocardiogram (ECG) to be contaminated by noise and artifacts. Worse yet, the noise frequency content often overlaps with the frequency band of interest (thus limiting denoising approaches in the frequency domain) or has morphology similar to the one of the ECG (thus limiting denoising approaches in the time domain) [98]. Figure 5.2 illustrates this frequency and temporal overlap by showing some simple noise from the Physionet Noise stress test database in the time and frequency domain. Identifying bad quality ECG signals is of tremendous importance when automated signal processing algorithms are to be used for screening or monitoring cardiac conditions. For example in the context of Holter ECG, which is often corrupted by a significant amount of motion artifact, or when considering telemonitoring systems. It is also essential in the context of non-invasive fetal ECG (NI-FECG) in order to exclude the noisy chest and abdominal ECG channels. The work presented in this chapter was published in Behar *et al.* [12, 99] following from the work of Clifford *et al.* [11]. A number of signal quality indices were evaluated on manually labelled ECG signals. Given the similarities between fetal ECG and adult ECG, and that adult ECG databases are more represented and more easily available, it was decided to tackle the signal quality index (SQI) problem using adult data. Moreover, studying the performance of SQI on pathological data was one of the aims of this work and it was therefore required to use adult ECG (as pathological fetal ECG are not currently available). Some findings of this work was further used in Chapter 6 in order to exclude bad quality maternal chest ECG segments.

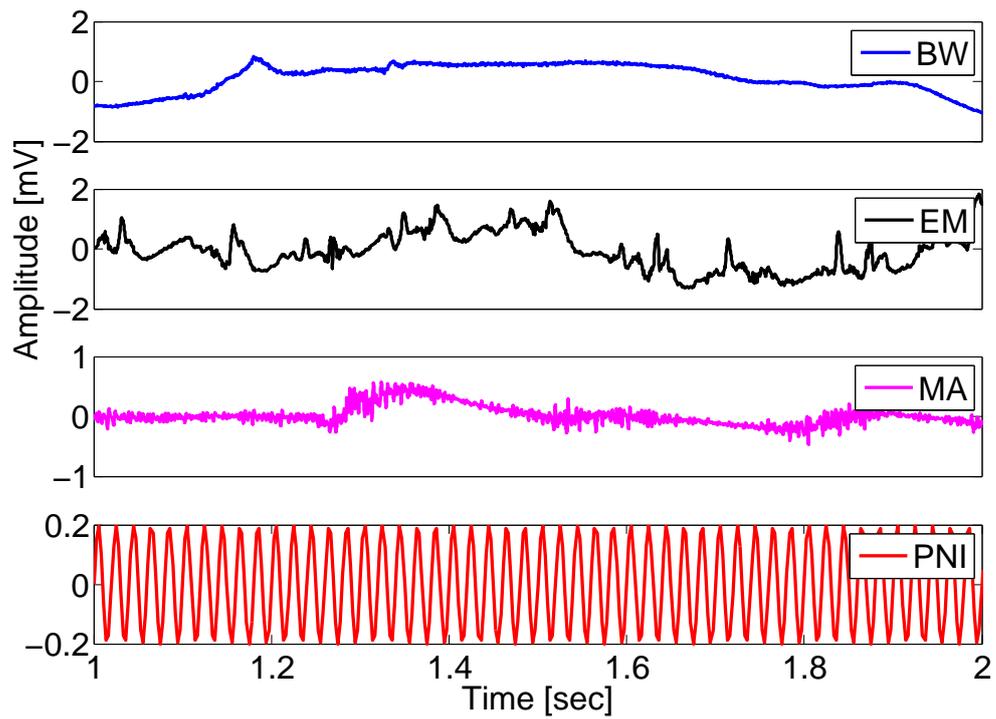

(a)

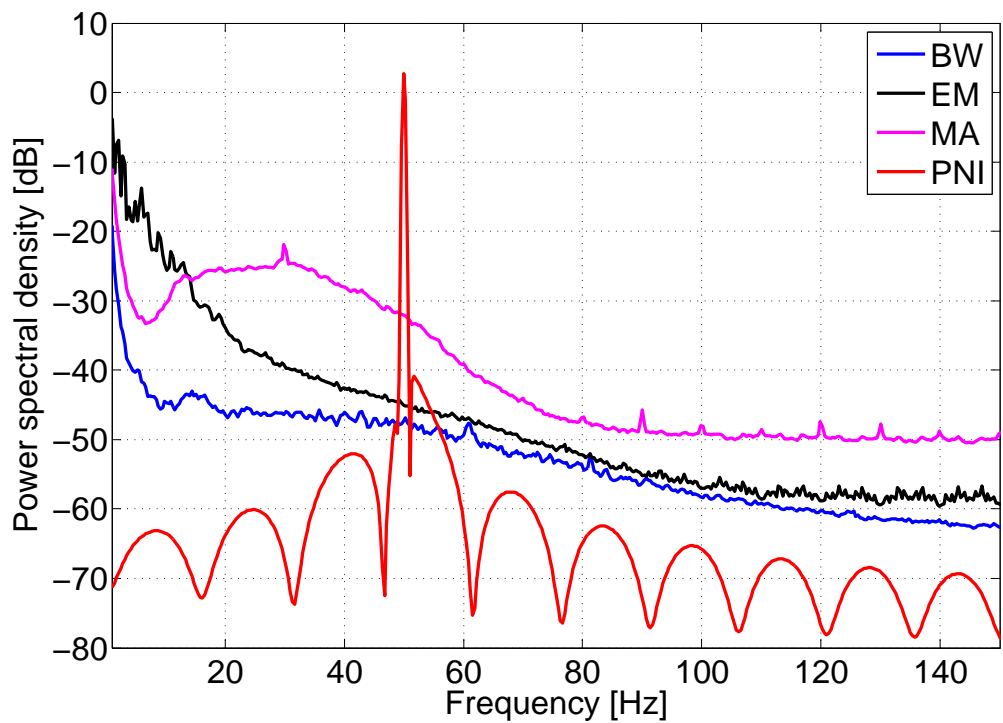

(b)

Figure 5.2: Sources of noise in the time (a) and frequency (b) domains. $BW$: baseline wander, $EM$: electrode motion, $MA$: muscle artifact and $PNI$: power noise interference.

| Databases | | Rythms |
|---|---|---|
| Extended CinC [100] | DB$_1$ | NSR |
| MIT-BIH Arrhythmia [101] | DB$_2$ | AFIB, SVTA, AFL, SBR, VT, VFL, A, V |
| MIMIC II [102] | DB$_3$ | ASYS, SBR, TACHY, VT, VFIB |

Acronyms; NSR: normal sinus rhythm, AFIB: atrial fibrillation, SVTA: supraventricular tachyarrhythmia, AFL: atrial flutter, SBR: sinus bradycardia, VT: ventricular tachycardia, VFL: ventricular flutter, A: atrial premature beat, V: premature ventricular contraction.

Table 5.1: SQI Databases and rhythms.

## 5.2 Methods

**Databases**

For this work, ECG records from three different databases were used: the Physionet/Computing in Cardiology (CinC) Challenge 2011 [100], the MIT-BIH arrhythmia [101] and the MIMIC II [102] databases denoted by DB$_1$, DB$_2$, DB$_3$ respectively (see Table 5.1). DB$_1$ was used to train an ECG quality assessment model. DB$_2$ was used to test how the model performed on arrhythmic records as well as on records of a different modality (Holter ECG). It was then used in the training set to account for pathological signals. Finally the resulting model was tested on ECG segments extracted from DB$_3$ in order to study how effectively ECG quality assessment can identify artefactual signals.

*DB$_1$*

The Physionet/CinC DB included 1500 ten-second (10 sec) recordings of standard twelve-lead ECG i.e. 18000 (12 × 1500) individual segments with full diagnostic bandwidth (0.05-100 Hz), sampled at 500 Hz with 16-bit resolution. The recordings were performed for a minimum of 10 sec by nurses, technicians, and volunteers who had varying amounts of training in recording the ECG. After human annotation, less than one third of the data was classified as poor quality.

It is well-known that building classifiers using imbalanced classes, i.e. when one class greatly outnumbers the other classes, causes bias and results in poor generalisation ability of the classification model. This can be accounted for by: (1) using a prior probabilities (and a Bayesian training paradigm); (2) weighting the penalty of the classes differently in the cost function or (3) balancing the classes. In this work the dataset was balanced by bootstrapping the unrepresented class to be equal to the more numerous class using additive real noise on clean data. In order to generate additional noisy records the PhysioNet Noise Stress Test Database (NSTDB) [65, 103] noise samples were used. The database contains samples for three types of noise: record $BW$ contains baseline wander noise, record $EM$ contains electrode motion artifact with a significant amount of baseline wander and muscle noise as well. Finally record $MA$ contains mainly muscle noise. $EM$ and $MA$ noise were used in the experiments. Baseline wander was not considered because it does not often render an ECG unusable. In order to prevent correlation between the noise added in the training and test sets, the $EM$ and $MA$ files were divided in two at the midpoint of the recordings with one half to be used with the training set and half with the test set.

As only two leads were provided with both the $EM$ and $MA$ records from PhysioNet, PCA was used to generate a third orthogonal lead along the dominant principal component from the two leads in the recording. Assuming the resultant leads form an orthogonal lead set with arbitrary orientation, the Dower transformation [104] was then used to create realistic correlated 12-lead sets of noise. The purpose of this step was to generate 12-lead noise records with realistically correlated, but not identical noise on the different leads. Herein, $EM$ and $MA$ refer to 12-leads records generated from the PhysioNet samples and using PCA and the Dower transformation.

Figure 5.3 introduces the workflow for generating the additional noisy records. A total of 18000 single leads were available from the challenge. All the leads were individually annotated, providing a training set composed of 2020 noisy and 9980 good quality leads. 4000 good quality leads were selected from a subset of 334 patients in the training set that had the 12-leads of good quality ($334 \times 12 = 4008$). For each of these patients, a calibrated amount of 10 sec noise from the $EM$ or $MA$ files was selected at random and added to the clean 12-lead record. The noise was added so that the SNR was equal to $-6db$ on each generated lead. After adding the noise and computing the SQI values, 8 leads were randomly dropped making 4000 10 sec ECG segments for each $EM$ and $MA$ noise. This resulted in 8000 additional leads

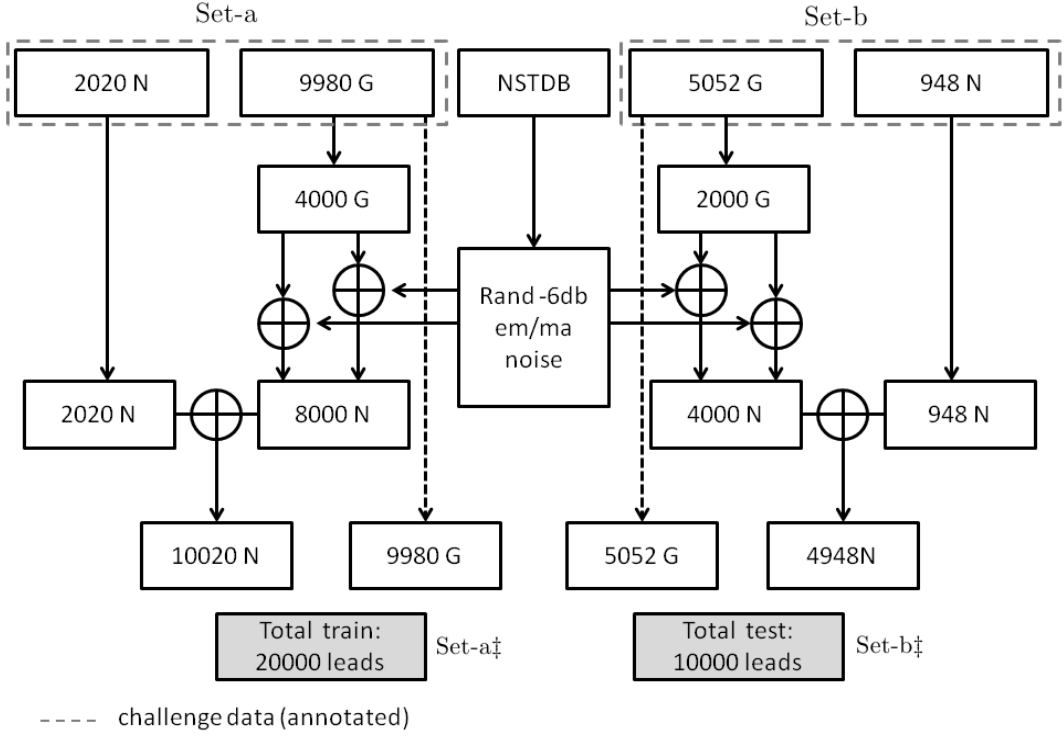

Figure 5.3: Flow diagram to illustrate how the data from Set-a and Set-b were generated to balance datasets Set-a‡ and Set-b‡. G: Good segments, N: Noisy segments, NSTDB: PhysioNet noise stress test database.

for the training set (4000 with $EM$ noise and 4000 with $MA$ noise).

The SNR, $S$, was controlled by computing the coefficient $p$ for each lead as follows:

$$y(t) = x(t) + p \cdot v(t)$$

$$p = \sqrt{\frac{P_x}{P_v}} \cdot 10^{\frac{-S}{20}}$$

where $y$ is the output signal (i.e. the noisy signal generated from the clean sample), $x$ is the initial clean signal, $v$ is a 10 sec noisy sample from $EM$ or $MA$ selected at random, $p$ is the amount of the noisy sample that was added to the clean signal, $P_x$ is the power of the clean signal and $P_v$ is the power of noisy signal.

The same steps were followed in order to generate 4000 bad quality leads from clean leads in the test set (Set-b). As a result, an overall of 20000, 10 sec ECG samples for the training set and 10000 for the test set were obtained. About half the segments were of good quality and half of poor quality, resulting in balanced classes. These segments constituted the first dataset, referred to as the extended CinC DB ($DB_1$, see Table 5.1). The training and test sets for this DB are denoted Set-a‡ and Set-b‡ respectively.

*$DB_2$*

The second dataset, $DB_2$, included 48 complete two-lead ECG records with arrhythmia reference annotations from the MIT-BIH arrhythmia DB. The records have a diagnostic bandwidth of 0.1-100 Hz with 12-bit resolution and were digitised at 360 Hz. Six arrhythmia types were analysed (see Table 5.1). Data were segmented for 10 sec after the onset of each arrhythmia. In the event of an arrhythmia being less than 10 sec, the segment was discarded to avoid mixed rhythms. Atrial premature beats (A) and premature ventricular contraction beats (V) were located and 10 sec segments centred on each beat, while ensuring no overlap between segments, were extracted. Both available leads were used. In most records, the first channel is the modified limb lead II and the second channel is usually a precordial lead V1. This resulted in 9452 individual 10 sec segments whose quality was manually evaluated. Within the 9452 signals, 269 were annotated as being of poor diagnostic quality and the remaining 9183 as good or medium quality. The proportion of good and bad quality signals from this dataset was not balanced because it was used as a first layer for evaluating the model on real arrhythmic data.

*$DB_3$*

The third dataset included a subset of over 4050 life-threatening HR-related arrhythmia alarms taken from $DB_3$. Five types of annotated arrhythmia were available (see Table 5.1). ECG were available at 125 Hz with a resolution of 8 to 10 bits. Segments of 10 sec were extracted from the records, starting 10 sec prior the alarm was triggered. The American Organization for the Advancement of Medical Instrumentation recommendations [105] specifies that a cardiac alarm should be triggered within 10 sec after an arrhythmia detection. This motivated the choice for working with 10 sec segments preceding each alarm. The number of ECG leads per record was variable and all flat line leads were removed from all records. Moreover, there were some intrinsic limitations to using this DB because the raw signals were not accessible from the monitoring system. Though the ECG signals were originally sampled at a high sampling rate with 12-bit resolution, they were scaled and re-sampled prior to capture. This reduced the resolution from 12 bits to 8 or 10 bits in most cases. Furthermore, only one in four samples was recorded using a turning-point compressor, thus providing a nonlinearly filtered signal. This resulted in reduced time and amplitude resolution of the ECG and, consequently, a high proportion of ECG signals manually labelled as medium quality.

**Labelling ECG signals**

The ECG signals were manually annotated using the classification scheme suggested in [106]. For this experiment only two classes were considered: good (A-B) and bad quality (D-E). Segments of class C were not considered because most of the time they were borderline, either because the level of noise was not high enough to cause bad QRS detections or because the noise was very localised within the 10 sec segment. Clinically speaking, these ECG were often interpretable but there were transient artifacts or a moderate amount of noise present making them prone to label disagreement between annotators. Therefore, it should not be considered wrong for the algorithm to classify such segments as either good or bad quality. As a consequence leads of type C were excluded in order to avoid confusing the classifier during model training.

Figure 5.4 shows examples of quality labels. A total number of 14968 bad quality (10020 in Set-a‡ and 4948 in Set-b‡) and 13055 good quality signals (8745 in Set-a‡ and 4310 in Set-b‡) from $DB_1$, as well as 269 bad quality and 8768 good quality signals from $DB_2$ were used. The distribution of per class signal quality is summarised in Table 5.2. Table 5.3 gives the distribution of signal quality per arrhythmia type for $DB_2$. The choice to work with 10 sec segments of a given rhythm was made in order to derive consistent statistics (e.g. kurtosis, skewness) that could be used to build up our model. No mixed rhythms were retained for $DB_2$ but segments with various numbers of A or V beats were considered. Furthermore, half of the records from $DB_3$ were manually annotated. This provided an additional 5634 annotated signals as detailed in Table 5.2.

Table 5.2: Distribution of signal quality across all databases.

| Database \Quality | # Bad | # Medium | # Good | Total # |
|---|---|---|---|---|
| Extended CinC (DB$_1$) | 14968 | 1977 | 13055 | 30000 |
| MIT-BIH arrhythmia (DB$_2$) | 269 | 415 | 8768 | 9452 |
| MIMIC II (DB$_3$) | 1080 | 780 | 3774 | 5634 |
| Overall | 16317 | 3172 | 25597 | 45086 |

Table 5.3: Distribution of signal quality per arrhythmia type for the MIT-BIH arrhythmia DB.

| Arrhythmia \Quality | # Bad | # Medium | # Good | Total # |
|---|---|---|---|---|
| AFIB | 24 | 87 | 1261 | 1372 |
| SVTA | 0 | 2 | 12 | 14 |
| AFL | 3 | 8 | 81 | 92 |
| SBR | 1 | 24 | 333 | 358 |
| VT | 1 | 0 | 11 | 12 |
| VFL | 1 | 1 | 14 | 16 |
| A | 55 | 106 | 1453 | 1614 |
| V | 184 | 187 | 5603 | 5974 |
| Overall | 269 | 415 | 8768 | 9452 |

Table 5.4: Repartition of record quality per arrhythmia type for TAs in the MIMIC II DB.

| Arrhythmia \Quality | # Bad | # Good | Total # |
|---|---|---|---|
| ASYS | 39 | 62 | 101 |
| SBR | 2 | 231 | 233 |
| TACHY | 46 | 835 | 881 |
| VT | 34 | 1370 | 1404 |
| VFIB | 7 | 34 | 41 |
| All TA | 128 | 2532 | 2660 |
| FA (all arrhythmias) | 952 | 1242 | 2194 |
| Overall | 1080 | 3774 | 4854 |

**Preprocessing and SQIs**

Each ECG channel was downsampled to 125 Hz when necessary for computing efficiency. QRS detection was performed on each channel individually using two open source QRS detectors (*eplimited* [107] and *wqrs* [108, 109]). The *eplimited* algorithm is less sensitive to noise than *wqrs* [110] and consequently discrepancies in their output is an indicator of noise (see bSQI and qSQI below). Seven SQIs, introduced in previous works [11, 99, 110], were calculated for every single-lead separately:

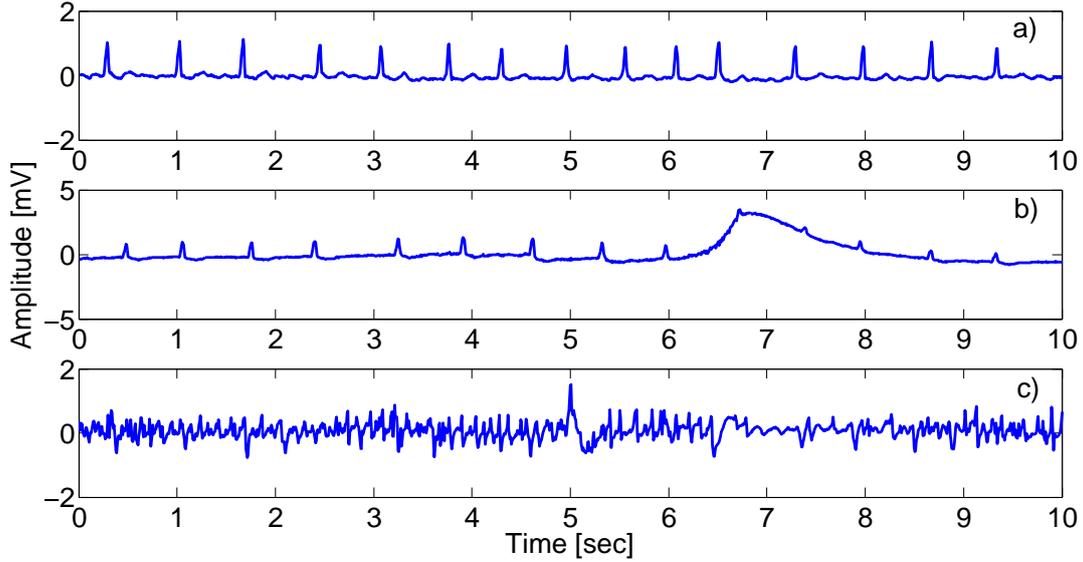

Figure 5.4: Example of lead quality. a) good quality, b) medium quality, c) bad quality. ECG segments are taken from the MIT-BIH arrhythmia DB.

1. kSQI: The fourth moment (kurtosis) of the signal is defined as $kSQI = E\{X - \mu\}^4/\sigma^4$, where $X$ is the signal vector considered as the random variable, $\mu$ is the mean of $X$, $\sigma$ is the standard deviation of $X$, and $E\{X - \mu\}$ is the expected value of the quantity $X - \mu$. It is expected that a good ECG will be highly non-Gaussian since it is not very random. Random noises such as muscle artifacts and baseline wander tend to have distributions tending to be normal and their kurtosis are less than five [98].

2. sSQI: The third moment (skewness) of the signal defined as $sSQI = E\{X - \mu\}^3/\sigma^3$. It is expected that the ECG to be highly skewed due to the QRS complex.

3. pSQI: The relative power in the QRS complex: $\int_{5Hz}^{15Hz} P(f)\,df \, / \, \int_{5Hz}^{40Hz} P(f)\,df$. Most of the power is expected to be in the 5-15Hz band for a clean ECG.

4. basSQI: The relative power in the baseline: $\int_{1Hz}^{40Hz} P(f)\,df \, / \, \int_{0Hz}^{40Hz} P(f)\,df$. A sudden 'low frequency ($\leq 1Hz$) bump' will result in low basSQI.

5. bSQI: The fraction of beats detected by *wqrs* that matched with beats detected by *eplimited*. Originally proposed by Li *et al.* [110].

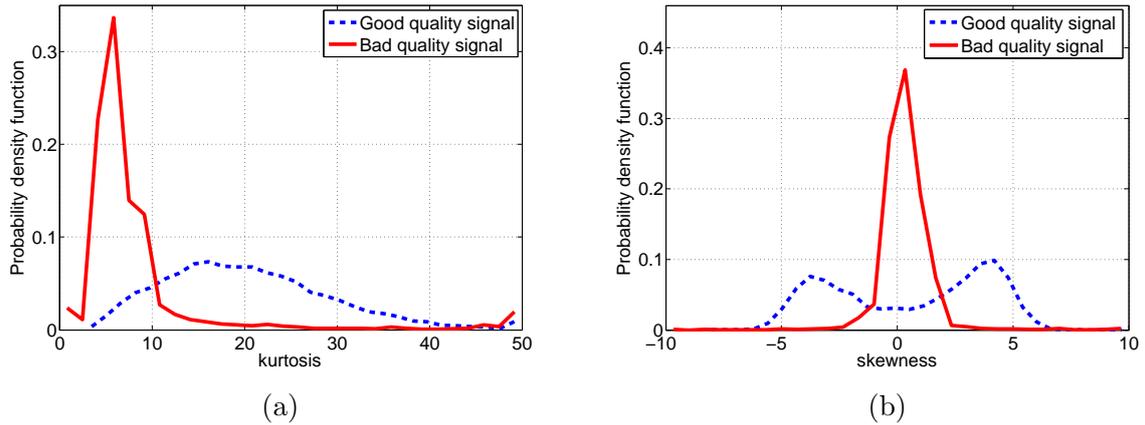

Figure 5.5: Probability density function of kurtosis (a) and skewness (b) for 21823 10 sec signals from the CinC DB (13055 good quality and 14968 bad quality) before normalisation.

6. qSQI: The ratio of the number of beats detected by *eplimited* and *wqrs*[1].

7. pcaSQI: A ratio comprising the sum of the eigenvalues associated with the five principal components over the sum of all eigenvalues obtained by PCA applied to the time-aligned ECG cycles detected in the window by the *eplimited* algorithm, segmented at 100 ms either side of the R-peak.

Note that baseline wander was filtered prior to computing sSQI and kSQI using a second order zero phase high pass filter with 0.7 Hz cut-off frequency. Power spectral density was evaluated using the Burg method [111] with eleven poles. Figure 5.5 shows the probability density function of kurtosis and skewness obtained for 21823 10 sec signals with 13055 good quality signals and 14968 bad quality signals before normalisation. The distributions suggest that the two statistical measures might provide information for distinguishing between good and bad quality signals because of the limited overlap between the distributions.

**Support vector machine classifier**

A support vector machine (SVM) classifier (libSVM library [112, 113]) was used with a Gaussian (nonlinear) kernel defined by: $k(\underline{\mathbf{x}}_n, \underline{\mathbf{x}}_m) = \exp\left(-\gamma \|\underline{\mathbf{x}}_n - \underline{\mathbf{x}}_m\|^2\right)$, where $\gamma$ controls the

---
[1]bSQI and qSQI are somewhat redundant but the addition of qSQI to bSQI was because of the original implementation of bSQI from Li *et al.* [110] which was not providing a score that was representative of the accuracy (i.e. taking into account $Se$ and $PPV$). bSQI was redefined in later experiments (see E.4), eliminating the need for the qSQI.

width of the Gaussian and plays a role in controlling the flexibility of the resulting classifier. $\underline{\mathbf{x}}_n$ and $\underline{\mathbf{x}}_m$ are two vectors expressed in the initial feature space. The SVM with a Gaussian kernel has two parameters: $C$ and $\gamma$. The parameter $C$ acts as a regularisation coefficient in that it controls the trade-off between minimising training errors and controlling the model complexity [114]. Based on cross-validation results performed on $DB_1$ (i.e. extended Challenge database) in [11] $C = 25$ and $\gamma = 1$ were used. A SVM classifier was chosen (over alternative machine learning methods) because the computation of the model parameters is a convex optimisation problem, and so there are no local solutions but only a global minimum [114]. Thus, given a set of input features derived from a dataset, and the $C$ and $\gamma$ parameters, the SVM will always converge to the same solution.

**Statistical measures**

In the context of the SQI experiments: sensitivity ($Se$) measures the proportion of poor quality signals that have been correctly identified as such. Specificity ($Sp$) measures the proportion of good quality signals that have been correctly identified as acceptable, and accuracy ($Ac$) corresponds to the proportion of signals that have correctly been classified.

**Quality assessment**

The classification quality assessment was conducted in three phases (see Figure 5.6):

1. The classifier model was built on $DB_1$ and the performance of the individual SQIs were evaluated, followed by the evaluation of combinations of the best individual SQIs: pairs, triplets, etc. Therefore the contribution of each SQI on the classification performance could be analysed.

2. The classifier was trained on all good and bad quality data from $DB_1$ and the performance of the trained classifier was assessed on $DB_2$. $DB_1$ and $DB_2$ were then merged and used as the new training set. 5-fold cross validation was performed on this set in order to assess the performance of the predictive model.

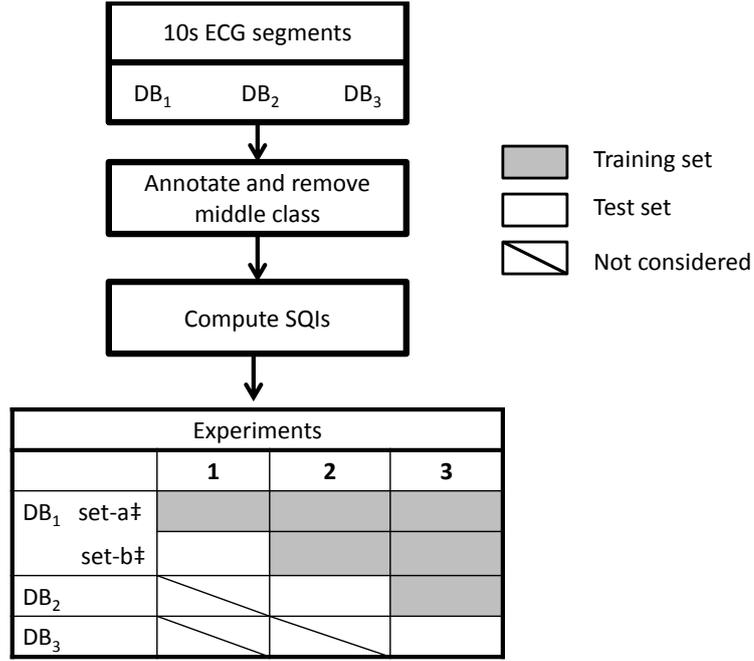

Figure 5.6: Flowchart of experiments. $DB_1$: extended CinC DB, $DB_2$: MIT-BIH arrhythmia DB, $DB_3$: MIMIC II DB. Set-a‡: training set of $DB_1$ and Set-b‡: test set of $DB_1$. Experiments 1,2,3 are described in Section 5.2. The databases are described in Tables 5.1-V.

3. The classifier was trained on the combination of $DB_1$ and $DB_2$, and performance was tested on $DB_3$.

## 5.3 Results

For each of the three experiments, the results were as follows;

1. Table 5.5 shows that bSQI and pcaSQI best distinguish between records of good and bad quality on training Set-a‡. The SVM was then trained with combinations of the best SQIs. The results for the best pair, triplet etc. of SQIs combinations are summarised in Table 5.6. The best accuracy on the training set (Set-a‡) was obtained when considering all SQIs ($Ac = 99.3\%$). However, the best accuracy ($Ac = 98.5\%$) on the test Set-b‡ was obtained using only the best six SQIs. Although using all SQIs provided a negligible drop in test performance (to 98.4%) compared to using the best combination of six SQIs, it was decided to keep all the SQIs for the later experiments, since the McNemar's test showed no statistically significant difference (with $\chi^2 = 1.32$ and $p = 0.25$).

Table 5.5: Single lead classification using SVM on individual SQIs and on Set-a‡.

|     | bSQI  | kSQI  | sSQI  | pSQI  | basSQI | qSQI  | pcaSQI |
|-----|-------|-------|-------|-------|--------|-------|--------|
| Ac  | 0.969 | 0.879 | 0.892 | 0.752 | 0.626  | 0.766 | 0.950  |
| Se  | 0.983 | 0.911 | 0.944 | 0.806 | 0.679  | 0.805 | 0.945  |
| Sp  | 0.953 | 0.842 | 0.831 | 0.690 | 0.564  | 0.721 | 0.956  |

Table 5.6: Single lead classification performance using SVM with combinations of SQIs. Classification is performed on $DB_1$.

|                                              | Ac train Set-a‡ | Ac test Set-b‡ |
|----------------------------------------------|-----------------|----------------|
| bSQI, pcaSQI                                 | 0.974           | 0.971          |
| bSQI, pcaSQI, qSQI                           | 0.982           | 0.980          |
| bSQI, pcaSQI, kSQI, sSQI                     | 0.986           | 0.981          |
| bSQI, pcaSQI, kSQI, basSQI, qSQI             | 0.988           | 0.983          |
| bSQI, pcaSQI, pSQI, sSQI, qSQI, basSQI       | 0.991           | 0.985          |
| bSQI, pcaSQI, kSQI, sSQI, qSQI, basSQI, pSQI | 0.993           | 0.984          |

2. Table 5.7 shows the results obtained on $DB_2$ per arrhythmia while training on $DB_1$. Note that the statistical measures $(Ac, Se, Sp)$ are not always significant due to the very small number of signals for a given type of arrhythmia. The overall accuracy obtained was 94.6% for $DB_2$. Repeating 5-fold cross validation 150 times on the combination of $DB_1$ and $DB_2$ resulted in a mean accuracy of $\mu = 0.991$ with standard deviation $\sigma = 2.86 \times 10^{-4}$ and 95% confidence interval $(CI)$, $CI = [0.990\ 0.991]$ on the training set and $\mu = 0.986, \sigma = 1.20 \times 10^{-3}, CI = [0.983\ 0.988]$ on the validation set. Note that cross-validation was repeated 150 times to ensure a stable value of accuracy and it was found that additional repetitions were not required as the variance estimates had converged.

3. Training the classifier on the combination of $DB_1$ and $DB_2$, and testing on $DB_3$ labelled signals resulted in $Ac = 89.3\%, Se = 74.4\%$ and $Sp = 93.5\%$ on the test set. Table 5.8 reports the results per arrhythmia type.

## 5.4 Discussion

Despite the promising results on healthy ECG, the accuracy for sinus bradycardia and asystole when tested on the MIMIC II database was very low. Outcomes for asystole were not

Table 5.7: Single lead classification using SVM. The classifier was trained on $DB_1$ and tested on $DB_2$.

|  | Quality | | | Statistics | | |
| --- | --- | --- | --- | --- | --- | --- |
|  | # Bad | # Good | Total # | Ac | Se | Sp |
| AFIB | 24 | 1261 | 1285 | 0.967 | 0.958 | 0.968 |
| SVTA | 0 | 12 | 12 | 1 | 0 | 1 |
| AFL | 3 | 81 | 84 | 0.833 | 1 | 0.827 |
| SBR | 1 | 333 | 334 | 0.961 | 1 | 0.961 |
| VT | 1 | 11 | 12 | 0.833 | 1 | 0.818 |
| VFL | 1 | 14 | 15 | 0.467 | 1 | 0.429 |
| A | 55 | 1453 | 1508 | 0.972 | 0.927 | 0.974 |
| V | 269 | 5603 | 5787 | 0.936 | 0.826 | 0.940 |
| Overall | 269 | 8768 | 9037 | 0.946 | 0.863 | 0.948 |

Table 5.8: Lead quality assessment per arrhythmia type on MIMIC II DB.

|  | Quality | | | Statistics | | |
| --- | --- | --- | --- | --- | --- | --- |
|  | # Bad | # Good | Total # | Ac | Se | Sp |
| TA-ASYS | 39 | 62 | 101 | 0.683 | 0.974 | 0.500 |
| TA-SBR | 2 | 231 | 233 | 0.708 | 0 | 0.714 |
| TA-TACHY | 46 | 835 | 881 | 0.943 | 0.804 | 0.951 |
| TA-VT | 34 | 1370 | 1404 | 0.961 | 0.677 | 0.968 |
| TA-VFIB | 7 | 34 | 41 | 0.902 | 0.571 | 0.971 |
| All TA | 128 | 2532 | 2660 | 0.902 |  |  |
| All FA | 952 | 1242 | 2194 | 0.857 | 0.737 | 0.949 |
| Overall | 1080 | 3774 | 4854 | 0.892 | 0.744 | 0.935 |

surprising as the system is likely to classify this type of arrhythmia as bad quality ($Sp$ not being closer to 0 might be explained by the 10 sec window involving mixed rhythm). Thus, asystole should be considered as a particular case. Further analysis of the outcomes for sinus bradycardia showed that the statistically based SQIs were the main factors resulting in failure to improve sinus bradycardia classification. The omission of pcaSQI increased accuracy for this arrhythmia type by 10%. Furthermore, removing kSQI and sSQI increased the accuracy by another 18% getting close to 100%. This strongly suggests two primary considerations for classifying arrhythmia: first, arrhythmia specific models are necessary since certain arrhythmias may behave very like noisy data according to certain SQIs. For example, sinus bradycardia was not expected to be classified well as the SQI behaviour is distinctly different in sinus bradycardia than for other rhythms[2]. Second, enough training data for the rhythms

---

[2]It would be interesting, in further experiments, to add a normalisation factor which would correspond to the number of peaks detected. This should diminish the effect of high/low rhythms on the evaluated statistics,

of interest would allow SQIs based on statistics, such as kSQI and sSQI, to adequately capture the statistical properties of the different types of arrhythmias. The classifier is then able to learn the distinction between good and bad quality data using the captured statistical information.

These observations underline the intrinsic limitation of the Challenge 2011 [100]; the Challenge dataset was limited as it was acquired from healthy subjects, and thus it did not include a wide range of pathological cardiac conditions. The results of this study motivated an upgraded framework for constructing a machine learning-based model for ECG quality assessment. The set of SQIs used should be rhythm dependent, and the applied model (a model is defined as a trained classifier for a given set of features) should be switched according to a rhythm call. Figure 5.7 shows the logic of such a model; the monitoring system identifies a rhythm and the corresponding model is selected to assess the signal quality (Model$_k$ in the figure) and outputs a probability estimate of the signal quality. Building such a system involves the following steps:

1. For each type of abnormal rhythm, 10 sec segments (no mixed rhythms) from all lead configurations are selected.

2. Experts annotate the quality of the signals and only keep good and bad quality ones in order to build the model. These labels constitute the gold standard.

3. Balance the classes using the approach introduced in the methods section 5.2 (this is because the dataset is likely to have a high prevalence of good quality signals in a given dataset) and divide into training and test sets.

4. For a given type of rhythm, train the classifier on the training set. This step includes selecting the set of SQIs to use for a given rhythm, which for a low number of SQIs can be performed by an exhaustive search.

5. Test validity of model on the test set of data and report the classifier performances.

---

particularly for sSQI and kSQI.

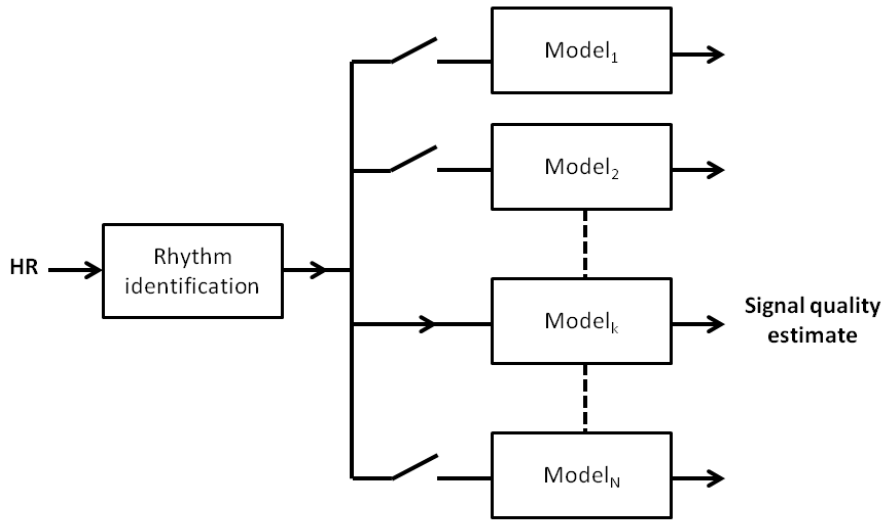

Figure 5.7: Model representation; the classifier model is chosen with respect to the identified rhythm and outputs a signal quality estimate.

The main limitation in building such a model is the large and diverse amount of manually labelled data that is required from a large and varied number of patients and rhythms (to account for the population variability). It should be noted that different preprocessing steps were applied to each database before being made available to the authors and the public. This may alter performance of the quality metrics. Moreover, many of the arrhythmias were under-represented even in the large databases used, and so further data and annotation is required.

Better results are also expected by improving the gold standard labels. This work was based on signal quality annotations from one annotator that attributed a quality label to the ECG signals. However, these annotations are likely to suffer from an expert's individual bias (even with strict guidelines and training).

In practice, bSQI showed the best individual results and was the most robust to arrhythmic rhythms. To strengthen the conclusion of the experimental results from this chapter, bSQI was used during the Physionet/Computing in Cardiology Challenge 2014 [115]. The Challenge 2014 addressed the problem of robust detection of heart beats in multimodal data, and bSQI was used as the indicator to switch from the ECG channel to the arterial blood pressure signal in instances where the quality of the ECG was low. Thanks to bSQI we finished first of the 2014 Challenge for phase III (the most challenging phase of the Challenge)[3]. The test set of the Challenge 2014 was constituted of 300, 10 min multimodal records. See Appendix E for more details on the work performed for this Physionet Challenge.

## 5.5 Conclusion

A total number of 45086, 10 sec ECG segments with quality reference annotations, including normal and pathological data were used to evaluate the performance of seven signal quality indices. It was shown that the SQIs were failing for some arrhythmias. This is because the number of training set instances would need to be much higher for the under represented arrhythmic rhythms in order to properly train a machine learning classifier. As a consequence, a more accurate and complex model that makes use of a set of features and a machine learning framework could be beneficial (as it was shown on normal sinus rhythms) but a large databases of annotated rhythms and signal quality is necessary for training such a classifier properly.

Until such a database is created, the best performing SQI for both normal and abnormal rhythms was found to be bSQI, and it was successfully used for the 2014 Challenge where we finished first of Phase III of the competition. bSQI was also used in Chapter 6 of this thesis.

---

[3]As a note to the reader: the code for bSQI used during the Physionet Challenge 2014 is available on Physionet.org. I would recommend to any interested reader to consider the following items when working on signal quality: bSQI outcomes will depend on what two QRS detectors you choose. If you choose two good QRS detectors then it means that you want to be more conservative. It is also meaningful to choose two QRS detectors that are based on different QRS detection methods (say a Pan and Tompkins and a wavelet detector for example) or that do not look for the same features on the QRS complex (e.g. R-peak and Q-onset). Finally, looking at some example records of the Challenge 2014 where bSQI failed, it was observed that in the case of partial saturation of the ECG (i.e. over say a period of 10 sec the ECG saturated every so often) then the two QRS detectors could end up picking the R-peaks that are not saturated and ignore the saturation parts of the ECG which would results in a bSQI of one (i.e. good quality signal). In order to tackle this problem I would highly recommend using a saturation SQI (that flags saturated segments or flat lines) as a complement to bSQI. Note that these considerations are very much application dependant. (You would not want to label an asystole from an ICU patient as bad quality when using a flat line quality index!)

# Part III

# New methods for NI-FECG extraction

# Chapter 6

# Fetal QRS detection with maternal chest reference

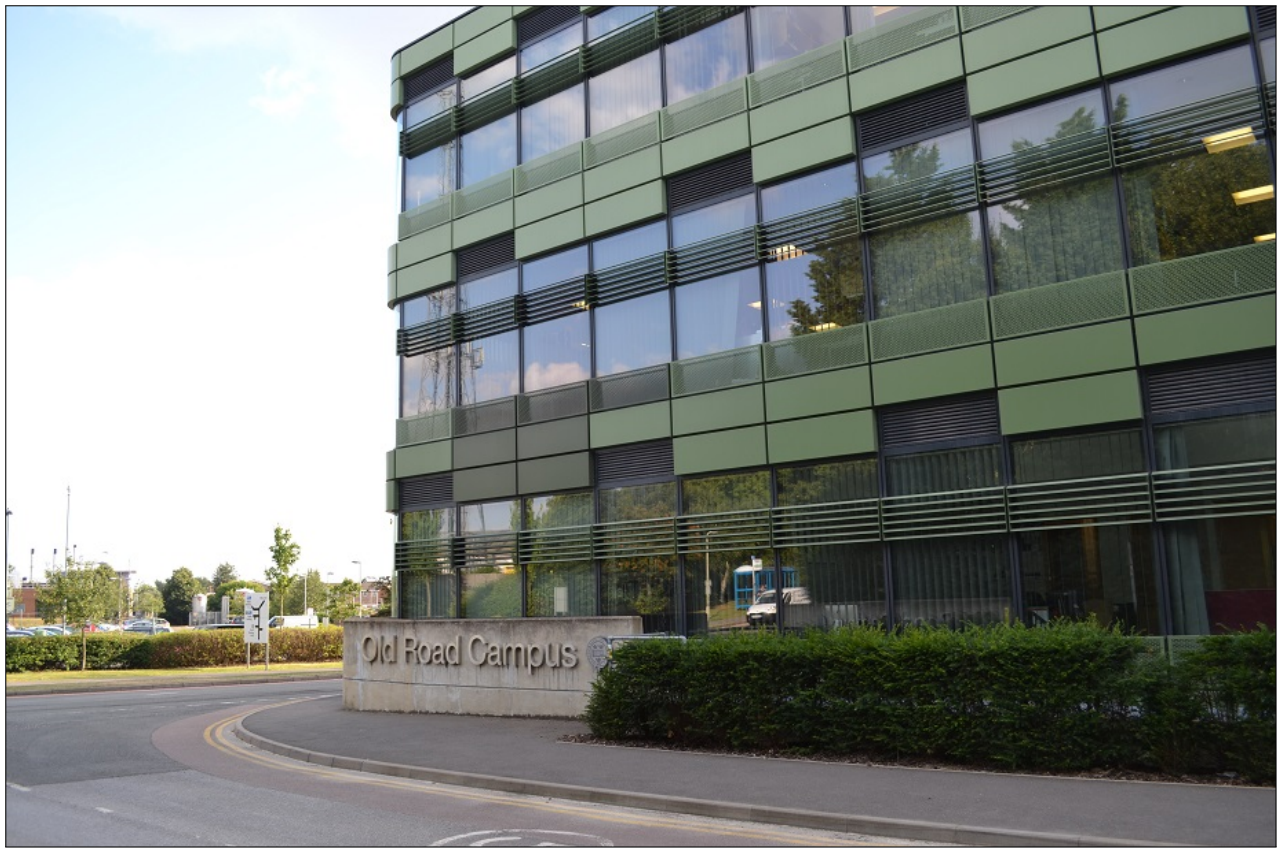

Figure 6.1: The Institute of Biomedical Engineering (IBME) is located at Old Road Campus, and is the host of over 100 academics and PhD students working in the field of biomedical engineering. It is within this green coloured facility that all the experiments started.

## 6.1 Introduction

This chapter compares adaptive temporal methods for extracting the NI-FECG and estimating heart rate. The adaptive techniques all use a reference electrode placed on the maternal chest. A particular focus is given to how the various parameters of the techniques are tuned (on a training set) in order to achieve optimal performances. A novel method for MECG cancellation using an echo state neural network (ESN) CRDB approach was introduced and compared with the least mean square (LMS), the recursive least square (RLS) adaptive filter and template subtraction (TS) techniques (used as baseline). These algorithms are described in Chapter 3, apart from the ESN which is introduced in this chapter. The motivation for using these methods is the adaptivity they provide in coping with non-stationarities such as the changing maternal beat morphology and noise. Such methods are also based on a limited number of channels, as a minimum of one abdominal and one chest channel are required. This last point (one abdominal channel) is motivated by instances where the number of channels available on the monitoring device is limited (which would be the case for portable monitors).

## 6.2 Methods

### 6.2.1 Database and statistics

The databases used in this work were presented in detail in Chapter 4, and are denoted $CRDB_1$ and $CRDB_2$ for the training and test set respectively. Data from $CRDB_1$ and $CRDB_2$ were downsampled to 250 Hz with an anti-aliasing filter prior to running extraction algorithms and tuning their parameters. The statistics reported are $Se$, $PPV$, $F_1$ and $HR_m$ (see definition in Chapter 4.4).

### 6.2.2 The echo state neural network

Recurrent neural networks (RNN) are a class of neural networks capable of non-linear modelling of dynamical systems. This is made possible by recurrent connections between neurons (or just the presence of multiple hidden layers), visualised as cycles in the network topology, that allow processing of temporal dependencies [116]. However, parameter estimation of the RNNs has proven to be a difficult task. Indeed, optimisation methods that were originally used for training feedforward neural networks, such as the error backpropagation algorithm, do not generally perform as well for training RNNs [117]. The ESN [118] is a recently introduced approach to RNN training, with the RNN (also called 'reservoir') being generated randomly (rather than the weights of the reservoir being optimised as in classical RNN approaches). The reservoir forms the dynamical system which will map the input of the system to a higher dimension. The reservoir is fixed, and only the weights of the output neurons are learnt and updated using online or offline linear regression. This method outperformed classic fully trained RNNs in many tasks [117]. The ESN is introduced in the context of noise cancelling as a non-linear medium into which the reference signals propagate before the "echo response", given by the network reservoir, is weighted by a readout layer.

In the configuration presented here, the MECG recorded on the chest is projected onto a set of non-orthogonal basis functions through the ESN reservoir, comparable to the kernel in kernel learning approaches. The input signal(s) (chest ECG) drive the nonlinear reservoir resulting in a high-dimensional dynamical "echo response" [116]. The reservoir also acts as a memory of the input signal thus providing temporal context [119]. In a second step, an adaptation algorithm is used to compute the weights of the output neurons. The RLS algorithm was used for this step. This readout layer maps the reservoir states to the output: the observed AECG.

For $K$ input units, $M$ internal units and $L$ output units: $\underline{\mathbf{u}}(n) = [u_1(n), ..., u_K(n)]$, $\underline{\mathbf{x}}(n) = [x_1(n), ..., x_M(n)]$, $\underline{\hat{\boldsymbol{\eta}}}(n) = [\hat{\eta}_1(n), ..., \hat{\eta}_L(n)]$, where $\underline{\mathbf{x}}(n)$ is the reservoir state vector, $\underline{\mathbf{u}}(n)$ is the vector of input signals and $\underline{\hat{\boldsymbol{\eta}}}(n)$ is the vector of output signals. We define the extended system state as $\underline{\mathbf{z}}(n) = [\underline{\mathbf{x}}(n) \ \underline{\mathbf{u}}(n)]$. The activation of internal units is updated using the following equation:

$$\underline{\mathbf{x}}(n+1) = \underline{\mathbf{x}}(n) + f(\mathbf{W}\underline{\mathbf{x}}(n) + \mathbf{W}_i \underline{\mathbf{u}}(n+1) + \mathbf{W}_b \underline{\hat{\boldsymbol{\eta}}}(n)), \qquad (6.1)$$

where $\mathbf{W} \in \Re^{M \times M}$ is the reservoir weight matrix and $\mathbf{W}_i \in \Re^{M \times K}$ is the input weight matrix (both randomly generated and fixed). $\mathbf{W}_b \in \Re^{M \times L}$ is the back projection weight matrix and $f$ is the reservoir neuron activation function, taken to be the hyperbolic tangent[1]. Considering a purely input-driven dynamical pattern recognition task, the system is simplified by setting $\mathbf{W}_b = 0$. The output is computed as:

$$\hat{\underline{\eta}}(n+1) = g(\underline{\mathbf{w}}_o(n)\underline{\mathbf{z}}(n)), \qquad (6.2)$$

where $g$ is the output neuron activation function (taken to be identity) and $\underline{\mathbf{w}}_o$ are the output weights which may be adaptive or fixed. An ESN with a leaky integrator neuron model [120] was used. This satisfies:

$$\underline{\mathbf{x}}(n+1) = (1-\alpha)\underline{\mathbf{x}}(n) + f(\mathbf{W}\underline{\mathbf{x}}(n) + \mathbf{W}_i\underline{\mathbf{u}}(n+1)), \qquad (6.3)$$

with $\alpha \in [0\ 1]$ being the leakage rate or forgetting factor. In the case $\alpha = 1$, the neurons do not retain any information about their previous state. Initial weights were generated from the uniform distribution on the interval $[-1\ 1]$ for $\mathbf{W}_i$ and $\underline{\mathbf{w}}_o$. $\mathbf{W} \in \Re^{N \times N}$ is a random $M \times M$ sparse matrix with approximately $\psi \times M \times M$ uniformly distributed non-zero entries ($\psi \in [0\ 1]$ is the sparsity of the reservoir). Figure 6.2 shows a representation of the ESN based FECG extraction algorithm. In the figure, one chest signals, $u(n)$, is used as the input of the ESN and the abdominal channel is used as the target signal. The reservoir and input weights are randomly initialised once, and the generated network is used independently for each abdominal channel available for each record. The predicted signal, $\hat{\eta}(n)$, is then subtracted from the jth abdominal signal, $y_j(n)$, giving the residual signal, $\hat{s}(n)$, containing the FECG. Using the ESN for NI-FECG extraction allows for a non-linear relationship between the maternal waveform recorded on the chest and recorded on the abdomen.

There are a number of global parameters that have an important influence on the algorithm's performance. The main three global parameters of the ESN are [119]: (1) input scaling $\gamma$, (2) the spectral radius $\rho$; (3) the leakage rate. The spectral radius determines how fast the influence of an input disappears in the reservoir with respect to time: choosing a small $\rho$ means that the output is more dependent on recent history [120]. It also affects how stable the reservoir activations are (see Equation 6.3). $\mathbf{W}$ is first rescaled by $\rho(\mathbf{W})$ (i.e. its dominant eigenvalue), and therefore has unit spectral radius. Next $\mathbf{W}$ is scaled by $\rho$. Input scaling determines the degree of non-linearity of the reservoir responses. Normalising the input signal so that it lies in the range [-1 1] plays a similar role as scaling $\mathbf{W}_i$ (see Equation 6.3). Table 6.1 summarises the ESN parameters.

---

[1]The hyperbolic tangent corresponds to the ratio between the hyperbolic sine and the cosine functions, $tanh = \frac{sinh(z)}{cosh(z)} = \frac{e^z - e^{-z}}{e^z + e^{-z}}$. It is a strictly increasing bijection from $\Re$ to $]-1, 1[$.

The weights can be allowed to evolve (adaptive filtering) or can be fixed (initialised on a sub-segment of the signal and kept constant). Both options were explored for the ESN approach; these will be denoted $\text{ESN}_a$ for the adaptive (i.e. weights are updated online) approach and $\text{ESN}_{na}$ for the nonadaptive (i.e. weights are determined on some initial training data and kept constant) approache. Online updating of the filter coefficients can then be assessed to see whether filter performance is improved.

### 6.2.3 Preprocessing and general experimental set up

For each of the available abdominal channels, baseline wander was first removed. In the context of NI-FECG extraction, it is common to use a larger than standard (for adult ECG filtering) low frequency cut-off before performing MECG cancellation (see [54, 61] for example), where larger is approximately 2 Hz. This cut-off is acceptable when the extraction is not aiming at ECG morphological analaysis. Yet, there is no comprehensive study that has assessed the prefiltering effect on outcomes for FQRS extraction. An exhaustive search was therefore conducted to determine $f_b$ (an 'optimal' cut-off frequency to remove the baseline using a high pass filter) and $f_h$ (an 'optimal' cut-off frequency to remove the high frequency content using a low pass filter). Two zero phase Butterworth digital filters were cascaded for that purpose: one 3rd order high pass filter and one 5th order low pass filter. The reference and input signals were then normalised according to the following procedure: (1) the first 5 sec of the signal were used to derive the amplitude range of the ECG signal, and the signal was then divided by this amplitude; (2) the mean was computed over the first 5 sec and subtracted from the ECG signal, and (3) the resulting signal was transformed using the hyperbolic tangent function. Step (3) was applied in order to avoid outliers which could result in the reservoir state $\underline{\mathbf{x}}(n)$ or LMS/RLS weights $\underline{\mathbf{w}}(n)$ taking unexpected values due to abnormally large values (likely attributable to signal artifact). Failing to perform this last step could result in loss of useful memory or a highly unpredictable output [119]. Following the preprocessing, each algorithm was used to filter out the MECG, resulting in a residual signal comprised of the FECG and some noise. FQRS detection was performed on the residual signal, $\hat{s}$, using a P&T QRS detector with a refractory period of 150 ms. All the parameter optimisation was performed on $\text{CRDB}_1$ while $\text{CRDB}_2$ was used as the independent test set. The $Se$, $PPV$, $F_1$ and $HR_m$ of each algorithm was evaluated. These statistical measures were described in Chapter 4. The NI-FECG extraction algorithms are comprised of the adaptive filters (LMS, RLS and ESN) that uses the reference chest ECG, as well as some baseline TS techniques for benchmarking purposes.

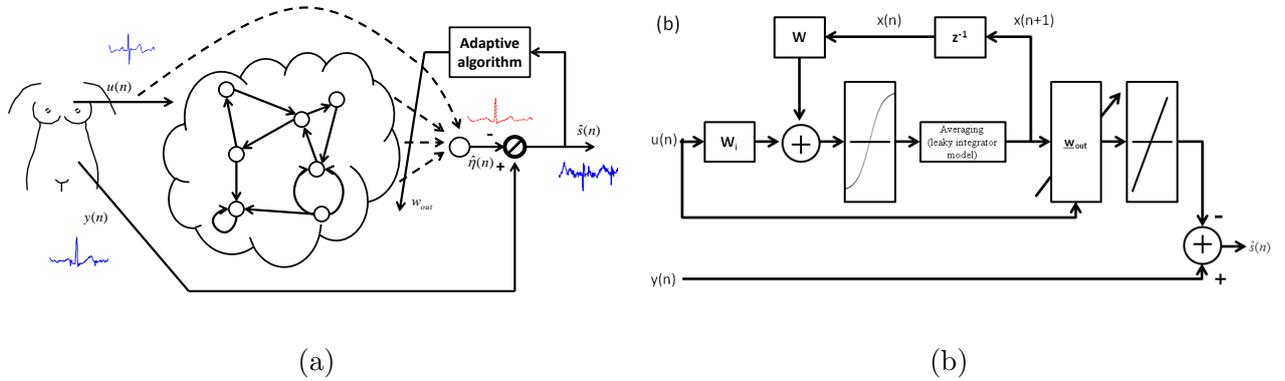

(a)                                                 (b)

Figure 6.2: The ESN based NI-FECG extraction algorithm showing the relationship between the chest signal $u$, predicted signal $\hat{\eta}(n)$, the jth abdominal ECG signal $y(n)$, and the residual signal $\hat{s}(n)$ containing the FECG. Dashed lines represent adaptive weights. Image of the women is adapted from [61]. (a) general overview; (b) process block diagram.

Table 6.1: Parameters of the ESN, LMS and RLS algorithms. † indicates the parameters that were not optimised.

|  | Symbol | Definition | Search range (step size) |
|---|---|---|---|
| ESN | $f_h$ | high frequency cut-off | [30, 120] (5) |
|  | $f_b$ | low frequency cut-off | [1, 49] (3) |
|  | $M$ | size/number of units of the reservoir | [10, 250] (20) |
|  | $\psi$ | sparsity of the reservoir. Percentage of connections between nodes of the reservoir | 20% † |
|  | $\rho$ | spectral radius of $W$, the reservoir connection matrix | [0, 1] (0.05) |
|  | $\gamma$ | input scaling of the input weight matrix $W_{in}$ | 1 † |
|  | $\alpha$ | leakage rate | [0, 1] (0.1) |
|  | $\lambda$ | forgetting factor of the RLS algorithm | 0.999 |
| LMS | $N$ | filter length | [1, 262] (5) |
|  | $\mu$ | step size | [0.01, 0.46] (0.05) |
| RLS | $N$ | filter length | [1, 61] (3) |
|  | $\lambda$ | forgetting factor | [0.8, 1] (0.0025) |

**TS techniques** MQRS detection was run on the chest channel because of its higher SNR and negligible FECG contribution. Each MQRS was then adjusted so that the R-peaks could be accurately located on each abdominal channel. This ensured a well constructed channel-specific MECG template, and good alignment between the ECG cycles and the template MECG.

**LMS technique** An exhaustive search over a range of values of $N \in [1 \ \ 262]$ and $\mu \in [0.01 \ \ 0.46]$ was performed. See Table 6.1.

**RLS technique** An exhaustive search over a range of values of $N \in [1 \ \ 61])$ and $\lambda \in [0.8 \ \ 1]$ was performed. See Table 6.1.

**ESN technique** The optimisation of the ESN parameters was performed using a stage-wise grid search as follows: (1) a standard prefiltering range using $f_b = 2$ Hz and $f_h = 100$ Hz was used to search for workable $\rho$ and $\alpha$ values, (2) given $\rho$ and $\alpha$, a search was performed over a range of preprocessing cut-off frequencies to find optimal $f_b$ and $f_h$, (3) using the optimised $f_b$ and $f_h$, the search for optimal values of $\rho$ and $\alpha$ was repeated, and finally (4) the number of reservoir neurons ($M$) was optimised as measured by the $F_1$ score with the parameters selected in (1-3). Note that (1-3) were performed with a high number of reservoir neurons ($M = 250$). In the case of adaptive filtering, an RLS algorithm was used to update the weights of the readout layer, which maps the reservoir states to the output of the observed AECG. In the case of non-adaptive filtering, both the Wiener-Hopf solution (corresponding to the direct pseudo-inverse) and ridge regression were used to determine the weights on a 30 sec signal epoch preceding the studied segments. When varying the number of neurons, $M$, the experiment was repeated 20 times in order to ensure that the variance of the $F_1$ measure, caused by the random initialisation of the ESN reservoir connections and weights, was negligible.

One chest channel was taken as the reference signal for the LMS, RLS and the ESN. Filter weights were initialised on 30 sec preceding the annotated data for each record (i.e. preceding the 1 min of $CRDB_1$ and the 5 min of $CRDB_2$).

### 6.2.4 Adding signal quality indices

In Chapter 5 different SQIs to identify bad quality ECG signals were evaluated. The most accurate SQI evaluates the agreement between two QRS detectors with different robustness to noise. This metric, termed bSQI $\in [0\ 1]$ (with 1 representing a good quality signal), was difficult to apply to the abdominal signal, as the MQRS and also some FQRS could be detected by the QRS detectors (although not systematically due to the refractory period of the detectors) in the case of high amplitude FECG traces. This makes bSQI a weak quality indicator if applied to the abdominal signal. Therefore, bSQI was only used on the chest channel (where there is no FECG contribution) with a 10 sec window and 9 sec overlap. This provided a second by second SQI for the chest channel. When bSQI was less than 0.8 the corresponding abdominal segments were removed.

## 6.3 Results

### 6.3.1 Parameter optimisation

**TS** For the TS class of techniques, the $TS_{pca}$ method gave the best results on the training set (see Table 6.2), thus only results for $TS_{pca}$ were reported in Table 6.3 for the performance on the individual records of $CRDB_2$.

**LMS** The parameters $\mu$ and $N$ of the adaptive LMS adaptive algorithm were searched (Figure 6.4). Selected parameters based on the grid search were $\mu = 0.1$ and $N = 20$, with corresponding performance statistics $Se = 95.8\%$, $PPV = 95.0\%$ and $F_1 = 95.4\%$ (Table 6.2).

Table 6.2: Overall statistics on CRDB$_1$ (training DB) and CRDB$_2$ (test DB). Best results in term of $F_1$ and $HR_m$ are underlined. 'a' stands for adaptive filtering and 'na' stands for non adaptive filtering. In the case of offline training the weights are determined on the 30 sec of signal preceding the studied ECG segment (for both CRDB$_1$ and CRDB$_2$).

|  | $Se(\%)$ | $PPV(\%)$ | $F1(\%)$ | $HR_m(\%)$ |
|---|---|---|---|---|
| TS$_c$-CRDB$_1$ | 91.2 | 90.5 | 90.8 | 57.9 |
| TS$_m$-CRDB$_1$ | 90.3 | 90.0 | 90.1 | 56.7 |
| TS$_{lp}$-CRDB$_1$ | 91.2 | 92.8 | 92.0 | 59.7 |
| TS$_{pca}$-CRDB$_1$ | 94.7 | 96.0 | 95.4 | 68.7 |
| LMS-CRDB$_1$ | 95.8 | 95.0 | 95.4 | 69.3 |
| RLS-CRDB$_1$ | 96.2 | 95.6 | 95.9 | 70.6 |
| ESN$_{na}$-CRDB$_1$ | 97.2 | 97.3 | <u>97.2</u> | 73.5 |
| ESN$_a$-CRDB$_1$ | 96.8 | 97.2 | 97.0 | 72.2 |
| TS$_c$-CRDB$_2$ | 86.8 | 85.3 | 86.0 | 68.1 |
| TS$_m$-CRDB$_2$ | 86.4 | 85.2 | 85.8 | 67.2 |
| TS$_{lp}$-CRDB$_2$ | 85.9 | 85.5 | 85.7 | 66.7 |
| TS$_{pca}$-CRDB$_2$ | 89.9 | 88.8 | 89.3 | 73.9 |
| LMS-CRDB$_2$ | 89.3 | 86.5 | 87.9 | 73.7 |
| RLS-CRDB$_2$ | 89.7 | 86.8 | 88.2 | 74.5 |
| ESN$_{na}$-CRDB$_2$ | 89.3 | 86.5 | 87.9 | 73.9 |
| ESN$_a$-CRDB$_2$ | 91.4 | 88.9 | <u>90.2</u> | 78.7 |

**RLS** The parameters $\lambda$ and $N$ of the RLS adaptive algorithm were searched (Figure 6.4). Parameter values selected based on the grid search were $\lambda = 0.999$ and $N = 20$, with corresponding performance statistics $Se = 96.2\%$, $PPV = 95.6\%$ and $F_1 = 95.9\%$ (Table 6.2).

**ESN** Figure 6.4 illustrates the exhaustive search results obtained on the training database CRDB$_1$ using the direct pseudo-inverse method for computing $\underline{\mathbf{w}}$. Based on this optimisation step, the preprocessing and ESN parameters were determined to be $f_b = 20$ Hz, $f_h = 95$ Hz, $\alpha = 0.4$, $\rho = 0.4$ and $M = 90$. The best results on the training set using the above mentioned parameters were $Se = 97.2\%$, $PPV = 97.3\%$ and $F_1 = 97.2\%$ (See Table 6.2). The same preprocessing parameters ($f_b = 20$ Hz, $f_h = 95$ Hz) were used for the other techniques (i.e. TS, LMS and RLS).

Table 6.3: Per record statistics on CRDB$_2$ (test DB). Best results in terms of the $F_1$ measure are underlined. Rec: record.

|     | TS$_{pca}$ | | LMS | | RLS | | ESN$_a$ | |
| --- | --- | --- | --- | --- | --- | --- | --- | --- |
| Rec | $F_1$ | $HR_m$ | $F_1$ | $HR_m$ | $F_1$ | $HR_m$ | $F_1$ | $HR_m$ |
| 123a | 84.1 | 51.7 | 83.4 | 63.8 | 83.3 | 51.7 | <u>84.3</u> | 65.0 |
| 123b | 95.9 | 88.4 | <u>96.3</u> | 92.4 | <u>96.3</u> | 92.5 | 95.7 | 90.2 |
| 125 | 87.2 | 71.6 | 78.7 | 54.5 | 80.7 | 59.3 | <u>88.2</u> | 75.6 |
| 126 | 86.2 | 69.4 | 88.4 | 75.8 | 88.2 | 76.1 | <u>88.9</u> | 76.5 |
| 172a | 98.1 | 90.2 | 99.1 | 96.4 | <u>99.3</u> | 97.8 | 99.2 | 97.3 |
| 172b | 94.1 | 85.4 | 94.1 | 88.7 | 94.5 | 89.9 | <u>94.7</u> | 90.7 |
| 210 | 86.2 | 63.4 | 84.2 | 51.2 | 85.4 | 62.0 | <u>88.9</u> | 70.8 |
| 261 | <u>77.2</u> | 58.7 | 73.6 | 53.1 | 73.5 | 52.9 | 75.9 | 56.4 |
| 265 | 77.4 | 54.5 | 70.7 | 42.9 | 70.5 | 43.0 | <u>78.6</u> | 56.7 |
| 299a | 98.0 | 89.0 | 98.9 | 95.3 | <u>99.0</u> | 96.8 | 98.0 | 93.8 |
| 299b | 97.4 | 90.5 | <u>98.8</u> | 96.7 | <u>98.8</u> | 97.2 | 98.0 | 93.2 |

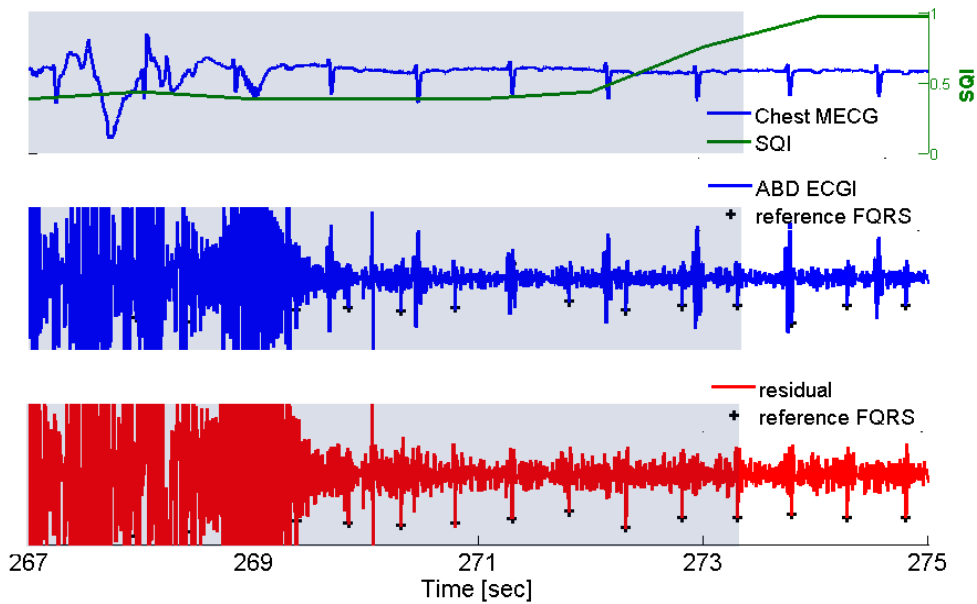

Figure 6.3: Signal quality for identifying bad quality regions from the maternal chest ECG (record 123a). From top to bottom: chest MECG, ABD ECG and residual after performing MECG cancellation using the ESN. In the shaded region is the part of the signal that is discarded because of its low quality (SQI ≤ 0.8 - window size for assessing quality is 10 sec with 9 sec overlap). Note that the SQI takes some time to get back to one because of the edge effect (10 sec window).

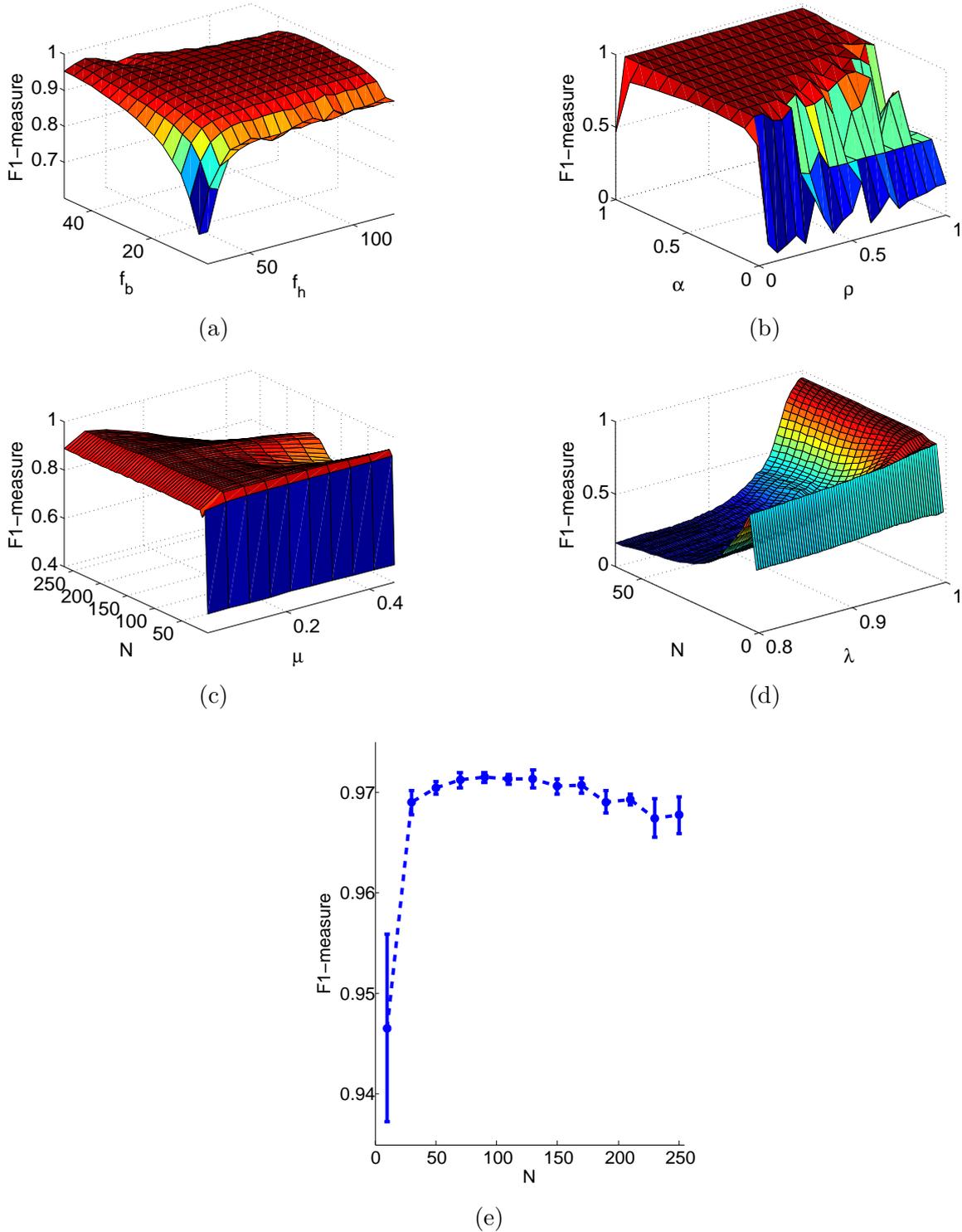

Figure 6.4: Search for the preprocessing parameters ($f_b$, $f_h$) the ESN specific parameters ($\alpha$, $\rho$) and the LMS parameters ($\mu$, $N$) on the training set database $CRDB_1$. (a) search for ESN preprocessing parameters ($\alpha = 0.4$, $\rho = 0.4$, $M = 90$), (b) search for ESN parameters $\alpha$ and $\rho$ ($f_b = 20$ Hz, $f_h = 95$ Hz, $M = 90$), (c) search RLS parameters $\lambda$ and $N$ ($f_b = 20$ Hz, $f_h = 95$ Hz), (d) search LMS parameters $\mu$ and $N$ ($f_b = 20$ Hz, $f_h = 95$ Hz), (e) the number of ESN neurons required ($f_b = 20$ Hz, $f_h = 95$ Hz, $\alpha = 0.4$, $\rho = 0.4$), repeated 20 times for each value of N to find the variance of the $F_1$-measure caused by the random initialisation of the ESN reservoir connections and weights).

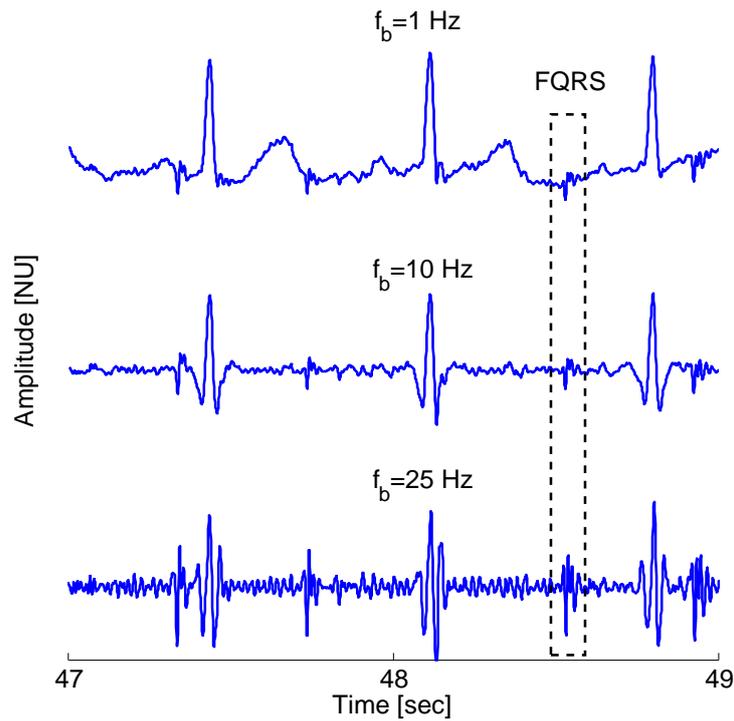

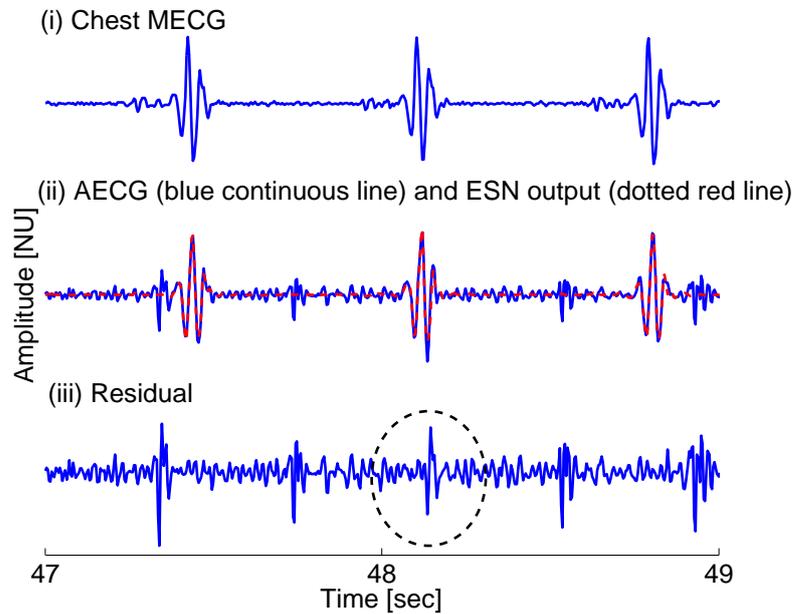

Figure 6.5: (a) Effect of varying $f_b$ (r154, $CRDB_1$, 3rd AECG, $f_h = 110$ Hz) for the preprocessing step. Note that at 10 Hz most of the frequency content of the P and the T-wave of the MECG have been filtered out, leaving only the FQRS and the MQRS. (b) Example of the ESN algorithm performance with $\alpha = 0.4$, $\rho = 0.4$, $M = 90$ (optimal parameters). The signal (r154, $CRDB_1$, 3rd AECG) was prefiltered using $f_b = 20$ Hz and $f_h = 95$ Hz. Notice in particular the extraction of the FQRS embedded in the MQRS at t=48.2 sec (circled by a broken black line).

Table 6.4: Results on CRDB$_2$ both with and without removing low quality record segments based upon a single SQI.

|  | $F_1$ no SQI (%) | $F_1$ SQI* (%) | $\Delta F_1$ (%) |
|---|---|---|---|
| TS$_{pca}$ | 89.32 | 90.39 | +1.07 |
| LMS | 87.88 | 88.83 | +0.95 |
| RLS | 88.23 | 89.27 | +1.04 |
| ESN$_a$ | 90.17 | <u>91.28</u> | +1.11 |

* with 3.6% of the overall signals being removed by bSQI

### 6.3.2 Results on training and test sets

The results in Figure 6.5 were produced with $f_b = 20$ Hz, $f_h = 95$ Hz. Figure 6.5b shows qualitative examples of the ESN algorithms performance on record 154 in CRDB$_1$ with optimal parameters. Table 6.2 summarises the results obtained on the training set CRDB$_1$. It can be seen that ESN based techniques performed better than the TS and adaptive filter techniques. Both ESN techniques gave similar results, with $F_1$ scores of 97.2% for ESN$_{na}$ and 97.0% for ESN$_a$. Among the TS techniques, TS$_{pca}$ gave $F_1 = 95.4\%$ and outperformed all other TS techniques. Finally, among the adaptive filter techniques the LMS algorithm ($F_1$ of 95.4%) was slightly outperformed by the RLS algorithm ($F_1$ of 95.9%). Table 6.2 also summarises the results obtained on the test set CRDB$_2$. The adaptive ESN (ESN$_a$) gave the best $F_1$ score of 90.2%, improving upon the TS$_{pca}$ technique (89.3%) and the RLS technique (88.2%). Table 6.3 presents the results for each record from CRDB$_2$ for the optimal ESN, LMS, RLS and TS$_{pca}$ algorithms as determined on the training set CRDB$_1$. Note that in Table 6.3 the performance of each algorithm was averaged over all individual channels of a given record. The best result in each case is underlined.

### 6.3.3 Results with signal quality indices

Table 6.4 presents the results for the test database in terms of $F_1$. Including the signal quality indices improved the results on CRDB$_2$ by 1.07% for the TS$_{pca}$ approach, 0.95% for the LMS approach, 1.04% for the RLS approach and 1.11% for the ESN approach, while excluding 3.6% of the overall signals' length. See Figure 6.3 for an example of a CRDB$_2$ excluded on the basis of low bSQI.

## 6.4 Discussion

The focus of this study was on assessing the accuracy of the ESN, RLS, LMS and TS techniques in extracting the FECG from the AECG, and facilitating robust FQRS detection. Better results may be achieved with an alternative QRS detector but the ordering of the algorithms is unlikely to change.

In practice, the FHR time series are smoothed prior to being displayed on the clinical monitor. This smoothing operation removes sudden drops or increases of FHR due to missed beat detections. Higher performances would be expected on $CRDB_1$ and $CRDB_2$, in terms of the $FH_m$ if this smoothing operation were applied (see Chapter 7).

A direct comparison between the various time-based techniques (that operate in the observation domain) can be made when studying the performance of the algorithms on individual channels without smoothing the extracted FQRS time series. Blind source separation techniques, such as ICA and PCA, were not considered in this work as its focus was on time based techniques using a single abdominal channel. However, in a system with multiple abdominal channels, any of the presented methods could be used to remove the MECG contribution to the different channels independently, before performing a BSS step (this is further explored in Chapter 7). In addition, the results in Table 6.3 show that no algorithm consistently outperformed the others. This suggests that the overall results could be improved by combining different approaches for NI-FECG extraction. This is examined in Chapter 7.

The importance of the signal preprocessing, particularly the choice of baseline wander cut-off frequency, was studied. Figure 6.5 shows an example of a segment (r154, $CRDB_1$, 3rd AECG) which was preprocessed with various $f_b$ values. Note that at 10 Hz most of the frequency content of the P-wave and the T-wave of the MECG have been filtered out, leaving only the FQRS and MQRS. This higher cut-off showed improvement in performance of both training and test sets for all methods. As an example, the $TS_{lp}$ technique gave $F_1 = 83.0\%$ and $F_1 = 80.0\%$ with $f_b = 2$ Hz, $f_h = 95$ Hz on $CRDB_1$ and $CRDB_2$ respectively and $F_1 = 92.0\%$ and $F_1 = 85.7\%$ with $f_b = 20$ Hz, $f_h = 95$ Hz on $CRDB_1$ and $CRDB_2$ respectively. The cut-off selected in this work was inferred from the grid search in Figure 6.4a. Although it was clear that choosing an unconventionally large $f_b$ was beneficial in this case, whether to choose $f_b = 20$ Hz (as in this work) or a lower value inevitably depends on the dataset and filter design considered.

One of the theoretical limitations of adaptive filtering approaches is that any noise contained in the abdominal, but not the chest ECG signal, will not be removed. The assumption behind adaptive noise cancelling is that the noise contaminants on the abdominal channels are also present on the chest channels (considered as the noise field). Nevertheless, the method is well suited for removing the main noise contaminant from the abdominal signal, namely the MECG, thus improving the FQRS detection.

The ESN algorithm requires a minimum of one reference and one abdominal channel. This is an important advantage over the blind source separation techniques which, while very popular for this application, require a relatively high number of channels (usually between 8 and 16 [9]). Furthermore, the ESN does not require any prior information about the MQRS location, unlike the TS or KF approaches. In particular, the KF requires a precise MQRS detection technique, making this technique particularly susceptible to noise in the MECG.

When using the $\text{ESN}_a$ with the traditional RLS implementation, presented in 3.2.2.2, to train the readout layer, a monotonic increase of the weights over time was observed. This led to large weights after a few minutes. When the weights are too large they amplify small differences among the dimensions of $\underline{\mathbf{x}}(n)$, which can in turn lead to instability in the presence of a small deviation from the conditions for which the network had been trained [117]. Therefore, the RLS algorithm introduced by Douglas *et al.* [121], which uses RLS in combination with least squares pre-whitening, was used. This implementation made the $\text{ESN}_a$ stable.

Within the TS techniques, $\text{TS}_{pca}$ performed the best on the training and test sets, and was certainly the most adaptive. Within the other methods (LMS, RLS, ESN), the RLS approach was found to perform slightly better than the LMS one, and the ESN approach better than the RLS one. This suggests that the MECG cycle was better removed with more adaptive algorithms.

It is important to note that the TS techniques include discontinuities due to the piece-wise template. The MECG template cycle was built to be centred on the maternal R-peak, using a duration of 0.20 sec, 0.10 sec and 0.40 sec for the P, QRS and T waves respectively [54]. The choice of each ECG segment's time interval is realistic but not optimal. However, the discontinuities mentioned are assumed to be minimal, with limited influence on FQRS detection (as opposed to FECG morphological analysis). Performance could be improved by varying the length of each of these intervals with the maternal heart rate rather than being constant.

The main drawback of the TS, with respect to the LMS and ESN techniques, is that it relies on accurate MQRS detection (the best adults QRS detectors typically achieve about 99% accuracy over a range of diverse databases). Indeed, a missed MQRS detection will most likely result in FPs (except if located in the FQRS refractory period) and possibly FNs (if the actual FQRS is located within the refractory period of the induced FP). Conversely, the main drawback of the LMS, RLS and ESN algorithms is that they are driven by, and subject to, poor signal quality in the chest signal. This particularly motivates the use of SQIs. It should also be noted that the number of coefficients for LMS/RLS adaptive filters, or neurons in the ESN, is a function of the signal sampling frequency. As a consequence, parameter optimisation should be conducted again if a different sampling frequency is to be considered.

Using a SQI on the chest reference channel improved the accuracy measure by approximately 1%, while suppressing approximately 3% of the overall record. It is expected that the added value in using SQI would be higher in the presence of noisier recordings, as the data used in this work were of relatively good quality with some minor local artifacts.

It is to be noted that the number of ESN parameters that need tuning, in addition to the randomness of reservoir initialisation and connectivity (that give no insight into the reservoir dynamic), makes the ESN design and implementation difficult. In an attempt to tackle these problems, Rodan *et al.* [122] compared the performance of the 'standard' ESN framework (the one used in this paper) with much simpler network structures, and showed that similar performance to the standard ESN could be achieved using these *deterministically* constructed reservoirs. An alternative approach is to use random search to find a set of acceptable parameters. This is described in Appendix E where the approach from Bergstra *et al.* [123] was used to optimise and better understand what parameters were playing an important role in the performance of the ESN in the context of NI-FECG extraction. Random search performed for the ESN and preprocessing parameters, showed similar performance to the exhaustive grid search presented in this work, but reduced the number of search iterations. The parameters that were assumed constant in this work, in order to restrict the grid search to an acceptable number of free-parameters (such as $\phi$ and $\gamma$- see their definition in Chapter 3), could be searched if a random search aporoach was used. In practice it is likely that random search would be used as an initialisation method to find which parameters are the most relevant and further parameter tuning would be performed with fine grid search.

The ESN performed slightly better than the LMS and RLS algorithms on the test database, but the results were not significantly different. One of the main differences between the two approaches is that the ESN approach allows for a non-linear relationship between the chest and abdominal ECG, whereas the LMS and RLS only consider a linear relationship. Thus, the results suggest that the mapping is mostly linear (or that the numerous ESN hyper-parameters were over-fitted on the training set database). Recall that the data in the training set ($CRDB_1$) and test set ($CRDB_2$) were recorded with different hardware, following a separate protocol and at different stages of pregnancy for different subjects.

Two records in the test set $CRDB_2$ (123a and 265) exhibited a relatively much higher $F_1$ statistic for the ESN technique compared to the RLS and LMS techniques. This is in contrast to the other records where the discrepancy in performance between the ESN and the RLS/LMS techniques is less pronounced. Investigation of these two records showed that this discrepancy was not due to the quality of the recordings but rather that in these two instances the ESN better managed to remove the MECG. More specifically, each record had four channels; for some channels the FECG amplitude was high enough so that the maternal residual left by the RLS and LMS did not affect the FQRS detection in the residual more than it was for the ESN residual. However, for other channels the maternal residual left when using the LMS and RLS were of similar amplitude to the FECG thus leading to a less accurate FQRS detection. The ESN MECG cancellation was superior even in the instances of low fetal to maternal amplitude ratio. In order to confirm the superior performances of the ESN in these instances it would be useful in future experiments to manually annotate the fetal to maternal amplitude ratio on a larger dataset and evaluate the $F_1$ measure as a function of this annotation to facilitate the comparison between algorithms. Since the LMS algorithm is conceptually simpler and computationally less expensive, it is most likely a more appropriate choice for low-cost devices with limited computational power. However, in the case of a more advanced hospital-based system, where computing power is generally not a concern, then there is more incentive to use the more computationally complex and accurate solution.

## 6.5 Conclusion

This work compared NI-FECG extraction methods. The methods were qualitatively and quantitatively evaluated. In addition, the preprocessing performed for baseline wander and high frequency removal was studied in some detail. These filters, along with various parameters for the LMS, RLS and ESN algorithms, were optimised through exhaustive grid search on a training database. The main findings of this research are: (1) the non-linear $ESN_a$ method performed slightly better than the LMS, RLS and TS methods; (2) using a high baseline wander cut-off frequency improved the performance of the extraction algorithms; (3) SQI improved the performance of all methods by excluding bad quality chest ECG, which would have led to an unexpected adaptive filter response or false QRS detection; (4) no one algorithm systematically outperformed all the others. Thus, there might be improvement by combining different NI-FECG extraction approaches as is explored in the next chapter. In addition, this work proposes a framework for assessing the algorithms performance in extracting the FQRS and FHR by using the $F_1$ and $HR_m$ measures. Future work requires assessing the algorithms on a larger dataset, benchmarking them against additional methods employed for this task and evaluating the alternative ESN reservoir design as for example suggested by Rodan *et al.* [122]. Open source code for the benchmark methods has been made available to allow comparison and reproducibility on the public domain data. The source code is available on Physionet at `http://physionet.org/`

# Chapter 7

# Fetal QRS detection with no maternal chest reference

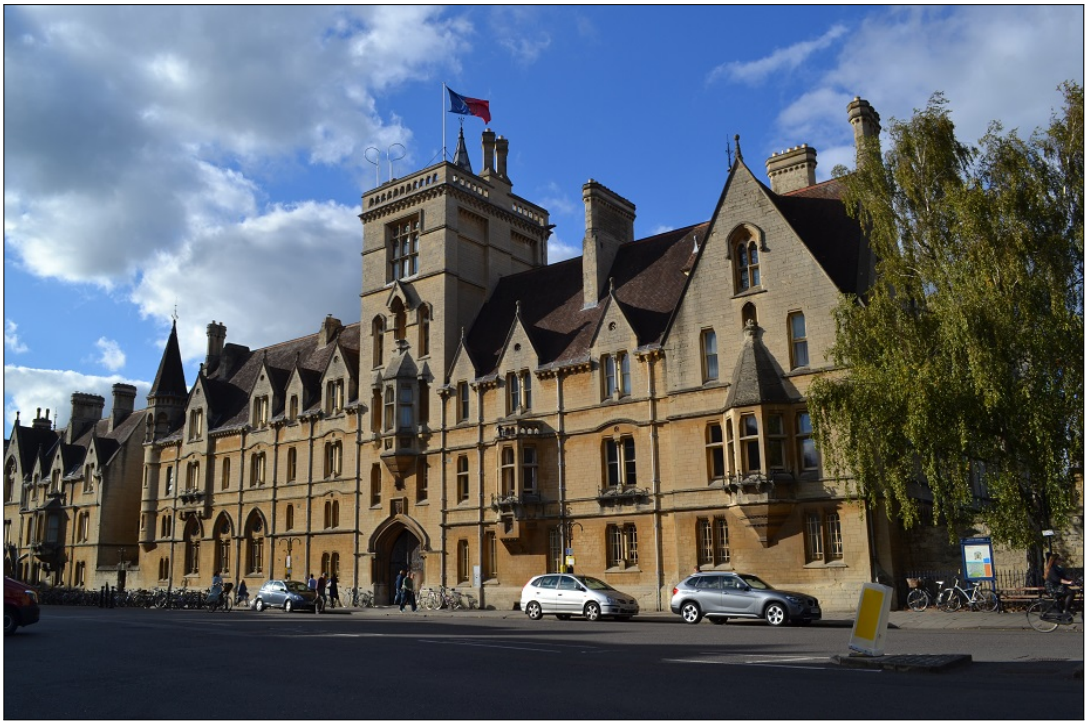

Figure 7.1: Balliol College was founded in 1263 and is one of the 44 colleges and private Halls constituting the university of Oxford. Balliol markets itself as the oldest Oxford college. However, this claim is disputed by University College and Merton. The definition of 'oldest' being variable (time at which students started living in the college? date a first lecture was delivered within the college walls? etc.). In science many publications market their novel method as being the best or better than alternatives. In the same way as this story, the definition of 'better' is variable and the absence of standards (metrics and databases) for evaluating the algorithms is problematic for objective assessment of their respective performances.

## 7.1 Introduction

This chapter presents the approach taken and the results achieved when participating in the Physionet Challenge 2013, addressing the topic of *Non-invasive fetal ECG*. The abdominal electrocardiogram (AECG) signals were first preprocessed with a band-pass filter in order to remove higher frequencies and baseline wander. A notch filter to remove power interferences at 50 Hz or 60 Hz was applied if required. The signals were then normalised before applying various source separation techniques to cancel the maternal ECG (MECG). These techniques included template subtraction, principal/independent component analysis, extended Kalman filter and a combination of a subset of these methods (FUSE method). Fetal QRS detection was performed on all residuals using a Pan and Tompkins QRS detector (See Appendix E.4 for more details on the performance of the QRS detector) and the residual channel with the most regular fetal heart rate time-series was selected (see Section 7.2.5).

Four abdominal channels (with no maternal chest reference) were available. This number was motivated by: (1) the observation that sometimes the signal quality was too low on the chest, and this could greatly (negatively) impact the adaptive filter algorithms such as the one evaluated in Chapter 6; (2) not having a chest electrode is appropriate for certain environments where it is not possible to place electrodes on the chest (i.e. it is culturally inappropriate); (3) if the number of channels available on the monitoring device is limited (which would be the case for portable monitors).

## 7.2 Methods

### 7.2.1 Database and statistics

The database and statistics used for evaluating the source separation algorithms were described in detail in Chapter 4. A total of 447, 1 min signals were available for the Challenge and the statistics reported are: $Se, PPV, F_1$ and $E$1-5 (see description of the statistics in Chapter 4).

## 7.2.2 Overview

Figure 7.2 presents the framework of the approach undertaken in this work for FQRS detection: (1) the four abdominal channels were preprocessed by removing the baseline wander and higher frequency components, and were also notch filtered if required; (2) MQRS detection was performed on each of the prefiltered channels as well as on the ICA transformed channel (to enhance the MECG contribution on some of the components); (3) one of the MQRS time-series was selected as being the reference time-series; (4) a source separation algorithm was applied to the AECG in order to extract the FECG. The MECG free channels after the source separation step were called residuals; (5) FQRS detection was performed on the residual signals containing the FECG; (6) one of the FQRS time-series detected on the residual channels was selected; (7) the RR time-series was smoothed to reduce the effect of missing and extra detected beats, and the extracted FQRS time-series was scored against the reference fiducials. Finally, the RR time-series and corresponding FHR were scored using the Challenge scoring system, and the statistics for evaluating QRS detectors' sensitivity ($Se$), positive predictive value ($PPV$) as well as an $F_1$ measure (see Chapter 4.4.1) were computed.

## 7.2.3 Prefiltering

ECGs were first preprocessed by cascading a low pass and a high pass filter in order to remove higher frequencies and baseline wander. Two zero phase Butterworth digital filters were used for that purpose with order 5 and 3 respectively. A notch filter to remove power interferences at 50 Hz or 60 Hz was then applied if required. This is because the data came from different databases, possibly a mixture of EU and US origin, thus 50 Hz and 60 Hz power-line interference had to be considered. The notch filter was applied if the ECG's power spectrum contained a peak which could likely be attributed to interference from mains power; more specifically, the filter was applied if the peak of the ECG power spectrum in the band $50/60 \pm 4$ Hz was located within $50/60 \pm 1$ Hz. The abdominal signals were then normalised by the amplitude range of the ECG signal in the interval 1-5 sec, subtracting the mean of the signal computed over the same interval and finally taking the hyperbolic tangent of the signal. Applying the hyperbolic tangent was intended to keep the signal in a controlled range, making the FQRS detection more accurate by shrinking the high amplitude artifacts in the ECG signal range while not affecting the centred ECG signal that was in range $[-0.5\ 0.5]$. Normalisation is also required before applying PCA/ICA based methods. Figure 7.3 shows an example of preprocessed AECG signal.

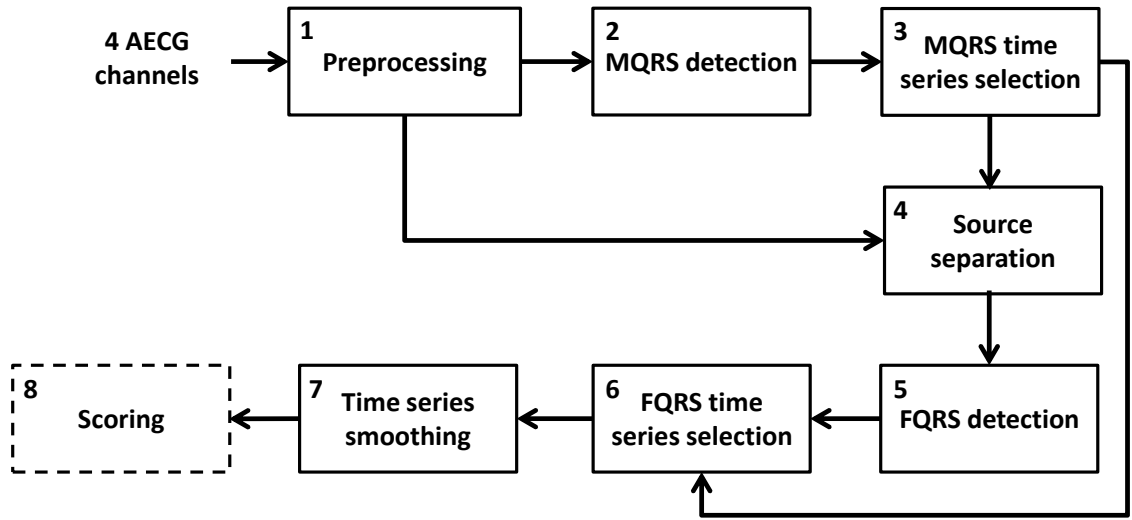

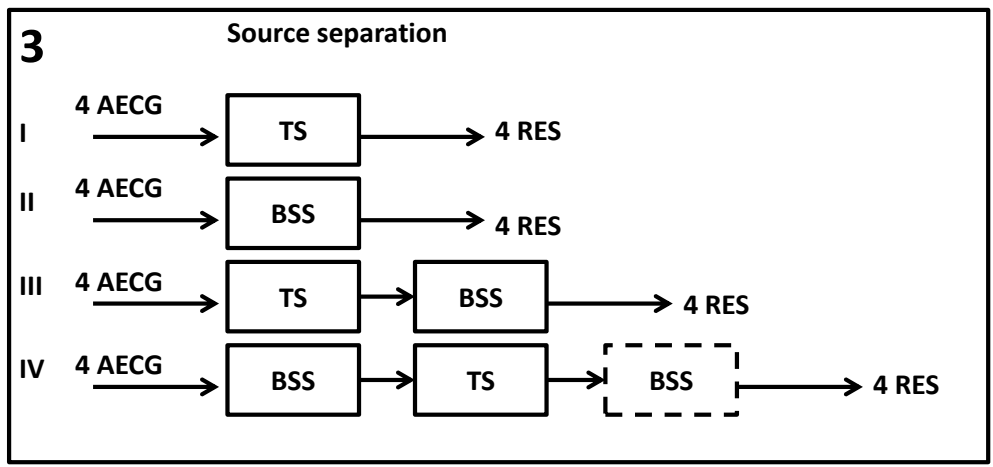

Figure 7.2: (a) NI-FECG extraction block diagram. (1) The four AECG channels are preprocessed; (2) MQRS detection; (3) one of the MQRS time-series is selected; (4) source separation to extract the FECG from the abdominal mixture; (5) FQRS detection; (6) one of the FQRS time-series is selected; (7) the resulting time-series is smoothed and finally the extracted time-series is scored against the reference FQRS. (b) Details of the source separation block. The source separation methods are divided in four classes (I-IV). AECG: abdominal ECG, BSS: blind source separation, RES: residual, dashed line: optional step.

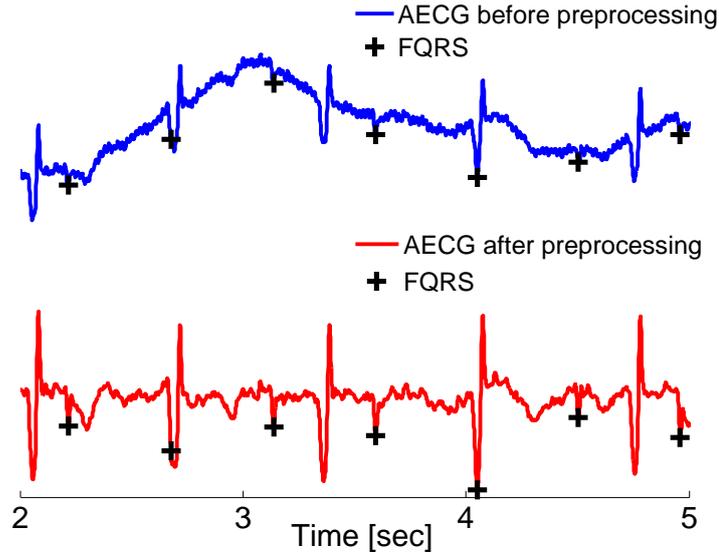

Figure 7.3: Example of preprocessing (channel 2, a39). Note in particular the effect of baseline wander removal as well as the 50 Hz noise suppression. $f_b$=2 Hz, $f_h$=100 Hz.

### 7.2.4 QRS detection

A QRS detector, based on an algorithm similar to that of P&T [66], was used to detect the MQRS and FQRS, with refractory periods of 250 ms and 150 ms respectively. Given the detected energy peaks, each R-peak's position was corrected so that the sign of the corresponding R-peaks were all positive or negative (taking the local maxima or minima on the ECG signal). This avoided changes in FHR when some R-peaks were detected positively and others negatively on a signal, and was also required for performing TS (building a template MECG cycle and accurately subtracting it from the subsequent MECG cycles requires the detected peaks to have the same sign).

The reference MQRS time-series was extracted from one of the abdominal channels. However, the use of a TS or EKF based technique required very precise locations of the maternal R-peaks for each individual channel in order to perform optimal MECG removal (see Figure 7.4). Indeed, this location can vary slightly from one channel to another due to the morphology of the ECG signal, which depends on the electrode location. For this reason, the maternal R-peaks were re-adjusted for each individual channel. Correction of maternal R-peak was performed by looking for a local maximum (or minimum) around each detected MQRS (30 ms each way).

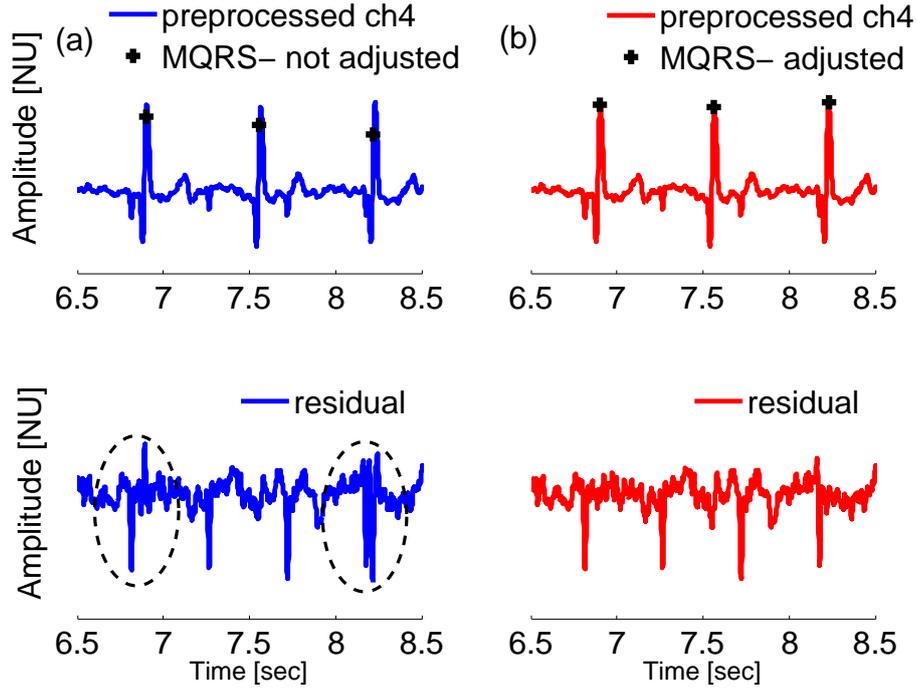

Figure 7.4: MQRS correction. MQRS detection was performed on channel 1 of a39 (reference MQRS) and used for TS on channel four without adjustment (a) and with adjustment (b). On (a) and (b) the upper plots correspond to the fourth abdominal channel (ch4) with the MQRS location (corrected/not corrected) and the lower plots corresponding to the residual obtained using TS. Note the MQRS residuals circled when the MQRS time-series was not adjusted.

### 7.2.5 Channel selection

Channel selection was required to choose: (1) the MQRS time-series (given that no reference chest channel was available) and (2) the FQRS time-series detected on the residuals. Selection of the QRS time-series extracted from the AECG (for MQRS detection) and the individual residual channels (for FQRS detection) was based on a smoothing indicator ($SMI$), defined as the number of occurrences, over each one-minute segment, where the absolute value of the change in instantaneous heart rate (CH) was more than 29 beats per minute (bpm). This threshold was empirically determined on the training set. The channel with the lowest $SMI$ (denoted $c_s$) was selected:

$$c_s = argmin_{c \in [1:4]}(SMI_c).$$

$$SMI_c = card(\{x \in \mathbf{CH}_c, x \geq 29 bpm\}),$$

where $\mathbf{CH}_c$ is the CH vector for channel $c$ and $card$ stands for cardinality.

In the case of the MQRS time series selection, the $SMI$ was required because, in some instances, the FECG was of comparable amplitude to the MECG. This resulted in a large proportion of FPs when such a channel was selected. Assuming that out of the four channels at least one had a dominant MECG amplitude, the $SMI$ selected a correct MQRS time-series.

In the case of the FQRS time series selection, the $SMI$ was used to select the smoothest FQRS time series. In addition, in order to avoid selecting an MQRS time-series instead of the FQRS time-series on the residual channels (which could happen if the MECG cancellation did not perform well or if the amplitude of the FECG on the residual channel was negligible with respect to the MECG residual), a beat comparison measure (BCM) was used. BCM corresponds to the fraction of detected beats on a residual channel that matched with the reference MQRS (within 50 ms). The closer the BCM was to one, the higher the probability of having detected the MQRS instead of the FQRS. A threshold, empirically set using the training set, of 40% was used (i.e if BCM $\geq$ 0.4 then the QRS time-series extracted from this residual channel was excluded). This implicitly assumed that the probability of the MQRS and FQRS being completely in phase (i.e. same HR and overlapping QRS complexes) was negligible. This measure is analogous to the output of the Physionet `bxb` function when using the MQRS time-series as the reference and the FQRS time-series as the input being tested. The algorithm for performing the FQRS selection is presented as pseudocode in Figure 7.5 and the corresponding MATLAB function is available on `Physionet.org`:

### 7.2.6 Source separation

There are many ways of classifying the methods for extracting the FECG from the AECG mixture. In this work the methods are classified in four categories (Figure 7.2 b): (1) Perform-

```
function chNb = select_channel(ecg)
for ch=1:NB_AECG
    // for each abdominal ECG channel
    ecgn = normalise_ecg(ecg(:,ch));
    QRSce{ch} = run_qrsdet_by_seg(ecgn);
    QRSce{ch} = smooth_rr(QRSce{ch});
    SMI(ch) = assess_regularity(QRSce{ch});
    // check that the FQRSs are not matching the MQRSs
    bxb(ch) = bsqi(MQRS,QRSce{ch});
end
// discard time series that are matching the maternal one
SMI(bxb>0.4) = Inf;
// select the most regular time series
chNb = select_min_SMI(SMI)
```

Figure 7.5: Pseudocode of the channel selection function.

ing TS in the observation domain (e.g. TS, $TS_c$ [53], $TS_m$ [54], $TS_{lp}$ [55], $TS_{pca}$ [48], $TS_{EKF}$ [9, 61]), (2) applying a BSS technique directly on the AECG (ICA, PCA), (3) performing TS and applying BSS on the residuals (denoted TS-ICA) and (4) moving to the source domain using BSS and performing TS in that domain with an eventual final BSS step (as for example in [9], denoted ICA-TS and ICA-TS-ICA). See Chapter 3 for an introduction to these algorithms.

The methods were implemented and evaluated in terms of the $F_1$ measure without post-processing (i.e. no RR smoothing). Based on these results, the FUSE method was introduced. FUSE was defined as the combination of a subset of the evaluated methods (ICA-TS, ICA-TS-ICA, TS-ICA, ICA, TS) where only one FQRS time-series (detected on one of the residual channels using one of the source separation methods) was selected. The reasoning for the FUSE approach was that the different source separation techniques each have their own strengths and weaknesses, and that combining some could lead to higher performance (given a good measure for selecting the 'best' extracted FQRS time-series). This was suggested by the results from Chapter 6, where no one technique outperformed all the others. The FUSE algorithm starts by prefiltering the data, then the MQRS time series derived from one of the abdominal channel is selected and used as fiducials for a given template subtraction NI-FECG separation technique. Different source separation techniques are run on the abdominal channels (using the MQRS). FQRS are detected on the residual signals using a P&T FQRS detector (see Appendix E.4 for more details). One of the FQRS time series is then selected based on the regularity of the time series and under the condition that the FQRS time series did not match the MQRS time series (the MQRS residual could sometimes be picked-up by the QRS detector). This last criterion, was assessed using the *bsqi* function which was used similar to the Physionet *bxb* function.

In what follows, FUSE-SMOOTH corresponds to the FUSE method when adding the time-series smoothing block (Figure 7.2), and FUSE-CHALL corresponds to the results of the FUSE method with a biased output toward the Challenge scoring system (i.e. in the case of FQRS detection failure the algorithm output a constant time-series at 143 bpm or at the dominant FHR mode). Detection failure was defined as $SMI > 26$ (no units). The 143 bpm and 26 cut-off values were empirically determined on the training set.

### 7.2.7 Smoothing

The algorithm then smooths the RR time-series in order to remove extra detected FQRS and fix missed FQRS. It deals with a number of cases, illustrated in Figure 7.7a, given some physiological prior on the FHR dynamics and range; for each detected FQRS the median RR interval over the past 5 beats was estimated as a predictor of the current RR interval. If this median RR (medRR) was physiologically sound (i.e. corresponding to a FHR in the interval 110-170 *bpm*) it was used to assess whether or not the current detection was accurate.

Extra detections were defined as an FQRS located less than 70% of the median RR away from the previous FQRS, and were suppressed when found. Missing beats were defined as an FQRS located more than 175% of the median RR away from the previous FQRS, and an additional FQRS annotation was inserted one instantaneous RR interval after the last reliable detection. When the median RR was not in the physiological range 110-170 *bpm*, no smoothing was performed. Figure 7.7b shows an example on a real signal where noise in the residual signal led to an extra detected fetal beat when no RR smoothing was performed. Figure 7.6 shows the pseudocode of the smoothing function and the corresponding MATLAB function is available on `Physionet.org`:

```
function FQRSc = smooth_rr(FQRS)
MIN_FRR = 0.35*fs;
MAX_FRR = 0.5*fs;
while qrsnb<length(FQRSc)-1
// median RR interval computed over the past 5 beats
med = median(diff(FQRSc(qrsnb-5:qrsnb)));
if med>MIN_FRR && med<MAX_FRR
    dTplus = FQRSc(qrsnb+1)-FQRSc(qrsnb); // RR forward
    dTminus = FQRSc(qrsnb)-FQRSc(qrsnb-1); // RR backward
    if dTplus<0.7*med && dTminus<1.2*med
        // extra beat
        FQRSc(qrsnb+1) = []; // remove extra beat
    elseif dTplus>1.75*med && dTminus>0.7*med
        // missed beat
        // use the median RR to predict missing beat location
        MissedFQRS = round(FQRSc(qrsnb)+med);
        FQRSc ← MissedFQRS // insert missed FQRS
    else
    // normal detection
    qrsnb = qrsnb+1;
    end
else
    qrsnb = qrsnb+1;
end
```

Figure 7.6: Pseudocode of the QRS smoothing function.

### 7.2.8 Parameter optimisation

There are several parameters that have an influence on the algorithm performance. Table 7.1 lists some of the important ones with their chosen values: the number of cycles for building the template ECG in TS (nbC), the gain of the **R** and **Q** covariance matrices ($G_R$ and $G_Q$) for the $TS_{EKF}$ and the baseline wander and high frequency cut-off ($f_b$, $f_h$) for the prefiltering parameters of all methods. For these parameters an exhaustive or random search was performed (see Table 7.1). For an introduction to random search, refer to Appendix E.3.

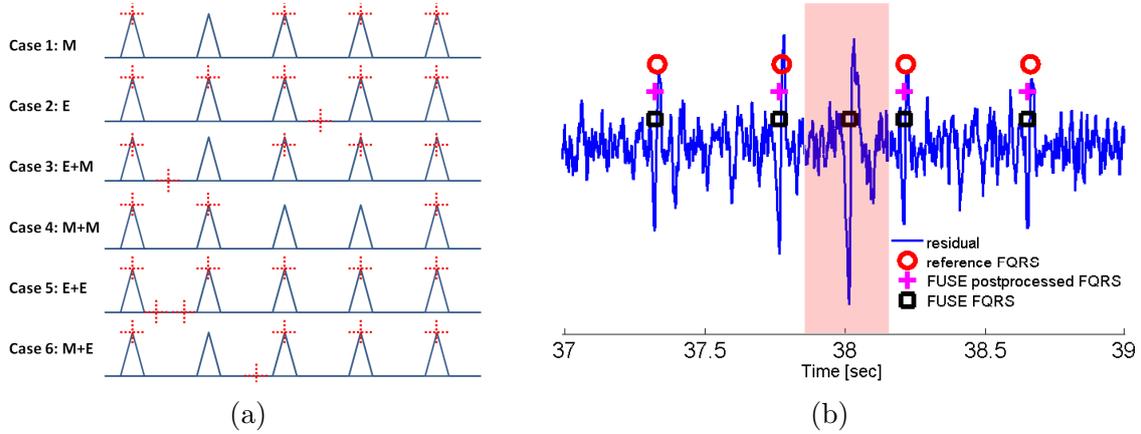

(a) (b)

Figure 7.7: (a) Smoothing FQRS time-series. Each triangle represents a FQRS and each cross a FQRS detected by the P&T detector. Six cases of missing/extra detection were handled by the smoothing function. M: missing detection, E: extra detection. (b) example of an extra FQRS detection on a real signal (a27); reference FQRS (circles), FQRS detected by the P&T detector on the extracted NI-FECG signal (residual) and the smoothed FQRS fiducial (crosses).

| Parameters | Definition | Values |
|---|---|---|
| nbC* | number of cycles for the template ECG in TS | 20† |
| nbPC | number of principal components for $TS_{PCA}$ | 2† |
| $G_R$ | gain of the R covariance matrix for the $TS_{EKF}$ | 100♮ |
| $G_Q$ | gain of the Q covariance matrix for the $TS_{EKF}$ | 5♮ |
| $f_b$ | baseline wander cut-off frequency | 10 Hz† |
| $f_h$ | high frequency cut-off | 99 Hz† |

Table 7.1: Key global parameters of the NI-FECG extraction algorithms. *with exception of $TS_{lp}$ for which 9 cycles were used as indicated in [57]. † indicates that the parameters were searched using grid search. ♮ indicates that the parameters were searched using random search.

## 7.3 Results

### 7.3.1 Parameter optimisation

#### 7.3.1.1 Prefiltering

A search for the prefiltering cut-off frequencies was carried out on the training set for the FUSE method (Figure 7.8a). From the grid search, $f_b$=10 Hz and $f_h$=99 Hz were selected.

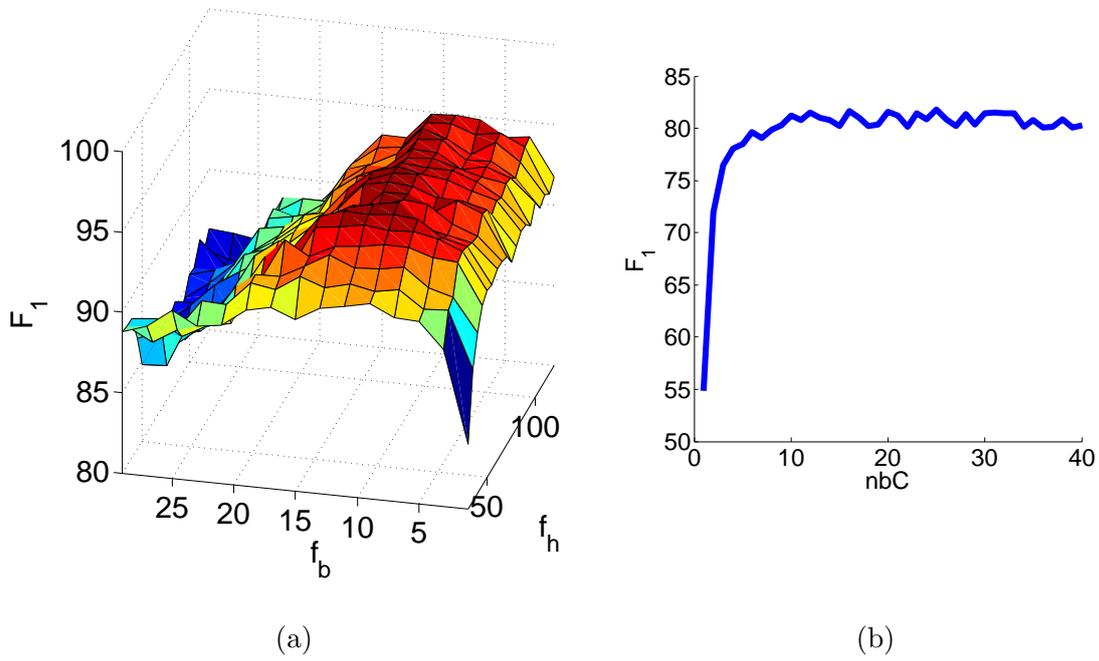

(a)                  (b)

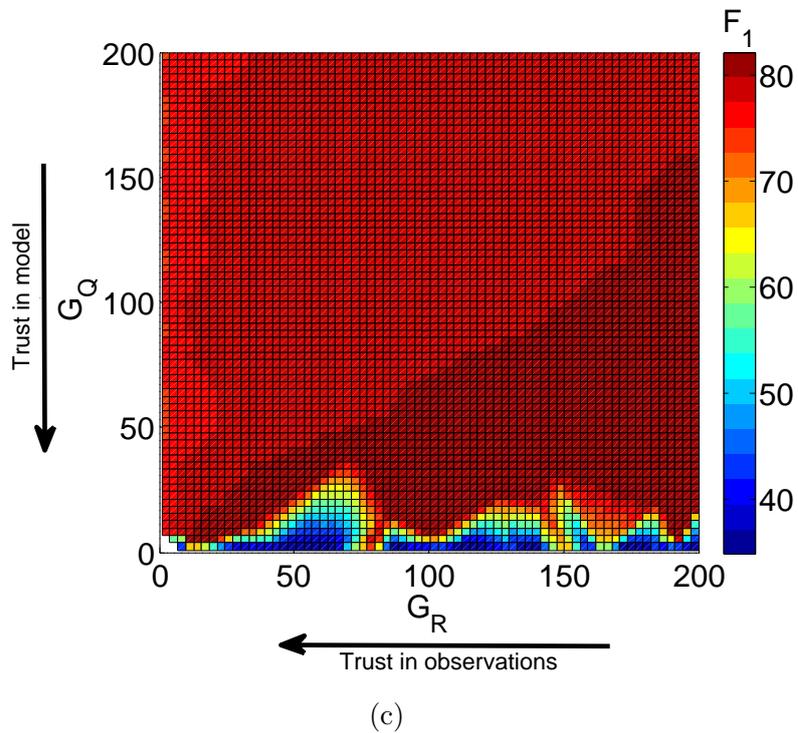

(c)

Figure 7.8: Search for parameters: (a) Grid search on prefiltering cut-off frequencies $f_b$ and $f_h$, (b) search on the number of cycles to use (nbC) for TS technique. (c) Random search performed on the EKF covariance matrices gains ($G_R$ and $G_Q$) with $f_b = 10$ Hz.

#### 7.3.1.2 Number of cycles and principal components

Figure 7.8 illustrates how the performance of TS changes with respect to the number of cycles used to build the maternal template ECG cycle. nbC=20 was chosen. For $TS_{pca}$ the best result was obtained for nbPC=2.

#### 7.3.1.3 EKF covariance matrices

Random search was performed similar to Behar *et al.* [124] (see Appendix E.3) over the EKF covariance matrix gains ($G_R$ and $G_Q$) with $f_b$=10 Hz. This was done by sampling the search space and fitting a hyper-surface to produce Figure 7.8c. $G_R$=100 and $G_Q$=5 were selected as gain parameters.

### 7.3.2 Challenge results

The best result in terms of $F_1$ measure was obtained using the FUSE method. Smoothing the FQRS time-series (FUSE-SMOOTH, post processing) improved the performance by 1% on set-a. The FUSE-SMOOTH algorithm performed better than all the individual methods on the training dataset. On the validation set, the best Challenge scores obtained were $E1$=179.4, $E2$=20.8, $E3$=153.1, $E4$=29.6 and $E5$=4.7 for events 1-5 respectively using the FUSE-CHALL method. These were the best Challenge scores for $E1$ and $E2$ and third and second best Challenge scores for $E4$ and $E5$ out of the 53 international teams that entered the Challenge. Table 7.2 lists the results obtained for all individual and combined methods on the training set data.

## 7.4 Discussion

### 7.4.1 Approach undertaken in this work

Using the $F_1$ measure to evaluate the algorithms and for parameter optimisation was motivated by a key problem with the Challenge scoring system. Generating a constant heart rate (or equivalently constant RR interval) that fell into an accurate physiological range (i.e. be-tween 120 bpm and 160 bpm [24]) provided superior results to most individual methods (see CONST-HR results in Table 7.2). This is because a root mean square (RMS) based measure will outweigh some inaccurate detection. Another limitation on the use of RMS values for scoring was its relevance to clinical practice; physicians are generally more interested in rapid variations of FHR, but the FHR tends to have an average mean around 120-160 bpm, while short-term FHR variations are much smaller in magnitude compared to this 120-160 range. Therefore, scoring functions that use RMS values are weighted more towards the average FHR component rather than its short-term variations. The associated flaw of the Challenge scoring system was mainly due to the fact that none of the HR and RR scores took into account the absolute FQRS location (only their relative location). An $F_1$-like measure would therefore be complementary in that it assesses the ability of an algorithm to extract an accurate R-peak location. Due to these limitations of the Challenge scoring system, optimisation of the various parameters used in the FUSE approach was performed using the $F_1$ accuracy metric on set-a (instead of the Challenge scores).

| CL | Method | $HRE$ | $RRE$ | $Se$ | $PPV$ | $F_1$ | $F_1$† |
|---|---|---|---|---|---|---|---|
| | | NU | NU | % | % | % | % |
| I | TS | 655.5 | 27.9 | 81.8 | 81.7 | 81.6 | 81.2 |
| I | $TS_c$ | 514.8 | 29.1 | 81.6 | 81.7 | 81.5 | 81.4 |
| I | $TS_m$ | 551.9 | 28.1 | 82.2 | 82.2 | 82.1 | 81.1 |
| I | $TS_{lp}$ | 902.0 | 46.2 | 82.1 | 81.9 | 81.8 | 78.5 |
| I | $TS_{pca}$ | 594.4 | 21.6 | 88.1 | 84.5 | 86.1 | 83.6 |
| I | $TS_{EKF}$ | 733.8 | 25.0 | 83.0 | 81.1 | 81.9 | 78.4 |
| II | ICA | 2852.1 | 39.3 | 69.1 | 60.0 | 63.7 | 61.7 |
| II | PCA | 3892.1 | 45.3 | 57.4 | 47.9 | 51.6 | 52.6 |
| III | TS-ICA | 272.7 | 17.1 | 93.0 | 91.1 | 92.0 | 91.3 |
| III | $TS_c$-ICA | 202.6 | 17.2 | 93.2 | 92.0 | 92.6 | 92.1 |
| III | $TS_m$-ICA | 251.9 | 18.4 | 91.7 | 90.8 | 91.2 | 92.3 |
| III | $TS_{lp}$-ICA | 399.1 | 37.9 | 88.4 | 88.7 | 88.4 | 85.2 |
| III | $TS_{pca}$-ICA | 153.2 | 16.9 | 93.8 | 92.2 | 93.0 | 92.4 |
| IV | ICA-$TS_{pca}$ | 396.9 | 27.1 | 90.1 | 88.8 | 89.2 | 89.4 |
| IV | ICA-$TS_{pca}$-ICA | 299.4 | 22.7 | 92.6 | 92.2 | 92.4 | 91.1 |
| | CONST-HR (143 bpm) | 172.2 | 8.9 | 23.2 | 23.1 | 23.0 | NA |
| | FUSE | 132.9 | 12.7 | 95.6 | 94.3 | 95.0 | 94.2 |
| | FUSE-SMOOTH | 19.1 | 6.3 | 95.9 | 96.0 | 96.0 | 95.2 |
| | FUSE-CHALL | 5.4 | 2.3 | NA | NA | NA | NA |

Table 7.2: Performance of the different algorithms on set-a. CL: class of the method (see Figure 7.2), HRE: score for the heart rate Challenge event, RRE: score for the RR Challenge event, NU: no unit, NA: not applicable. Statistics are given for $f_b$=10 Hz apart from the last column on the right (indicated $F_1$†) which is for $f_b$=2 Hz.

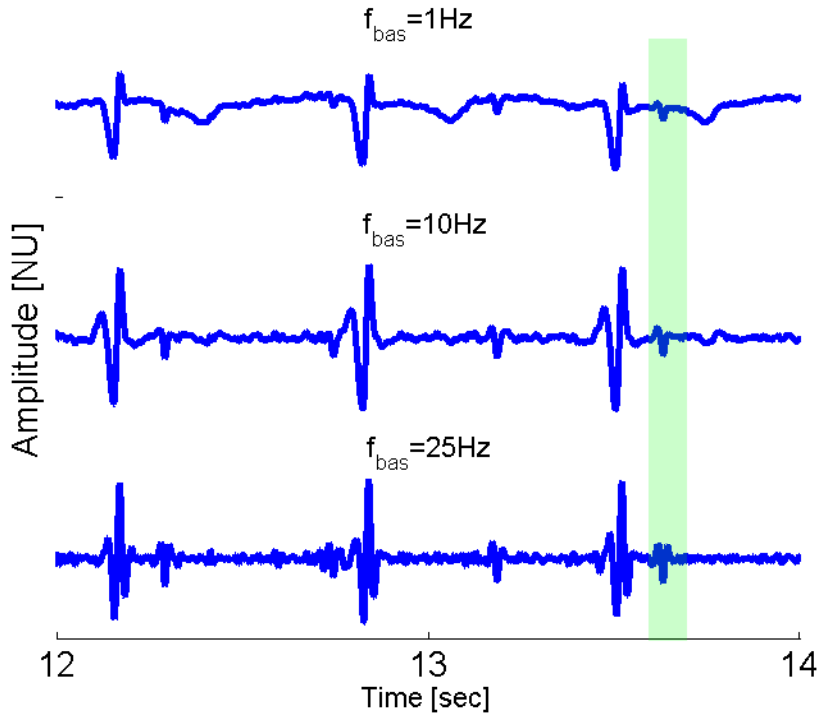

Figure 7.9: Example of high pass filtering effect on record a39, first abdominal channel. At 10 Hz most of the frequency content of the P and the T-wave of the MECG have been filtered out leaving only the FQRS and MQRS and some noise in the signal. Note that the amplitude scale is not the same for the three alternative filtered signals.

As noted in Chapter 6, a high cut-off frequency for baseline wander removal led to improved results (compare $F_1$ for $f_b$=10 Hz and $F_1$ for $f_b$=2 Hz for all methods in Table 7.2). This is because the high cut-off frequency results in a large reduction in the amplitude of the P- and T-waves, leaving only the MQRS and FQRS (and some noise) in the AECG mixture (Figure 7.9). However, such a high cut-off frequency cannot be used for FECG morphological analysis, as features of clinical interest (such as the T-wave, ST segment) might be highly distorted [125]. As a consequence, a high $f_b$ cut-off can be used in order to accurately detect the fetal R-peaks initially, and a fetal QT (FQT) extraction algorithm can be used in a second run with a lower $f_b$ given the location of the fetal R-peaks. In addition, the viability of a FQT measure performed on PCA/ICA transformed signals is to be assessed. When BSS is performed, the fetal T-wave (or part of it) could be projected onto some other components (than the one selected) making the FQT measure in the source domain inaccurate. It is therefore necessary to assess whether FQT measurement performed in the source domain can be a surrogate for the 'true' FQT measurement. This will be assessed in Chapter 8.

There are some intrinsic limitations with the PCA approach: (1) PCA looks for new orthogonal axes, but there is no reason for the FECG and MECG sources to be orthogonal. PCA is not likely to provide the 'best' new basis onto which to re-express the data; (2) the range of the input signal plays an important role; PCA is sensitive to the scaling of the variables. As a consequence, it was important to ensure that the AECG channels were normalised; (3) PCA aims to decorrelate the data, i.e. remove second-order dependencies. However, it is still under investigation whether this is the best way of revealing the underlying FECG. In addition, it was estimated by Sameni *et al.* [126], that 4 to 6 statistically uncorrelated dimensions are required to represent typical ECG signals, and so it is likely that four abdominal channels are not enough to make the most of the BSS approaches.

The different extraction methods described in this work and in the literature vary greatly in their ability to adapt to the changing ECG morphology; high adaptability is usually synonymous with better MECG removal but a lower FECG residual amplitude which might not be preferable, particularly in the case of overlapping MQRS and FQRS. In addition, the performance of most of the evaluated methods would be negatively affected by a change in beat type, as in the presence of ectopic beats. One way to tackle this problem would be to identify different modes corresponding beat types (e.g. normal sinus rhythms, premature ventricular contraction), and build a template ECG from each of these beat types (see, for example, the work on the Kalman switching in the paper we recently submitted for publication: Oster *et al.* [127]). Then the template could switch depending on what beat type is the most likely.

Within the TS techniques, $TS_c$, $TS_m$, $TS_{lp}$ did not improve upon the simple TS (i.e. simply subtracting the MECG template to each subsequent MECG cycle without any scaling). Only $TS_{pca}$ showed improvement on the training set data (Table 7.2), confirming the results observed in Chapter 6. Using PCA or ICA directly on the preprocessed AECG signals did not perform well. This may be due to the low number of abdominal channels available (only four), which is not enough to represent the dimensionality of the MECG, FECG and noise. As a consequence, performing a TS step before applying PCA/ICA increased the $F_1$ accuracy measure (see Table 7.2). Performing TS prior to BSS (for removing the MECG signal) reduces the dimensionality of the problem, making the components more separable. TS$_{EKF}$ did not perform better than simpler TS techniques. This might be due to the automated Gaussian initialisation procedure used, or due to the fact that the covariance matrices were considered to be stationary. The random search performed in Figure 7.8c highlighted that the evaluation of **R** and **Q** was critical to the performance of the EKF method. In particular, the results suggest that an increased G$_R$ and a low G$_Q$ improved the results. This corresponds to high trust in the ECG model and low trust on the observation, tending to a TS-like method with the advantage of having no constraint on the P, QRS and T wave lengths and some overall adaptability.

The FUSE method introduced in this work was a combination of a subset of the methods implemented. The selected methods were run independently and the output with the smoothest detected FQRS time-series was selected. This improved the results in comparison to using any individual method. However, this is a very simplistic way of fusing the information, and there is room for improvement in the approach to combine the FQRS detection performed on different channels using different methods (of different categories I-IV). Some ideas have been suggested by other Challenge participants such as in Lipponen *et al.* [128] and Andreotti *et al.* [76], who were the winners for events 4-5.

It was noticeable that beyond the selected source separation algorithm used, the FQRS detector, the method for performing channel selection and the time-series smoothing algorithm made a difference to the results. This is likely to be the main element that discriminated between the top Challenge participants.

Only two entries, including the one presented in this work, managed to produce an output for $E3$ of the Challenge addressing the question of FQT detection. However, due to the low number of FQT references, the high variability of the FQT length over a given record (FQT varying with HR) and the poor performance obtained, the method and results for this event were not reported here. Further work was performed as a second attempt to extract the FQT from the NI-FECG (see Chapter 8).

The Physionet/Computing in Cardiology Challenge 2013 was the first major publicly

available database for NI-FECG algorithm evaluation, with annotations and gold standard scalp data, including an independent test set which is not publicly available (so overfitting the public data was avoided). However, the Challenge still possessed a number of limitations: (1) the limited number of simultaneously acquired signals (some adaptive source separation approaches, which are very promising, require more than four AECG channels [26]), (2) the absence of chest ECG which could have been helpful for the MQRS detection step (where the MECG SNR is much higher on the chest than on the abdomen), and so the evaluation of adaptive source separation approach would have been possible, such as the methods evaluated in Chapter 6, (3) the presence of errors in some of the reference annotations, and (4) the lack of pathological examples: no arrhythmia, ST deviations, QT prolongation, late or early decelerations or contractions were present in the data.

### 7.4.2 Competitors' approach

A large number of algorithms for FHR and RR series estimation were proposed for the Challenge. The aim of this section is to present several of the different signal processing techniques from other entrants that led to successful FECG estimation and that were published in the subsequent Physiological measurement special issue [3]. The techniques presented at the Challenge were unique and original, but in general each had a five step approach as follows: (1) pre-processing; (2) estimation of maternal component; (3) removal of maternal component; (4) estimation of FHR and RR time-series; (5) post-processing.

The articles in the special issue [3] fall into three general categories based upon their signal processing approaches; temporal in the observation domain, spatial, and frequency (or time-frequency) approaches. The following review attempts to group these articles following this classification. However, several articles combined multiple of these approaches and so do not fall neatly into a single category.

Andreotti *et al.* [129] won events 2 and 5. The authors used kernel density estimation for fusing detection algorithms on the different channels for MQRS detection. The use of differential channels to augment the set of the four abdominal channels was also studied. Template adaption and an extended Kalman smoother for removing the maternal contribution were employed. An evolutionary algorithm was used in order to correct for FQRS detection, where weights were chosen based on signal periodicity and signal morphology. ICA was only used for MQRS detection, but all the processing for extracting the FECG and detecting the FQRS was performed in the time domain using temporal methods on the available abdominal channels and possible differentials. The authors also used a 470 min private dataset recorded from 10 pregnant women to further evaluate their extraction algorithms. The authors reported that the TS approach performed better than the Kalman filter approach on the Challenge dataset but lower on their additional private dataset.

Haghpanahi *et al.* [130] used the deflation approach from Sameni *et al.* [9] (iterative subspace decomposition and Kalman filtering) in order to remove the MECG. They also used PCA on the four abdominal channels and selected the best FQRS time-series out of the two approaches (deflation/PCA). The authors used kurtosis as a surrogate for signal quality in order to rank the residual signals from the deflation methods and combine a subset of these to infer the FQRS time-series.

Varanini *et al.* [131] removed the MECG from the abdominal signal using a PCA-based template subtraction algorithm and then applied ICA on the residuals. One of the residuals was selected based on knowledge of typical FHR, the number of detected FQRS, mean of absolute RR first derivative and mean of absolute RR second derivative. This was a very similar approach to the $TS_{pca}$ technique followed by an ICA step (Table 7.2, see $TS_{pca}$-ICA), and the results in this and the previous chapter have shown that using $TS_{pca}$ was better than all alternative template subtraction techniques. In addition, subsequent application of ICA improved the results.

Dessi *et al.* [132] used a template subtraction approach followed by an ICA step, FQRS detection and correction, and channel selection. The authors noted that performing the template subtraction step at a higher frequency (they upsampled the data to 8 kHz) was important to align each MECG cycle with the template and get the best cancellation. In order to build the template MECG cycle, the authors selected beats based on correlation thresholding to avoid averaging abnormal beats.

Lipponen *et al.* [133] used a PCA-based template subtraction approach in order to remove the MECG. They built the design matrix for the P, QRS and T waves separately and then applied PCA to identify the prinicipal components. The most significant eigenvectors were fitted back to individual wave epochs from the MECG in order to remove them. The approach for suppressing the MECG is similar to Varanini *et al.* [131] and Behar *et al.* [2] (i.e. the TS$_{pca}$ approach presented in Chapter 3 and used in this chapter), although Lipponen *et al.* [133] separated the MECG cycles into P, QRS and T-waves.

Di Maria *et al.* [134] took a very standard PCA and template subtraction approach. The main focus of the paper explored picking the best principal component in order to identify the best MECG channel and the best FECG channel after performing MECG cancellation.

Liu *et al.* [78] performed prefiltering, then MQRS detection, then template subtraction and finally FQRS detection on the residual. It is important to note that they used a quality index (sample entropy) in order to exclude bad quality channels, which is theoretically more sensible than performing FQRS detection on each channel and making the decision based on the regularity of the RR interval (as most entrants did). The authors also showed that by adjusting the MECG template to each cycle (in contrast to performing the simple construction with the template centred on the MQRS location), a performance improvement can be found. This second point was also illustrated in Behar *et al.* [2], although Liu *et al.* [78] provided an interesting quantification of this phenomenon.

Lukosevicius *et al.* [135] focused on the application of a QRS detector using an Echo state neural network (ESN), a data-driven statistical machine learning approach. The ESN is trained with the four residual signals (obtained using the MECG cancellation method from [54]) as channels of the input stream, and a probability of QRS detection was the output. It should be noted that the authors did not focus on the extraction algorithms but on the QRS detector using multiple channels.

Rodrigues [136] employed a Wiener filter which took, as the input, the three abdominal channels with a number of coefficients (91) in order to filter out the MQRS from the fourth channel. The authors also used the MIT 'Abdominal and direct fetal electrocardiogram database' in order to train their algorithm, which may have led to a bias in the results as this database was included in set-a, set-b (and possibly a few records in set-c).

Christov *et al.* [137] described a template subtraction method, with the template length being heart rate dependant, followed by an enhancement method that combined the four abdominal channels. The combined lead was obtained using (1) PCA, (2) RMS or (3) Hotelling T-squared. The final combined lead was obtained by taking a mean over these three methods.

The final article in this collection is by Almeida *et al.* [138], who took a wavelet approach to denoising and extracting the fetal ECG. Although a time-frequency analysis seems very promising, the large cross-over in the spectral domain between the maternal and fetal signals and the noise, means this approach appeared to be limited.

### 7.4.3 Competitors score

The Challenge participants were provided with an open source sample entry. This entry corresponds to the implementation of $TS_m$ presented in this thesis (see description in Chapter 3). The competitors were invited to either improve the sample entry or generate their entry following the same interface as that of the sample entry. A total of 53 teams participated in the Challenge, yielding to a total of 93 open source entries, with the vast majority outperforming the sample entry (Figure 7.10). The top scores for all the events (E) were: 179.4 $(beats/minute)^2$ ($E1$), 20.8 $ms$ ($E2$), 18.1 $(beats/minute)^2$ ($E4$), and 4.3 $ms$ ($E5$). Results presented at the Computing in Cardiology conference 2013 are presented in Table 7.3. Following the Challenge some participants further refined their algorithms and their updated scores are reported in Table 7.4. Note that only the scores from the participants that submitted a paper in the follow-up special issue were listed in the tables.

## 7.5 Summary and conclusion

The FUSE algorithm performed better than all the individual methods on the training dataset. On the validation and test sets, the best Challenge scores (as defined in Chapter 4) achieved were $E1 = 179.4$ $(beats/minute)^2$, $E2 = 20.8$ $ms$, $E3 = 153.1$ $ms^2$, $E4 = 29.6$ $(beats/minute)^2$ and $E5 = 4.7$ $ms$ using the FUSE method for events 1-5 respectively. These were the best Challenge scores for $E1$ and $E2$ and third and second best

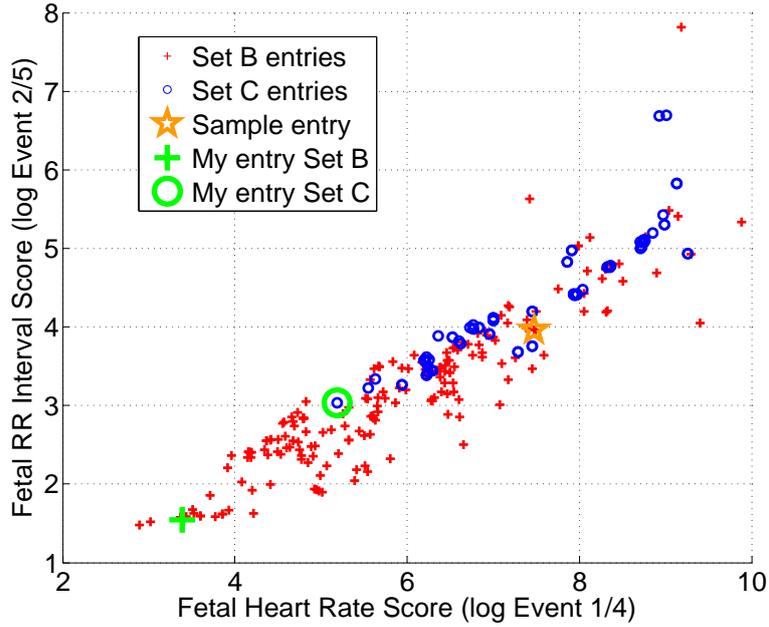

Figure 7.10: Scatter plot of the scores for the Challenge (best scores are in the lower left corner). Scores for Set B and Set C are marked with red crosses and blue circles respectively. The score for the sample entry is highlighted by a star. The best entries from this thesis work are highlighted with a green cross and green circle for the scores on Set B and Set C respectively.

| Participants\Events | $E1$ | $E2$ | $E4$ | $E5$ |
|---|---|---|---|---|
| Andreotti et al. [76] | NA | NA | <u>18.1</u> | <u>4.3</u> |
| Behar et al. [139] (non-official) | <u>179.4</u> | <u>20.8</u> | 29.6 | 4.7 |
| Haghpanahi et al. [140] | 6298.1 | 159.9 | 50.1 | 9.1 |
| Varanini et al. [141] | <u>187.1</u> | <u>21.0</u> | 34.0 | 5.1 |
| Dessi et al. [142] | 684.2 | 48.0 | 639.5 | 23.8 |
| Lipponen et al. [128] | NA | NA | 28.9 | 4.8 |
| Di Maria et al. [77] | NA | NA | 223.2 | 19.3 |
| Liu et al. [143] | 2782.3 | 81.7 | 264.9 | 9.0 |
| Lukoševičius et al. [144] | NA | NA | 66.3 | 8.2 |
| Rodrigues et al. [145] | 278.8 | 28.2 | 124.8 | 14.4 |
| Christov et al. [146] | NA | NA | 285.1 | 20.0 |
| Almeida et al. [147] | NA | NA | 521.4 | 33.0 |

Table 7.3: Results presented at the Computing in Cardiology conference 2013 for events 1-5 ($E1$-$E5$) for all the papers published in the follow-up special issue. NA: Not Available, because the corresponding participants did not enter the open source events $E1$-$E2$. $E1$ and $E4$ in bpm$^2$ and $E2$, $E5$ in ms.

Challenge scores for $E3$, $E4/E5$ out of the 53 international teams that entered the Challenge. The results demonstrated that existing standard approaches for fetal heart rate estimation can be improved by fusing estimators together. Open source code for the benchmark methods have been made available on `physionet.org/challenge/2014/`, allowing for comparison and reproducibility on the public domain data. The work presented in this chapter was published in Behar *et al.* [2], the co-authored conference paper presenting the competition in Silva *et al.* [148] and the subsequent co-authored Physiological Measurement special issue editorial in Clifford *et al.* [3].

| Participants\Events | $E1$ | $E2$ | $E4$ | $E5$ |
|---|---|---|---|---|
| Andreotti *et al.* [129] | NA | NA | <u>15.1</u> | <u>3.3</u> |
| Behar *et al.* [2] | <u>179.4</u> | <u>20.8</u> | 29.6 | 4.7 |
| Haghpanahi *et al.* [130] | NA | NA | 50.1 | 9.1 |
| Varanini *et al.* [131] | 187.0 | 21.0 | 34.0 | 5.1 |
| Dessi *et al.* [132] | 281.1 | 25.93 | 134.5 | 12.4 |
| Lipponen *et al.* [133] | NA | NA | 28.9 | 4.8 |
| Di Maria *et al.* [134] | NA | NA | 142.7 | 19.9 |
| Liu *et al.* [78] | NA | NA | 47.5 | 7.6 |
| Lukoševičius *et al.* [135] | NA | NA | 66.3 | 8.2 |
| Rodrigues *et al.* [136] | 278.8 | 28.2 | 124.8 | 14.4 |
| Christov *et al.* [137] | NA | NA | 305.7 | 23.1 |
| Almeida *et al.* [138] | NA | NA | 513.1 | 35.3 |

Table 7.4: Challenge results for the algorithms presented in the follow-up special issue i.e. considering further development after the Challenge deadline. NA: Not Available (i.e. not reported in the journal paper publication). $E1$ and $E4$ in bpm$^2$ and $E2$, $E5$ in ms.

In conclusion, this work: (1) evaluated a wide variety of standard and state of the art methods used for FECG extraction on the (to date) largest NI-FECG public dataset, (2) benchmarked these algorithms on the same database, with the same experimental set-up, (3) showed that performance can be improved by combining different methodologies, (4) introduced a method for performing channel selection to ensure that the detected beats do not correspond to the MQRS time-series, (5) evaluated the effect of the preprocessing on the algorithms' outcomes and (6) suggested some metrics to evaluate the algorithms' accuracy. The broader contributions made in this thesis, through the Challenge, to the field of NI-FECG signal processing, were that it:

- provided the (to date) largest open source database of NI-FECG signals with FQRS references to the scientific community for evaluating NI-FECG extraction algorithms

- defined a standard for evaluating NI-FECG extraction algorithms using a training, validation and hidden test set, together with a number of statistics for quantifying the performance of the algorithms

- showed that the selection of abdominal channels based on signal quality improved the performance of the BSS NI-FECG extraction techniques [145]

- highlighted the importance of aligning the MQRS between beat and template MECG for the TS based techniques [2, 77]

- showed that using a high baseline wander cut-off frequency $f_b$ improved the performance of the FQRS extraction algorithms [2]

- showed that $TS_{pca}$ performed better than any benchmark TS-like techniques [2, 78, 132]

- introduced an advanced version of the NI-FECG simulator [5] including a series of non-stationary effects in the modelling [6]. (See Chapter 4 for a description of the simulator)

- provided a number of open-source QRS detectors specifically designed for FQRS detection

- showed that the extraction methods had strengths and weaknesses and that combining them improved the overall FQRS detection performances in [2]

- showed that the database was not suitable for QT measurement with the annotations as they were.

The scores obtained in the Challenge, and associated $F_1$ values, reflect the success of the general approach undertaken in this work to accurately extract the FHR and fetal RR intervals as well as locate the FQRS (fetal peak detection being an essential first step in morphological analysis of the FECG). The results demonstrated that existing standard approaches for fetal heart rate estimation can be improved by fusing estimators together. Open source code has been made available on Physionet at `http://physionet.org/` to enable benchmarking for each of the standard approaches described.

# Chapter 8

# NI-FECG morphology analysis

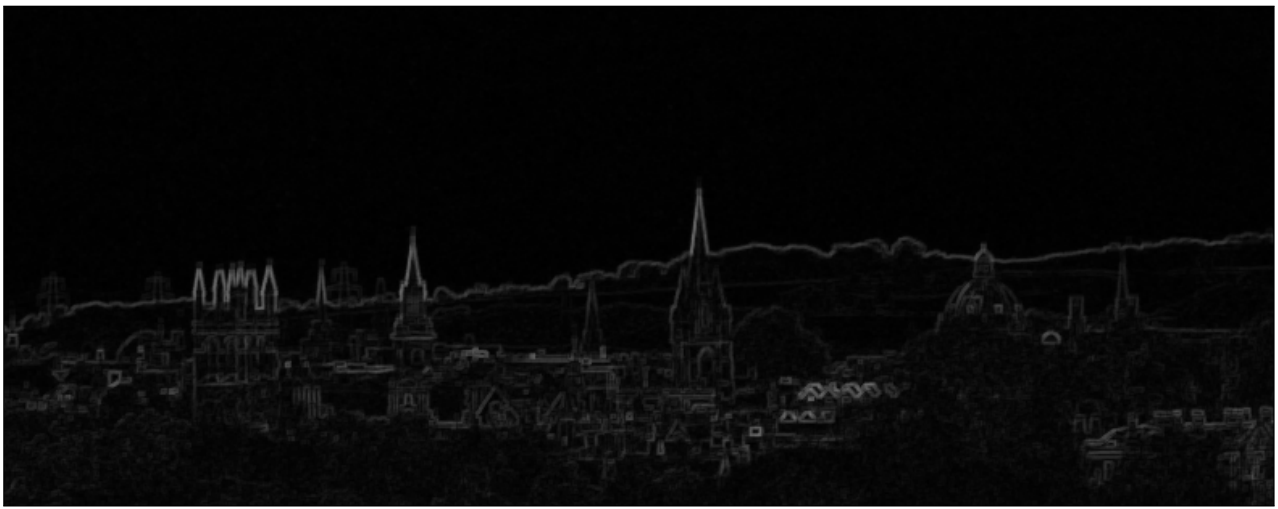

Figure 8.1: Oxford has an amazing architectural heritage. However, because the city has restricted the height of any new building in the city, there are a limited number of places where it is possible to capture this amazing Oxford architectural morphology 'from above'.

## 8.1 Introduction

Commercial non-invasive fetal electrocardiogram (NI-FECG) monitors, such as the Meridian monitor from MindChild Medical (North Andover, MA), have proved to be accurate in detecting the fetal heart rate (FHR), and early works on extracting morphological information have been published [46, 47]. These recent advances in the field are very exciting, but the studies are still limited in number and population size because the algorithms and associated hardware for extracting useful FECG are relatively new. Moreover, the commercial algorithms were designed to accurately extract the FHR and not specifically the FECG morphology. Thus, a novel class of algorithms focused on morphological extraction may be necessary as an adjunct to the existing algorithm targeted at FHR estimation. This was also highlighted during the Challenge 2013 [148] where the QT event was unsuccessful due to both (1) the difficulty for obtaining meaningful reference morphological QT annotations on the FECG data; (2) the scientific challenge in extracting the morphology of the FECG from abdominal recordings; (3) the low dimensionality of the data and lack of maternal reference. Research into developing methods specifically focused at accurate morphological extraction of the NI-FECG is ongoing (see for example the recent contribution from Niknazar *et al.* [149]), but few studies have demonstrated the possibility of extracting any clinically relevant information from the reconstructed morphology. In this chapter the focus is given to the measurement of the fetal QT (FQT) for (1) the output NI-FECG signal provided by a commercial monitor, the Meridian monitor, whose output signal is in the source domain (i.e. following a BSS step), and (2) the extraction of the NI-FECG in the observation domain using a new formulation of the Bayesian filtering framework for accurately extracting the FECG morphology from a single abdominal channel. It is important to note that, depending on the category of NI-FECG extraction algorithm that is used, the morphology reconstruction can be attempted in the source domain (experiment 1) or in the observation domain (experiment 2). In this chapter, both options were evaluated and although this work is exploratory, the feasibility of both approaches is assessed, as well as identifying their advantages and drawbacks.

## 8.2 Recovering the FQT in the source domain

### 8.2.1 Introduction

In this section, the extraction of the FQT from the NI-FECG output by the MindChild monitor is studied by comparing manual annotations from the NI-FECG and the scalp ECG (considered as ground truth).

### 8.2.2 Methods

The database used for this experiment was presented in Chapter 4 (section 4.2.4) as the Real QT database (RQTDB). For SET1 and for each 1 min segment, the median QT annotation of the five QT annotations performed by a given annotator was obtained for both the SECG (in the observation domain) and the NI-FECG (referring in this section to the output of the MindChild monitor in the source domain). For SET2 the variation between paired measurements of the QT segment (i.e. the annotation performed on the average SECG and corresponding average NI-FECG) was analysed. For both SET1 and SET2, the annotations from the NI-FECG were compared to the ones of the SECG for each independent annotator and when fusing the annotator's annotations. Three methods for fusing the annotations were investigated: mean, median and an expectation maximisation (EM) algorithm as used in Zhu *et al.* [150] (see Chapter 4.3.2). In addition, Wilcoxon signed rank test was applied to test the hypothesis that the difference between scalp and abdominal QT estimations were sampled from continuous distributions with no differences in the medians of SET1 and SET2.

### 8.2.3 Results

Figure 8.2 shows the empirical probability density estimate for the FQT intervals annotated by the three annotators for SET1 and SET2. On this plot, AQT refers to the QT annotated on the NI-FECG extracted using the MindChild monitor and SQT refers to the QT annotated on the SECG. For SET2, the two distributions (AQT and SQT) superimpose perfectly, while the AQT distribution has a lower median and is more platykurtic (broader) for SET1. Figure 8.3 shows an example of averaged FECG waveforms (FSE and NI-FECG) after annotation by the three experts. The figure shows the close agreement between the experts' annotations on the FSE and on the NI-FECG. A total of 3150 annotations were performed for SET1 (1050 per annotator) and 630 for SET2 (210 per annotator).

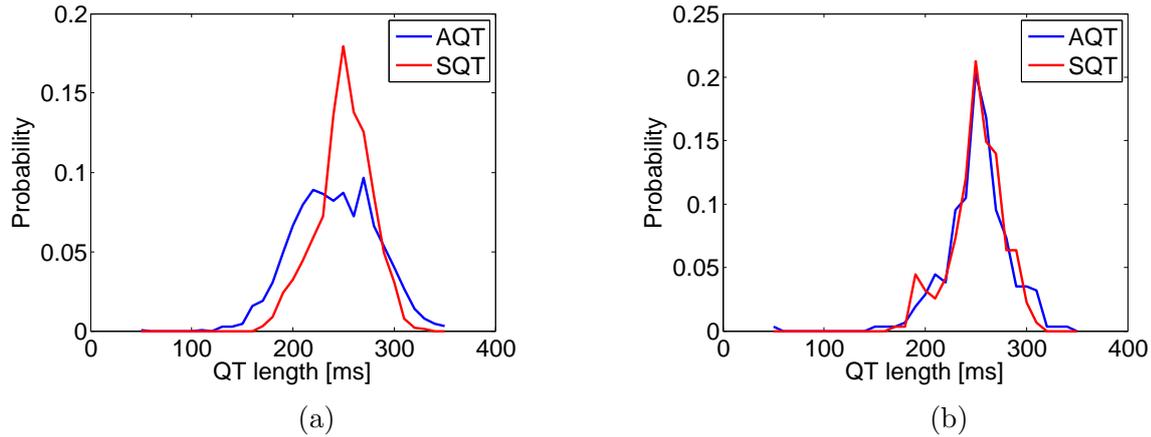

Figure 8.2: Probability density function for the median FQT interval annotated by the three annotators for: (a) SET1 (i.e. annotation on the raw signals), 3150 annotations and (b) SET2 (annotation on the averaged waveforms) 630 annotations. For SET2, the two distributions (AQT and SQT) superimpose closely, while the AQT distribution has a lower median and is more platykurtic (broader) than the SQT distributions for SET1.

For SET1, the null hypothesis was rejected under the 5% significance level whereas the null hypothesis could not be rejected for SET2 (p = 0.58). This statistical test confirms that the distributions for scalp and abdominal annotations only matched (i.e. were not significantly different) when the averaged waveform of the NI-FECG and SECG signals were used. Table 8.1 and 8.2 present the results for SET1 and SET2 respectively when considering each individual annotator and all the annotators combined. The lowest absolute error when combining all three annotators for SET1 and SET2 was 14.2 ms and 10.4 ms respectively. Figure 8.4 shows that combining the annotations from the three experts resulted in a lower bias, a slope closer to one and higher goodness of fit ($R^2 = 0.61$) than any of the three annotators taken individually.

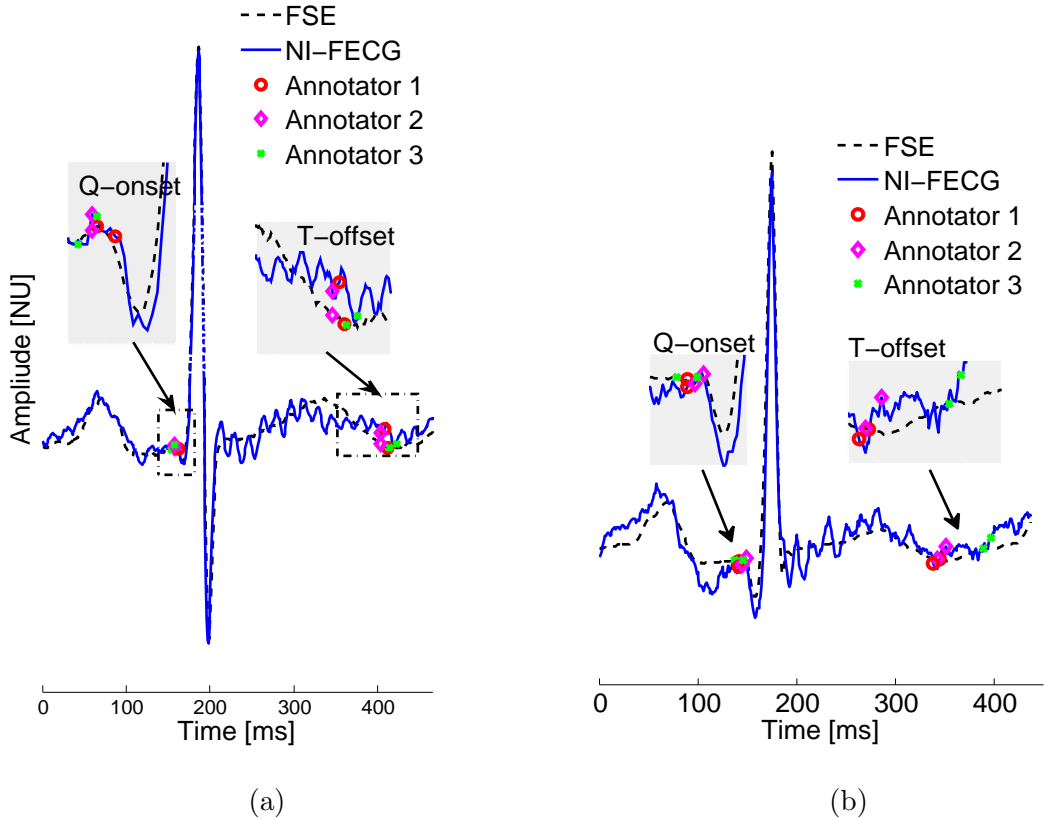

Figure 8.3: Comparison of annotations performed on an average FSE and NI-FECG monitor ECG beat by three experts. (a): note the close correspondence between experts on both the FSE and NI-FECG signal. (b): note the disagreement between annotator 3 and the other two experts. This last example illustrates the importance of combining annotators.

| Method/Stats | RMSE | AE | RMSE (95%) | AE (95%) |
|---|---|---|---|---|
| A1-SET1 | 27.5 | 22.5 | 25.0 | 20.7 |
| A2-SET1 | 41.3 | 32.8 | 35.9 | 29.5 |
| A3-SET1 | 21.6 | <u>17.1</u> | 19.2 | <u>15.5</u> |
| A1-SET2 | 33.2 | 20.3 | 22.0 | 16.3 |
| A2-SET2 | 22.7 | 16.6 | 17.8 | 14.1 |
| A3-SET2 | 18.3 | <u>14.8</u> | 16.2 | <u>13.4</u> |

Table 8.1: Individual annotators (A1-A3) and annotations for EVENT1 and EVENT2. Reference: SQT from annotator A. Measure: AQT from annotator A. 95%: 95% confidence interval. All values are expressed in ms.

| Method/Stats | RMSE | AE | RMSE (95%) | AE (95%) |
|---|---|---|---|---|
| Mean-SET1 | 17.9 | 14.1 | 15.1 | <u>12.4</u> |
| Median-SET1 | 21.3 | 17.1 | 18.7 | 15.5 |
| EM-SET1 | 18.0 | <u>14.2</u> | 15.3 | 12.7 |
| Mean-SET2 | 15.4 | 11.5 | 12.1 | 9.9 |
| Median-SET2 | 18.8 | 14.2 | 15.8 | 12.5 |
| EM-SET2 | 13.6 | <u>10.4</u> | 11.4 | <u>9.2</u> |

Table 8.2: Combining cardiologists annotations to get SQT and AQT for SET1 and SET2. The error is assessed for the mean/median/EM AQT against mean/median/EM SQT approaches for fusing the annotations. 95%: 95% confidence interval. All values are expressed in ms.

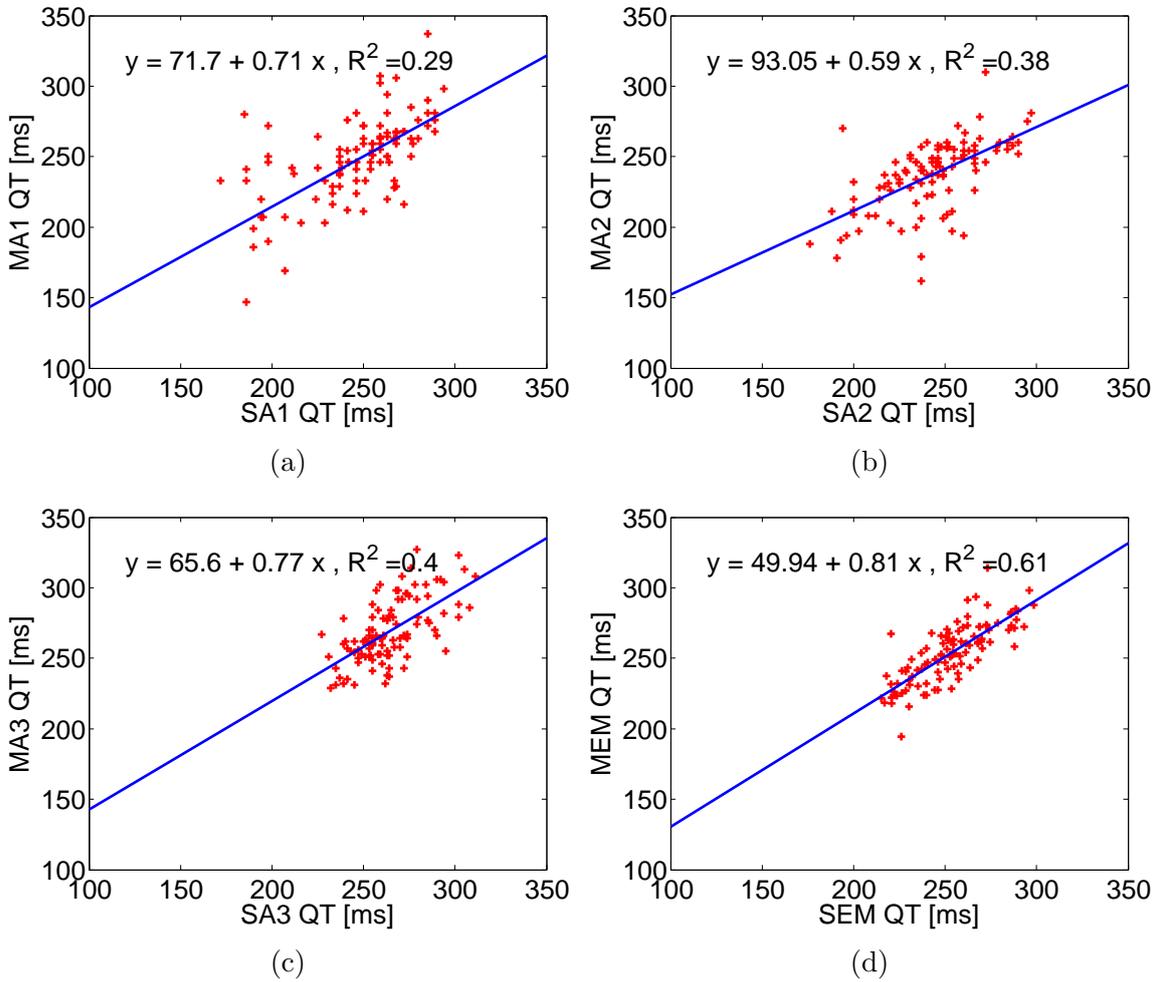

Figure 8.4: Plot of QT annotations from the output NI-FECG from the Meridian monitor against QT annotations from the the SECG signal, 22 fetuses (105, 1-min segments). M: data output from the Meridian monitor, A: annotator, S: scalp data. (e.g. MA1 QT refers to the QT annotated by annotator one on the NI-FEGC output from by the Meridian monitor). EM: crowd sourced annotations from the three clinicians using the EM algorithm (e.g. SEM QT refers to the scalp QT annotations merged using the EM algorithm). Line fit: y = intercept + gradient $\times$ x and $R^2$ is the corresponding coefficient of determination (goodness of fit).

### 8.2.4 Discussion

This is the first study that demonstrates that the FQT interval can be reliably estimated from ECG data recorded non-invasively, using electrodes on the maternal abdomen, by validating these measurements against invasive SECG data.

The annotators generated QT intervals with excellent correlation between the abdominal data and the corresponding FSE signal when averaged waveforms were used. In contrast, when unaveraged waveforms were annotated, the distortion inherent in the waveforms led the annotators to generally identify shorter QT intervals when annotating the NI-FECG than when annotating the FSE signal. These findings suggest that the most accurate approach to QT annotation will be to use a waveform created from a running average of several heart beats. This was confirmed by the quantitative analysis presented in Tables 8.1 and 8.2, where the results for the experiments on SET2 were consistently better than for SET1. Combining the annotations from the three electrophysiologists resulted in improved (i.e. lower) RMSE and AE than when taking each individual annotator. In the case of the experiment on SET2, the EM algorithm gave the best results for fusing the annotations.

The FQT estimation errors obtained in this study (17.9 ms RMSE, SET1 and 13.6 ms RMSE, SET2) compare well to the RMSE obtained when fusing adult QT annotations from three annotators (RMSE of 16.07 ms found in Zhu *et al.* [150]).

A number of published studies have attempted to extract the FQT (and other ECG morphology based quantities) from the NI-FECG or fetal magnetocardiography. However, these studies did not validate their measurements with invasive data: Stinstra *et al.* [151] used magnetocardiography recordings from 582 healthy patients at different stages of pregnancy (gestational age 17-41 weeks) and manually annotated the PR, PQ, QRS and QT intervals, averaging over 100 cardiac cycles per recording. They found the QT length was in the interval [149 339] ms, but did not have SECG data to validate their measurements. Brambati *et al.* [152] used NI-FECG to record 421 pregnant women (17-41 weeks), and performed a similar set of measurements, averaging 50 cardiac cycles per measurement, again without simultaneous measurement of invasive SECG. Two other publications performed similar analyses and reported QT to ranges from [207 338] ms [153] and [197 306] ms [154]. Although these studies did not validate their measurements with invasive data, the QT ranges were similar to the ones obtained in this work. The FQT measured on the scalp data were in the range [192 303] ms when all annotations from each annotator in SET1 were used at the 95% CI (1575 datapoints). The interval was [203 295] ms when fusing individual annotations and annotators for SET1 (105 QT). The interval was [190 295] ms for all individual annotators in SET2 (315 annotations, 95% CI) and [215 299] ms after fusing the annotators for SET 2 (105 annotations).

Overall, the FQT range at delivery was found to be in the range 200-300 ms. When removing the extreme values (by using a 95% CI or by elaborating the statistics fusing multiple annotators), these ranges matched with other studies on human annotations of FQT intervals [151, 152, 154]. The work presented in this section showed that it is possible to recover the FQT from extracted NI-FECG in the source domain using the output of a commercial monitor. Of course, the results need to be confirmed on a larger data set, for a range of pathologies such as screening for long QT syndrome. (None of the subjects in this chapter had electrocardiographic abnormalities.) This is because there is a risk that the signals obtained with ICA (or any alternative BSS transform) may not preserve the FECG morphology while moving into the source domain.

The study presented here is unique in that it validated the non-invasive measurements for each subject with invasive scalp data. The scalp data are not overly prone to potential artifact or error due to the extraction process. Therefore measurements performed on SECG can be used as ground truth.

Similar to prior studies, the QT interval was most accurately measured by averaging a series of cardiac cycles (SET2). This created a high quality ECG averaged waveform, and increased the signal to noise ratio. However, the question which needs to be asked is how much 'averaging' should be allowed, given that the ECG is a non-stationary signal and that the purpose of these measurements is to track changes over time of a given morphological quantity.

In Sameni's thesis [9], Chapter 6, the author raised the question of the influence of a PCA or ICA transformation on the extracted FECG morphology. The author used a subset of signals from one 12-lead ECG, and evaluated the mean square error between manually annotated P, QRS and T waves on the original data and the data that were transformed using a BSS step, followed by dimensionality reduction and finally back projected onto the original basis. The author showed that, provided enough components were kept before performing the back projection step, the morphology (as quantified by the MSE) could be recovered. Although insightful, this experiment was limited in that it did not quantify if a given interval (e.g. QT) could be accurately recovered (c.f. discussion on the difference between using a MSE and some physiological interval in Section 8.3) and did not address whether these physiological intervals could be measured in the source domain directly (i.e. without having to project the data back in the observation domain).

## 8.3 Dual extended Kalman filter

### 8.3.1 Introduction

The ECG Bayesian filtering framework was originally introduced by Sameni *et al.* in 2008 [9] for its application to noise cancellation in the adult ECG. It was later used in the context of NI-FECG extraction, beat segmentation [81, 155, 156], signal compression [157], denoising of the ECG recorded in an MRI [158, 159] and detection of premature ventricular contractions [82] (see Appendix A). The underlying strength of the framework is that it relies on an ECG dynamical model [4] that is used as a prior on the ECG morphology. As such, the ECG Kalman filter framework is a good tool to estimate the MECG and FECG morphology in an online or recursive fashion. The framework is also valuable since it allows to use a prior on the MECG and FECG morphology (through the dynamical model); as such it is suited to NI-FECG morphological extraction.

Historically, the work presented in this thesis was developed in parallel to the work of Niknazar *et al.* [149] from the GIPSA laboratory in Grenoble, France. In their work, the authors presented an extension of the EKF framework that incorporated the prior on the FECG morphology. The main idea behind the dual extended Kalman filter, presented in this chapter, is similar to the work of Niknazar *et al.* in that the intention is to incorporate a prior on the morphology of the FECG within the ECG Bayesian filtering framework. However, there are some conceptual differences in how the equations were formulated and this will be highlighted in the discussion (Section 8.3.4).

The work presented in this section has been partially published in [10]. More explanation on the original EKF framework [9] can be found in Appendix A.

### 8.3.2 Methods

A Bayesian filtering framework based on an Extended Kalman Filter (EKF) for extracting the FECG from a single abdominal channel is described. It is evaluated on a training database of 20, one-minute maternal-fetal mixtures and 200, one-minute mixtures as test signals. The overall database (training and test set) is called the AQTDB (and is described in detail in Chapter 4). Data is generated using the simulator, *fecgsyn* (see the description of the simulator in Chapter 4). Background on the ECG Bayesian filtering framework can be found in Chapter 3 and in more detail in Appendix A.

#### 8.3.2.1 Design considerations

The design specifications for deciding on the formulation of the equations were as follows:

- the ECG model should be able to simultaneously extract both the MECG and FECG morphology and as such require two priors (one on the MECG and one on the FECG morphology)

- since the application of this filter is targeted at morphological analysis (i.e. tracking clinically relevant intervals such as the QT segment) over time, the Gaussian parameters of the FECG were left to evolve i.e. $\alpha_{i,k}^f, b_{i,k}^f, \xi_{i,k}^f$ were considered state variables following a random walk with perturbation term $\varepsilon$

- the filter must require no or a minimal number of user inputs, which means that the Gaussian initialisation has to be automated

- the resulting filter must exhibit a stable behaviour.

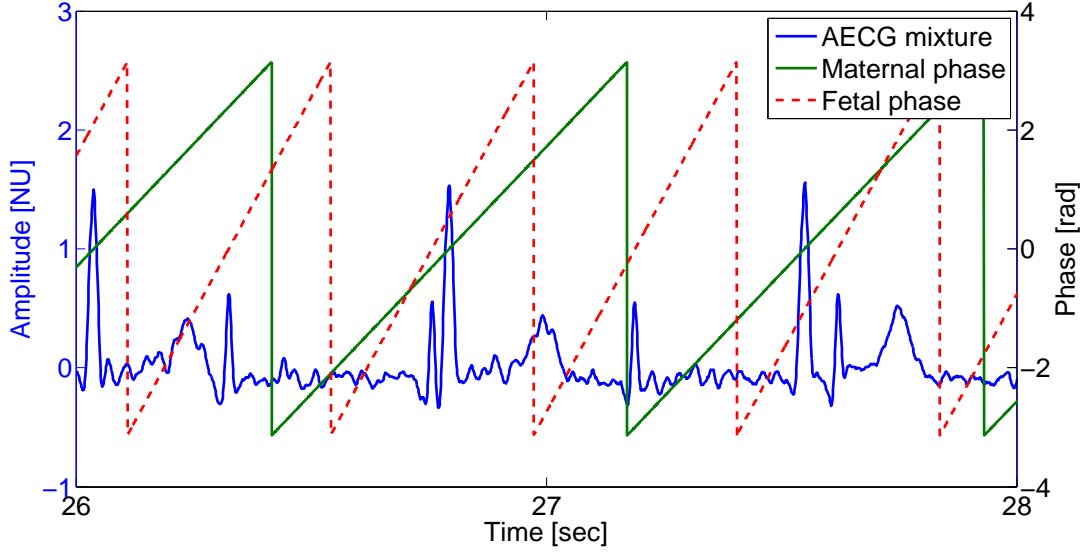

Figure 8.5: EKFD phases: two phases are used within the EKFD Bayesian filtering framework; one for the MECG ($\varphi^m$, in green) and one for the FECG ($\varphi^f$, in dotted red). Both phases are used as observations. AECG: abdominal ECG.

Considering these constraints the Dual Extended Kalman Filter (EKFD) is introduced:

**State-space equations** ECG template cycle construction and Gaussian fitting were performed as detailed in Chapter 3. Following the design considerations specified above, the variables of the optimal state estimation problem were defined as:

$$\begin{aligned}
\underline{\mathbf{x}}_k &= [\theta_k^f, \theta_k^m, f_k, m_k, \{\alpha_{i,k}^f\}, \{b_{i,k}^f\}, \{\xi_{i,k}^f\}], \forall i \in [1\ N] \\
\underline{\mathbf{y}}_k &= [\varphi_k^f, \varphi_k^m, s, s] \\
\underline{\mathbf{w}}_k &= [\omega^f, \omega^m, \eta_k^f, \eta_k^m, \{\varepsilon_{\alpha,i}\}, \{\varepsilon_{b,i}\}, \{\varepsilon_{\xi,i}\}, \{\alpha_{i,k}^m\}, \{b_{i,k}^m\}, \{\xi_{i,k}^m\}], \forall i \in [1\ N] \\
\underline{\mathbf{v}}_k &= [v_{1,k}, v_{2,k}, v_{3,k}, v_{4,k}],
\end{aligned} \quad (8.1)$$

where, $N$ is the number of Gaussians used to map the averaged ECG cycle, $\underline{\mathbf{x}}_k \in \Re^{3N+4}$ is the state variable vector at time instant $k$, $\underline{\mathbf{y}}_k \in \Re^4$ is the observation vector, $\underline{\mathbf{w}}_k \in \Re^{6N+4}$ is the process noise vector and $\underline{\mathbf{v}}_k \in \Re^4$ is the observation noise vector. Figure 8.5 illustrates the observations used; the maternal phase $\varphi^m$, the fetal phase $\varphi^f$ and the AECG mixture amplitude $s$. The corresponding state equations for the EKFD are:

$$\begin{cases} \theta_{k+1}^f \equiv (\theta_k^f + \omega^f \delta) \mod 2\pi \\ \theta_{k+1}^m \equiv (\theta_k^m + \omega^m \delta) \mod 2\pi \\ f_{k+1} = f_k - \sum_{i=1}^{N} \delta \frac{\alpha_i^f \omega^f}{(b_i^f)^2} \Delta\Theta_{i,k} exp(-\frac{\Delta\Theta_{i,k}^2}{2(b_i^f)^2}) + \eta_k^f \\ m_{k+1} = m_k - \sum_{i=1}^{N} \delta \frac{\alpha_i^m \omega^m}{(b_i^m)^2} \Delta\theta_{i,k} exp(-\frac{\Delta\theta_{i,k}^2}{2(b_i^m)^2}) + \eta_k^m \\ \alpha_{i,k+1}^f = \alpha_{i,k}^f + \varepsilon_{\alpha,i} \\ b_{i,k+1}^f = b_{i,k}^f + \varepsilon_{b,i} \\ \xi_{i,k+1}^f = \xi_{i,k}^f + \varepsilon_{\xi,i}. \end{cases} \quad (8.2)$$

The corresponding observation equations are:

$$\begin{cases} \varphi_k^f = \theta_k^f + v_{1,k} \\ \varphi_k^m = \theta_k^m + v_{2,k} \\ s_k = m_k + f_k + v_{3,k} \\ s_k = m_k + \sum_{i=1}^{N} \alpha_{i,k}^f \exp(-\frac{\Delta\Theta_{i,k}^2}{2(b_i^f)^2}) + v_{4,k}, \end{cases} \quad (8.3)$$

where the superscripts $f$ and $m$ stand for fetal and maternal respectively, $\Delta\theta_{i,k} = \theta_k - \xi_i^m$ and $\Delta\Theta_{i,k} = \theta_k - \xi_i^f$; $\varphi_k$ correspond to the observed phases and $s_k$ to the observed amplitude of the ECG; $f_k$ and $m_k$ correspond to the fetal and maternal amplitudes which were considered as state variables. Since a special interest is given to morphological changes (quantities such as the QT interval) over time, the Gaussian parameters of the FECG were allowed to evolve, i.e. $\alpha_{i,k}^f, b_{i,k}^f, \xi_{i,k}^f$ were considered as state variables following a random walk with a perturbation term $\varepsilon$. In order to limit computation time, the Gaussian parameters of the MECG were not considered as state variables. The observation equations contain information on the phase (the 'true' phase is equal to the observed phase with some noise) and on the amplitude (the sum of the amplitude between the FECG and the MECG correspond to the observed amplitude plus some noise). However, in the amplitude equation (3rd equation in the system of equations 8.3) there is no discrimination between the contributions of the maternal and fetal ECGs. Take for example a virtual increase of +0.1mV for the MECG and -0.1mV for the FECG then these will sum to zero and the 3rd equation (in the system of equations 8.3) will still be satisfied despite the estimation of the MECG and FECG will have drifted by ±0.1mV. This phenomenon was observed empirically as shown in Figure 8.10. To fix it, a 4th observation equation was introduced (to the system of equations 8.3) in which a synthetic signal was created by subtracting the prior on the FECG signal to the raw observed ECG mixture. The fourth equation can thus be read as: $s_k - \sum ... = m_k + v_{4,k}$ i.e. MECG is equal to the observed ECG mixture minus the prior on the FECG contribution. This equation was thus added to act as a virtual observation of the MECG to help stabilise the filter, similar to what was suggested by Oster *et al.* [160] in the context of filtering out the MagnetoHydroDynamic effect induced by the Magnetic Resonance Imaging (MRI) machine on the ECG. The following assumptions were also made: (1) The FQRS and MQRS locations are known; (2) Seven Gaussians describe the template ECG cycle[1].

**Linearisation** The KF formalism only applies to linear systems. In practice, systems are rarely linear or if so only in a given range [161]. Many non-linear systems can be piecewise approximated by a linear system with satisfactory results. In order to use the KF framework it is necessary to linearise the state and observation equations (see Appendix A for more details).

**Linearising state equations** The model equations $\underline{\mathbf{x}}_{k+1} = g_k(\underline{\mathbf{x}}_k, \underline{\mathbf{w}}_k)$ were linearised using the Taylor series expansion around $(\hat{\underline{\mathbf{x}}}_k, \hat{\underline{\mathbf{w}}}_k)$ in order to use the KF framework:

$$\underline{\mathbf{x}}_{k+1} = g_k(\underline{\mathbf{x}}_k, \underline{\mathbf{w}}_k) \approx g_k(\hat{\underline{\mathbf{x}}}_k, \hat{\underline{\mathbf{w}}}_k) + \mathbf{G}_k(\underline{\mathbf{x}}_k - \hat{\underline{\mathbf{x}}}_k) + \mathbf{L}_k(\underline{\mathbf{w}}_k - \hat{\underline{\mathbf{w}}}_k) \quad (8.4)$$

with $\mathbf{G}_k \in \Re^{(3N+4)\times(3N+4)}$ and $\mathbf{L}_k \in \Re^{(3N+4)\times(6N+4)}$,

---
[1] This number of Gaussian was motivated in Sayadi *et al.* [82]

$$\mathbf{G}_k = \left.\frac{\partial g}{\partial \underline{\mathbf{x}}}\right|_{\hat{\underline{\mathbf{x}}}_k^+, \hat{\underline{\mathbf{w}}}_k} = \begin{bmatrix} \dfrac{\partial g_1}{\partial x_1} & \dfrac{\partial g_1}{\partial x_2} & \cdots & \dfrac{\partial g_1}{\partial x_{3N+4}} \\ \vdots & & & \vdots \\ \dfrac{\partial g_{3N+4}}{\partial x_1} & \dfrac{\partial g_{3N+4}}{\partial x_2} & \cdots & \dfrac{\partial g_{3N+4}}{\partial x_{3N+4}} \end{bmatrix}$$

$$= \begin{bmatrix}
1 & 0 & \cdots & & & & 0 \\
0 & 1 & 0 & \cdots & & & 0 \\
g_{3,1} & 0 & 1 & 0 & g_{3,5} & \cdots & g_{3,3N+1} \\
0 & g_{4,2} & 0 & 1 & 0 & \cdots & 0 \\
0 & \cdots & & 0 & 1 & 0 & \cdots \\
0 & \cdots & & & 0 & \ddots & 0 \\
0 & \cdots & & & 0 & \cdots & 1
\end{bmatrix}$$

(8.5)

$$\mathbf{L}_k = \left.\frac{\partial g}{\partial \underline{\mathbf{w}}}\right|_{\hat{\underline{\mathbf{x}}}_k^+, \hat{\underline{\mathbf{w}}}_k} = \begin{bmatrix} \dfrac{\partial g_1}{\partial w_1} & \dfrac{\partial g_1}{\partial w_2} & \cdots & \dfrac{\partial g_1}{\partial w_{6N+4}} \\ \vdots & & & \vdots \\ \dfrac{\partial g_{3N+4}}{\partial w_1} & \dfrac{\partial g_{3N+4}}{\partial w_2} & \cdots & \dfrac{\partial g_{3N+4}}{\partial w_{6N+4}} \end{bmatrix}$$

$$= \begin{bmatrix}
\delta & 0 & \cdots & & & & & & 0 \\
0 & \delta & 0 & \cdots & & & & & 0 \\
l_{3,1} & 0 & 1 & 0 & \cdots & & & & 0 \\
0 & l_{4,2} & 0 & 1 & 0 & \cdots & l_{4,3N+5} & \cdots & l_{4,6N+4} \\
0 & \cdots & & 0 & 1 & 0 & \cdots & & 0 \\
0 & \cdots & & & 0 & \ddots & 0 & \cdots & 0 \\
0 & \cdots & & & & 0 & 1 & 0 & \cdots & 0
\end{bmatrix}$$

with

$$\begin{aligned}
g_{3,1}[k] &= \left.\frac{\partial g_3}{\partial \theta^f}\right|_{\hat{\underline{x}}_k^+,\hat{\underline{w}}_k} = \sum_{i\in\{f_1,...,f_N\}} \frac{\delta\alpha_{i,k}^f \omega^f}{(b_{i,k}^f)^2}[1-\frac{\Delta\Theta_{i,k}^2}{(b_{i,k}^f)^2}]exp(-\frac{\Delta\Theta_{i,k}^2}{2(b_{i,k}^f)^2}) \\
g_{3,i=5:N+4}[k] &= \left.\frac{\partial g_3}{\partial \alpha_i^f}\right|_{\hat{\underline{x}}_k^+,\hat{\underline{w}}_k} = -\delta\frac{\omega^f \Delta\Theta_{i,k}}{(b_{i,k})^2}exp(-\frac{\Delta\Theta_{i,k}^2}{2(b_{i,k}^f)^2}) \\
g_{3,i=N+5:2N+4}[k] &= \left.\frac{\partial g_3}{\partial b_i^f}\right|_{\hat{\underline{x}}_k^+,\hat{\underline{w}}_k} = \delta\alpha_{i,k}^f \frac{\omega^f \Delta\Theta_{i,k}}{(b_{i,k}^f)^3}[2-\frac{\Delta\Theta_{i,k}^2}{(b_{i,k}^f)^2}]exp(-\frac{\Delta\Theta_{i,k}^2}{2(b_{i,k}^f)^2}) \quad (8.6)\\
g_{3,i=2N+5:3N+4}[k] &= \left.\frac{\partial g_3}{\partial \xi_i^f}\right|_{\hat{\underline{x}}_k^+,\hat{\underline{w}}_k} = \delta\alpha_{i,k}^f \frac{\omega^f}{(b_{i,k}^f)^2}[1-\frac{\Delta\Theta_{i,k}^2}{(b_{i,k}^f)^2}]exp(-\frac{\Delta\Theta_{i,k}^2}{2(b_{i,k}^f)^2}) \\
g_{4,2}[k] &= \left.\frac{\partial g_4}{\partial \theta^m}\right|_{\hat{\underline{x}}_k^+,\hat{\underline{w}}_k} = \sum_{i\in\{m_1,...,m_N\}} \frac{\delta\alpha_{i,k}^m \omega^m}{(b_{i,k}^m)^2}[1-\frac{\Delta\theta_{i,k}^2}{(b_{i,k}^m)^2}]exp(-\frac{\Delta\theta_{i,k}^2}{2(b_{i,k}^m)^2})
\end{aligned}$$

and

$$\begin{aligned}
l_{4,i=3N+5:4N+4}[k] &= \left.\frac{\partial g_4}{\partial \alpha_i^m}\right|_{\hat{\underline{x}}_k^+,\hat{\underline{w}}_k} = -\delta\frac{\omega^m \Delta\theta_{i,k}}{(b_{i,k}^m)^2}exp(-\frac{\Delta\theta_{i,k}^2}{2(b_{i,k}^m)^2}) \\
l_{4,i=4N+5:5N+4}[k] &= \left.\frac{\partial g_4}{\partial b_i^m}\right|_{\hat{\underline{x}}_k^+,\hat{\underline{w}}_k} = \delta\alpha_{i,k}^m \frac{\omega^m \Delta\theta_{i,k}}{(b_{i,k}^m)^3}[2-\frac{\Delta\theta_{i,k}^2}{(b_{i,k}^m)^2}]exp(-\frac{\Delta\theta_{i,k}^2}{2(b_{i,k}^m)^2}) \\
l_{4,i=5N+5:6N+4}[k] &= \left.\frac{\partial g_4}{\partial \xi_i^m}\right|_{\hat{\underline{x}}_k^+,\hat{\underline{w}}_k} = \delta\alpha_{i,k}^m \frac{\omega^m}{(b_{i,k}^m)^2}[1-\frac{\Delta\theta_{i,k}^2}{(b_{i,k}^m)^2}]exp(-\frac{\Delta\theta_{i,k}^2}{2(b_{i,k}^m)^2}) \quad (8.7)\\
l_{4,2}[k] &= \left.\frac{\partial g_4}{\partial \omega_i^m}\right|_{\hat{\underline{x}}_k^+,\hat{\underline{w}}_k} = -\sum_{i\in\{m_1,...,m_N\}} \frac{\delta\alpha_{i,k}^m \Delta\theta_{i,k}}{(b_{i,k}^m)^2}exp(-\frac{\Delta\theta_{i,k}^2}{2(b_{i,k}^m)^2}) \\
l_{3,1}[k] &= \left.\frac{\partial g_3}{\partial \omega_i^f}\right|_{\hat{\underline{x}}_k^+,\hat{\underline{w}}_k} = -\sum_{i\in\{f_1,...,f_N\}} \frac{\delta\alpha_{i,k}^f \Delta\Theta_{i,k}}{(b_{i,k}^f)^2}exp(-\frac{\Delta\Theta_{i,k}^2}{2(b_{i,k}^f)^2})
\end{aligned}$$

**Linearising space equations** The space equations $h_k(\underline{x}_k,\underline{v}_k)$ were linearised in order to use the EKF framework:

$$\underline{y}_{k+1} = h(\underline{x}_k,\underline{v}_k) \approx h_k(\hat{\underline{x}}_k,\hat{\underline{v}}_k) + \mathbf{H}_k(\underline{x}_k - \hat{\underline{x}}_k) + \mathbf{U}_k(\underline{v}_k - \hat{\underline{v}}_k) \quad (8.8)$$

with $\mathbf{H}_k \in \Re^{(4\times 3N+4)\times(3N+4)}$ and $\mathbf{U}_k \in \Re^{4\times 4}$,

$$\mathbf{H}_k = \left.\frac{\partial h}{\partial \underline{\mathbf{x}}}\right|_{\hat{\underline{\mathbf{x}}}_k^+,\hat{\underline{\mathbf{w}}}_k} = \begin{bmatrix} 1 & 0 & \cdots & & & & 0 \\ 0 & 1 & 0 & \cdots & & & 0 \\ 0 & 0 & 1 & 1 & 0 & \cdots & 0 \\ h_{4,1} & 0 & 0 & 1 & h_{4,5} & \cdots & h_{4,3N+4} \end{bmatrix} \quad (8.9)$$

$$\mathbf{U}_k = \left.\frac{\partial h}{\partial \underline{\mathbf{v}}}\right|_{\hat{\underline{\mathbf{x}}}_k^+,\hat{\underline{\mathbf{w}}}_k} = \begin{bmatrix} 1 & 0 & 0 & 0 \\ 0 & 1 & 0 & 0 \\ 0 & 0 & 1 & 0 \\ 0 & 0 & 0 & 1 \end{bmatrix} = \mathbf{I}_{4\times 4} \quad (8.10)$$

with

$$\begin{aligned}
h_{4,1}[k] &= \left.\frac{\partial h}{\partial \theta^f}\right|_{\hat{\underline{\mathbf{x}}}_k^+,\hat{\underline{\mathbf{w}}}_k} = -\sum_{i=1}^{N} \alpha_{i,k}^f \frac{\Delta\Theta_{i,k}}{(b_{i,k}^f)^2} exp(-\frac{\Delta\Theta_{i,k}^2}{2(b_{i,k}^f)^2}) \\
h_{4,i=5:N+4}[k] &= \left.\frac{\partial h}{\partial \alpha_i^f}\right|_{\hat{\underline{\mathbf{x}}}_k^+,\hat{\underline{\mathbf{w}}}_k} = exp(-\frac{\Delta\Theta_{i,k}^2}{2(b_{i,k}^f)^2}) \\
h_{4,i=N+5:2N+4}[k] &= \left.\frac{\partial h}{\partial b_i^f}\right|_{\hat{\underline{\mathbf{x}}}_k^+,\hat{\underline{\mathbf{w}}}_k} = \alpha_{i,k}^f \frac{\Delta\Theta_{i,k}^2}{(b_{i,k}^f)^3} exp(-\frac{\Delta\Theta_{i,k}^2}{2(b_{i,k}^f)^2}) \\
h_{4,i=2N+5:3N+4}[k] &= \left.\frac{\partial h}{\partial \xi_i^f}\right|_{\hat{\underline{\mathbf{x}}}_k^+,\hat{\underline{\mathbf{w}}}_k} = \alpha_{i,k}^f \frac{\Delta\Theta_{i,k}}{(b_{i,k}^f)^2} exp(-\frac{\Delta\Theta_{i,k}^2}{2(b_{i,k}^f)^2}).
\end{aligned} \quad (8.11)$$

**Protocol**

Figure 8.6 shows the block diagram of the algorithm denoted EKFD: (1) one abdominal ECG channel is prefiltered (pass band [0.7 100] Hz); (2) the single EKF (EKFS [162]) was applied in order to remove the MECG contribution to the mixture; (3) the template FECG was built based on the residual of the EKFS step; (4) the EKFD was applied to separate the three components of the mixture (MECG, FECG and noise); (5) a postfiltering step (pass band [0.7 100] Hz using two cascaded Butterworth filters with zero phase) was applied, (6) finally the signal to noise ratio (SNR) was computed. The SNR in dB, between a reference $r$

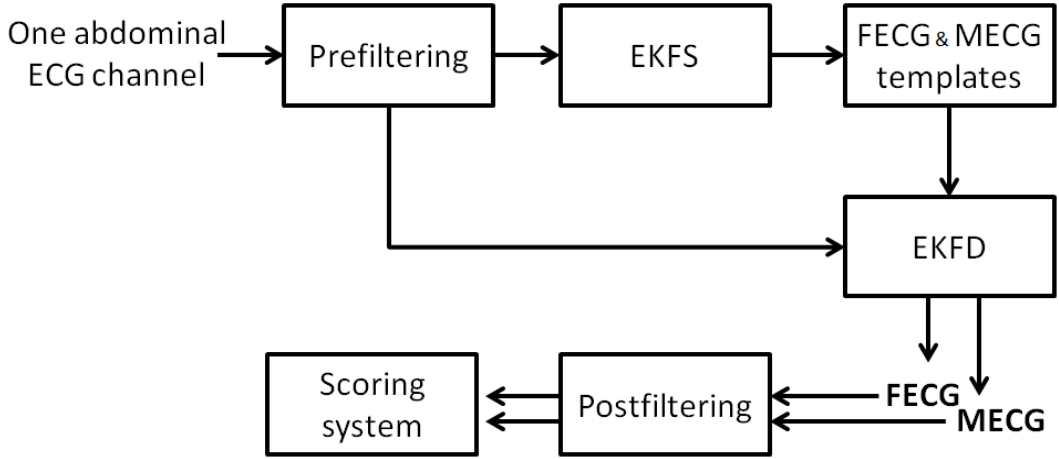

Figure 8.6: Block diagram of the NI-FECG extraction procedure using the EKFD.

and an extracted signal $f$ with $K$ samples was defined as:

$$SNR = 20 \cdot log \left( \sqrt{\sum_{i=1}^{K} r_i^2 / \sum_{i=1}^{K} (f_i - r_i)^2} \right) \qquad (8.12)$$

The first and last five seconds of the records were not included in the SNR evaluation to account for border effects.

In addition, benchmarking was performed against the EKFS, i.e. taking the residual of the first EKF pass to be the FECG. Benchmarking was also performed against EKFSS which corresponds to two single pass of EKF, one for removing the MECG and taking the residual as being the FECG, and another single pass on the FECG residual in order to clean the signal. Finally, benchmarking was performed against the EKF model introduced in Niknazar *et al.* [149] (denoted EKFN).

A random search [123] was performed on the training set to optimise the values of the entries of the observation and process noise covariance matrices. One hundred random search iterations were performed for the EKFD (ten free parameters), 90 random iterations for the EKFN (nine free parameters) and 50 for the EKFSS (five free parameters). Mean and median SNR across all training and test records were reported for the optimised parameters.

In addition to the SNR computation, automated QT measurement was performed using the open source *ecgpuwave* software [163] ran on the EKFS, EKFN and EKFD filtered signals. *ecgpuwave* was run on the 1 min segments and QT intervals extracted from the interval 5-55 sec were kept (to account for possible edge effects). Since *ecgpuwave* was not designed for fetal data and is imperfect, records with at least three successful measures on the reference FECG and EKFD and EKFS were kept. For each filtering method, the QT interval estimate was taken to be the median of the individual QT measurements on all beats in the 50 sec interval taken into account in the recording.

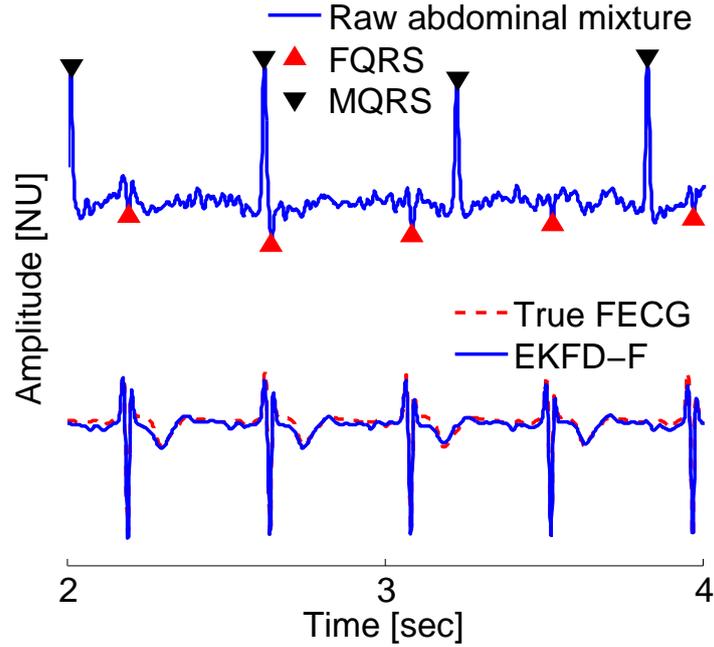

Figure 8.7: From top to bottom: abdominal mixture, EKFD pass on top of the true FECG. Note that the FECG morphology is reconstructed well using the EKFD.

### 8.3.3 Results

Tables 8.3 and 8.4 present the quantitative results of the analysis and Figures 8.7 and 8.8 give a qualitative example of the algorithm performances. The EKFD SNR on the test set was 14.14 dB vs. 1.97 dB for the EKFS, 8.28 dB for the EKFSS and 13.01 dB for the EKFN.

A QT measure could be extracted from a total of 196 files of the test set (out of the 200) on the reference FECG, EKFS, EKFN and EKFD outputs. QT estimates from the EKFD and EKFS were compared to the true QT (see Figure 8.9). AE and RMSE errors were (19.8 ms, 24.6 ms) for the EKFS, (8.5 ms, 14.2 ms) for EKFN and (7.6 ms, 12.7 ms) for EKFD. The EKFD had a smaller bias than the EKFS and EKFN, with a slope closer to unity, an offset closer to zero and a higher goodness of fit ($R^2 = 0.64$).

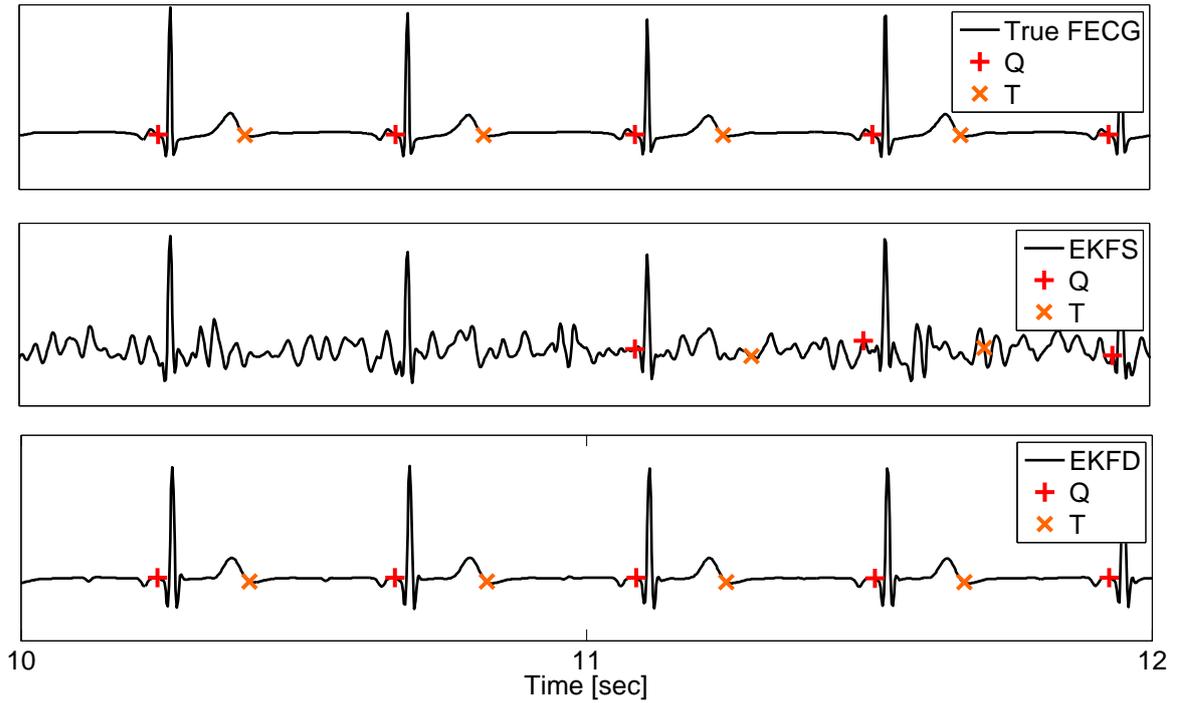

Figure 8.8: QT detection using *ecgpuwave*. This plot highlights the added value of using a framework that has a prior on both the MECG and FECG morphology (EKFD or EKFN) compared to performing morphological analysis on the residual signal from the EKFS (or equivalently any alternative methods reviewed in Chapter 3 that would consider the FECG as noise).

Table 8.3: Measured QT after source separation against true QT. EKFS: single EKF extracted ECG, EKFSS: ECG extracted after a second pass of EKFS, EKFD: dual EKF. M: maternal, F: fetal. SNR is in decibels.

|         | Train |  | Test |  |
| --- | --- | --- | --- | --- |
| Params | SNR (mean) | SNR (med) | SNR (mean) | SNR (med) |
| EKFS-M | 14.64 | 14.86 | 16.53 | 16.70 |
| EKFN-M | 18.89 | 18.74 | 16.28 | 15.98 |
| EKFD-M | 16.91 | 16.56 | 14.96 | 14.51 |
| EKFS-F | 1.50 | 2.95 | 1.32 | 1.97 |
| EKFSS-F | 8.86 | 9.42 | 7.74 | 8.28 |
| EKFN-F | 14.96 | 15.47 | 13.00 | 13.01 |
| EKFD-F | 15.06 | 16.43 | 13.71 | 14.14 |

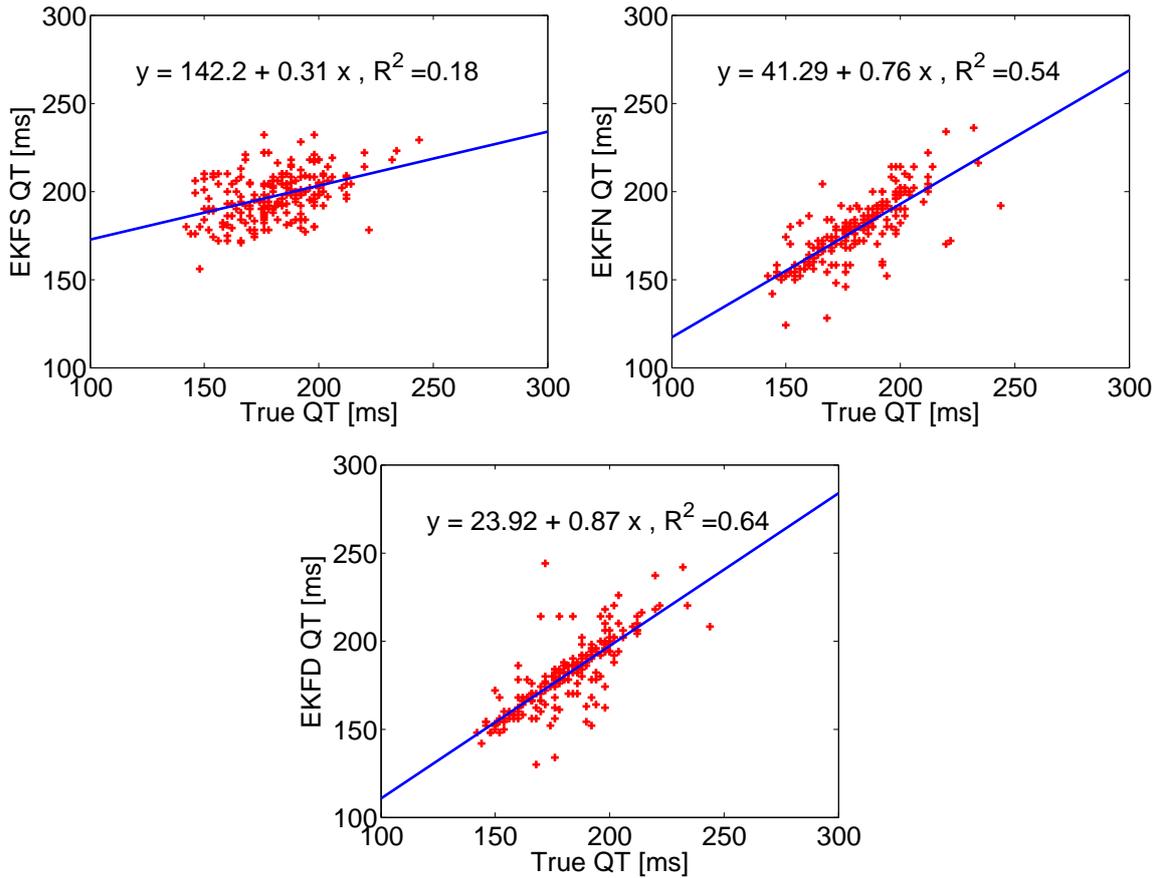

Figure 8.9: Plot of QT extracted from the EKFS (a), EKFN (b) and EKFD (c) vs. the true QT for 196 1 min artificial segments. The blue line corresponds to the linear regression (y = intercept + gradient × x) and $R^2$ is the corresponding coefficient of determination (goodness of fit).

### 8.3.4 Discussion

The EKFD method relies on accurate and precise FQRS and MQRS detection. Methods such as the ones evaluated in Chapters 6 and 7 can be used for that purpose. The residual signal from any of these methods can then be used (instead of the single-pass EKFS) to build the template FECG and use it as a prior for the EKFD. It is important to note that the frequency cut-off $f_b$, was taken to be 0.7 Hz, unlike in Chapter 6 and 7 where $f_b = 10$ Hz was used. Taking too high a frequency cut-off to remove baseline wander affects the T-wave morphology, and consequently the QT measurement.

Ultimately, the ability of an algorithm to extract an accurate FECG morphology should be assessed in terms of clinically significant parameters, such as QT segment length and ST elevation. This is because RMS based similarity measure gives an overall 'picture' of the extraction, but provides no insights into whether it is possible or not to recover clinically important measures from the extracted NI-FECG (which is the ultimate application).

Table 8.4: Absolute error (AE) and root mean square error (RMSE) on QT measurement using EKFS, EKFN, EKFD separation methods.

|      | AE (ms) | RMSE (ms) |
|------|---------|-----------|
| EKFS | 19.8    | 24.6      |
| EKFN | 8.5     | 14.2      |
| EKFD | 7.6     | 12.7      |

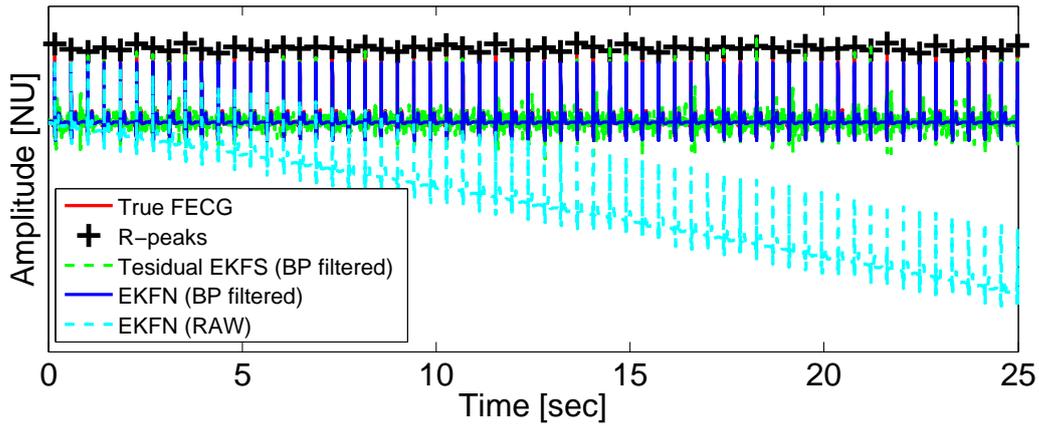

(a)

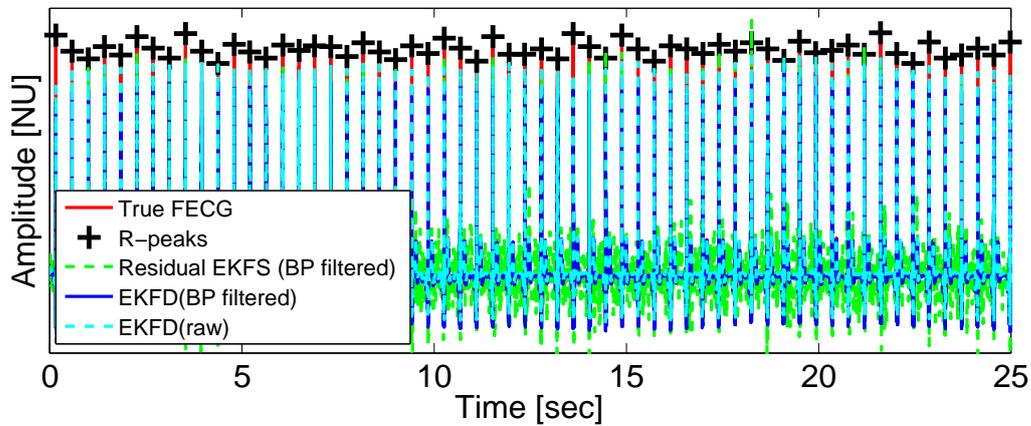

(b)

Figure 8.10: Instability in the EKFN that was corrected by reformulation of the equations in the EKFD framework. (a) the filtered output of the EKFN starts drifting if no post-filtering is performed; (b) no drifting when using the EKFD on the same example.

The EKFD performed better than the EKFS, EKFSS and EKFN, and allowed more accurate QT measurement than the EKFS and EKFN. Although it improved only slightly over EKFN, it was noticeable that the set of equations used in EKFN resulted in a NI-FECG with a baseline drift, whose behaviour is not expected for a stable filter (and for this particular application). This is highlighted in Figure 8.10, where the EKFN filtered FECG linearly drifts from the baseline. (However, this drift is not systematically linear). This phenomenon is inconvenient both for morphological analysis (by introducing some unwanted modulation) and for long records where this could lead to signal saturation. The effect of this phenomenon was minimised by the postfiltering step (see Figure 8.6) that removed the drift (by bandpass filtering the extracted signals) and the limited size of the records (one-minute long). By adding the fourth observation equation this phenomenon was not observed in EKFD, resulting in a more stable set of equations.

During the development of the EKFD, and because of the automated initialisation of the Gaussian parameters, it was noted that in some instances the filtered signals exhibited some singularities (spikes). Although this behaviour might remain unnoticed when looking at the postfiltered signal or summary statistics (which it may not affect substantially), it might sometime cause the system to fail and clearly highlights some internal instabilities of the filter that need to be fixed. Investigation of this behaviour showed that this happened in instances where the standard deviation of some of the Gaussian functions was too small. This seems to be due to some of the EKF propagation equations in the case where $\Delta\Theta_{i,k}$ was very close to zero and $b_i$ (i.e. the standard deviation of the corresponding Gaussian) was very small. Constraining the minimum value of the standard deviation during the Gaussian fitting was thus required in order to limit this effect. In future work it will be important to consider elaborating a Gaussian fitting function that evaluates the number of Gaussian (i.e. non-necessarily seven) that are required to best fit an ECG cycle morphology while not getting instances of Gaussian which can be problematic such as a low standard deviation or Gaussians in opposite orientation (i.e. of opposite sign and mostly compensating each other thus not contributing much to the representation of the ECG cycle).

In conclusion, a SNR improvement of 1.97 dB after EKFS and 14.14 dB after EKFD on the test set were achieved. AE and RMSE for FQT evaluation were (19.8 ms, 24.6 ms) for the EKFS and (7.6 ms, 12.7 ms) for the EKFD. This work is a proof of concept that accurate extraction of the FECG morphology from abdominal recordings using the EKFD is possible. The method is fully automated and can be run online in real-time with relatively low processing time and initialisation (to determine the initial average beat morphology). The EKFD opens the research route for beat-to-beat morphological analysis of the NI-FECG. However, evaluation on real data is required and is presented in the following section.

## 8.4 Recovering the FQT in the observation domain

### 8.4.1 Introduction

In this section, the accurate extraction of the FQT from the EKFD extracted NI-FECG is studied by comparing manual annotations performed on the output of the EKFD and the SECG (considered as ground truth).

### 8.4.2 Methods

The RQTDB, presented in Chapter 4, was used to evaluate the EKFD on real data. All abdominal channels were processed individually following the framework depicted in the block diagram in Figure 8.6. The parameters obtained by the random search procedure on the training set of the AQTDB (in Section 8.3) were used for the RQTDB. One abdominal channel was manually selected and one manual QT annotation was performed for each 60 sec segment output by the EKFD for each record of the RQTDB. Manual selection, by the author, of the channel was based on visually inspecting the template ECG to determine if a T wave was identifiable to the human eye.

### 8.4.3 Results

Figure 8.11 shows an example of filtered real abdominal ECG using the EKFD. One can clearly observe the extracted MECG and FECG output by the EKFD. An overall mean absolute error of 13.0 ms and RMSE of 17.3 ms between annotated and reference QT was achieved. It was not possible to successfully annotate one of the records because only one abdominal channel contained some ECG. Another record was not processed as no FECG could be identified by eye on the abdominal channels, making the exact alignment of the peaks with the FQRS time series extracted from the SECG impossible. The goodness of fit, between QT annotated on the EKFD output and the true QT, was $R^2 = 0.6$, a bias of 15.4 ms and a slope close to one (0.94) - See Figure 8.12.

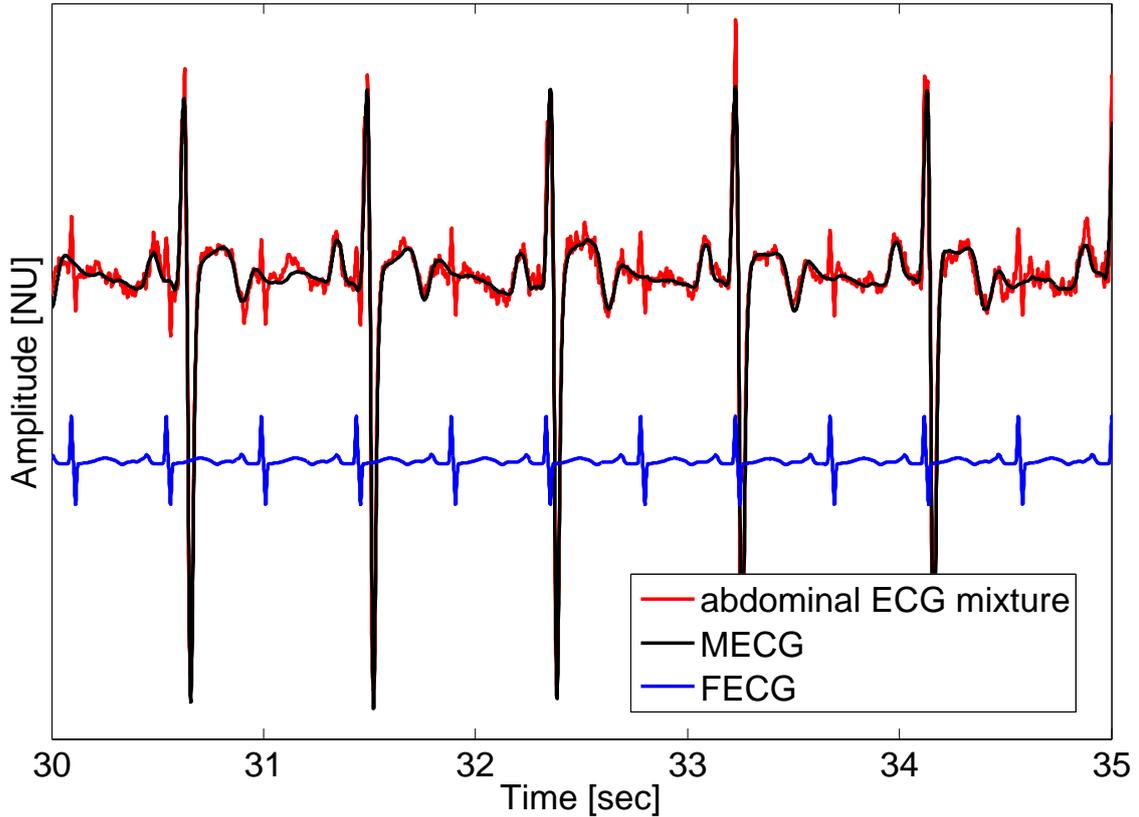

Figure 8.11: Example of abdominal ECG filtered using the EKFD. The extracted FECG has been given an negative baseline offset for display purposes.

### 8.4.4 Discussion

While the results of the EKFD on real data are promising, a number of important challenges require more research. These are listed below:

The QT annotation will depend greatly on which abdominal channel is selected. Figure 8.13 shows a series of extracted NI-FECG channels using the EKFD and where the T-wave morphology is somewhat different from a channel to another. This impact on the error in evaluating the QT should be studied, as no standard exists for which channel to annotate.

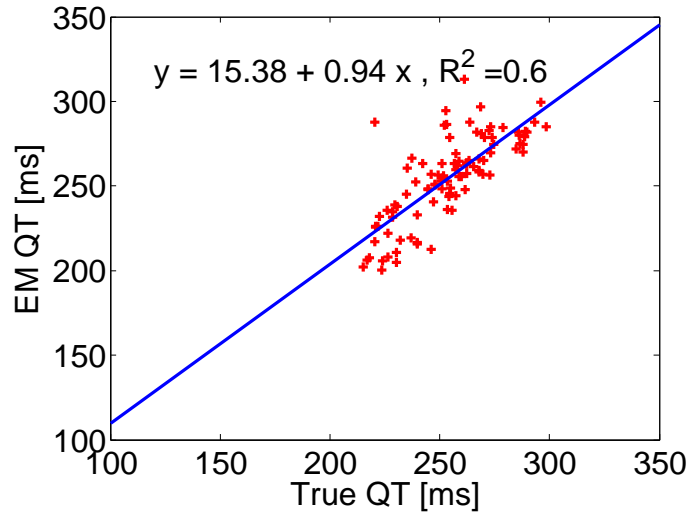

Figure 8.12: Regression on annotated QT output by the EKFD algorithm against the reference QT obtained from the SECG of 20 fetuses (97, 1-min segments). Line fit: y = intercept + gradient × x and $R^2$ is the corresponding coefficient of determination (goodness of fit).

For adults, it is recommended that the lead exhibiting the longest QT should be used, which usually corresponds to $V_2$ or $V_3$ [164, 165]. Variation in the QT measurements can be the result of [166]: (1) the projection of the resultant cardiac activity onto the varying lead axis or (2) the dispersion of the ventricular depolarisation or repolarisation across leads. The second effect has been subject to intense research in the past two decades in the adult ECG. The T-wave end was reported to vary with the ECG channel as early as 1934 by Wilson *et al.* [167]. It was suggested that this effect could be due to the dispersion of the myocardium and the QTd quantity, corresponding to the difference between the maximum and minimum QT intervals in a 12-lead ECG [166], has been intensively studied as a marker of this dispersion. However, it is still unclear how much of the variation of the QT length across channel due to (1) or (2) above. This is mainly caused by the fact that the precision of measurement is in the same order of magnitude of the so called dispersion effect [168]. Nevertheless, it is important to highlight that variations in the QT length across channels has been reported to have a mean dispersion as high as 48 ± 18 ms [169] when measured on 21 healthy individuals, and to be even higher in individuals with myocardial infarction. Mirvis [170] reported the difference between the longest and shortest interval to be 59.4 ± 12.9 ms in 50 normal subjects[2]. In the case of FECG analysis, the impact of electrode placement on the variation of the measured FQT length has not been studied. Study of the fetal QTd could be indicative of fetal cardiac defect or quantify the effect of projecting the cardiac activity onto different abdominal channels.

---

[2]This actually corresponds to the QTd quantity but was not called that at the time of this publication.

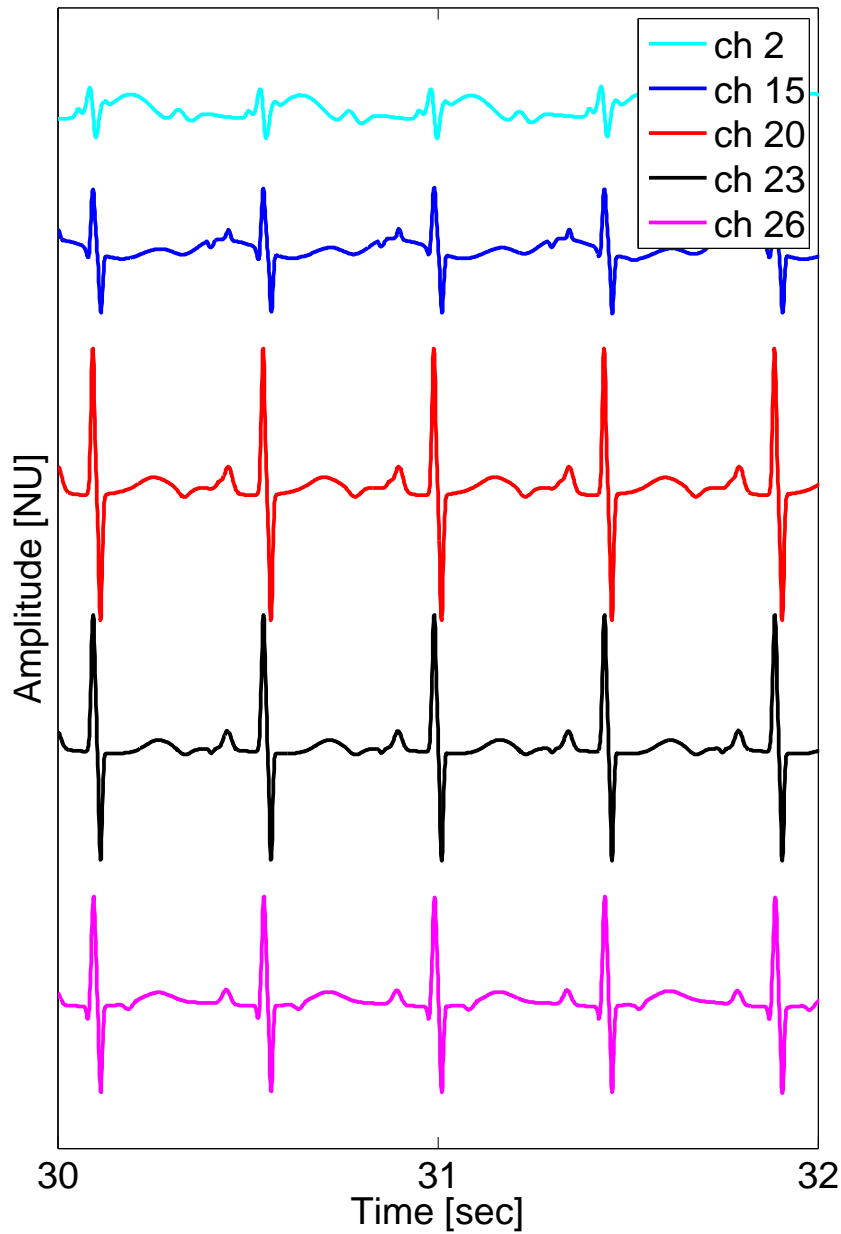

Figure 8.13: Series of extracted NI-FECG channels (from the same record) using the EKFD on real data. The ECG were recorded from a set of abdominal sensors superimposed on the maternal abdomen. This figures illustrates that the T-wave morphology is somewhat different from a channel to another which will have an impact the FQT.

The method for building the average ECG waveform from the residual signal is a vital contributor to the success to the EKFD framework. An acceptable template construction was obtained in this work, but it can be improved further. In particular the method of rejecting noisy cycles, the amount of baseline wander filtering $f_b$ and the RR interval variability are important parameters to consider when building this template. This is because failing to accurately remove the baseline wander, that is rather significant in abdominal recordings, will results in inaccurate vertical alignment of the ECG cycles. In addition, building a template with ECG cycles that have very different RR intervals will result in inaccurate horizontal alignment of the cycles if any linear time-wrapping is used and is likely to affect the morphology of the template cycle.

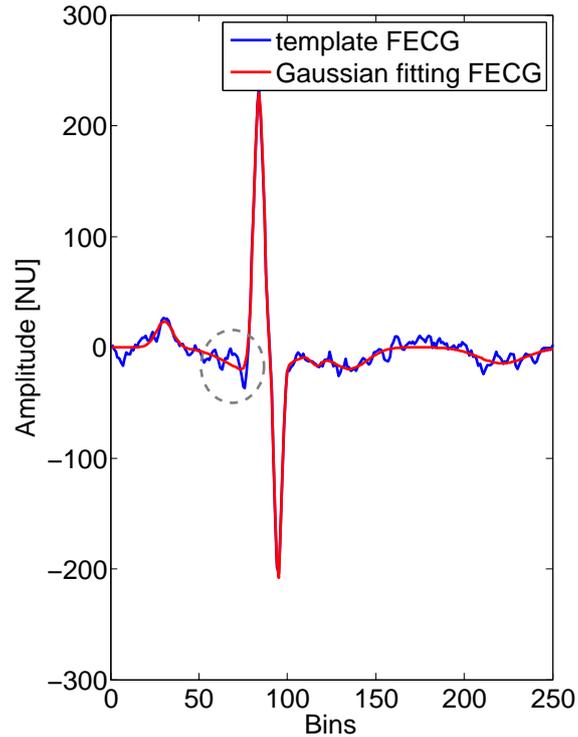

Figure 8.14: Example of Gaussian fitting on an average template FECG. Note that the Q onset is not well mapped. Bins: corresponds to the number of bins (250) the ECG cycles are mapped to using phase-wrapping and in order to build an average waveform.

The initialisation procedure for the Gaussian fitting will play an important role in the EKFD outcomes; Figure 8.14 shows an example where the Q onset does not appear to be well described by the Gaussian fitting. Although minor, this defect will affect the QT evaluation. (This example also highlights the difference in using the RMS approach or some clinical intervals as markers of success.)

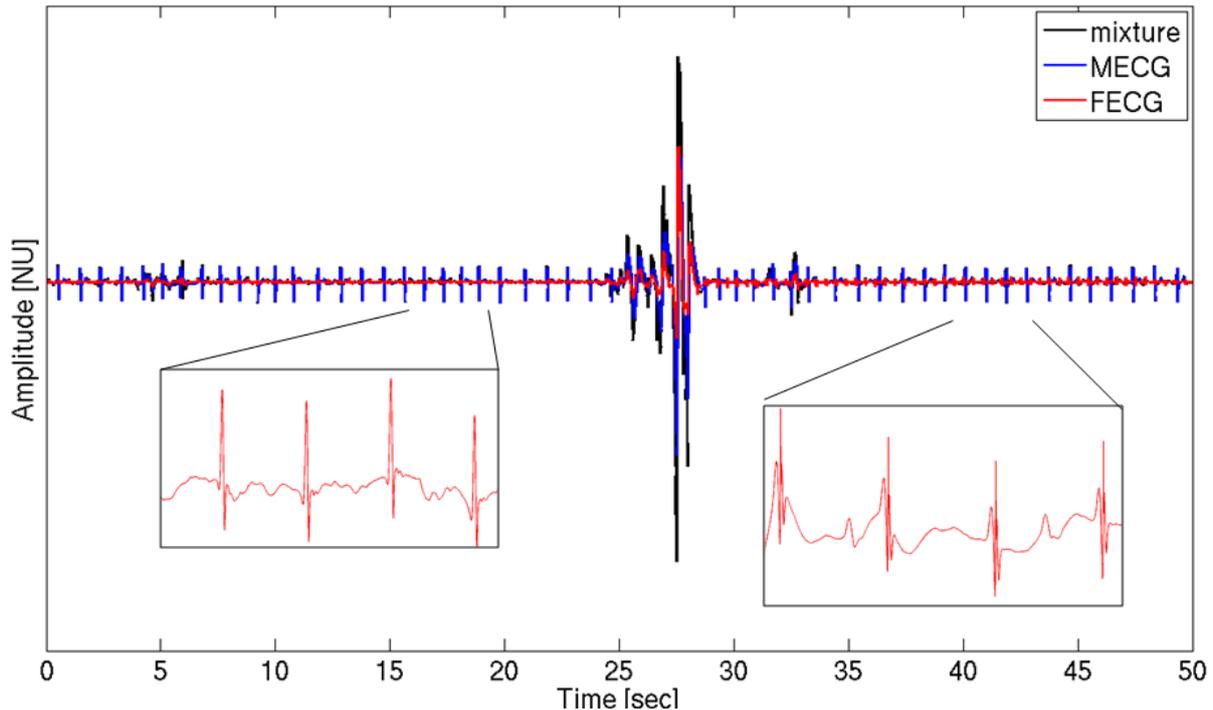

Figure 8.15: Example showing the importance of the EKFD calibration of the covariance matrix. Because the Gaussian parameters of the model are allowed to change for the FECG (i.e. they are part of the state equations), in the instance of transient noise and moderate trust in the observations, the Gaussian parameters change drastically, thus causing the filtering process to fail and produce poor morphology after the transient noise.

Setting the covariance matrices (of the Kalman filter) right will be essential for a successful filtering using the EKFD. It must be calibrated correctly (something that can be done using a random search on the artificial data as suggested in the previous section) but also updated in the instance of transient noise. In particular, trusting the ECG model too much will result in failing to follow the FECG morphology (and so change in the FQT), and the dynamical model would update the fetal Gaussian parameters incorrectly (which are part of the state equation - see Equations 8.2), as illustrated in Figure 8.15. There is a motivation in leaving these parameters to evolve over time and consequently an index on signal quality could be used to flag transient noise and avoid updating the Gaussian parameters during such noise episode. This could be done by modulating the gains of the covariance matrices with the signal quality index.

Clifford *et al.* [171] demonstrated the fitting of Gaussian for automated QT estimation on adult QT. However, the fetal T-wave tends to have a lower amplitude than the adult one and the non-standardisation of the fetal heart view through the abdominal electrodes will certainly influence the results. Nevertheless, it would be the right approach since it would make the EKFD framework self-contained, in that the algorithm would both be used to separate the NI-FECG morphology but also provide standard quantities (such as the QT) from its internal parameters. Figure 8.16 illustrates how the FQT can be derived from the Gaussian functions fitted to the template ECG.

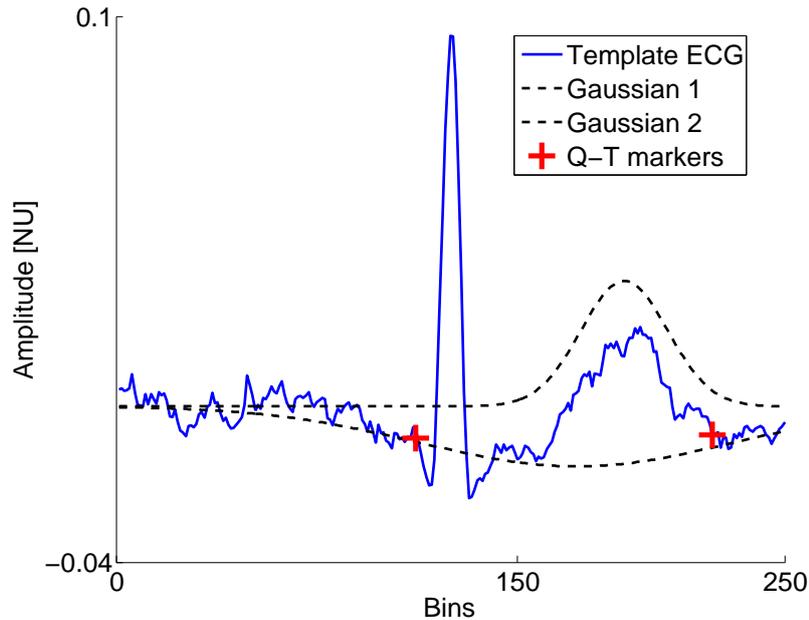

Figure 8.16: QT measurement and Gaussian fitting; this example shows an example of FQT obtained by using the two Gaussian mapping the fetal T-wave on the template FECG extracted from an abdominal recording. Bins: corresponds to the number of bins (250) the ECG cycles are mapped to in order to build an average waveform.

For the experiments performed in this section, the QT intervals were not varying substantially within a given record. Although this was a design choice, it will be necessary in the future to annotate longer data segments with varying QT lengths, in order to explore the potential of the EKFD to adapt to a changing FECG morphology (built into the state Equations 8.3), and so track changes in the QT interval. This could be first explored by using *fecgsyn* and simulating pathologies such as QT shortening.

Table 8.5: Summary table of QT experiments. EXP: experiment. EXP1: experiment conducted in Section 8.2, 22 fetuses, three annotators (A1-A3); EXP2: experiment conducted in Section 8.3, automated annotations; EXP3: experiment conducted in Section 8.4, 20 fetuses, one annotator. EM: fusion of the annotators annotations using the EM approach, AR: absolute error, RMSE: root mean square error, a: bias, b: gradient.

|         | AE   | RMSE | $R^2$ | a    | b    | data type  |
|---------|------|------|-------|------|------|------------|
| EXP1-A1 | 20.3 | 33.2 | 0.29  | 71.7 | 0.71 | real       |
| EXP1-A2 | 16.6 | 22.7 | 0.38  | 93.1 | 0.59 | real       |
| EXP1-A3 | 14.8 | 18.3 | 0.40  | 65.6 | 0.77 | real       |
| EXP1-EM | 10.4 | 13.6 | 0.61  | 49.9 | 0.81 | real       |
| EXP2    | 7.6  | 12.7 | 0.64  | 23.9 | 0.87 | artificial |
| EXP3    | 13   | 17.3 | 0.60  | 15.4 | 0.94 | real       |

## 8.5 Conclusion

This chapter explored the possibility of obtaining an accurate estimate of the FQT from the extracted NI-FECG. This feasibility study showed that it is possible, both in the observation and source domains, to measure the FQT with an error that is clinically acceptable (see Table 8.5). A number of challenges remain to make the FQT extraction fully automated and robust. In conclusion, this chapter provides strong evidence for research on accurate FQT extraction from the NI-FECG, and more generally on the morphological analysis of the NI-FECG.

# Chapter 9

# Conclusion and future work

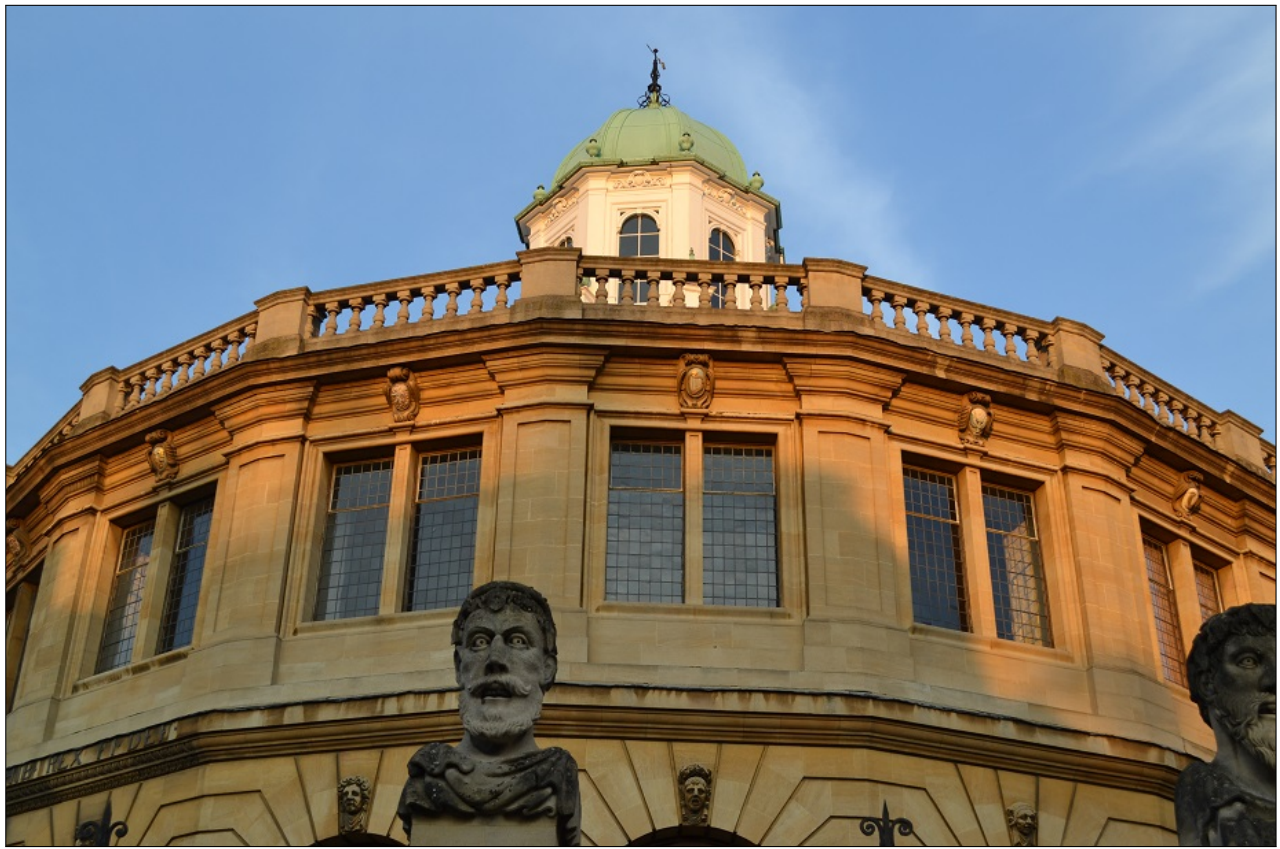

Figure 9.1: The Sheldonian theatre was built in the 17th century following the design of Christopher Wren. The building can accommodate between 600 and 1000 people and is nowadays used for lectures, conferences, matriculation and graduation ceremony at the University of Oxford.

## 9.1 Summary of contributions

This thesis provides a number of contributions to the field of NI-FECG extraction and analysis; (1) review of existing NI-FECG extraction approaches; (2) introduction of the Echo State Neural network as an adaptive filtering technique for its application to NI-FECG extraction and benchmark with existing techniques; (3) co-development of the Physionet/Computing and Cardiology Challenge 2013; (4) clarification of the field with respect to the performance of NI-FECG source separation methods; (5) demonstrated that using a high baseline wander cut-off frequency improved the overall performance of the NI-FECG extraction techniques and that by combining some of these techniques it was possible to improve the overall results. (6) development of the *fecgsyn* simulator; (7) accurate QT measurement by comparing the manually annotated NI-FECG from a commercial monitor to the scalp ECG; (8) novel joint Bayesian extraction approach to enable the study of beat-to-beat morphological analysis of the NI-FECG; (9) Development of signal quality indices and evaluation of them on large databases of normal and pathological adult ECG; (10) Open source code that was made available on `Physionet.org`.

## 9.2 Database and evaluation

### 9.2.1 Database

Together with providing the first significant open NI-FECG database, the Physionet Challenge 2013 highlighted the need for a larger and more complete database to be created. This database should ideally have a number of characteristics which are summarised below.

Part of the database should have a reference SECG and at least one chest MECG channel. The scalp recordings can be used as the *silver* standard for FHR and ECG morphological analysis. The chest recording can be used to ensure accurate MQRS detection, and for some source extraction techniques that require a reference channel free of any FECG contribution. It can also be used to avoid confusing the FHR and MHR, as this mistake sometimes happens with the ultrasound transducer and can be life threatening for the fetus [31].

The database should include at least eight abdominal channels (see Section 6.5 of [9]) to capture the multidimensional nature of both the fetal and maternal ECG. This is to allow better performance of BSS based methodologies and also because an optimal positioning of the electrodes with respect to the fetal heart position is unknown (i.e. channel selection might be required). The sampling frequency and amplitude resolution need to be high enough: $fs$ over 1 kHz, 16-bits. In particular, low frequency resolution can cause misalignment problems in removing the MECG using the TS methods.

A number of pieces of information such as outcome of delivery, maternal history (e.g. preeclampsia) should be reported. Information on outcome of the delivery can include metabolic acidosis, neonatal encephalopathy, Apgar score, neonatal intubation, etc. This is to assess whether the information extracted from the NI-FECG is predictive of delivery outcomes. It is also important to obtain rhythm annotations as well as information on early and late decelerations from experts since these events are typically the ones where the NI-FECG monitor should not fail to extract the FECG.

The database should be 'large enough' both in term of the recordings length and the number of fetuses. In particular, the records should be long enough in order to assess the adaptability of the approaches with respect to the non-stationarity nature of the FECG (intrinsic to the ECG signal but also due to fetal motion, resulting in a change of the fetal heart orientation).

Recordings should be performed at different stages of pregnancy, in order to assess when NI-FECG extraction is feasible (and in what proportion), and also for studying HRV/ECG morphological analysis as a predictor of cardiovascular diseases in infants. However when assessing the performance of the antepartum NI-FECG it is important to compare the success of the extraction against an alternative technique such as ultrasound, for FHR, rather than manually annotating fetal R-peaks on the abdominal recordings. Indeed, manually annotating the fetal R-peaks (in order to use them as the reference annotations) implies that these peaks must be clearly identifiable on at least one abdominal ECG channel. However, as it has been observed, it happens that the NI-FECG can also be extracted from abdominal recordings where no FECG can be identified by human inspection of the surface electrode recordings (see example in Figure 3.5).

The extraction of the FECG might fail because of poor quality signals. The FECG monitor should be able to report a failure to extract the FECG to the technician, and suggest checking/repositioning of the electrodes or using an alternative monitoring technique. Signal quality annotations can be used to test the robustness of the extraction algorithm with respect to noise and/or its ability to reject bad quality segments from the analysis. It can be used to exclude some abdominal channels before performing a BSS step (e.g. this showed improvement in the work of Liu *et al.* [143]).

Because there is currently no such database in existence, an artificial NI-FECG model such as *fecgsyn* introduced in Chapter 4, can be used until such databases is created, or in addition to such databases as stress tests/proof of concept for any algorithm.

### 9.2.2 Evaluation

There are a number of statistics that should be reported when assessing the performance of any NI-FECG extraction algorithm; classical statistics for QRS detection algorithm assessment should include: $Se$, $PPV$, $F_1$ computed with a tolerance of 50 ms around the reference annotations. These statistics reflect the performance of the algorithm in extracting the FQRS from the abdominal mixture. These statistics should also be reported at different stages of pregnancy. Although the NI-FECG can be recorded from 20 weeks onward [27], the vernix caseosa forms around 28th-32nd weeks and dissolves in 37th-38th weeks in normal pregnancies [19], theoretically limiting successful NI-FECG recording during this time period. To what extend the vernix caseosa influences accurate FECG extraction is to be further quantified.

In addition to reporting statistics on FQRS detection, it is necessary to report the performance of the algorithms in terms of FHR (i.e. after any smoothing has been performed to compute the FHR from the detected FQRS), since ultimately the FHR is used in clinical practice. However, a RMS based score between the reference and the extracted FHR is not the best option. Rather a $HR_m$, as introduced in Chapter 6, is preferable (or alternatively a percentage ratio of the absolute error between the reference and the measured FHR divided by the reference FHR). This metric should reflect the clinical imperative of extracting an accurate FHR. Statistics could also be grouped by HRV range or events (e.g. accelerations/decelerations): in clinical practice physicians are interested in rapid variations of FHR; as such algorithms should be evaluated in their capacity to adapt to highly variable FHR rather than when there is no event (and thus less non-stationarity).

The ability of an algorithm to extract an accurate FECG morphology should be assessed in terms of clinically significant parameters, such as QT length and ST elevation as opposed to a SNR measure sometimes used to compare the 'overall' morphology to some reference signal morphology. The reference morphological annotations can be obtained from the scalp electrode, although it is important to keep in mind that this silver standard will be limited by the projection of the resultant cardiac activity and QT dispersion (see Section 8.4.4).

## 9.3 Algorithms for FQRS detection

A series of NI-FECG algorithms were implemented and benchmarked in this thesis. The motivation for doing so was that the published algorithms addressing the topic of NI-FECG extraction are rarely thoroughly evaluated. A number of general items were identified from this research and are summarised below.

Using a high baseline wander cut-off frequency $f_b$ improved the performance of the extraction algorithms. This has been demonstrated in Chapter 6 and Chapter 7.

Combining multiple NI-FECG extraction methods, improved the performance over a single method, and is perhaps a better approach than trying to fine tune a specific algorithm. This has been shown through the Physionet Challenge 2013 results in Chapter 7, where the FUSE method obtained the best overall Challenge results on the hidden test set. All methods have their advantages and limitations, and running them in parallel (while being able to switch to the best performing algorithm) can provide better results. A promising alternative to switching from one channel to another would be to combine the FQRS annotations provided by the different algorithms.

It is important to evaluate the various methods in a sequential manner i.e. following the approach taken in Chapter 7, and more specifically in Table 7.2. The added value of each step within the overall framework can then be quantified. In particular, it was noticeable that adding a BSS step following the time based cancellation of the MECG improved the $F_1$ measure by 2-10% (see Table 7.2 in Chapter 7). This very clearly highlighted the added value of using BSS techniques in the context of NI-FECG extraction.

The $TS_{EKF}$ method generally provided lower results than other benchmarks methods in extracting the FQRS. This was mainly due to the adaptive characteristic of the EKF, which typically misses the overlapping FQRS and MQRS. However, as demonstrated in Chapter 7 there is a lot of room for improvement with this Bayesian approach by further tuning and updating online (with respect to the varying noise level) the Kalman filter covariance matrix. The effect of the number of Gaussian on the EKF performances (in accurately extracting the FQRS) has not been evaluated.

In Chapter 6 adaptive filtering methods (LMS, RLS and ESN) were benchmarked on real data. The results showed that the $ESN_a$ had the highest performance on the training and test sets. This suggests that using non-linear adaptive filtering can improve the performance over the linear LMS and RLS.

An important element that was not possible to evaluate in this thesis is the relative performance of the extraction algorithms in the presence of a variety of maternal and fetal rhythms. In particular, it would be expected that the least-adaptive techniques, such as TS or $TS_c$, would fail in these cases. An example of abdominal MECG with frequently occurring ventricular beats is shown in Figure 9.2. This Figure highlights the need for algorithms that can handle pathological cases. Different templates may be built depending on the beat morphology, or a filtering technique must be adaptive enough to cancel out the different morphologies of maternal beats. Another option is to use ICA as a preprocessing step in order to isolate the different types of maternal beats, (e.g. Figure 9.2b). The non-stationary nature of the ECG (understood here to be the changing ECG beat morphology) is particularly important given the prevalence of some arrhythmia in the adult population. For example, it was demonstrated that frequent (>60/h or 1/min) and complex PVCs can occur in healthy subjects with an estimated prevalence of 1-4% [172]. Moreover, the incidence and severity of atrial and ventricular ectopy were reported to be higher during pregnancy [173, 174]. Work (see [7]) is ongoing to compare the extraction techniques on simulated data that were generated using *fecgsyn* and which contain the physiological events modelled in Chapter 4.5.2.2.

The FQRS detectors developed during the Challenge 2013 were evaluated on normal sinus rhythm fetal ECG. However, fetal arrhythmia such as extrasystoles, atrioventricula blocks have been reported in the literature [175], and there is currently no insight into how these detectors would perform in such circumstances. The question also holds for the channel selection routine which, for the Challenge 2013, were mainly based on some assessment of the regularity of the RR time series derived from the FQRS detected on each residual (of the individual abdominal channels).

Focus was given to the study of single pregnancy in this thesis. No twin, or more generally speaking 'multiple pregnancy' data were available. As a consequence the experimental design of the methods were targeted at single pregnancies. The prevalence of multiple pregnancy is estimated to be 3% [176]. This number is non-negligible and consequently there is interest in evaluating methods that can handle these cases. A number of techniques (all TS, adaptive filter techniques) are expected to still perform well in removing the maternal ECG waveform but FQRS detection cannot be made in a straightforward manner on the residual anymore. This is because a FQRS detector would pick up the FQRS from the different fetuses indiscriminately and also miss a non-negligible number of them due to the in-built refractory period of the detector. A blind source separation approach seems natural for breaking down the FECG mixture into the individual sources coming from the different fetal hearts. As an example, a case study has been demonstrated in Sameni [9], where the author used ICA to separate the FECG sources after performing maternal ECG cancellation. Taking into consideration multiple pregnancy cases will not only impact the source separation choice, but also the number of abdominal electrodes that are necessary to separate the increased number of sources. Finally, an NI-FECG extraction method that could manage multiple pregnancies would be of particular interest given that the positioning of ultrasound transducers (one for each fetal heart) becomes challenging in these instances.

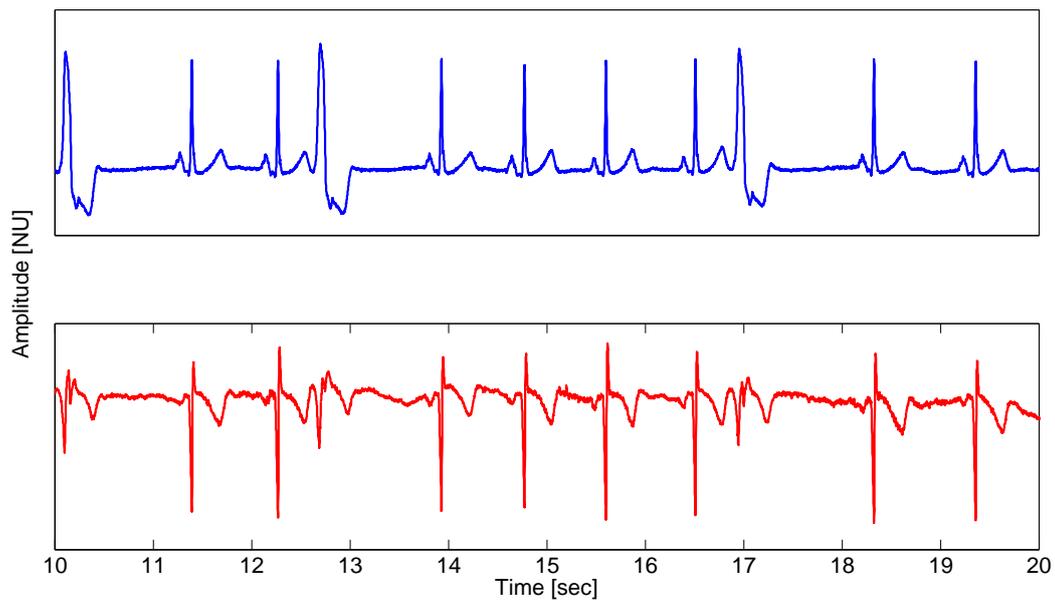

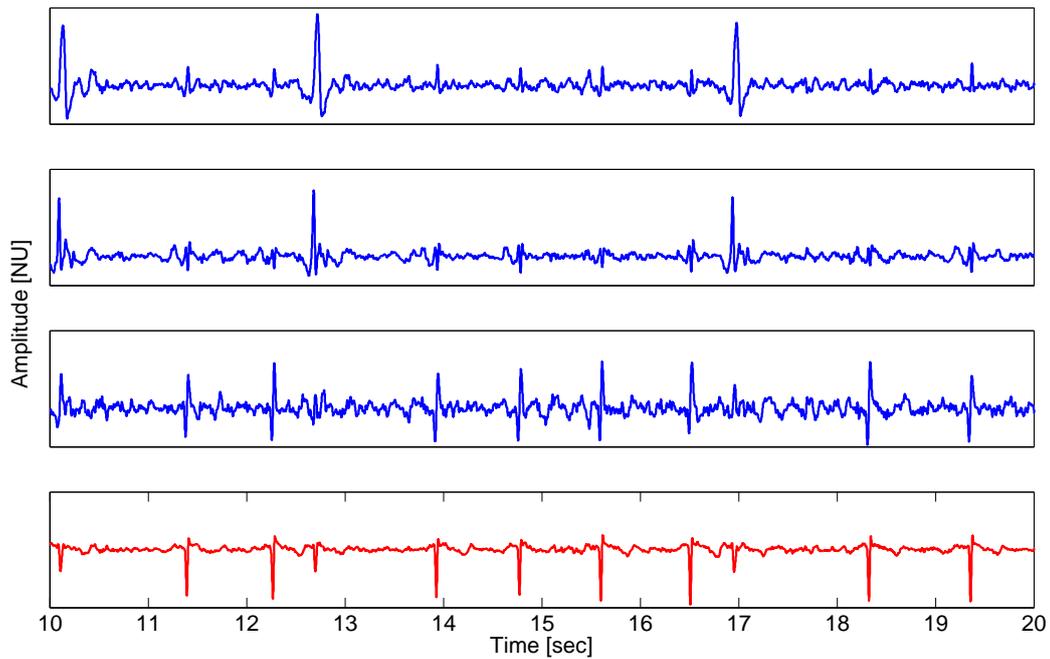

Figure 9.2: (a) Example of abdominal ECG with the presence of frequently occuring maternal ventricular beats. Two channels are plotted; (b) Example of ICA output for eight abdominal ECG with the presence of maternal ventricular beats. Note that ICA is breaking down the beats type into different components (normal beats and bygeminy - the FECG could not be isolated on this example). The data were prefiltered ($f_b$=3 Hz, $f_h$=45 Hz) before applying the JADE ICA algorithm. A subset of the eight channels were represented for display purposes.

## 9.4 Morphological analysis

In the current literature, very little work has been done in reconstructing the FECG morphology from the extracted NI-FECG signal. In this thesis a number of points were highlighted that, if addressed, could push the research forward. These are discussed below.

The effect of the $f_b$ on FQT extraction is yet to be determined. It is known that for adults a high $f_b$ will distort the T-wave morphology, negatively affecting the QT estimation. However, the baseline wander is rather substantial on the abdominal recordings and needs to be heavily removed. The trade-off between least QT distortion and good baseline removal needs to be studied.

It was shown, in Chapter 8, that QT can be extracted from both the source and observation domain. However, a number of challenges associated with the two approaches need to be tackled; while working in the source domain, is QT affected by the BSS transform? while working in the observation domain, which abdominal channel should be selected? In particular, the abdominal channel selection when working in the observation domain is likely to have an important impact on the measured QT due to the associated projection and QT dispersion.

The covariance matrix of the EKFD should be updated online with the signal quality. Updating the covariance matrix with SQI makes the most sense in the case where the Gaussian parameters are left to evolve. This is to prevent them from diverging in the presence of noise.

Figure 9.3 describes what is thought to be the reference procedure for FHR and morphological information extraction (here QT) from the abdominal ECG recording. A number of ECG electrodes are placed to cover the maternal abdomen; the signal quality of these channels is assessed and the bad quality channels are rejected before applying the FUSE method to extract the FQRS; the FQRS, together, with the good quality abdominal recordings are then used to run the EKFD to extract the morphology of the FECG. In particular, note that a different $f_b$ cut-off frequency might be used for the FQRS extraction step and for the EKFD step. This would preserve the P and T wave morphology.

One of the main limitation in performing morphological analysis of the NI-FECG is to be able to accurately locate the R-peaks. Indeed, the fetal R-peak locations are used as fiducual points in many recently introduced approaches such as the EKFD introduced in this thesis, Gaussian processes [177], tensor decomposition[178] and $\pi$CA [51]. As a consequence there may be utility in using additional non-invasive techniques such as the phonocardiogram (PCG) in order to help locate fetal R-peaks to further improve fiducial point location and improve extraction of the FECG morphology. However, it is yet to be proven that usage of multimodal data for this purpose can be beneficial; Springer *et al.* [179] demonstrated accuracies of 80-86% for adult HR estimation from PCG recordings. Varadi *et al.* [180] demonstrated an 83% accuracy in estimating the fetal HR from fetal PCG by comparing the PCG obtained FHR to the ultrasound obtained FHR. Whether this is enough to improve the NI-FECG extraction by improving FQRS detection and fetal R-peak location is yet to be demonstrated.

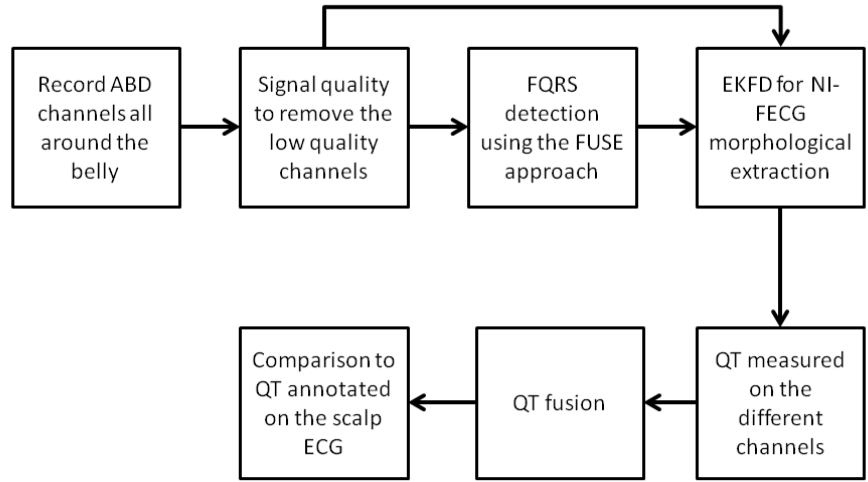

Figure 9.3: Block diagram of the suggested procedure for extraction of both FHR and morphological information from the abdominal ECG recording. ABD: abdominal.

## 9.5 Signal quality

**Signal quality for maternal chest channels** The SQI developed in this work can be used on the maternal chest reference channel, as in Chapter 6, where it can be valuable for NI-FECG extraction techniques that require a maternal reference.

**Signal quality for abdominal MECG** The signal quality indices presented in this thesis were assessed on adult ECG and were used to ignore low quality maternal chest ECG. Despite the good performance on these data, they did not perform as well, in general, on abdominal ECG data. This was because the statistical properties of the AECG are different (generally lower SNR and mixture of MECG and FECG) than an adult ECG. In particular, bSQI which was identified as being the best performing quality index, fails on abdominal data in the instances where the FECG is of high enough amplitude (because it would result in detecting both the maternal and fetal beats although not consistently because of the QRS detector's refractory period). Thus the development and evaluation of signal quality index for abdominal recordings from pregnant women is required. For example, during the Physionet Challenge 2013, Liu *et al.* [78] used sample entropy as quality index for excluding abdominal channels that were of poor quality. In their publication they showed improved performance when including this channel selection step.

**Signal quality for the residual ECG** An element that was not attempted, but that would be of interest, is to use bSQI (or any alternative robust SQI) on the residual signal (i.e. ECG residual obtained after MECG cancellation). This way, agreement between two robust FQRS detectors would be synonymous with successful FQRS detection. This approach could then be used to choose 'the best' extracted residual channel instead of using a regularity measure (as was done in Chapter 7). Switching channel based on quality assessment (using bSQI) has proved to be effective when entering (and eventually winning) the Physionet Challenge 2014 addressing the problematic of *Robust Detection of Heart Beats in Multimodal Data*.

## 9.6 Other promising applications of the NI-FECG

Although most of this thesis was focused on intrapartum monitoring, it is important to note that there is a great potential for the NI-FECG to be used toward identifying intrauterine growth restricted (IUGR) fetuses. It is known that IUGR is associated with stillbirth, neonatal death and perinatal morbidity, as well as cerebral palsy and adult conditions [181]. Very often, IUGR fetuses are not identified antenatally. In routine clinical practice, up to three quarters of babies at risk of IUGR are not identified and this number raise to up to 85% of missed identification in low-risk pregnancy [181]. Enabling early identification of IUGR fetuses would have a tremendous impact on perinatal care. EFM techniques have the potential to be used to accurately identify IUGR fetuses (prepartum). Indeed, smallness defined as the weight of a fetus below a threshold at a given gestational age, has been shown to be a weak identifier of IUGR fetuses. This means that the association between pathological smallness and adverse outcome is unclear [181]. In contrast screening for reduced heart rate variability has shown great promise in identifying IUGR fetuses [34]. As it was shown in this thesis, it is possible to obtain a reliable extraction of the FQRS from the NI-FECG, enabling accurate antenatal FHR estimation and HRV studies.

## 9.7 Conclusion

This thesis focused on the processing of the ECG recorded from a set of abdominal sensors recorded on pregnant women, and the extraction of clinically relevant information from this signal. Indeed, NI-FECG offers many advantages over the alternative fetal monitoring techniques, the most important one being the opportunity to enable morphological analysis of the FECG which is instrumental in determining whether an observed FHR event is physiological or pathological during delivery. Measurement of the QT could also be used prepartum to identify whether any drugs the mother takes prolongs the QT of the fetus during the pregnancy.

For that purpose, a number of novel and state of the art extraction techniques were implemented and extensively evaluated on artificial and real databases. Part of the work presented in this thesis was also benchmarked against the approaches of 53 research teams that entered the Physionet Challenge 2013 and about the same number for the 2014 Challenge. For both competitions, the algorithms presented in this thesis finished in first place. Finally, an extension of the ECG Bayesian filtering framework was introduced and evaluated in the context of morphological analysis of the NI-FECG. Results indicate that an accurate fetal QT interval can be extracted from abdominal ECG, and hence a range of possibilities for antenatal safety monitoring are possible.

# Appendix A

# Background on Kalman Filter

This section presents the basic principles of the Kalman Filter (KF) that are necessary to understand its strengths and limitations in the context of ECG processing.

## A.1 Introduction

**Control theory** is an engineering and applied mathematics area which deals with the behaviour of dynamical systems. The aim is to control the 'trajectory' of a system using some defined control input and control feedback. The difference between the output of the system and the desired output is called error, and it is the purpose of control theory to minimise it. **State-space estimation** is recognised as being the cornerstone of control theory. The main feature of the so-called state-space method is that the processes are described by a set of coupled differential equations [182]. **The Kalman filter** estimates the state of a noisy system (i.e. its states and observations are uncertain) where not all state variables are necessarily observed (i.e. they cannot all be measured and used for control feedback). It is an optimal estimator in the mean square error sense, and it is a recursive filter which means that the measurements can be processed online. The filter is named after Rudolf E. Kálmán and was published in his work entitled "A new approach to linear filtering and prediction problems" [183] in 1960. The idea emerged when Kálmán visited NASA and found its first application on the Apollo navigation computer. This appendix section deals with the Kalman filter ECG model originally introduced in Sameni *et al.* [184].

## A.2 Mathematical basis

Lets assume that a set of variables, called state variables, provide a complete representation of the internal conditions or status of the system of interest at a given instant of time. Bayesian filtering is a probabilistic approach that recursively estimates the posterior distribution $P(\underline{\mathbf{x}}|\underline{\mathbf{y}}_{1:k})$ of a hidden state variable, $\underline{\mathbf{x}}$, at each time, $k$, by using incoming measurements

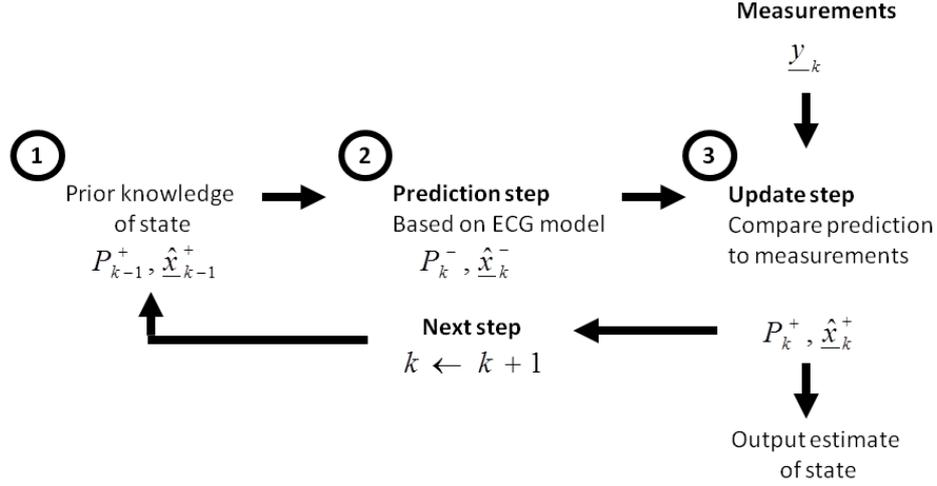

Figure A.1: Steps of the KF in estimating the state of the system in the context of ECG filtering.

$\{\underline{\mathbf{y}}_k\}$ and a mathematical process model that plays the role of a prior on the system being observed. The random variables are assumed to evolve according to an *evolution equation* (the mathematical process model) and observed through measurements that are related to the state by a *measurement equation*. The KF is a particular case of Bayesian filtering. The KF equations, which allow for the recursive estimation of the hidden state in the restricted case of linear systems with Gaussian variables, are first introduced. Subsequently the KF is extended to non-linear dynamical systems. Finally, the application of Bayesian filtering to ECG filtering is introduced. The Kalman filter assumes that:

- the system can be described through a linear model
- the system and measurement noises are white and Gaussian[1].

### A.2.1 Bayesian framework

The Bayesian filtering framework allows for recursive estimation of the state variables. Given the KF equations, the evaluated model parameters, the state $\hat{\underline{\mathbf{x}}}^+_{k-1}$ and covariance matrix $\mathbf{P}^+_{k-1}$ at time step $k-1$, the ECG model can be used to predict the value of the covariance matrix and state vector at time step $k$ i.e $(\mathbf{P}^-_k, \hat{\underline{\mathbf{x}}}^-_k)$. Given the measurements $\underline{\mathbf{y}}_k$ (phase and signal amplitude) the estimate of $\mathbf{P}$ and $\underline{\mathbf{x}}$ are updated, resulting in $(\mathbf{P}^+_k, \hat{\underline{\mathbf{x}}}^+_k)$ before $k$ is incremented. Figure A.1 displays a diagram summarising the key steps of the KF procedure in the context of ECG filtering.

---

[1]We assume that the variables are Gaussian in the following for simplicity, and because it gives a better insight into the Kalman filter process. However, as noted by Simon [161]: "It is often asserted in books and papers that the Kalman filter is not optimal unless the noise is Gaussian. However, as our derivation in this chapter has shown, that is simply untrue. Such statements arise from erroneous interpretations of Kalman filter derivations. Even if the noise is not Gaussian, the Kalman filter is still the optimal *linear* filter".

## A.2.2 The discrete-time Kalman filter

Consider the dynamical system with the following state-space mathematical representation:

$$\begin{cases} \underline{\mathbf{x}}_k = \mathbf{G}_{k-1}\underline{\mathbf{x}}_{k-1} + \underline{\mathbf{w}}_{k-1} & (\textit{evolution equation}) \\ \underline{\mathbf{y}}_k = \mathbf{H}_k\underline{\mathbf{x}}_k + \underline{\mathbf{v}}_k & (\textit{measurement equation}), \end{cases} \quad (A.1)$$

where $\mathbf{G}_{k-1}$ is the state transition matrix applied to the previous state $\underline{\mathbf{x}}_{k-1}$, $\underline{\mathbf{w}}_k$ and $\underline{\mathbf{v}}_k$ correspond to the state and observation noises which are assumed to be white, zero-mean, uncorrelated ($E[\underline{\mathbf{v}}_k\underline{\mathbf{w}}_j^T] = 0$) and have covariance matrices $\mathbf{Q}_k = E[\underline{\mathbf{w}}_k\underline{\mathbf{w}}_k^T]$ and $\mathbf{R}_k = E[\underline{\mathbf{v}}_k\underline{\mathbf{v}}_k^T]$ respectively. $\mathbf{H}_k$ is the observation matrix that maps state variables to the observations. The goal is to estimate the state $\underline{\mathbf{x}}_k$ based on the knowledge of the system dynamics and the noisy measurements $\{\underline{\mathbf{y}}_k\}$. The initial state estimate and covariance matrix are respectively given by: $\hat{\underline{\mathbf{x}}}_0^+ = E[\underline{\mathbf{x}}_0]$ and $\hat{\mathbf{P}}_0^+ = E[(\underline{\mathbf{x}}_0 - \hat{\underline{\mathbf{x}}}_0^+)(\underline{\mathbf{x}}_0 - \hat{\underline{\mathbf{x}}}_0^+)]$. The following KF equations are then recursively used:

*Prediction*

$$\begin{cases} \mathbf{P}_k^- = \mathbf{G}_{k-1}\mathbf{P}_{k-1}^+\mathbf{G}_{k-1}^T + \mathbf{Q}_{k-1} \\ \hat{\underline{\mathbf{x}}}_k^- = \mathbf{G}_{k-1}\hat{\underline{\mathbf{x}}}_{k-1}^+ \end{cases} \quad (A.2)$$

*Correction*

$$\begin{cases} \mathbf{K}_k = \mathbf{P}_k^-\mathbf{H}_k^T(\mathbf{H}_k\mathbf{P}_k^-\mathbf{H}_k^T + \mathbf{R}_k)^{-1} \\ \hat{\underline{\mathbf{x}}}_k^+ = \hat{\underline{\mathbf{x}}}_k^- + \mathbf{K}_k(\underline{\mathbf{y}}_k - \mathbf{H}_k\hat{\underline{\mathbf{x}}}_k^-) \\ \mathbf{P}_k^+ = (\mathbf{I} - \mathbf{K}_k\mathbf{H}_k)\mathbf{P}_k^-(\mathbf{I} - \mathbf{K}_k\mathbf{H}_k)^T + \mathbf{K}_k\mathbf{R}_k\mathbf{K}_k^T, \end{cases} \quad (A.3)$$

where $\hat{\underline{\mathbf{x}}}_k^- = E[\underline{\mathbf{x}}_k|\underline{\mathbf{y}}_{1:k-1}]$ and $\hat{\underline{\mathbf{x}}}_k^+ = E[\underline{\mathbf{x}}_k|\underline{\mathbf{y}}_{1:k}]$ are the *a priori* and *a posteriori* state estimates respectively. $\mathbf{P}_k^- = E[(\underline{\mathbf{x}}_k - \hat{\underline{\mathbf{x}}}_k^-)(\underline{\mathbf{x}}_k - \hat{\underline{\mathbf{x}}}_k^-)^T]$ and $\mathbf{P}_k^+ = E[(\underline{\mathbf{x}}_k - \hat{\underline{\mathbf{x}}}_k^+)(\underline{\mathbf{x}}_k - \hat{\underline{\mathbf{x}}}_k^+)^T]$ are the covariance matrices of the estimation error of $\hat{\underline{\mathbf{x}}}_k^-$ and $\hat{\underline{\mathbf{x}}}_k^+$ respectively. $\mathbf{K}_k$ is the KF gain and represents how much an new observation is going to change the estimate of the states. Note that the expression used in equation A.3 for $\mathbf{P}_k^+$ is known as the Joseph stabilised version of the covariance measurement update equation [185]. It can be shown that this is more stable than alternative formulations, because it guaranties that $\mathbf{P}_k^+$ is always symmetric positive definite as long as $\mathbf{P}_k^-$ is symmetric positive definite [161].

In this system, the random variables are Gaussians since the linear transform of Gaussian variables gives Gaussian variables. This means that all the variables in the equations are fully characterised by their first and second order momentum. Thus the KF can been seen as a recursive procedure of prediction and correction during which only the necessary statistics (i.e. mean and variance) are propagated [186].

## A.2.3 The discrete-time extended Kalman filter

The KF formalism only applies to linear systems. In practice, systems are rarely linear or are only linear in a given range [161]; many non-linear systems can be piecewise approximated by a linear system with satisfactory results. The key idea behind the extended KF (EKF) is to linearise the system of equations in the vicinity of the previous estimated instant (see Equation A.4), compute the covariance matrices $\mathbf{P}_k^-$ and $\mathbf{P}_k^+$ and the Kalman gain, $\mathbf{K}_k$, from the linearised system while propagating the state variable using the initial non-linear equation [187]. Consider the model defined by the following non-linear equations:

$$\begin{cases} \underline{\mathbf{x}}_k = g_{k-1}(\underline{\mathbf{x}}_{k-1}, \underline{\mathbf{w}}_{k-1}) \\ \underline{\mathbf{y}}_k = h_k(\underline{\mathbf{x}}_k, \underline{\mathbf{v}}_k), \end{cases} \quad (A.4)$$

where $g$ and $h$ are the non-linear state and measurement functions. A first order Taylor series expansion around $\underline{\mathbf{x}}_{k-1} = \hat{\underline{\mathbf{x}}}_{k-1}^+$ and $\underline{\mathbf{w}}_{k-1} = 0$ is performed for the state-space equation and around $\underline{\mathbf{x}}_k = \hat{\underline{\mathbf{x}}}_k^-$ and $\underline{\mathbf{v}}_k = 0$ for the measurement equation (recall that $\hat{\underline{\mathbf{v}}}_k = E[\underline{\mathbf{v}}_k] = 0$ and $\hat{\underline{\mathbf{w}}}_k = E[\underline{\mathbf{w}}_k] = 0$):

$$\begin{cases} \underline{\mathbf{x}}_k = g_{k-1}(\hat{\underline{\mathbf{x}}}_{k-1}^+, 0) + \mathbf{G}_{k-1}(\underline{\mathbf{x}}_{k-1} - \hat{\underline{\mathbf{x}}}_{k-1}^+) + \mathbf{L}_{k-1}\underline{\mathbf{w}}_{k-1} \\ \underline{\mathbf{y}}_k = h_k(\hat{\underline{\mathbf{x}}}_k^-, 0) + \mathbf{H}_k(\underline{\mathbf{x}}_k - \hat{\underline{\mathbf{x}}}_k^-) + \mathbf{U}_k\underline{\mathbf{v}}_k, \end{cases} \quad (A.5)$$

where

$$\mathbf{G}_{k-1} = \left.\frac{\partial g_{k-1}}{\partial \underline{\mathbf{x}}}\right|_{\hat{\underline{\mathbf{x}}}_{k-1}^+, 0}, \mathbf{L}_{k-1} = \left.\frac{\partial g_{k-1}}{\partial \underline{\mathbf{w}}}\right|_{\hat{\underline{\mathbf{x}}}_{k-1}^+, 0}, \mathbf{H}_k = \left.\frac{\partial h_k}{\partial \underline{\mathbf{x}}}\right|_{\hat{\underline{\mathbf{x}}}_k^-, 0}, \mathbf{U}_k = \left.\frac{\partial h_k}{\partial \underline{\mathbf{v}}}\right|_{\hat{\underline{\mathbf{x}}}_k^-, 0}. \quad (A.6)$$

Note that the reason to expand $h(\underline{\mathbf{x}}_k)$ around $\hat{\underline{\mathbf{x}}}_k^-$ is because it is our best estimate of $\underline{\mathbf{x}}_k$ before the measurement at time $k$ is taken into account [161]. This results in the EKF:

*Prediction*

$$\begin{cases} \mathbf{P}_k^- = \mathbf{G}_{k-1}\mathbf{P}_{k-1}^+\mathbf{G}_{k-1}^T + \mathbf{L}_{k-1}\mathbf{Q}_{k-1}\mathbf{L}_{k-1}^T \\ \hat{\underline{\mathbf{x}}}_k^- = g_{k-1}(\hat{\underline{\mathbf{x}}}_{k-1}^+, 0) \end{cases} \quad (A.7)$$

*Correction*

$$\begin{cases} \mathbf{K}_k = \mathbf{P}_k^-\mathbf{H}_k^T(\mathbf{H}_k\mathbf{P}_k^-\mathbf{H}_k^T + \mathbf{U}_k\mathbf{R}_k\mathbf{U}_k^T)^{-1} \\ \hat{\underline{\mathbf{x}}}_k^+ = \hat{\underline{\mathbf{x}}}_k^- + \mathbf{K}_k(\underline{\mathbf{y}}_k - h_k(\hat{\underline{\mathbf{x}}}_k^-, 0)) \\ \mathbf{P}_k^+ = (\mathbf{I} - \mathbf{K}_k\mathbf{H}_k)\mathbf{P}_k^-(\mathbf{I} - \mathbf{K}_k\mathbf{H}_k)^T + \mathbf{K}_k\mathbf{U}_k\mathbf{R}_k\mathbf{U}_k^T\mathbf{K}_k^T. \end{cases} \quad (A.8)$$

## A.3 ECG Bayesian filtering framework

### A.3.1 Dynamical ECG model

*"The most important task confronting the control system analyst is developing a mathematical model of the process of interest"* [182].

A dynamical model for synthetic ECG generation was introduced by McSharry *et al.* in 2003 [4]. This original model is based on three coupled ordinary differential equations (in the Cartesian coordinates system), and was initially intended to generate realistic synthetic ECGs for assessing biomedical signal processing techniques. The model uses a set of Gaussian functions to approximate ECG cycles. Sameni *et al.* [184] suggested a change of coordinate system and further simplification of the synthetic model in order to use it as the dynamic model of a KF. An ECG cycle is approximated by a pseudo periodic signal represented by its phase $\theta$ and its amplitude $z$. The continuous set of the equations of the model can be written:

$$\begin{cases} \dot{\theta} = \omega \\ \dot{z} = -\sum_{i=1}^{N} \frac{\alpha_i \omega}{b_i^2} \Delta\theta_{i,k} exp(-\frac{\Delta\theta_{i,k}^2}{2b_i^2}), \end{cases} \tag{A.9}$$

where $\{\alpha_i\}$, $\{b_i\}$, $\{\xi_i\}$ correspond to the peak amplitude, width and centre parameters of the $N$ Gaussian functions respectively. $\Delta\theta_{i,k} \equiv (\theta - \xi_i) \mod 2\pi$ and $w = 2\pi f$ is the velocity of the trajectory with $f$ equal to the beat-to-beat heart rate (HR). The equations can be discretised as follows by considering a small sampling period $\delta$:

$$\begin{cases} \theta_{k+1} = (\theta_k + \omega\delta) \mod 2\pi \\ z_{k+1} = z_k - \sum_{i=1}^{N} \delta\frac{\alpha_i \omega}{b_i^2} \Delta\theta_{i,k} exp(-\frac{\Delta\theta_{i,k}^2}{2b_i^2}) + \eta_k, \end{cases} \tag{A.10}$$

where $\theta_k$ and $z_k$ are the discrete phase and amplitude, $\Delta\theta_{i,k} = \theta_k - \xi_i$ and $\eta$ is a perturbation term that corresponds to a random additive noise modelling the error made by substituting the model for a real ECG. This includes the modelling approximation as well as physiological noises such as baseline wander, electrode noise and respiration modulation [188]. Figure A.2 illustrates how an ECG cycle is approximated by a set of $N = 7$ Gaussian functions with the parameters given in Table A.1.

### A.3.2 Model integration into the Bayesian framework

If $\theta_k$ and $z_k$ are considered to be the state variables and $\omega_k, \{\alpha_{i,k}\}, \{\xi_{i,k}\}, \{b_{i,k}\}, \eta_k$ to be process noises, we can therefore define the state and state noise vector as:

Table A.1: Parameters for the Gaussian functions in Figure A.2.

| Index(i) | 1 | 2 | 3 | 4 | 5 | 6 | 7 |
|---|---|---|---|---|---|---|---|
| $\alpha_i(mV)$ | 0.27 | 0.04 | -0.13 | -0.13 | 0.71 | 0.37 | 0.06 |
| $\xi_i(rads)$ | 1.96 | 2.16 | 1.97 | 0.04 | 0.02 | -0.05 | -0.94 |
| $b_i(rads)$ | 0.37 | 0.12 | 0.59 | 0.38 | 0.06 | 0.06 | 0.13 |

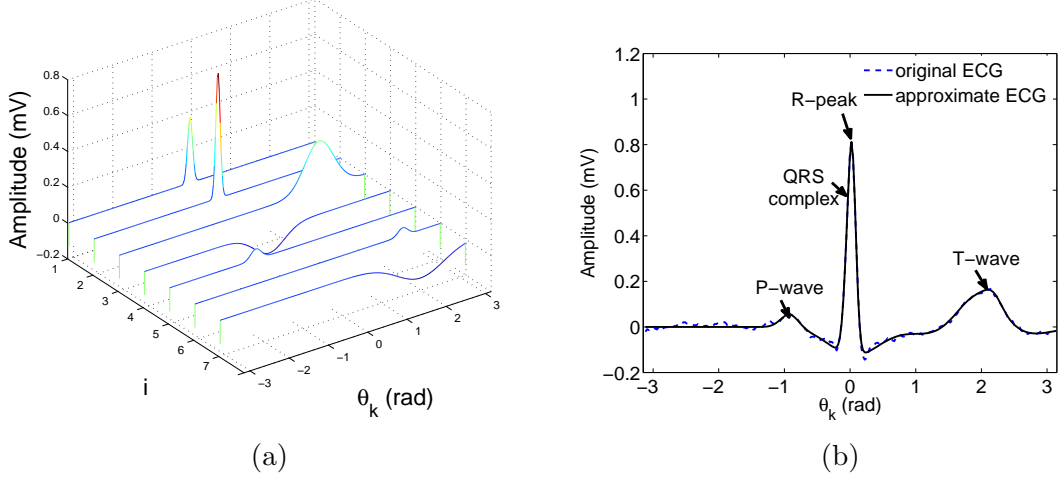

(a)  (b)

Figure A.2: (a) The individual seven Gaussians (index i) used to represent an ECG cycle. (b) Original ECG cycle and the reconstructed ECG cycle obtained by summing the seven Gaussian functions in (a).

$$\underline{\mathbf{x}}_k = [\theta_k, z_k]^T, \tag{A.11}$$

$$\underline{\mathbf{w}}_k = [\alpha_{1,k}, ..., \alpha_{N,k}, b_{1,k}, ..., b_{N,k}, \xi_{1,k}, ..., \xi_{N,k}, \omega_k, \eta_k]^T. \tag{A.12}$$

In this formulation of the state vector, only two states are considered: phase and amplitude.

### A.3.3 State equations

The model equations $\underline{\mathbf{x}}_{k+1} = g(\underline{\mathbf{x}}_k, \underline{\mathbf{w}}_k)$ are linearised in order to use the EKF. We write:

$$\begin{cases} g_k^1(\theta_k, \omega_k) = (\theta_k + \omega_k \delta) \bmod 2\pi \\ g_k^2(\theta_k, z_k, \alpha_{i,k}, \theta_{i,k}, b_{i,k}, \eta_k, \omega_k) = -\sum_{i=1}^{N} \delta \frac{\alpha_{i,k}\omega_k}{b_{i,k}^2} \Delta\theta_{i,k} exp(-\frac{\Delta\theta_{i,k}^2}{2b_{i,k}^2}) + z_k + \eta_k. \end{cases} \tag{A.13}$$

With these notations, and by using the Taylor series expansion around $(\hat{\underline{\mathbf{x}}}_k^+, \hat{\underline{\mathbf{w}}}_k)$:

$$\underline{\mathbf{x}}_{\mathbf{k+1}} = \begin{bmatrix} g_k^1(\hat{\underline{\mathbf{x}}}_k^+, \hat{\underline{\mathbf{w}}}_k) \\ g_k^2(\hat{\underline{\mathbf{x}}}_k^+, \hat{\underline{\mathbf{w}}}_k) \end{bmatrix} + \mathbf{G}_k(\underline{\mathbf{x}}_k - \hat{\underline{\mathbf{x}}}_k^+) + \mathbf{L}_k(\underline{\mathbf{w}}_k - \hat{\underline{\mathbf{w}}}_k), \quad (A.14)$$

where

$$\mathbf{G}_k = (a_{i,j}) = \left.\frac{\partial g_k}{\partial \underline{\mathbf{x}}}\right|_{\hat{\underline{\mathbf{x}}}_k^+, \hat{\underline{\mathbf{w}}}_k}, \mathbf{L}_k = (c_{i,j}) = \left.\frac{\partial g_k}{\partial \underline{\mathbf{w}}}\right|_{\hat{\underline{\mathbf{x}}}_k^+, \hat{\underline{\mathbf{w}}}_k},$$

and

$$a_{1,1}[k] = \left.\frac{\partial g_k^1}{\partial \theta_k}\right|_{\hat{\underline{\mathbf{x}}}_k^+, \hat{\underline{\mathbf{w}}}_k} = 1$$

$$a_{2,1}[k] = \left.\frac{\partial g_k^2}{\partial \theta_k}\right|_{\hat{\underline{\mathbf{x}}}_k^+, \hat{\underline{\mathbf{w}}}_k} = -\sum_{i=1}^N \delta \frac{\alpha_i \omega}{b_i^2}[1 - \frac{\Delta \theta_{i,k}^2}{b_i^2}] \exp\left(-\frac{\Delta \theta_{i,k}^2}{2b_i^2}\right)$$

$$a_{2,2}[k] = \left.\frac{\partial g_k^2}{\partial z_k}\right|_{\hat{\underline{\mathbf{x}}}_k^+, \hat{\underline{\mathbf{w}}}_k} = 1$$

$\forall i \in [1\ N]$,

$$c_{2,i}[k] = \left.\frac{\partial g_k^2}{\partial \alpha_i}\right|_{\hat{\underline{\mathbf{x}}}_k^+, \hat{\underline{\mathbf{w}}}_k} = -\delta \frac{\omega \Delta \theta_{i,k}}{b_i^2} \exp\left(\frac{-\Delta \theta_{i,k}^2}{2b_i^2}\right)$$

$$c_{2,i+N}[k] = \left.\frac{\partial g_k^2}{\partial \theta_i}\right|_{\hat{\underline{\mathbf{x}}}_k^+, \hat{\underline{\mathbf{w}}}_k} = \delta \frac{\alpha_i \omega}{b_i^2}[1 - \frac{\Delta \theta_{i,k}^2}{b_i^2}] \exp\left(-\frac{\Delta \theta_{i,k}^2}{2b_i^2}\right)$$

$$c_{2,i+2N}[k] = \left.\frac{\partial g_k^2}{\partial b_i}\right|_{\hat{\underline{\mathbf{x}}}_k^+, \hat{\underline{\mathbf{w}}}_k} = 2\delta \frac{\alpha_i \omega \Delta \theta_{i,k}}{b_i^3}[1 - \frac{\Delta \theta_{i,k}^2}{2b_i}] \exp\left(-\frac{\Delta \theta_{i,k}^2}{2b_i^2}\right)$$

$$c_{1,3N+1}[k] = \left.\frac{\partial g_k^1}{\partial \omega}\right|_{\underline{\mathbf{x}}_k^+, \hat{\underline{\mathbf{w}}}_k} = \delta$$

$$c_{2,3N+1}[k] = \left.\frac{\partial g_k^2}{\partial \omega}\right|_{\underline{\mathbf{x}}_k^+, \hat{\underline{\mathbf{w}}}_k} = -\sum_{i=1}^N \delta \frac{\alpha_i \Delta \theta_{i,k}}{b_i^2} \exp\left(-\frac{\Delta \theta_{i,k}^2}{2b_i^2}\right)$$

$$c_{2,3N+2}[k] = \left.\frac{\partial g_k^2}{\partial \eta}\right|_{\underline{\mathbf{x}}_k^+, \hat{\underline{\mathbf{w}}}_k} = 1,$$

with $\Delta \theta_{i,k} = \theta_k - \hat{\xi}_i$ and the assumption that $(\underline{\mathbf{w}}_k - \hat{\underline{\mathbf{w}}}_k)$ is a white and zero-mean random variable.

### A.3.4 Observation equations

The observation vector can be related to the state variables as follows:

$$\underline{\mathbf{y}}_k = \begin{bmatrix} \phi_k \\ s_k \end{bmatrix} = \begin{bmatrix} 1 & 0 \\ 0 & 1 \end{bmatrix} \cdot \begin{bmatrix} \theta_k \\ z_k \end{bmatrix} + \begin{bmatrix} v_{1,k} \\ v_{2,k} \end{bmatrix} = \mathbf{H}_k \underline{\mathbf{x}}_k + \mathbf{U}_k \underline{\mathbf{v}}_k. \quad (A.15)$$

With the notations introduced in Section A.2:

$$\mathbf{H}_k = \left.\frac{\partial h_k}{\partial \underline{\mathbf{x}}}\right|_{\hat{\underline{\mathbf{x}}}_k^+, \hat{\underline{\mathbf{v}}}_k} = \mathbf{I}, \quad \mathbf{U}_k = \left.\frac{\partial h_k}{\partial \underline{\mathbf{v}}}\right|_{\hat{\underline{\mathbf{x}}}_k^+, \hat{\underline{\mathbf{v}}}_k} = \mathbf{I},$$

where $\mathbf{I}$ the identity matrix and with the assumption that $\underline{\mathbf{v}}_k$ is a white and zero-mean random variable. The observation equation A.15 represents the relationship between the system state and the observations. Two observations are considered: $\phi_k$ and $s_k$, the phase and amplitude of the ECG signal respectively with their measurement noises $v_{1,k}$ and $v_{2,k}$. In order to evaluate the phase $\phi_k$, peak detection is performed and a phase is assigned linearly to each sample of a given ECG cycle with $\phi_k = 0$ at the peak location. The phase signal is defined to have a saw-tooth-shape with phase $\phi_k \in [-\pi; \pi]$ assigned linearly to ECG samples that are between two successive R-peaks (see Figure A.3).

### A.3.5 Model parameters

The described KF framework needs to define the observation noise covariance matrix $\mathbf{R}_k = E[\underline{\mathbf{v}}_k \underline{\mathbf{v}}_k^T]$ and the process noise covariance matrix $\mathbf{Q}_k = E[\underline{\mathbf{w}}_k \underline{\mathbf{w}}_k^T]$. The noise covariance matrices are inherent to the system, but there is a need to set the values of these matrices appropriately. If it is further assumed that the components of the noise processes are uncorrelated then $\mathbf{R}_k$ and $\mathbf{Q}_k$ are diagonal:

$$\mathbf{R}_k = diag(\sigma_{v_{1,k}}^2, \sigma_{v_{2,k}}^2), \tag{A.16}$$

$$\mathbf{Q}_k = diag(\sigma_{\alpha_{1,k}}^2, \ldots, \sigma_{\alpha_{N,k}}^2, \sigma_{b_{1,k}}^2, \ldots, \sigma_{b_{N,k}}^2, \sigma_{\xi_{1,k}}^2, \ldots, \sigma_{\xi_{N,k}}^2, \sigma_{\omega_k}^2, \sigma_{\eta_k}^2) \tag{A.17}$$

where $\mathbf{Q}_k \in \Re^{(3N+2)\times(3N+2)}$ and $\mathbf{R}_k \in \Re^{2\times 2}$ and $\sigma_x^2 = E[(x-\hat{x})^2]$, where $\hat{x}$ is the mean of the random variable $x$. The parameters are evaluated from a phase-wrapped average ECG waveform as described in Sameni *et al.* [184]. Table A.2 describes how the parameters can be evaluated from the average ECG waveform.

Figure A.4a illustrates the typical filtering results obtained after applying the EKF framework on a real, noisy ECG segment. The Bayesian framework denoises the signal while preserving the important features of the ECG, in particular the QRS complexes which would be distorted with traditional bandpass filtering for example (see Figure A.4b).

## A.4 EKF in the context of ECG signal processing

Sameni *et al.* [184] introduced the above Kalman filtering framework to ECG filtering in 2007. The authors originally used a set of five Gaussian functions for the ECG model, and benchmarked the Kalman filtering approach with wavelet denoising, adaptive filtering and classical finite impulse response bandpass filters. The algorithms were evaluated on 190, 30 sec ECG segments taken from the MIT-BIH normal sinus rhythm database (NSRDB) [65], to which they added real and synthetic noise with signal to noise ratio (SNR) in the range $[-5\ 30]$ dB. The KF outperformed the other methods in terms of SNR improvement. The main limitations of this first study were that the SNR improvement has no clinical significance by itself. Moreover, the framework was tested on normal rhythm ECG with only one type of real noise: muscle artifact. The EKF used in their paper is referred as EKF2, where 2 stands for the 2 states variables used $(\theta, z)$.

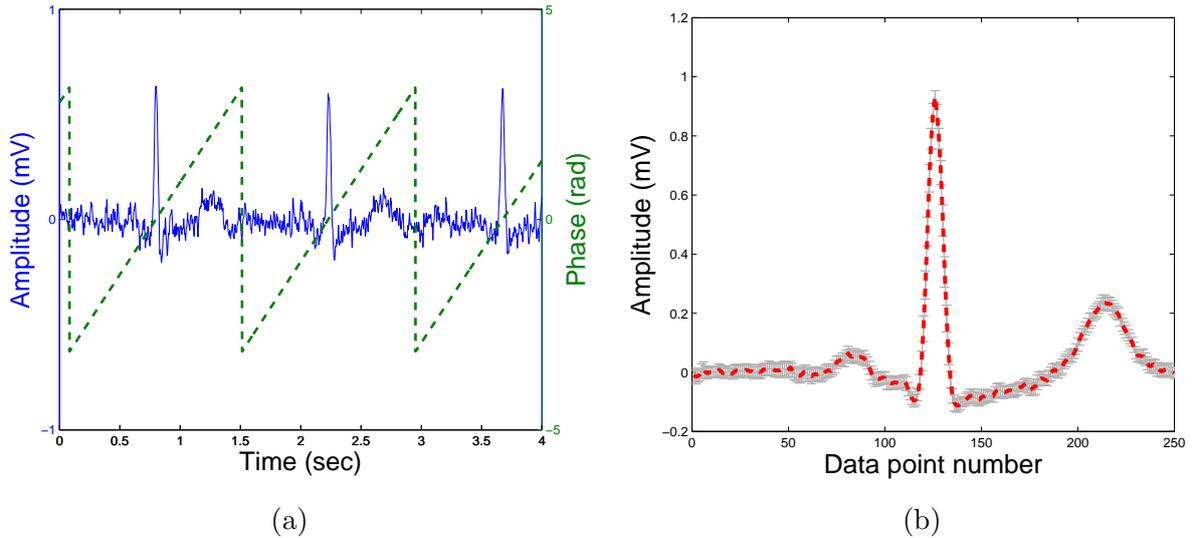

Figure A.3: (a) Illustration of phase assignment. The phase signal is defined to have a sawtooth shape with varying period and phase $\phi_k \in [-\pi\ \pi]$ and is used as one of the observations for each time step $k$. (b) Average ECG waveform on which the noise covariance matrices are evaluated. The dashed red line corresponds to the mean ECG and the grey error bars represent the standard deviation of ECG cycles around the mean ECG.

| Variable | Definition | Estimation |
| --- | --- | --- |
| $\xi_i(rads)$ | Gaussian kernel centre | Gaussian kernel centre $\pm 0.05\pi$ |
| $\alpha_i(mV)$ | Gaussian kernel peak | Gaussian kernel peak $\pm 10\%$ of the peak amplitude |
| $b_i(rads)$ | Gaussian kernel width | Gaussian kernel width $\pm 0.05\pi$ |
| $\eta(mV)$ | Observation noise | 1% of maximum ECG peak |
| $\omega(rads/sec)$ | Angular phase | Mean beat-to-beat angular frequency $(\hat{\omega}) \pm \sigma_\omega$ |
| $v_1(rads)$ | Phase observation noise | $0.0 \pm (\hat{\omega}\delta)/\sqrt{12}$ |
| $v_2(mV)$ | Amplitude observation noise | Ranges over different SNRs |

Table A.2: Evaluating the parameters of the noise covariance matrices. Note that the phase observation noise is assumed to be only dependant on the sampling frequency which means that R-peak detection is assumed to be 'very' accurate.

Sayadi *et al.* [157] modified the KF framework in order to consider the Gaussian parameters $\{\alpha_i\}, \{b_i\}, \{\xi_i\}$ as state variables (denoted EKF17). The purpose was to make the artificial ECG model adaptable to variations in the ECG signals. As the ECG is a non-stationary pro-cess, the parameters (and thus the prior on the model) should be allowed to evolve. The evolution equation for these parameters assumes a random walk evolution: $p_{i,k} = p_{i,k} + u_{i,k}$, where $p_i$ is any of the Gaussian parameters. The authors reported improvement over the performance of the EKF2. Moreover, the authors assessed the performance of the Gaussian model as a compression technique using the weighted diagnostic distortion ($WDD$) measure [189], which is based on comparing a weighted sum of clinically relevant features of the original and reconstructed ECGs. $WDD$ is defined as $WDD(\beta, \hat{\beta}) = \Delta\beta^T \frac{\mathbf{\Lambda}}{tr(\mathbf{\Lambda})} \Delta\beta \cdot 100$, where $\beta$ and $\hat{\beta}$ are two vectors containing 18 clinically diagnostic features of the original and reconstructed cycles, $\Delta\beta = |\beta - \hat{\beta}|$ and $\mathbf{\Lambda}$ is a diagonal matrix of weights as in [189]. The authors tested the algorithm on the MIT-BIH compression test database and reported $WDD < 1.73\%$ while providing a compression ratio (CR) of 11.37 : 1. Improvement of the $WDD$ was possible with the addition of more Gaussians for a better description of the signal, but resulted in a decrease of the CR.

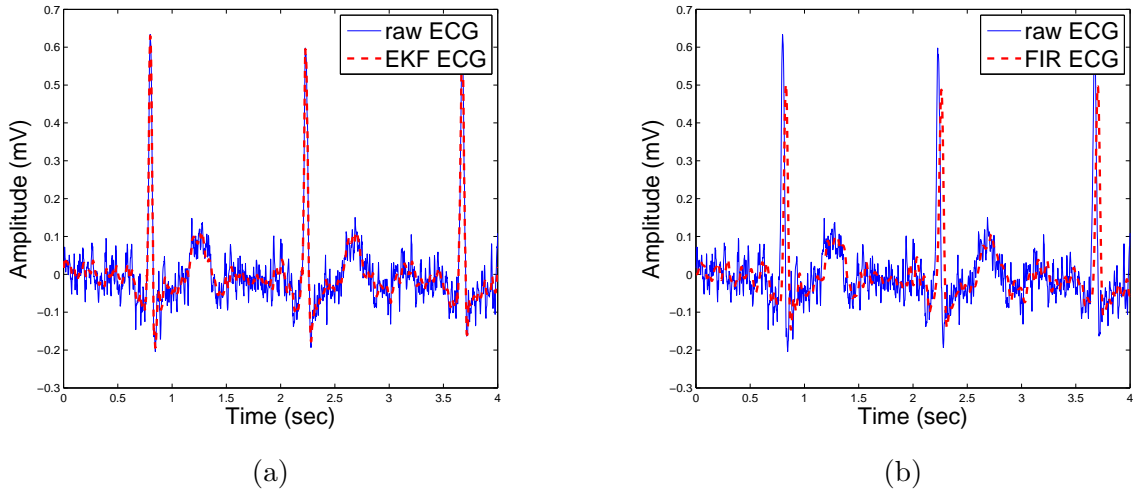

Figure A.4: (a) output example for the EKF framework. (b) output example using a standard ECG finite impulse response filter with passband [0 45] Hz.

In Sayadi *et al.* [156], the authors used the EKF17 algorithm for ECG beat segmentation and extraction of fiducial points by relating the Gaussian functions parameters, $\{\alpha_i\}, \{\beta_i\}, \{\xi_i\}$, to the ECG fiducial points, which was originally proposed by Clifford *et al.* [81, 155]. Five Gaussian kernels were used in the work of Sayadi *et al.* (one to represent each of the turning points in a beat). The algorithm was evaluated on 80 low noise segments of 30 sec clean ECG segments taken from the Physionet Noise stress test database, and showed a very accurate detection of fiducial points whose location was compared with expert label annotations. Although the authors claimed that their method performed better than the other benchmark methods, no quantitative comparison of fiducial point detection accuracy was performed on the particular ECG segments from this study. It should be noted that the method was tested for normal rhythms only and considered all the data as training set.

In 2010, Sayadi *et al.* [82] applied the ECG Kalman framework for detecting abnormal cardiac cycles knows as premature ventricular contractions (PVCs). The equations were modified in order to track the ECG characteristic waveforms (CWs) by assuming that the signal could be divided into three components representing the different CWs: the P wave, QRS complex and T wave (EKF4). Seven Gaussian kernels were used to model an ECG cycle: 2 for the P wave, 3 for the QRS complex and 2 for the T wave (see Figure A.5). This number was motivated by Clifford *et al.* [81] who demonstrated that asymmetric turning points (P and T) require more than one (symmetric) Gaussian function to be approximated (two being the minimum). Sayadi *et al.* used the signal fidelity, along with a polar representation of the ECG signal (*Polargram*) to identify PVCs. Figure A.6 illustrates how the innovation signal varies when a PVC beat is encountered. Note that the signal fidelity measure acts as an abnormality detector (important morphological changes in the signal with respect to the ECG model) but does not differentiate between noise and abnormality. This is why the *Polargram* was used as a complement in classifying the abnormality as PVC or non PVC. The authors reported an accuracy of 99.10% for PVC detection performed on 40 records taken from the MIT-BIH Arrhythmia Database. The EKF4 model was further assessed in [188] for its denoising performance and compared with EKF17 and EKF2; the authors obtained a better SNR improvement with the CWs model when testing the filters on 13 hand-selected records of 60 min each taken from the NSRDB with various level of real electromyographic noise ($SNR \in [-6\ 24]db$). However, it is unclear whether the Gaussian fitting was performed in an automated or semi-automated fashion.

In later work, Sayadi *et al.* [190] used the CWs based model for arrhythmia detection and classification. In this work, the dynamical model was applied to the ECG, as well as other waveforms such as arterial blood pressure and the photoplethysmograph, in order to combine the information provided by the different signals. Two steps were performed: abnormality detection (using the signal fidelity of the different waveforms) and abnormality determination (using the *polargram*). The authors tested their algorithm on a subset of records from the MIMIC II database [191] containing 1500 life-threatening arrhythmias. Sensitivity of the arrhythmia detection and classification ranged between 93.5% and 100% for the different types of arrhythmias. The process was further used for determining the status of the alarm, as a true alarm (TA) or a false alarm (FA), reducing the overall FA rate from 42.3% to 9.9% while suppressing 1.5% of the TA (all of which were ventricular tachycardia alarms).

In the context of magnetic resonance imaging (MRI) ECG, Oster *et al.* [158, 159] combined the ECG model with a magnetic field gradient model (MFG) in order to remove the severe distortion induced by the MFG. The combined model was evaluated on 14681 QRS complexes, representing about 3.5 hours of MR-ECG acquisition recorded from 13 healthy patients. The method allowed significant noise reduction and improved the quality of QRS detection over state of the art methods. In more recent work, Oster *et al.* [192] worked on suppressing the magneto-hydro-dynamic (MHD) effect distortion from the ECG. In this context an additional equation was added to the ECG observation equations modelling the MHD contribution on the signal waveform. Promising outcomes were obtained using synthetic data.

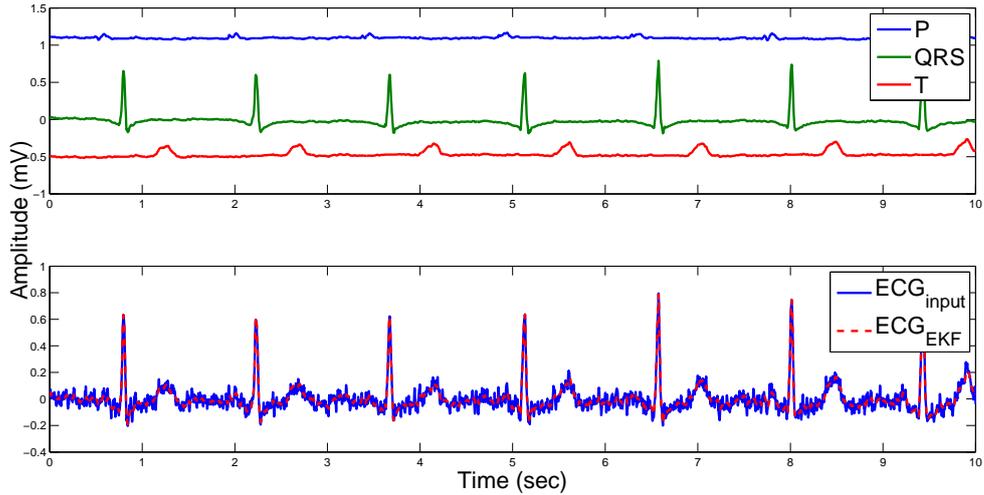

Figure A.5: Example of CWs decomposition (top). The sum of the individual waves gives $\text{ECG}_{EKF}$ (bottom).

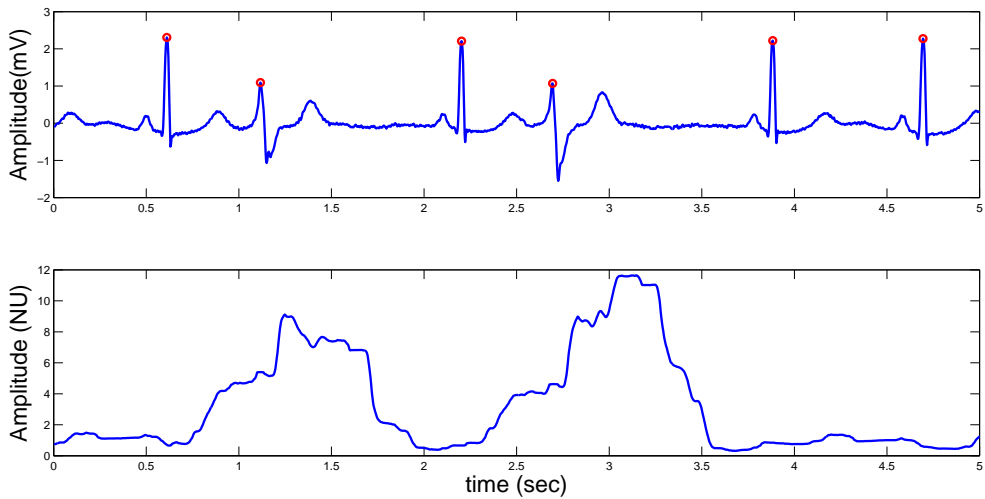

Figure A.6: Innovation signal illustration for PVC identification. Upper plot: ECG signal with two PVC beats. Bottom: signal fidelity (no unit). Note the signal fidelity increases when the PCV beats are encountered.

In the context of NI-FECG, the EKF has been used for filter out the MECG from the abdominal mixture as explained in Chapter 3. It was used first by Sameni *et al.* [9] and was further used for this application by a number of researchers (e.g. [61]). However, the method was not rigorously benchmarked to alternative NI-FECG extraction methods.

# A.5 Conclusion

In conclusion, the ECG Bayesian filtering framework takes advantage of prior knowledge of the ECG morphology to accurately denoising and track the non-stationarity of the signal. The framework has been used in the context of data filtering, compression, classification of ECG rhythms, beat typing and source separation. For these works, the Bayesian framework has had different variations as to the number of Gaussian to use, whether some of the Gaussian are forced to represent the P-QRS-T waves, and whether the prior (Gaussian parameters) are left to evolve i.e. are part of the state equations. The following points have been identified for further the research:

- the initialisation of the Gaussian parameters was often performed manually in the reviewed papers. This is not realistic within an automated system, and so a robust automated initialisation procedure is required. This was addressed in Chapter 7.

- the performance of the KF framework has not been satisfactorily quantitatively evaluated for its application to NI-FECG extraction and benchmarked against alternative methods. This was addressed in Chapter 7.

- it was important to note that the states of the KF were chosen differently from one study to another. Some considered only two states (EKF2, phase and amplitude), some considered four states (EKF4, amplitude of P/QRS/T waves separately and phase) and others even more states (considering the Gaussian parameters as states). It is important to clarify what choice make sense in the context of NI-FECG extraction. This was discussed in Chapter 8.

- the covariance matrices $\mathbf{Q}$ and $\mathbf{R}$ are usually estimated from the first few ECG cycles, but an online extension that would update the values in $\mathbf{Q}$ and $\mathbf{R}$ from the most recent cycles, or use some measure of the signal quality would, most likely, improve the performances. This item was discussed in Chapter 7.

- a varying number of Gaussian kernels have been used for Gaussian fitting. However, some quantitative analysis relating the performance of NI-FECG extraction to this variable (number of Gaussian kernels) would provide better insight into its importance. Some preliminary work of this problem has been performed [193], but deeper analysis is required.

# Appendix B

# Background on PLA Algorithm

This Appendix presents the model and mathematical derivation of the Probabilistic Label Aggregator (PLA) algorithm used in the context of medical annotation crowd-sourcing and, more specifically, QT annotations fusion in the context of this thesis.

## B.1  Introduction

The expectation maximisation (EM) algorithm is an iterative procedure that is used in statistics to find the maximum likelihood (ML) or maximum a posteriori (MAP) estimates of some latent parameters of a statistical model in the presence of missing or hidden data. The EM estimates some unknown parameters $\theta$ given some observed data $\mathbf{D}$. The EM algorithm is divided in two steps: the expectation step (or E-step) and the maximisation step (M-step). The E-step is used to estimate the expectation of the log-likelihood given an estimate of the latent variables. The M-step is used to compute the ML estimate of the latent variables. The E and M steps are repeated iteratively until convergence, or when a user specified number of iterations has been reached. One of the earliest reference for the EM is the work by Hartley [194] but the formalism and proof of convergence of the algorithm is attributed to Dempster *et al.* [195]. The so called PLA algorithm presented in this section is adapted from the algorithm presented in Raykar *et al.* [73] for its application to QT annotation crowd-sourcing.

## B.2  Mathematical basis

In this work, the EM was used to compute the ML of the PLA model for combining crowd-sourced medical QT annotations. In this context, it is assumed that $R$ annotators have annotated a series of $N$, QT observations. We write $z_i$ the true QT annotation for each individual record $i \in [1\ R]$ and $y_i^j$ is the annotation of annotator $j$ performed on record $i$. The distribution of $y_i^j$ is modelled as:

$$P[y_i^j \mid z_i, \tau^j] = \mathcal{N}(y_i^j \mid z_i, 1/\tau^j) \tag{B.1}$$

where $\tau^j$ is the precision of the $j$th annotator, defined as the estimated inverse-variance of annotator $j$ ($\tau^j = 1/(\sigma^j)^2$) and $\mathcal{N}(. \mid \mu, \sigma)$ stands for the normal distribution with mean $\mu$ and standard deviation $\sigma$. Note that the $\tau^j$ are considered to be constants, i.e. all annotators are assumed to be consistent throughout records. In words, this equations means that for a given annotator, $j$, the annotation of this annotator on record $i$ is described as a distribution centred on the *true* annotation, with a variance that is inversely proportional to his performance (i.e. how good this specific annotator is).

Furthermore, we assume that $z_i$ can be predicted using a linear regression model as $z_i = \underline{\mathbf{w}}^T \underline{\mathbf{x}}_i + \epsilon$, where $\underline{\mathbf{w}} \in \Re^N$ is the regression coefficients; $\epsilon$ is a zero-mean Gaussian noise with precision $\gamma$; $\underline{\mathbf{x}} \in \Re^N$ is the feature vector for all annotation of annotator $i$. The precision represents the incertitude in the prediction model. The probability of $z_i$ may be defined as:

$$P[z_i \mid \underline{\mathbf{w}}, \underline{\mathbf{x}}_i, \gamma] = \mathcal{N}(z_i \mid \underline{\mathbf{w}}^T \underline{\mathbf{x}}_i, 1/\gamma). \tag{B.2}$$

By combining both the annotator (B.1) and the predictive model (B.2), the distribution of $y_i^j$ can be estimated as being the following posterior:

$$
\begin{aligned}
P[y_i^j \mid \underline{\mathbf{w}}, \underline{\mathbf{x}}_i, \tau^j, \gamma] &= \int_{z_i} P[y_i^j, z_i \mid, \underline{\mathbf{w}}, \underline{\mathbf{x}}_i, \tau^j, \gamma] dz_i \\
&= \int_{z_i} P[y_i^j \mid z_i, \tau^j] P[z_i \mid \underline{\mathbf{w}}, \underline{\mathbf{x}}_i, \gamma] dz_i \\
&= \int_{z_i} \mathcal{N}(y_i^j \mid z_i, 1/\tau^j) \mathcal{N}(z_i \mid \underline{\mathbf{w}}^T \underline{\mathbf{x}}_i, 1/\gamma) dz_i \\
&= \int_{z_i} \frac{1}{\sqrt{2\pi/\tau^j}} e^{-(y_i^j - z_i)^2 / (\frac{2}{\tau^j})} \frac{1}{\sqrt{2\pi/\gamma}} e^{-(z_i - \underline{\mathbf{w}}^T \underline{\mathbf{x}}_i)^2 / (\frac{2}{\gamma})} dz_i \\
&= \frac{1}{\sqrt{2\pi(\frac{1}{\gamma} + \frac{1}{\tau^j})}} e^{-(y_i^j - \underline{\mathbf{w}}^T \underline{\mathbf{x}}_i)^2 / (2(\frac{1}{\gamma} + \frac{1}{\tau^j}))} \\
&= \mathcal{N}(y_i^j \mid \underline{\mathbf{w}}^T \underline{\mathbf{x}}_i, 1/\lambda^j)
\end{aligned}
\tag{B.3}
$$

where $\underline{\lambda}$ is the precision for all annotators with $\underline{\lambda} = \{\lambda^1, ..., \lambda^R\}$ and $1/\lambda^j = 1/\tau^j + 1/\gamma$. Under the assumption that records are independent, the likelihood of the parameter $\theta = \{\underline{\mathbf{w}}, \underline{\lambda}\}$ for a given dataset $\mathbf{D}$ can be formulated as:

$$P[\mathbf{D} \mid \theta] = \prod_{i=1}^{N} P[y_i^1, \cdots, y_i^R \mid \underline{\mathbf{x}}_i, \theta] \tag{B.4}$$

Assuming $y_i^1, \cdots, y_i^R$ are conditionally independent given the feature $\underline{\mathbf{x}}_i$ (i.e. each annotator works independently), the likelihood of the parameter $\theta = \{\underline{\mathbf{w}}, \underline{\lambda}\}$ for a given dataset $\mathbf{D}$ can be written as:

$$P[\mathbf{D} \mid \theta] = \prod_{i=1}^{N} \prod_{j=1}^{R} \mathcal{N}(y_i^j \mid \underline{\mathbf{w}}^T \underline{\mathbf{x}}_i, 1/\lambda^j) \tag{B.5}$$

The parameters $\theta$ can be estimated by maximising the log-likelihood ($\ln P[\mathbf{D} \mid \theta]$). The *log* is a monotonically increasing function and so the likelihood and log-likelihood will be maximum on the same points. However, the log-likelihood is more convenient to work with because it transforms the multiplications into additions before computing the gradient. The log-likelihood can then be rewritten as:

$$\ln P[\mathbf{D} \mid \theta] = -\frac{1}{2} \sum_{i=1}^{N} \sum_{j=1}^{R} [\ln(\frac{2\pi}{\lambda^j}) + (y_i^j - \underline{\mathbf{w}}^T \underline{\mathbf{x}}_i)^2 \lambda^j]. \tag{B.6}$$

The maximum likelihood estimator is found by:

$$\hat{\theta}_{ML} = \{\hat{\underline{\mathbf{w}}}, \hat{\underline{\lambda}}\} = \operatorname{argmax}_\theta \{\ln P[\mathbf{D} \mid \theta]\}. \tag{B.7}$$

The $\theta_{ML}$ can be obtained by estimating the gradient of the log-likelihood respectively:

$$\frac{d \ln P[\mathbf{D} \mid \theta]}{d \lambda^j} = -\frac{1}{2} \sum_{i=1}^{N} [(y_i^j - \underline{\mathbf{w}}^T \underline{\mathbf{x}}_i)^2 - \frac{1}{\lambda^j}]$$

$$\frac{d \ln P[\mathbf{D} \mid \theta]}{d \underline{\mathbf{w}}} = \sum_{i=1}^{N} \sum_{j=1}^{R} y_i^j \underline{\mathbf{x}}_i \lambda^j - \underline{\mathbf{w}}^T \underline{\mathbf{x}}_i \underline{\mathbf{x}}_i^T \lambda^j$$

By stetting derivatives to zero, the parameters can be estimated as:

$$\frac{1}{\hat{\lambda}^j} = \frac{1}{N} \sum_{i=1}^{N} (y_i^j - \hat{\underline{\mathbf{w}}}^T \underline{\mathbf{x}}_i)^2 \tag{B.8}$$

$$\hat{\underline{\mathbf{w}}} = (\sum_{i=1}^{N} \underline{\mathbf{x}}_i \underline{\mathbf{x}}_i^T)^{-1} \sum_{i=1}^{N} \underline{\mathbf{x}}_i \hat{z}_i \tag{B.9}$$

with

$$\hat{z}_i = \frac{\sum_{j=1}^{R} y_i^j \hat{\lambda}^j}{\sum_{j=1}^{R} \hat{\lambda}^j}$$

This likelihood problem can be maximised and solved using the EM algorithm in a two-step iterative process:
(1) The E-step, which estimates the expected true annotations for all records, $\hat{\underline{\mathbf{z}}}$, as a weighted

sum of the provided annotations with their precisions, can be written as:

$$\hat{\underline{z}} = \frac{\sum_{j=1}^{R} \underline{y}^j \hat{\lambda}^j}{\sum_{j=1}^{R} \hat{\lambda}^j} \qquad (B.10)$$

(2) The M-step is based on the current estimation of $\hat{\underline{z}}$ and given dataset $\mathbf{D}$. The model parameters $\hat{\underline{\mathbf{w}}}$ and $\hat{\lambda}$ can be updated using equations (B.8) and (B.9) accordingly in a sequential order.

## B.3 EM and ECG annotations

A first experiment was published in joint work in Zhu *et al.* [150]. In this experiment, the EM algorithm was used to combine annotations from human annotators and automated algorithms, and managed to achieve better results than the best entry to the Physionet Challenge 2006 that addressed the problem of adult QT measurement.

A second experiment published by Zhu *et al.* [71] was conducted in-house at the Oxford Institute of Biomedical Engineering where over 7,307 FQT annotations were gathered from a total of 23 annotators (PhD students and post-docs), who annotated scalp ECG segments from 15 pregnant women. This experiment was targeted at stress-testing the LightWAVE crowd-sourcing annotation interface, and evaluating the EM algorithm in fusing annotations from a number of independent annotators in the context of FQT measurement. Figure B.1 illustrates some of the analysis performed on these data to ensure that the EM approach, in the context of FQT aggregation, was behaving as expected, given that the underlying ground truth is unknown.

## B.4 Limitations of the EM approach

An element that was found to be missing in the presented EM model is to account for annotators bias. It is possible for an annotator to be accurate, but systematically over- or underestimate the QT interval. Figure B.2 demonstrates this effect for the in-house trial conducted on FQT annotations; some annotators, like 15 and 18, are biased but their boxplot suggests that they are consistent in the way they annotate. Similar bias can be expect from automated algorithms, depending on how they were trained and if they tend to over- estimate or underestimate the *true* QT. New work in jointly estimating the crowd sourced QT and annotator bias is about to be submitted [196].

The annotators were assumed to act independently. However, this is not be necessarily true, especially in cases where annotations from automated algorithms are used. Many such algorithms (e.g. QT detector) may be partially dependent variations of the same approach. However, this choice was made to simplify the model and subsequent derivation of

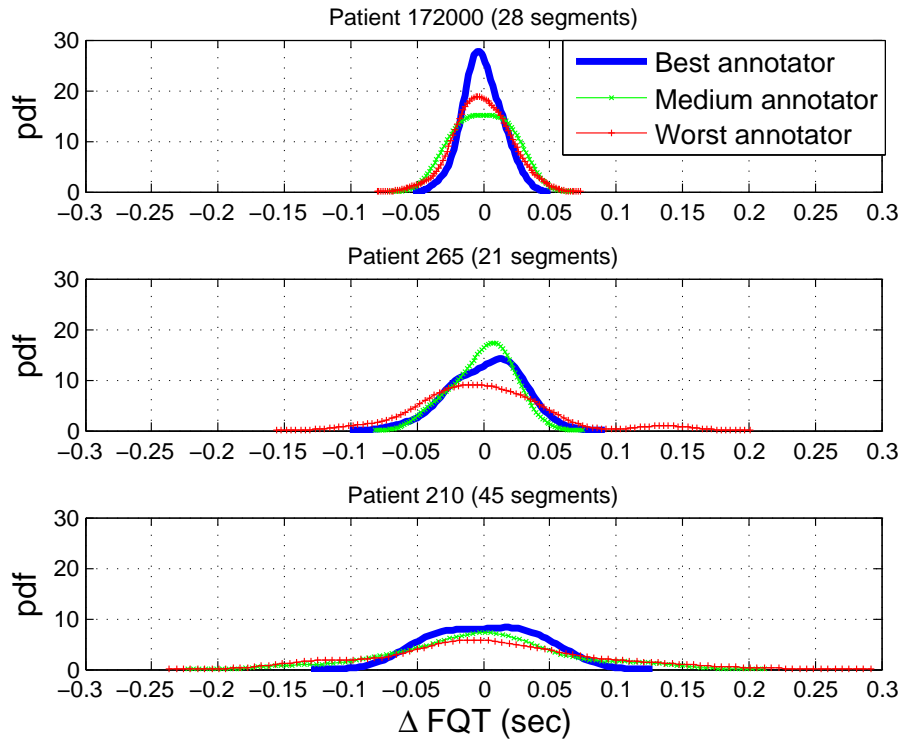

Figure B.1: The variance of FQT annotations ($\triangle$FQT) for three annotators identified by the EM algorithm as being good, mediocre and poor are plotted for segments with very good, medium, and bad signal quality from top to bottom respectively. pdf: probability density function.

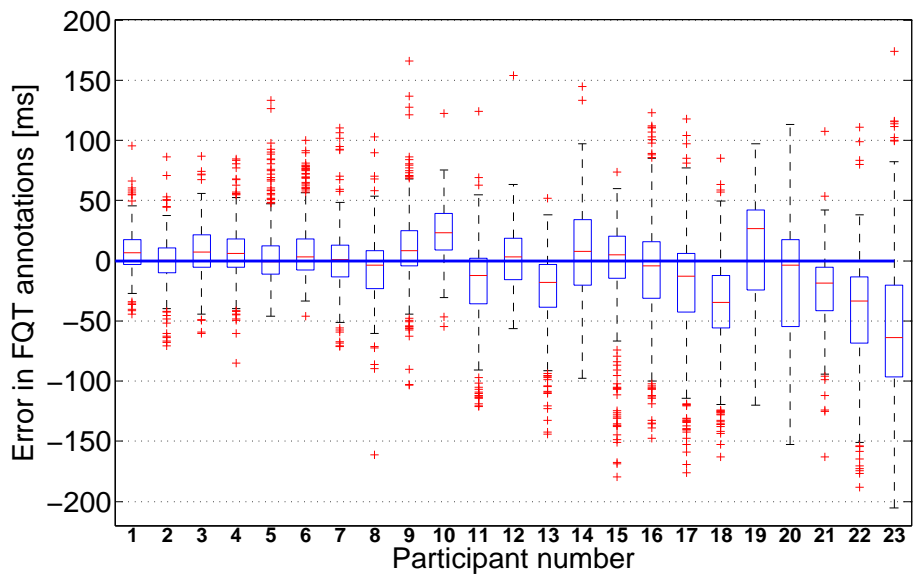

Figure B.2: A boxplot of the error of submitted FQT annotations when compared with annotations inferred by the EM algorithm. This plot illustrates that some annotators (such as 10 and 18) are biased. Error corresponds to the root mean square error.

the likelihood. (In the context of this thesis independent human annotators were used.)

The prediction model $\underline{\mathbf{y}} = \underline{\mathbf{w}}^T \underline{\mathbf{x}}$ assumes, in the context of QT annotations, that it is possible to predict the QT length using some features extracted from the signal. In [150] the heart rate and signal quality were used as features. In particular, the HR (or equivalently the RR interval) has been shown to be related to the QT length [68] (although there exist different flavour for this relationship).

One problem with the PLA algorithm presented in this section is that the sequences of data to fuse need to be aligned, and the algorithm cannot easily be used for instance in combining FQRS detections from multiple methods (e.g. in the case of the FUSE method presented in this thesis) or channels. Thus its use was limited to QT annotation fusion in this thesis.

## B.5   Conclusion

The PLA algorithm is a probabilistic framework that can be used to fuse biomedical annotations without priors on annotators performance. It assumes that the annotators are independent and identically distributed. It relies on a predictive model which is application specific. The PLA algorithm was used without taking into account the potential bias from the annotators, and work is ongoing for modelling it and reformulating the EM equations.

## B.6   Clarification on contribution

My contribution to the work on the EM was: (1) mathematical re-derivation of the equations presented above in the context of QT; (2) implementation of an initial version of the algorithm; (3) organisation and data formatting for the in-house trial (*experiment 2* - see Section B.3) and ongoing support to the first author of the published papers, including suggesting that the bias be incorporated into the model in future work (to be submitted in [196]). Further development of the algorithm, experiment and data analysis for *experiment 1/2* (see Section B.3) and implementation of the CrowdLabel interface used for crowd sourcing human annotations were the work of Tingting Zhu. Publications co-authored are [71, 70, 150, 196].

# Appendix C

# Background on PCA

This section presents the basic principles of Principal Component Analysis (PCA) that are necessary to understand its strengths and limitations in the context of NI-FECG extraction.

## C.1 Introduction

The goal of PCA is to identify a meaningful basis to re-express a data set. In this basis, it is expected that hidden structures will be revealed, or that the important structure will be better highlighted. Principal component analysis was first discovered by Karl Pearson, an English mathematician. The first formulation of PCA was published in Pearson's work entitled "On Lines and Planes of Closest Fit to Systems of Points in Space" [197]. Since then it has been given a variety of names depending on the field and mathematical formulation. These names include the Hotelling transform in multivariate quality control, the Karhunen-Love transform in signal processing and singular value decomposition (SVD). PCA has been intensively used in the field of Biomedical Engineering for dimensionality reduction, visualisation and source separation. The purpose of this appendix is to derive PCA and highlight the assumptions and associated limitations of the method in the context of NI-FECG extraction. Different approaches [198, 199] for deriving PCA are presented in what follows.

## C.2 Mathematical basis

### C.2.1 Iterative procedure

This first derivation defines the goal of PCA, while highlighting some of its main assumptions. We define, $\underline{\mathbf{m}} = [m_1^k, ..., m_p^k]$ a vector of $p$ observed random variables at time instant $k$. The corresponding design matrix $\mathbf{M}$ is defined as:

$$\mathbf{M} = \begin{bmatrix} m_{11} & \cdots & m_{1q} \\ \vdots & & \vdots \\ m_{p1} & \cdots & m_{pq} \end{bmatrix}, \tag{C.1}$$

where $\mathbf{M} \in \Re^{p \times q}$, $p$ variables, $q$ observations. In the case of NI-FECG extraction, $p$ corresponds to the number of channels (or abdominal sensors) and $q$ corresponds to the observations i.e. the samples observed over time. The covariance measures the degree of the linear relationship between two variables. Highly correlated is synonymous with high redundancy. Let $x$ and $y$ be two random variables with zero mean that are observed $n$ times. The covariance is defined as $cov(x,y) = E[x\,y]$ where $E[z]$ is the expected value of $z$. The covariance can be estimated by:

$$\hat{\sigma}_{xy} = \frac{1}{n-1} \sum_{i=1}^{n} x_i y_i \tag{C.2}$$

The covariance matrix is defined as $E[\mathbf{m}\,\mathbf{m}^T]$ and can be estimated by:

$$\mathbf{C} = \frac{1}{n-1} \mathbf{M}\mathbf{M}^T \tag{C.3}$$

Large off-diagonal terms of the covariance matrix $\mathbf{C}$ are synonymous with large redundancy, a large diagonal term corresponds to interesting structures (see Figure C.1). Thus the PCA problem becomes: how to transform $\mathbf{C}$ into a diagonal matrix? In other words, is it possible to find a basis on which components are uncorrelated between each other, that is the off-diagonal components of the covariance matrix expressed in the new domain are zero? The mathematical problem can be formulated as follow:

$$\underline{\mathbf{w}}_1 = arg\,max_{|\underline{\mathbf{w}}|=1}\{\underline{\mathbf{w}}^T \mathbf{M}\mathbf{M}^T \underline{\mathbf{w}}\}, \tag{C.4}$$

where $\underline{\mathbf{w}}_1$ is the direction of the first principal component (PC). This is highlighted in Figure C.1 where the first PC describes most of the signal content. Equation C.4 states that we are looking for the vector $\underline{\mathbf{w}}$ so that the variance along this axis $\underline{\mathbf{w}}^T \mathbf{M}\mathbf{M}^T \underline{\mathbf{w}}$ is maximal (see Figure C.1 for a graphical representation). Assuming we have determined the first $i-1$ PCs, PC$_i$ is the first principal component of the covariance matrix of the residual design matrix:

$$\mathbf{M}_{res} = \mathbf{M} - \sum_{j=1}^{i-1} \underline{\mathbf{w}}_j \underline{\mathbf{w}}_j^T \mathbf{M} \tag{C.5}$$

The problem becomes:

$$\underline{\mathbf{w}}_i = arg\,max_{|\underline{\mathbf{w}}|=1}\{\underline{\mathbf{w}}^T \mathbf{M}_{res} \mathbf{M}_{res}^T \underline{\mathbf{w}}\} \tag{C.6}$$

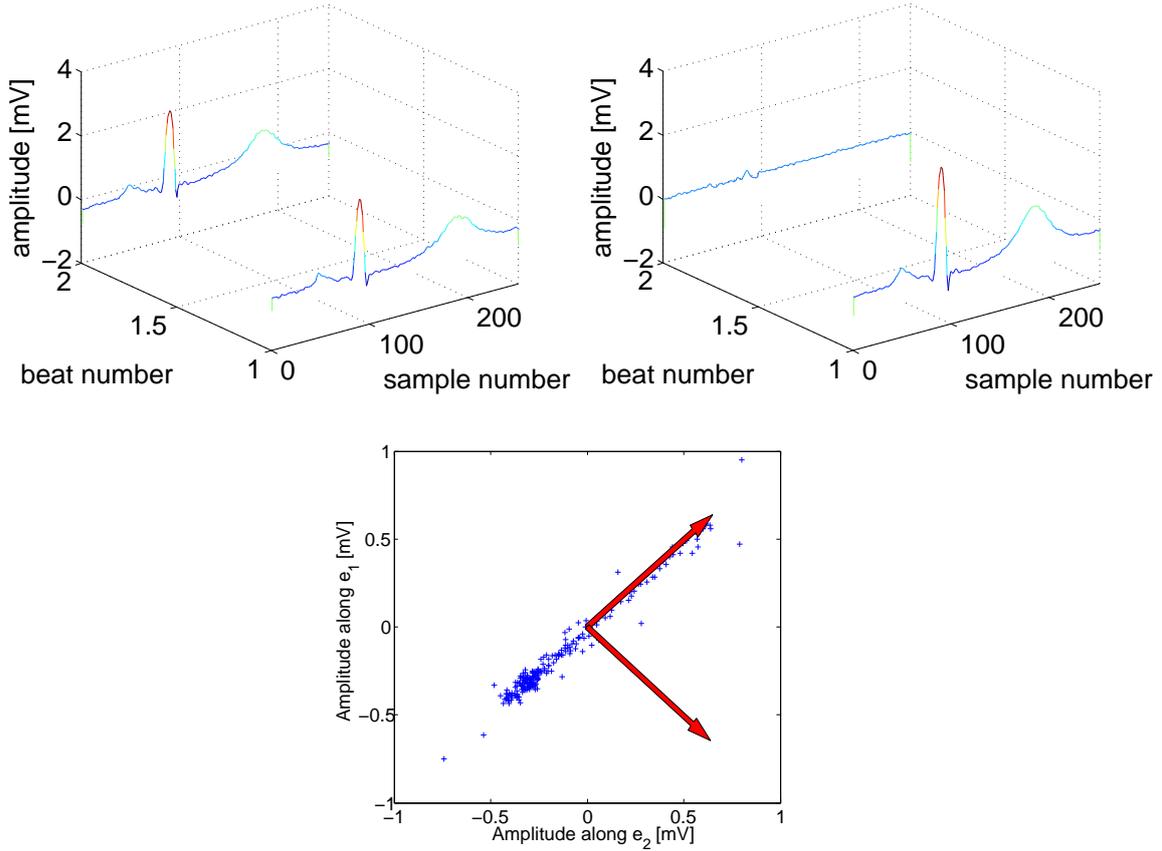

Figure C.1: Change of basis using PCA. PCA finds an algebraic transformation that maximises the entries along the diagonal of the covariance matrix $\mathbf{C}$, i.e. that maximises the variance along the axis of the new basis, while minimising the covariance. This is in order to minimise redundancy. On these plots two consecutive adult ECG cycles of the same ECG channel were selected. From left to right and top to bottom: the two ECG cycles are stacked, plot of the two ECG cycles in the PCA domain i.e. after change of basis, cycles plotted against each other revealing the high redundancy in the two signals. The PCA basis is represented by the red arrows.

We then use the Lagrange multipliers to find the local maxima and minima of the function subject to equality constraints. The Lagrange multiplier is defined:

$$L(\underline{\mathbf{w}}, \lambda) = \underline{\mathbf{w}}^T \mathbf{M}\mathbf{M}^T \underline{\mathbf{w}} - \lambda(\underline{\mathbf{w}}^T \underline{\mathbf{w}} - 1) \tag{C.7}$$

with $\lambda \in \Re$. Taking the first derivative with respect to each of the two variables gives the following system:

$$\frac{\partial L(\underline{\mathbf{w}}, \lambda)}{\partial \underline{\mathbf{w}}}\bigg|_{\underline{\mathbf{w}}=\underline{\mathbf{w}}_1} = 2(\mathbf{M}\mathbf{M}^T \underline{\mathbf{w}}_1 - \lambda \underline{\mathbf{w}}_1) = 2(\mathbf{M}\mathbf{M}^T - \lambda \mathbf{I}_p)\underline{\mathbf{w}}_1 = 0$$

$$\frac{\partial L(\underline{\mathbf{w}}, \lambda)}{\partial \lambda} = \underline{\mathbf{w}}^T \underline{\mathbf{w}} - 1 = 0$$

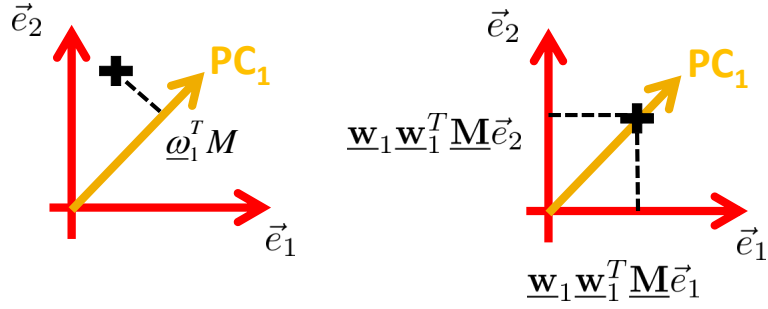

Figure C.2: PCA visuals for understand equations C.4 and C.5. The plot on the right is useful for understanding equation C.4, and the left plot is useful to understand equation C.5, where the contribution of the ith PCs are removed from the design matrix.

Solving the system provides the answer to the mathematical problem in equation C.4: $\mathbf{MM}^T\underline{\mathbf{w_1}} = \lambda\underline{\mathbf{w_1}}$ i.e. the eigenvector of the covariance matrix $\mathbf{C}$ corresponds to the principal component. Using equation C.5 we understand how to find the following principal components iteratively using this procedure. However, using such an iterative routine would hide one of the most important features of PCA, which is that it is possible to obtain a closed form solution to the mathematical problem. This solution really takes advantage of the assumptions that were made, and will be discussed in the next section.

When the principal components have been computed, it is possible to use equation C.5 to remove the contribution of a number of PCs to the signal. This was illustrated in Figure 3.5 where the contribution of the first PC (along the red arrow), corresponding to the MECG, was removed.

As can be seen in Figure C.2. PCA and ordinary least square (used when doing linear regression) are not equivalent. In ordinary least square the assumption is that predictor variables are measured exactly and only the response variable has an error component. In PCA, variables $\underline{\mathbf{x}}$ and $\underline{\mathbf{y}}$ are assumed to be measured with some error.

### C.2.2 Algebraic derivation

The purpose of PCA is to find a linear transformation $\mathbf{P}$, such that the data, $\mathbf{M}$, expressed in the initial basis are re-expressed as $\mathbf{Y}$ in a new basis, where the variance along the new basis vector is maximal. The covariance matrix $\mathbf{C} = \frac{1}{n-1}\mathbf{MM}^T$ is a positive-semidefinite and symmetric matrix. A symmetric matrix can be diagonalised by an orthogonal matrix of its eigenvectors. Using linear algebra we know that $\mathbf{C}$ can be diagonalised i.e. there exist $\mathbf{Q}$ such that $\mathbf{J} = \mathbf{QCQ}^T$ is diagonal:

$$\mathbf{J} = \mathbf{QCQ}^T = \frac{1}{n-1}\mathbf{QMM}^T\mathbf{Q}^T = \frac{1}{n-1}\mathbf{YY}^T \qquad (C.8)$$

In conclusion, we have demonstrated that it is possible to find a linear transformation that solves the PCA problem. The change of basis for this transformation corresponds to the one used for diagonalisation of the covariance matrix $\mathbf{C}_M$.

### C.2.3 Link with SVD

This paragraph highlights the link between PCA and singular value decomposition (SVD). SVD is a factorisation of real and complex matrices. The SVD of the design matrix $\mathbf{M}$ can be written:

$$\mathbf{M} = \mathbf{U}\boldsymbol{\Sigma}\mathbf{V}^T \tag{C.9}$$

where $\mathbf{U}$ is a unitary matrix, $\boldsymbol{\Sigma}$ is a rectangular diagonal matrix and $\mathbf{V}$ is an unitary matrix. The diagonal entries of $\boldsymbol{\Sigma}$ are the singular values of $\mathbf{M}$. We now show that $\mathbf{U}$ contains the eigenvectors and $\boldsymbol{\Sigma}$ the eigenvalues of the covariance matrix:

$$\begin{aligned}\mathbf{M}\mathbf{M}^T &= \mathbf{U}\boldsymbol{\Sigma}\mathbf{V}^T(\mathbf{U}\boldsymbol{\Sigma}\mathbf{V}^T)^T \\ &= \mathbf{U}\boldsymbol{\Sigma}\mathbf{V}^T\mathbf{V}\boldsymbol{\Sigma}^T\mathbf{U}^T \\ &= \mathbf{U}\boldsymbol{\Sigma}\boldsymbol{\Sigma}^T\mathbf{U}^T\end{aligned}$$

$$\begin{aligned}\mathbf{M}\mathbf{M}^T\underline{\mathbf{u}}_1 &= \mathbf{U}\boldsymbol{\Sigma}\boldsymbol{\Sigma}^T\mathbf{U}^T\underline{\mathbf{u}}_1 \\ &= \lambda_1^2\underline{\mathbf{u}}_1\end{aligned}$$

where $\underline{\mathbf{u}}_1$ an eigenvector of the covariance matrix and $\mathbf{U} = [\underline{\mathbf{u}}_1, \underline{\mathbf{u}}_2, ..., \underline{\mathbf{u}}_p]$. The last equality is obtained by considering that $\boldsymbol{\Sigma}$ is a diagonal matrix and that $\mathbf{U}$ is unitary ($\mathbf{U}\mathbf{U}^T = \mathbf{U}^T\mathbf{U} = \mathbf{I}$).

## C.3 Conclusion

The purpose of PCA is to decorrelate the dataset i.e. remove second-order dependencies. It is used for visualisation, dimensionality reduction and source separation. PCA assumes: **linearity**, **large variance represent interesting structures** and that the **principal components are orthogonal**. This enables to have an analytical solution to the PCA problem (different from ICA). It is a simple and non-parametric method (no assumption on the structure of the model). However, non-linear relationships between variables are not handled. The purpose of PCA is to decorrelate the data, however is it the best way of revealing the underlying data? **Looking for orthogonal axis can be limiting** (see Figure C.3) and the **relationship between variables might be non-linear**. Ways of addressing some of these limitations are ICA (looking at higher order statistics for identifying independent sources) and kernel PCA (looking at non-linear relationships between variables).

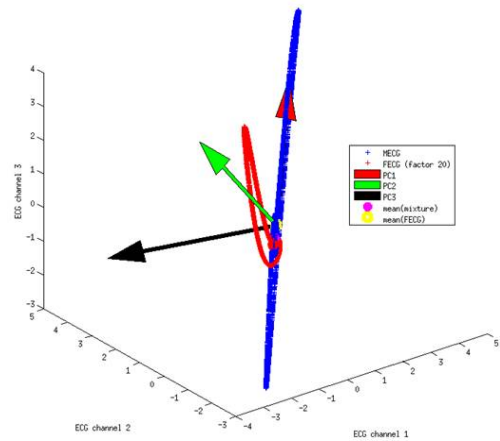

Figure C.3: Illustration of the limitation imposed by the PCA orthogonality assumption; the maternal and fetal VCGs are plotted together with the principal components that are found by PCA. The component with the highest variance (in red) is along the main maternal VCG plane. The FECG will be somewhat present along the second and third PCs (green and black arrows), although it would be best to resolve the FECG along a non-orthogonal axis in the fetal VCG plane.

# Appendix D

# Background on ICA

This section presents the basic principles of Independent Component Analysis (ICA) that are necessary for understanding its strengths and limitations in the context of NI-FECG extraction. In the following, the design matrix, **M**, is defined as in Appendix C.

## D.1  Introduction

Blind source separation (BSS) approaches have been employed for NI-FECG analysis many times over the last decade. BSS can be formulated by assuming an unobservable set of sources, $\underline{\mathbf{s}} = [s_1^k, s_2^k, ..., s_p^k]$, (such as the maternal dipole, fetal dipole and noise) are linearly mixed through some unknown mixing matrix **A** such that:

$$\underline{\mathbf{m}} = \mathbf{A}\underline{\mathbf{s}}, \tag{D.1}$$

where $\mathbf{A} \in \Re^{p \times p}$ and $\underline{\mathbf{s}} \in \Re^p$ and $\underline{\mathbf{m}} \in \Re^p$ with $p$ as the number of sources. This holds when it is assumed that the number of sources and observed variables are equal or, in other words, that there are as many sources as sensors (or ECG channels in the context of this thesis). The elements of **A** are found through a process of extracting sources as statistically independent as possible, by maximising some measure of non-Gaussianity or by minimising the mutual information between each of the sources.

The framework for ICA was introduced by Jeanny Herault and Christian Jutten in 1986 [200] and was more clearly stated by Pierre Comon in 1994 [201]. The purpose of ICA is to find a linear transformation that will minimise the statistical dependence between the components of the data. In the context of NI-FECG, the purpose of ICA is to find a transformation that will separate the multivariate signal recorded on the abdomen into its additive subcomponents, which include the NI-FECG and the MECG. Figure D.1 illustrates ICA effect on some simulated sinusoid mixture; Figure D.1a shows the initial sinusoidal functions, Figure D.1b the linear mixture of the two sinusoidal functions and Figure D.1c the de-mixed signals using ICA. Classical ICA algorithms assume that: **the source signals are independent** and that **the distribution of each source signal is non-Gaussian**. Typically, ICA algorithms can be broken down into the following steps:

1. centring: subtract the mean of the signal

2. whitening: uncorrelate the data ($E[xx^T]$). This means to treat all dimensions equally, and significantly simplify the ICA problem

3. dimensionality reduction: this step is optional and can be used to simplify the ICA problem by removing the principal components (obtained with PCA) that have the least variance

4. iterative algorithm: after the data have been pre-processed (centring, whitening, dimensionality reduction), an iterative algorithm is used to find the independent components. There exists a number of formulations for this procedure. Example of classical ICA algorithms are JADE [63] (which is iterative with a defined number of iterations) and FastICA [202] (which is iterative with an adaptive number of iterations).

The different steps of the ICA procedure are illustrated in Figure D.2; Figure D.2a shows two independent variables $x$ and $y$ plotted against each other, Figure D.2b shows two dependant variables, $x_2$ and $y_2$, plotted against each other, where $x_2$ and $y_2$ are linear combinations of $x$ and $y$. Figure D.2c shows the data after the whitening procedure; Figure D.2d illustrates the transformations that is found by ICA in order for the distributions of each of the sources to be the least Gaussian (D.2d left, before transformation and D.2d right, after the ICA rotation). These plots also illustrate the geometric intuition for understanding the whole transformation; whitening restores the initial shape of the data and the ICA iterative algorithm looks for the rotation that will maximise the non-Gaussianity of the distributions.

## D.2 Mathematical basis

### D.2.1 Whitening procedure

The purpose of the whitening procedure is to transform the observation matrix, $\mathbf{M} \in \Re^{p \times q}$, linearly so that we obtain a new observation matrix, $\tilde{\mathbf{M}}$, which is white, i.e. its components are uncorrelated. $\tilde{\mathbf{M}}\tilde{\mathbf{M}}^T = \mathbf{I}$ [64]. This treats all dimensions equally, and significantly simplifies the ICA problem. Whitening can be performed using eigenvalue decomposition (EVD) of the covariance matrix $\mathbf{C}$[1]:

$$\mathbf{C} = E[\mathbf{M}\mathbf{M}^T], \quad E[\mathbf{C}] = \mathbf{PDP^T}, \tag{D.2}$$

---

[1]Or equivalently SVD. In the special case of a normal matrix, SVD and EVD are the same transform. The covariance matrix $\mathbf{C}$ is normal.

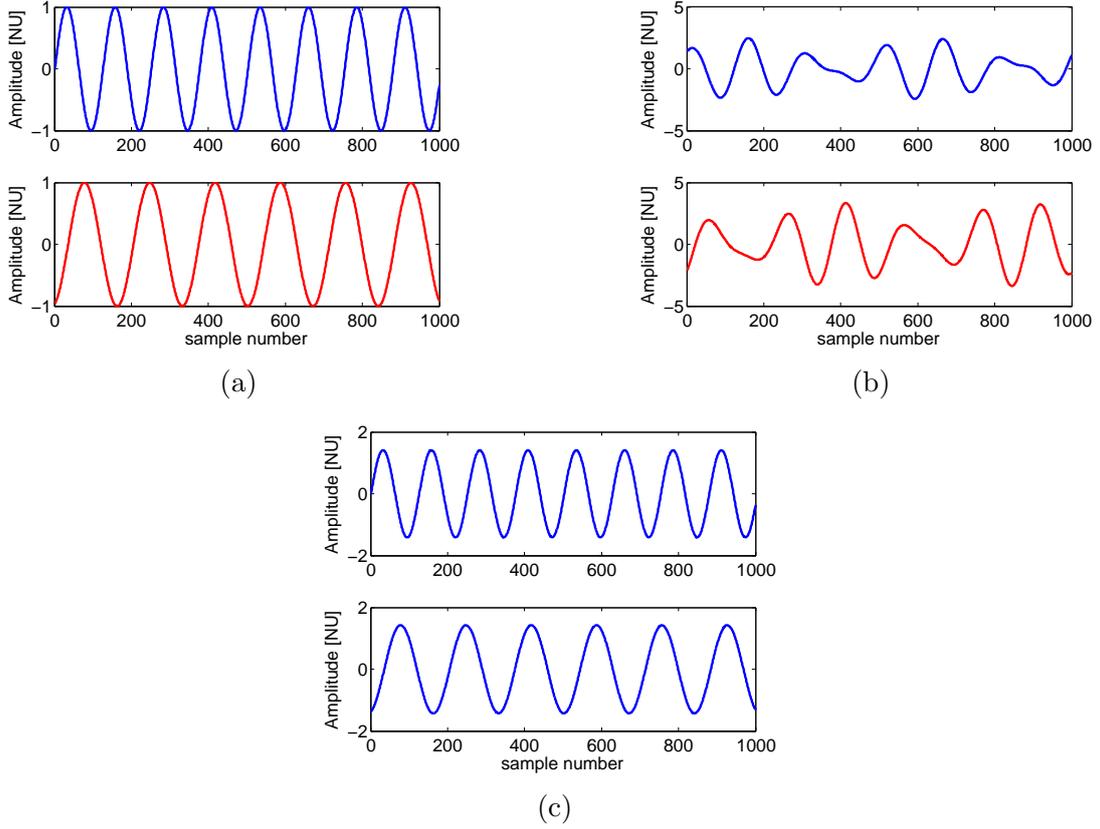

Figure D.1: ICA illustration: (a) the initial sinusoidal functions, (b) the linear mixture of the two sinusoidal functions and (c) the de-mixed signals using ICA. Plots inspired from `http://sccn.ucsd.edu/~arno/indexica.html`

where $\mathbf{P} \in \Re^{p \times p}$ is a square matrix whose ith column is the eigenvector $\underline{\mathbf{w}}_i$ of $\mathbf{C}$. $\mathbf{D}$ is a diagonal matrix whose diagonal elements are the corresponding eigenvalues. We write:

$$\tilde{\mathbf{M}} = \mathbf{P}\mathbf{D}^{-1/2}\mathbf{P}^T\mathbf{M} \tag{D.3}$$

This transformation moves the eigenvector basis to the new PCA domain, rescales the eigenvalues in this domain and finally returns to the original basis.

$$\tilde{\mathbf{M}} = \mathbf{P}\mathbf{D}^{-1/2}\mathbf{P}^T\mathbf{A}\underline{\mathbf{s}} = \tilde{\mathbf{A}}\underline{\mathbf{s}} \tag{D.4}$$

$\tilde{\mathbf{A}}$ is now orthogonal:
$$E\{\tilde{\mathbf{M}}\tilde{\mathbf{M}}^T\} = \tilde{\mathbf{A}}E\{\underline{\mathbf{s}}\,\underline{\mathbf{s}}^T\}\tilde{\mathbf{A}}^T = \tilde{\mathbf{A}}\tilde{\mathbf{A}}^T = \mathbf{I}$$

$\mathbf{A} \in \Re^{p \times p}$ so it has $p^2$ parameters to estimate

$\tilde{\mathbf{A}} \in \Re^{p \times p}$ is orthogonal so it has $p(p-1)/2$ degree of freedom.

In a sense whitening already solves half of the ICA problem (complexity is $p(p-1)/2$ in-stead of $p^2$ after whitening). After whitening, only a rotation is left to be found by the ICA algorithm. As the ICA procedure is more complex than whitening, it is useful to whiten the data as a preprocessing step. Note that it was assumed that $E\{\underline{s}\,\underline{s}^T\} = I$ i.e. $\underline{s}$ is white. It is also important to note that systematically whitening the data is equivalent to assuming that second-order statistics are "infinitely more reliable" than any alternative statistics which is arguably false [62].

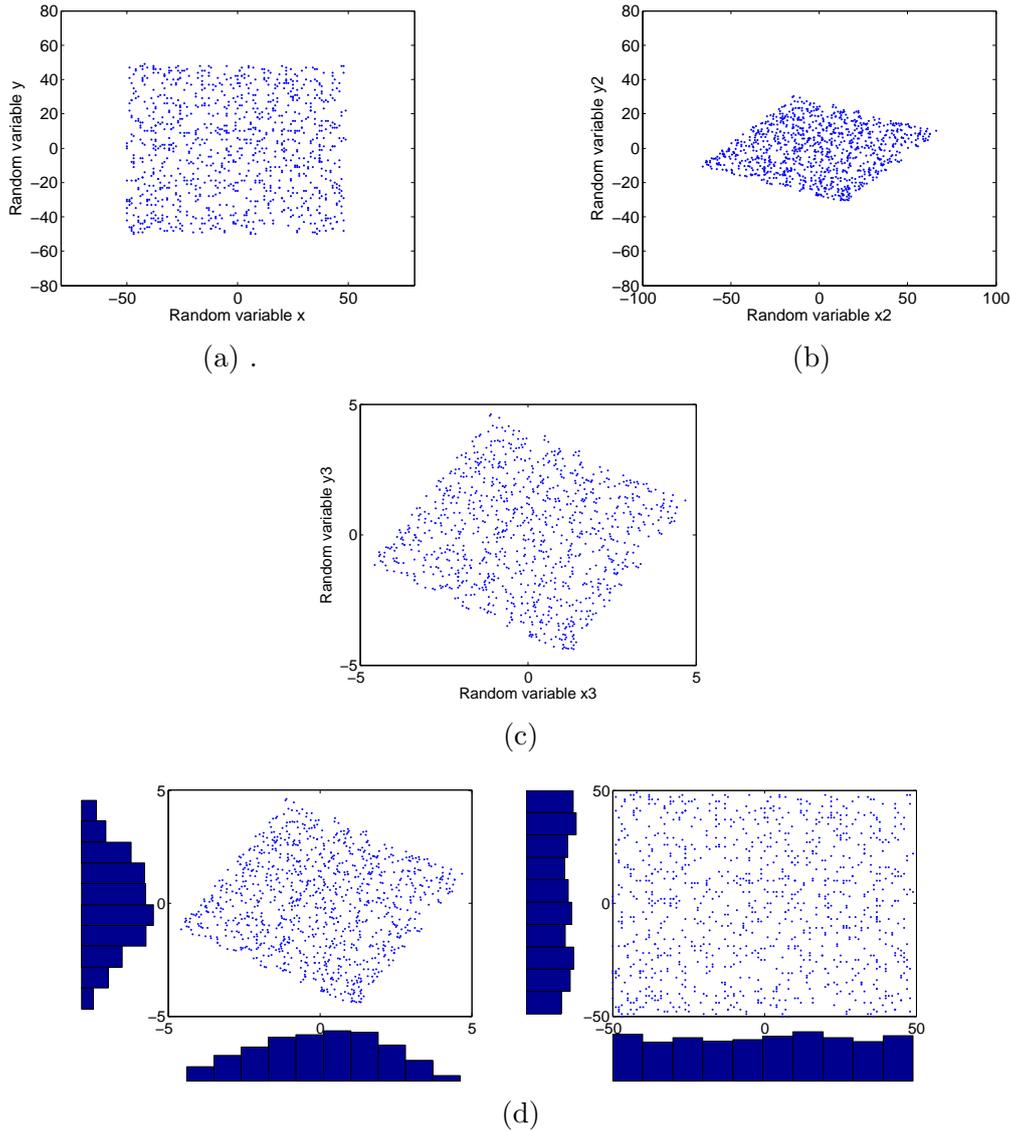

Figure D.2: The different steps of the ICA procedure; figure D.2a shows two independent variables $x$ and $y$ plotted against each other, figure D.2b shows two dependant variables $x_2$ and $y_2$ plotted against each other, where $x_2$ and $y_2$ are a linear combination of $x$ and $y$. Figure D.2c shows the data after the whitening procedure and figure D.2d illustrates the transformation that is found by ICA so that the distributions of each source is least Gaussian. Plots inspired by http://sccn.ucsd.edu/~arno/indexica.html

## D.2.2 Defining independent components

The ICA procedure looks for independent components by maximising a statistical measure of independence for the estimated components. There are different categories of such statistical measures in the context of non-parametric ICA. For example, minimisation of mutual information (e.g. Negentropy approximated: $J(y) \approx \frac{1}{12}E\{y^3\}^2 + \frac{1}{48}kurt(y)^2$) and higher moments[2], such as Kurtosis $kurt(y) = E\{y^4\} - 3(E\{y^2\})^2$. These measures are often called **contrast functions**. Their optimisation estimates the independent components.

## D.3 ICA and NI-FECG

ICA was used for the first time in the context of NI-FECG extraction by De Lathauwer *et al* [203]. The authors showed that it was possible to isolate the FECG from the abdominal mixture on a single recording. It has been substantially used since then for this application. However, the benefit of ICA in leading to an improved detection of the FQRS was not sufficiently quantitatively evaluated. This shortcoming was addressed in Chapter 7.

## D.4 Conclusion

ICA is a powerful source separation technique and it has been widely used in biomedical engineering over the past decade. ICA relies on the assumption that **the source signals are independent** and that **the distribution of each source signal is non-Gaussian**. The main limitations of ICA are:

- it is not possible to determine the variances of the independent components. For this reason the variance is set to one for each independent component, and so ICA does not preserve scaling, which can be seen in Figure D.1

- it is not possible to determine the order of the independent components as in PCA where the components are ordered with respect to the energy they contain

- traditional ICA techniques do not handle non-linear relationships between observed variables

- no phase information: the temporal structure of the signal is not taken into account by traditional ICA approaches. In the context of NI-FECG extraction, $\pi CA$ has been used as a measure of periodicity for the contrast function in the work of Sameni *et al* [51]. However, it has not been demonstrated that this alternative contrast function improves the separation of the NI-FECG.

---

[2]Higher order moments $E(X^n) = \int_{-\infty}^{+\infty} x^n f(x)dx$. Mean, variance, skewness, kurtosis, hyperskewness, hyperflatness

# Appendix E

# Open source code

While publishing scientific papers as well as this thesis, it is important to enable other researchers in the field to reproduce the work that has been done. This is both proof of good science and helps to accelerate the research in the field by avoiding duplicates (and sometimes inexact re-implementation) of code. This section presents some of the code produced in the context of this thesis, which were contributed to `Physionet.org`.

## E.1 Fetal ECG synthetic simulator *fecgsyn*

URL: `http://www.physionet.org/physiotools/ipmcode/fecgsyn/`

The *fecgsyn* code described in detail in Chapter 4 and in Behar *et al* [6]. The graphic user interface *fecgsyngui* for easy access to the underlying *fecgsyn* code was published in Alvi *et al.* [204].

### E.1.1 Presentation

*fecgsyn* is a realistic non-invasive fetal ECG (NI-FECG) generator that uses the Gaussian ECG model originally introduced by McSharry et al. [4]. The code generates synthetic NI-FECG mixtures with user-settable noise, heart rate and heart rate variability, rotation of the maternal and fetal heart axes due to respiration, fetal movement, contractions, ectopic beats and multiple pregnancy. Any desired number of electrodes can be placed on the maternal abdomen. The synthetic ECG simulator is a good tool for modelling realistic FECG-MECG mixtures and specific events, such as abrupt heart rate increase, in order to benchmark signal processing algorithms on realistic data for clinically important scenarios.

*fecgsyn* is the result of a collaboration between the Department of Engineering Science, University of Oxford (DES-OX) and the Institute of Biomedical Engineering, TU Dresden (IBMT-TUD). The authors are Joachim Behar (DES-OX), Fernando Andreotti (IBMT-TUD), Julien Oster (DES-OX), Sebastian Zaunseder (IBMT-TUD) and Gari Clifford (DES-OX).

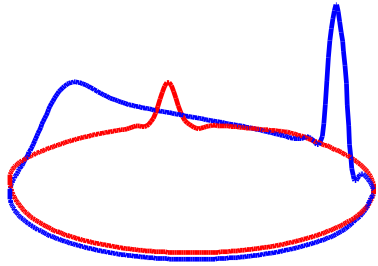

Figure E.1: *fecgsyn* software logo. The logo shows two ECG cycles in polar coordinates, one representing the MECG and the second one, with a smaller amplitude, represents the FECG. The ECG cycles are overlapping as is the case on the abdominal ECG mixture. The ECG are represented in polar coordinates because of the ECG model which is defined as such.

All information and code related to *fecgsyn* has been upload to Physionet.org. The code is freely available under the GNU GPL (General Public License). The overall length of the source code is over 2500 lines (not including the graphic user interface).

### E.1.2 Graphic user interface (*fecgsyngui*)

A friendly MATLAB user interface, *fecgsyngui*, was created to facilitate access to the source code and to allow interested researchers to familiarise themselves with the different physiological effects modelled. ECG mixtures can be generated with realistic beat-to-beat variability, morphology, amplitude modulation due to respiration and fetal movement using *fecgsyngui*. The GUI lets the user to visualise and generate a set of realistic scenarios to stress test and train separation and detection algorithms. Electrodes can be placed on the maternal abdomen using an interactive geometric representation. Simulation parameters and generated signals can be exported to .mat files for further processing.

Figure E.2 displays the main page where the generated ECG are plotted. Using the options on the left hand side box, it is possible to select what signals to display. Figure E.3 displays the "custom page" used to specify the parameters of the model.

### E.1.3 Clarification on contribution

The *fecgsyn* software was the product of joint work with Fernando Andreotti (IBMT-DUD). Fernando contributed the 'modelling fetal movement' (4.5.2.8) as well as the contraction effects on the modulation of the heart position and noise on the AECG (4.5.2.9). The implementation of the code relative to these two items as well as some of the code related to signal calibration were contributed by Fernando. The user interface was implemented by Mohsan Alvi (DES-

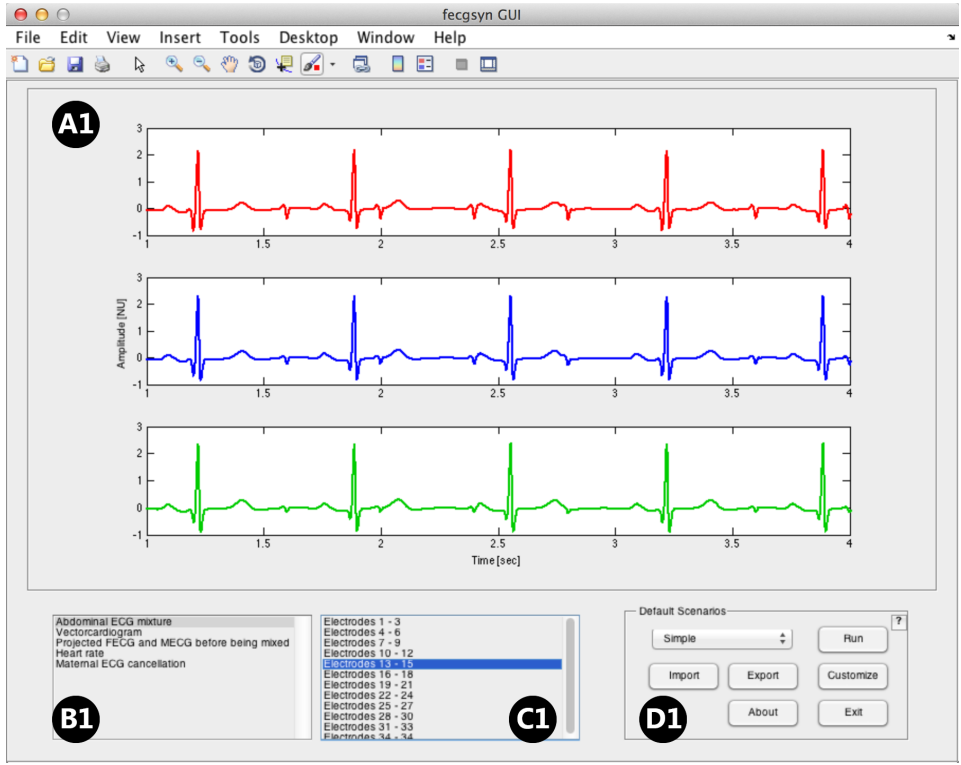

Figure E.2: *fecgsyn* GUI - display window. A1: plot display panel of maternal-fetal mixtures for three electrodes, B1: list of visualisation options, C1: subplot list, if multiple plots are available (e.g. for different electrodes), D1, GUI controls.

OX) under my initiative and supervision.

## E.2 NI-FECG source separation code

URL: http://physionet.org/challenge/2013/sources/
The NI-FECG source separation code was contributed during the Challenge 2013. The separation methods available were presented in Chapter 7 and published in [2].

### E.2.1 Presentation

Open source code from the Physionet Challenge 2013 is available on `Physionet.org`. In this package the source code for the techniques used in Behar *et al.* [2] (and presented in Chapter 7) for NI-FECG source separation are available. The code implementing the LMS, RLS and ESN techniques of the publication in Behar *et al.* [1] (presented in Chapter 6) is available on my personal webpage. Figure E.4 shows the example of using one of the implemented techniques ($TS_{pca}$) for cancelling the MECG.

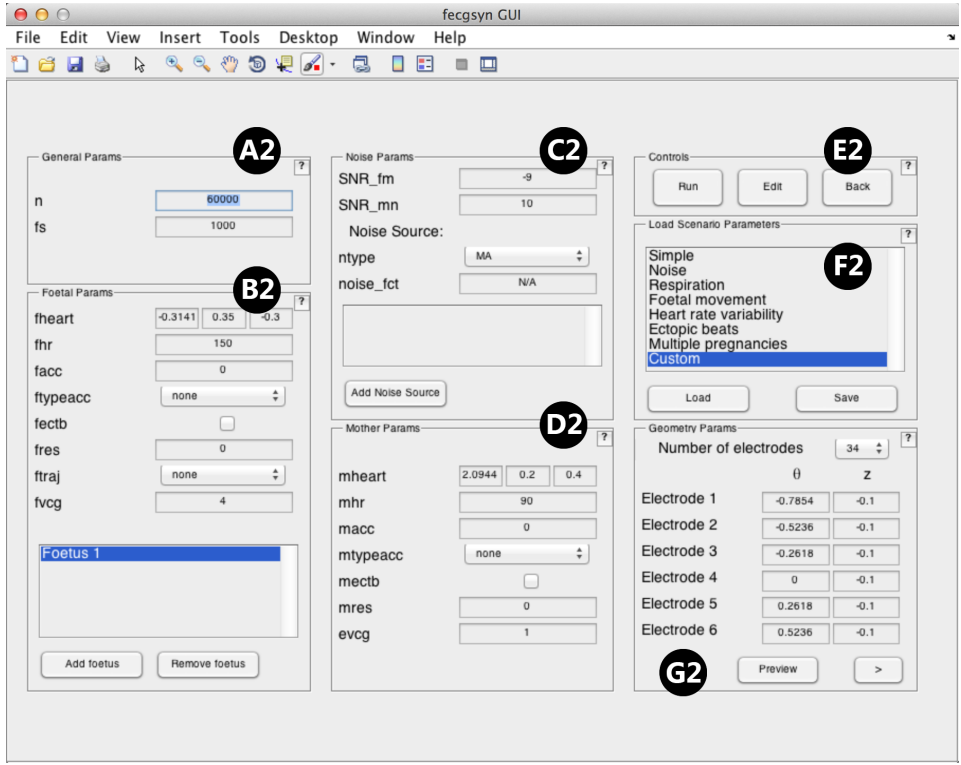

Figure E.3: *fecgsyn* GUI - customisation window. A2: general signal parameters, B2: fetal parameters, C2: noise parameters, D2: maternal parameters, E2: GUI controls, F2: simulation scenario list, G2: electrode geometry parameters.

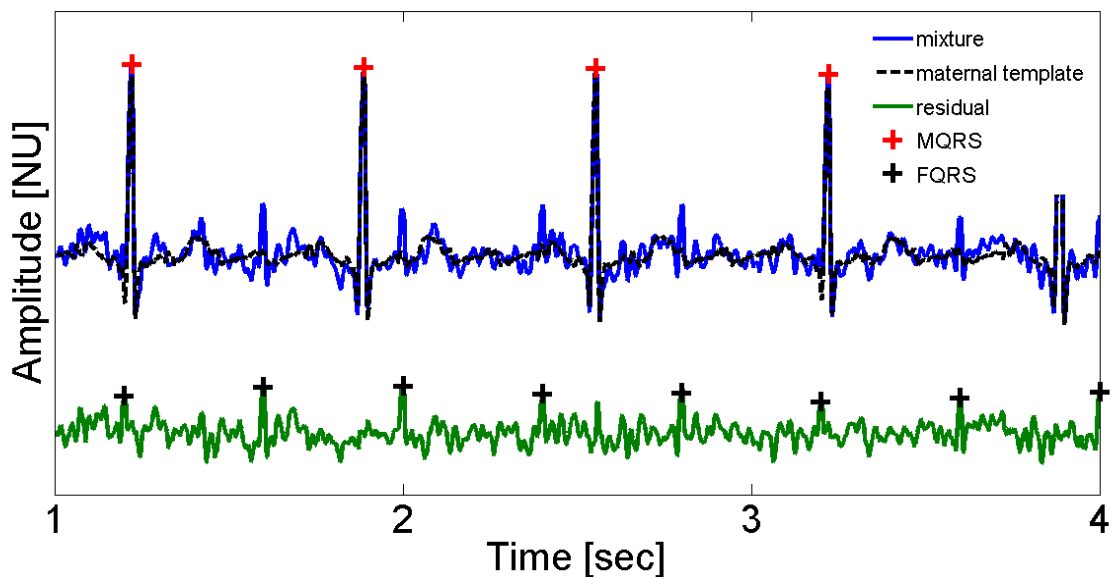

Figure E.4: Example of template subtraction using $TS_{pca}$ as implemented in [1]. $TS_{pca}$ was performed on an abdominal ECG signal generated with *fecgsyn*

# E.3 Random search toolbox *rs-toolbox*

URL: `http://www.physionet.org/physiotools/random-search/`

Random search was used in Chapters 7 and 8.

### E.3.1 Presentation

A major issue with many signal processing and machine learning algorithms is the lack of optimisation methods for determining the numerous parameters associated with the model, as well as, the knowledge of which parameters are relevant. These parameters are usually tuned by trial and error (manual search) or by grid search. In 2012, Bergstra *et al.* [123] empirically showed that random search performed as well as, or better, than grid search, while reducing the computational cost substantially.

The RS-toolbox (Figure E.5) is contributed by Joachim Behar, Alistair Johnson, Julien Oster and Gari D. Clifford. All authors are part of the Institute of Biomedical Engineering at the University of Oxford. The RS-toolbox code provides functions for creating Efficiency Curves (EC, e.g. Figure E.6) and Automatic Relevance Determination (ARD) plots (e.g. Figure E.7). This code, together with its example of parameter search for training an echo state neural network [124], should provide users with a quick insight on how to use random search and produce EC and ARD from the searched parameters.

The work presented at NIPS in 2013 *"An Echo State Neural Network for fetal Electrocardiogram Extraction Optimised by Random Search"* (Behar *et al.* [124]) was the first time random search was used in our group. Since then it has been successfully used in other works that we published or recently submitted [2, 10, 127].

### E.3.2 Clarification on contribution

The work and code published in Behar *et al.* [124] was joint work with Alistair Johnson. I contributed the code and *know how* relative to the ESN and Alistair the code for the ARD and EC plots.

# E.4 Signal quality and QRS detector

URL: `http://physionet.org/challenge/2014/`

### E.4.1 Presentation

The bSQI signal quality index and the QRS detector used in this thesis were open sourced when participating to the Physionet Challenge 2014 addressing the topic of "Robust Detection

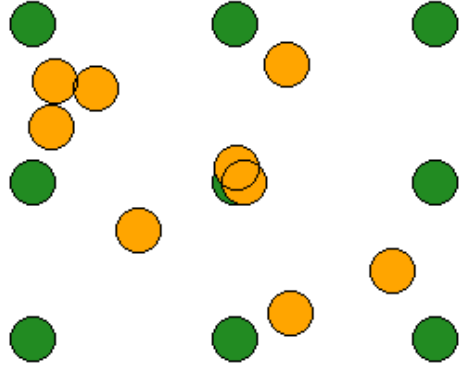

Figure E.5: *rs-toolbox* logo. The logo represents the samples from a grid (green) and random search (orange) of nine trials for optimising a function $g$ when considering two variables. Assuming the y-variable does not change the performance of the function $g$ (i.e. $g$ is independent of y over the sampled interval), then random search will likely find a better set of parameters for optimising $g$ given that it spans over more values along the x-axis. For a high dimensional search, this phenomena often happens.

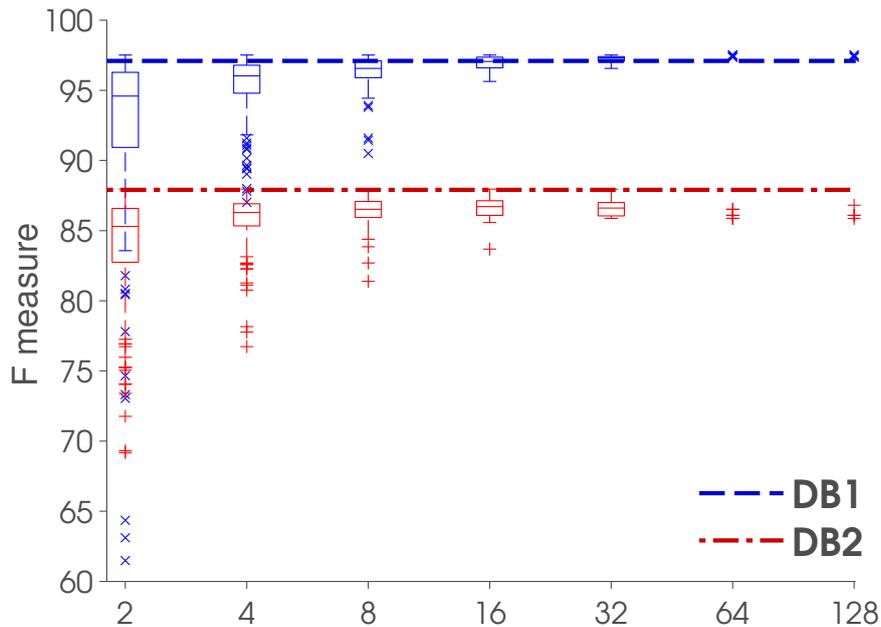

Figure E.6: Performance of the ESN with respect to number of random search iterations. As each search is i.i.d., the overall 512 searches can be subsampled to replicate as many as $\frac{512}{N}$ searches each of size $N$. Blue 'x's are training set performances ($DB_1$). Red '+'s are test set performances ($DB_2$). This plot is useful for analysing how many random search iterations are necessary to obtain similar results as grid search optimisation.

of Heart Beats in Multimodal Data". For further reference see the paper Johnson *et al.* [14].

The MGH/MF Waveform database [205] was used to evaluate the QRS detectors. The MGH/MF database is a collection of 250 recordings from hemodynamic (ABP, CVP, PAP) and electrocardiographic (three leads) signals acquired from patients in critical care units, operating rooms, and catheterisation labs. These recordings were usually one hour in length, but could vary between 12 to 86 minutes. All recordings had manually corrected R-peak annotations. The MGH/MF database was reformatted into 5274 10-minute recordings, containing one ECG and one ABP channel. Three QRS detectors were benchmarked: (1) "gqrs" (available on Physionet [65]), which consists of a QRS matched filter with a custom built set of heuristics (such as search back). (2) "coqrs" [206, 207, 208] based on the peak energy (no search back). (3) "jqrs" (this thesis) consists of a window-based peak energy detector, replacing the original band-pass filter with a QRS matched filter (Mexican hat) and an addition of heuristic ensuring that no detections were made during flat lines. The QRS detector developed in this thesis performed best (see Table E.1). Unlike gqrs and coqrs, the jqrs detector works on 15 sec windows (i.e. offline or with a delay of 15 sec). Note that coqrs has no search back routine implemented (as opposed to gqrs and jqrs) since it was designed to be run as an online QRS detector.

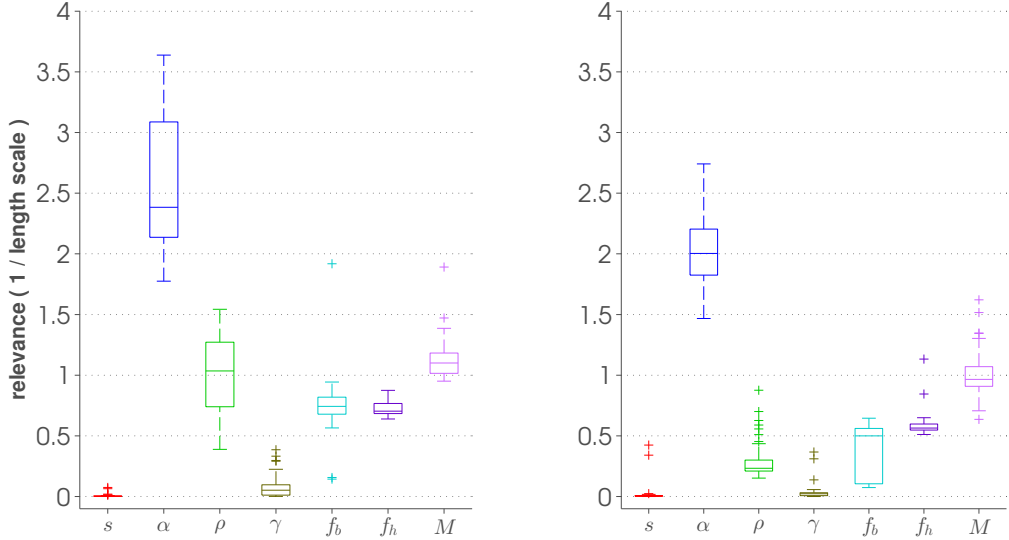

Figure E.7: Automatic relevance determination (ARD) applied to the ESN hyper-parameters and the prefiltering parameters. Left: ARD performed on $DB_1$. Right: ARD performed on $DB_2$. This plot can be used to assess the importance of a variety of hyperparameters of the ESN over the intervals the parameters were sampled from. The left most parameters $s$ is the seed value for a Mersenne twister pseudorandom number generator, which is used as the baseline (because its relevance should be zero).

### E.4.2 Clarification on contribution

My contribution to this work has been: to provide the signal quality evaluation for the ECG channel (see Figure E.8 for a visual example and Chapter 5 for the work on signal quality), the QRS detector denoted jqrs (which corresponds to the Pan and Tompkins detector I developed in the context of this thesis). I also contributed to the technical work for formatting and submitting the source code to be scored on the test set by the MIT Physionet team, as well as to the methodology that we implemented for our entries. bSQI and jqrs were instrumental in enabling us to win Phase III of the Challenge which was the most difficult event of this competition. Lastly, bSQI built upon the idea originally introduced by Li *et al.* [110], however although the idea was kept, the implementation of this SQI was reworked over the course of the PhD finding its most recent form in the code open sourced during the Challenge 2014.

|  | | Average | | Gross | | |
| --- | --- | --- | --- | --- | --- | --- |
|  | Algorithm | Se(%) | PPV(%) | Se(%) | PPV(%) | $F_1$(%) |
| Single detectors | gqrs | 91.75 | 96.19 | 91.87 | 96.63 | 94.19 |
|  | jqrs | **92.66** | **96.51** | **93.00** | **96.72** | **94.82** |
|  | coqrs | 91.19 | 93.46 | 91.77 | 93.41 | 92.58 |

Table E.1: QRS detector benchmarked on the MGH database. Both average and gross statistics are indicated. Note that coqrs has no search back routine implemented (as opposed to gqrs and jqrs) since it was designed to be run as an online QRS detector.

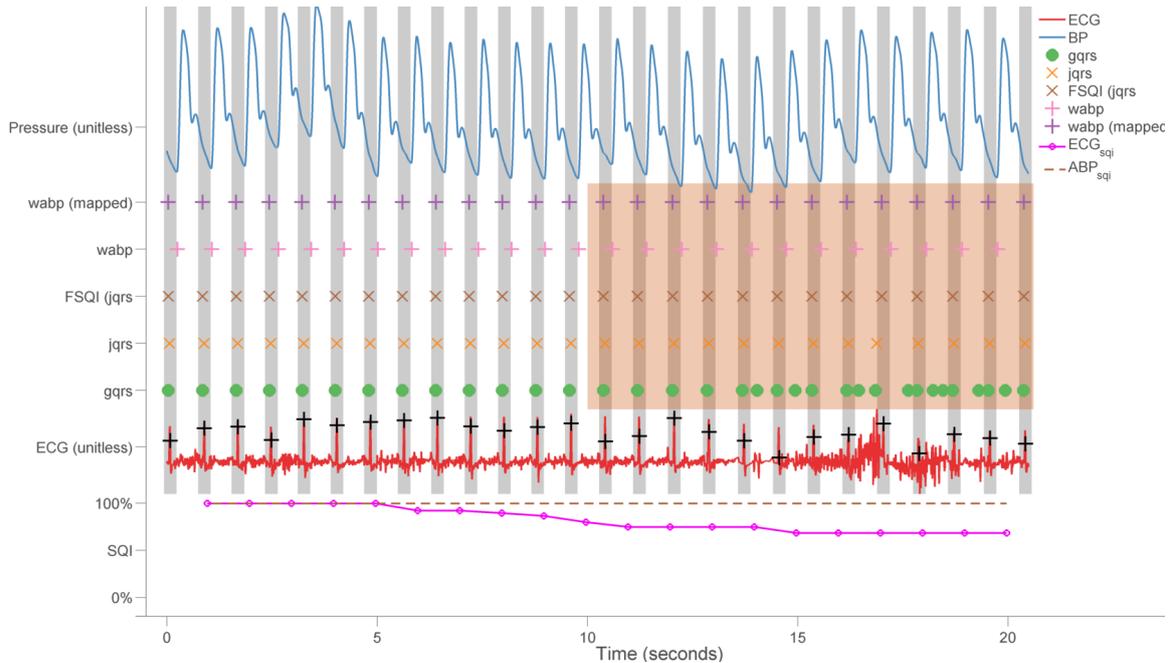

Figure E.8: Example of SQI switching when ECG is low quality. The top row contains the ABP signal, while the ECG signal is depicted in red in the third row from the bottom, with the reference annotations marked by black crosses. Vertical grey lines represent the windows for acceptable detected peaks. The two bottom graphs represent both bSQI (plain line with circle markers) and aSQI (dotted lines). The annotations of the individual detectors are depicted in the middle rows, and the region that uses ABP annotations instead of ECG annotations is shaded red.